\documentclass[final,5p,times,twocolumn,nopreprintline]{elsarticle}

\usepackage{framed,multirow}

\usepackage{amssymb} 
\usepackage{latexsym}

\usepackage{placeins}
\usepackage{booktabs}
\usepackage{graphicx}
\usepackage{subcaption}
\usepackage[table]{xcolor} 
\usepackage{lipsum} 
\newcolumntype{P}[1]{>{\centering\arraybackslash}p{#1}}
\usepackage[percent]{overpic}
\usepackage{wrapfig}
\usepackage{amsmath}
\usepackage{dblfloatfix}

\usepackage{pgfplots}
\definecolor{darkblue}{RGB}{31, 119, 180}
\definecolor{darkorange}{RGB}{244, 104, 20}
\definecolor{darkred}{RGB}{214, 39, 40}
\definecolor{darkgreen}{RGB}{44, 160, 44}
\definecolor{darkpurple}{RGB}{148, 103, 189}

\usepackage{tablefootnote}

\usepackage{tikz} \usetikzlibrary{calc, arrows.meta, intersections, patterns, positioning, shapes.misc, fadings, through,decorations.pathreplacing,shadows.blur, decorations.pathmorphing}

\tikzstyle{phase} = [align=center, outer sep=1pt, minimum width=3cm, minimum height=1cm, text=white, font=\footnotesize]
\tikzstyle{event} = [align=center, black, font=\footnotesize]

\definecolor{timeline_col1}{RGB}{108, 43, 108}
\definecolor{timeline_col2}{RGB}{142, 50, 112}
\definecolor{timeline_col3}{RGB}{176, 59, 108}
\definecolor{timeline_col4}{RGB}{208, 73, 97}
\definecolor{timeline_col5}{RGB}{226, 104, 92}
\definecolor{timeline_col6}{RGB}{233, 140, 107}

\usepackage{adjustbox}

\definecolor{reg_phase}{HTML}{4C72B0}  
\definecolor{train_phase}{HTML}{DD8452} 
\definecolor{sub_phase}{HTML}{55A868}   
\definecolor{eval_phase}{HTML}{C44E52}  
\definecolor{timeline_bg}{HTML}{F9F9F9} 

\tikzset{
    phase/.style = {
        rectangle, minimum width=3cm, minimum height=1.5cm,
        align=center, text=white, font=\small, rounded corners=2mm
    },
    event/.style = {
        align=center, text=black, font=\footnotesize
    },
    arrow/.style = {thick, ->, >=Stealth, black},
}

\usepackage{pifont}
\usepackage{tabularx}
\usepackage{siunitx}

\usepackage{suffix}		
\newcommand\sref[1]{\protect\subref{#1}}
\WithSuffix\newcommand\sref*[1]{\protect\subref*{#1}}

\usepackage{url}
\usepackage{xcolor}
\usepackage{hyperref}
\usepackage{makecell}

\DeclareRobustCommand{\legendsquare}[1]{%
  \textcolor{#1}{\rule{2ex}{2ex}}%
}

\usepackage{adjustbox}
\usepackage{graphicx}

\usepackage{cleveref}

\usepackage{float}

\usepackage{gensymb}

\usepackage[absolute,overlay]{textpos}
\usepackage[protrusion=true, expansion=false]{microtype}
\usepackage{titlesec}
\usepackage{etoolbox}
\usepackage{ragged2e}
\usepackage{relsize}

\usepackage[most]{tcolorbox}
\definecolor{main}{HTML}{5989cf}    
\definecolor{sub}{HTML}{e6e6e6}     

\newtcolorbox{box_phase_recognition}{
    float, 
    floatplacement=t!, 
    colback = sub, 
    boxrule = 1pt  
}

\newtcolorbox{box_instance_segmentation}{
    float, 
    floatplacement=b!, 
    colback = sub, 
    boxrule = 1pt  
}

\newtcolorbox{box_keypoint_estimation}{
    float, 
    floatplacement=b!, 
    colback = sub, 
    boxrule = 1pt  
}

\newcolumntype{Y}{>{\centering\arraybackslash}X}
\newcolumntype{R}{>{\raggedleft\arraybackslash}X}
\newcolumntype{P}[1]{>{\centering\arraybackslash}p{#1}}
\newcolumntype{Z}[1]{>{\raggedleft\arraybackslash}p{#1}}

\usepackage{fancyhdr}
\begin{document}

\begin{frontmatter}

\title{Comparative validation of surgical phase recognition, instrument keypoint estimation, and instrument instance segmentation in endoscopy: Results of the PhaKIR 2024 challenge}



\author[1,2,3]{Tobias Rueckert}

    \author[1]{David Rauber}
    \author[1]{Raphaela Maerkl}
    \author[1,4,5]{Leonard Klausmann}
    \author[1,4,5]{Suemeyye R. Yildiran}
    \author[1]{Max Gutbrod}
    \author[1]{Danilo Weber Nunes}
    
    \author[10,11]{Alvaro Fernandez Moreno}
    \author[10]{Imanol Luengo}
    \author[10,11]{Danail Stoyanov}
    \author[10]{Nicolas Toussaint}
    
    \author[15]{Enki Cho}
    \author[15]{Hyeon Bae Kim}
    \author[15]{Oh Sung Choo}
    \author[15]{Ka Young Kim}
    \author[15]{Seong Tae Kim}

    \author[20]{Gonçalo Arantes}

    \author[25]{Kehan Song}
    \author[25]{Jianjun Zhu}
    \author[25]{Junchen Xiong}
    \author[25]{Tingyi Lin}

    \author[30]{Shunsuke Kikuchi}
    \author[30]{Hiroki Matsuzaki}
    \author[30]{Atsushi Kouno}

    \author[35]{João Renato Ribeiro Manesco}
    \author[35]{João Paulo Papa}

    \author[40]{Tae-Min Choi}
    \author[40]{Tae Kyeong Jeong}
    \author[40]{Juyoun Park}

    \author[45]{Oluwatosin Alabi}
    \author[45]{Meng Wei}
    \author[45]{Tom Vercauteren}

    \author[50]{Runzhi Wu}
    \author[50]{Mengya Xu}
    \author[50]{An Wang}
    \author[50]{Long Bai}
    \author[50]{Hongliang Ren}

    \author[55]{Amine Yamlahi}
    \author[55]{Jakob Hennighausen}
    \author[55]{Lena Maier-Hein}

    \author[60]{Satoshi Kondo}
    \author[61]{Satoshi Kasai}
    \author[62]{Kousuke Hirasawa}

    \author[65]{Shu Yang}
    \author[65]{Yihui Wang}
    \author[65,66,67]{Hao Chen}

    \author[70]{Santiago Rodríguez}
    \author[70]{Nicolás Aparicio}
    \author[70]{Leonardo Manrique}
    \author[70]{Juan Camilo Lyons}
    \author[70]{Olivia Hosie}
    \author[70]{Nicolás Ayobi}
    \author[70]{Pablo Arbeláez}

    \author[75]{Yiping Li}
    \author[75]{Yasmina Al Khalil}
    \author[75,76]{Sahar Nasirihaghighi}

    \author[80,81]{Stefanie Speidel}

    \author[90,91]{Daniel Rueckert}
    \author[3,93]{Hubertus Feussner}
    \author[3,93]{Dirk Wilhelm}
    \author[1,4,5]{Christoph Palm}

    \address[1]{Regensburg Medical Image Computing (ReMIC), OTH Regensburg, Regensburg, Germany}
    \address[2]{AKTORmed Robotic Surgery, Neutraubling, Germany}
    \address[3]{Research Group MITI, TUM University Hospital, School of Medicine and Health, Technical University of Munich, Munich, Germany}
    \address[4]{Regensburg Center of Biomedical Engineering (RCBE), OTH Regensburg and Regensburg University, Regensburg, Germany}
    \address[5]{Regensburg Center of Health Sciences and Technology (RCHST), OTH Regensburg, Regensburg, Germany}

    \address[10]{AI Centre of Excellence, Medtronic Ltd., Watford, UK}
    \address[11]{Engineering Sciences, University College London, London, UK}

    \address[15]{Augmented Intelligence Lab, Kyung Hee University, Seoul, South Korea}

    \address[20]{University of Minho, Braga, Portugal}

    \address[25]{Hanglok Tech, Zhuhai City, China}

    \address[30]{Jmees Inc., Kashiwa City, Japan}

    \address[35]{School of Sciences, São Paulo State University (UNESP), Bauru, Brazil}

    \address[40]{KIST HARILAB, Center for Humanoid Research, Artificial Intelligence and Robot Institute, Korea Institute of Science and Technology (KIST), Seoul, South Korea}

    \address[45]{King's College London, London, UK}

    \address[50]{The Chinese University of Hong Kong, Hong Kong, China}

    \address[55]{Division of Intelligent Medical Systems, German Cancer Research Center (DKFZ), Heidelberg, Germany}

    \address[60]{Muroran Institute of Technology, Hokkaido, Japan}
    \address[61]{Niigata University of Health and Welfare, Niigata, Japan}
    \address[62]{Konica Minolta, Inc., Osaka, Japan}

    \address[65]{Department of Computer Science and Engineering, The Hong Kong University of Science and Technology, Hong Kong, China}
    \address[66]{Department of Chemical and Biological Engineering, The Hong Kong University of Science and Technology, Hong Kong, China}
    \address[67]{HKUST Shenzhen-Hong Kong Collaborative Innovation Research Institute, Shenzhen, China}

    \address[70]{Center for Research and Formation in Artificial Intelligence (CinfonIA), Los Andes University, Bogota, Colombia}

    \address[75]{Department of Biomedical Engineering, Medical Image Analysis Group, Eindhoven University of Technology, Eindhoven, Netherlands}
    \address[76]{Institute of Information Technology (ITEC), Klagenfurt University, Klagenfurt, Austria}

    \address[80]{Center for Tactile Internet with Human-in-the-loop (CeTI), TU Dresden, Dresden, Germany.}
    \address[81]{Department of Translational Surgical Oncology, National Center for Tumor Diseases (NCT/UCC), Partner Site Dresden, Dresden, Germany}

    \address[90]{Chair for AI in Healthcare and Medicine, Technical University of Munich (TUM) and TUM University Hospital, Munich, Germany}
    \address[91]{Biomedical Image Analysis Group, Department of Computing, Imperial College London, London, UK}
    \address[93]{Department of Surgery, TUM University Hospital, School of Medicine and Health, Technical University of Munich, Munich, Germany \newline \newline Corresponding author: Tobias Rueckert. E-mail address: tobias.rueckert@re-mic.de}

\begin{abstract}
Reliable recognition and localization of surgical instruments in endoscopic video recordings are foundational for a wide range of applications in computer- and robot-assisted minimally invasive surgery (RAMIS), including surgical training, skill assessment, and autonomous assistance.  
However, robust performance under real-world conditions remains a significant challenge.  
Incorporating surgical context -- such as the current procedural phase -- has emerged as a promising strategy to improve robustness and interpretability.

To address these challenges, we organized the Surgical Procedure Phase, Keypoint, and Instrument Recognition (PhaKIR) sub-challenge as part of the Endoscopic Vision (EndoVis) challenge at MICCAI 2024.  
We introduced a novel, multi-center dataset comprising thirteen full-length laparoscopic cholecystectomy videos collected from three distinct medical institutions, with unified annotations for three interrelated tasks: surgical phase recognition, instrument keypoint estimation, and instrument instance segmentation.  
Unlike existing datasets, ours enables joint investigation of instrument localization and procedural context within the same data while supporting the integration of temporal information across entire procedures.  

We report results and findings in accordance with the BIAS guidelines for biomedical image analysis challenges.  
The PhaKIR sub-challenge advances the field by providing a unique benchmark for developing temporally aware, context-driven methods in RAMIS and offers a high-quality resource to support future research in surgical scene understanding.
\end{abstract}

 
\end{frontmatter}

\section{Introduction}
\label{introduction}

In recent years, significant progress has been made in the field of computer- and robot-assisted minimally invasive surgery (RAMIS) with the aim of supporting the surgical team during laparoscopic interventions with innovative assistance systems~\cite{Dagnino_Kundrat_2024}.
The foundation for the development of such applications often relies on the determination of the current phase of a procedure, the segmentation of the surgical instruments in endoscopic images, and the localization of certain keypoints and the associated estimation of the instruments' pose.
Knowing the current intervention phase offers further possibilities, such as improving safety in the operating room (OR) by providing early context-sensitive warnings~\cite{katic2013context}, by optimizing the OR management~\cite{tanzi2020intraoperative}, or by providing accurate procedure time predictions~\cite{aksamentov2017deep}, e.~g., for anesthesia planning.
Segmentation of surgical instruments provides a variety of application areas, such as surgical navigation systems~\cite{ahmed2024deep}, surgical skill assessment~\cite{lam2022machine}, or autonomous endoscope guidance~\cite{gruijthuijsen2022robotic}.
Based on the localization of certain keypoints that describe the pose of an instrument, the distance to risk structures can be determined~\cite{bernhardt2017status} and the assessment of surgical skills of the surgeon can further be automated~\cite{law2017surgeon}.
Purely image-based processing of endoscopic data is of particular importance for all three tasks, as no changes need to be made to the surgical environment and the procedure workflow, allowing a simple integration into existing clinical practice~\cite{rueckert2024methods}.

In the past, the development of methods was often based on the collection and annotation of application-specific data.
In order to avoid the disadvantages of this procedure, such as the lack of reproducibility or the inability to compare the results between the methods, the concept of \textit{challenges} was introduced.
The most well-known platform for conducting such challenges in the area of medical image processing is the annual conference of the Medical Image Computing and Computer Assisted Intervention (MICCAI) society, which enables the organization of a large number of challenges for various medical applications.
One of them is the Endoscopic Vision Challenge (EndoVis), which has been held annually since 2015. It coordinates a series of sub-challenges, all focused exclusively on endoscopic image data.
Within the domain of laparoscopic phase recognition, three dedicated sub-challenges have been conducted to date~\cite{speidel2017surgical, bodenstedt2021heichole, wagner2019comparative}. For the task of surgical instrument segmentation, eight sub-challenges have been performed~\cite{bodenstedt2015comparative, allan2017robotic, allan2018robotic, ross2019comparative,bodenstedt2021heichole,psychogyios2022sar,malpani2023synthetic,bodenstedt2023}. As for surgical instrument keypoint estimation, one dedicated sub-challenge has been held so far~\cite{bodenstedt2015comparative}.

\begin{figure*}[t]
    \centering

    \small
    \begin{tabularx}{\textwidth}{p{1.7cm} X X X }

        \toprule
        \multicolumn{1}{c}{Task} & \multicolumn{1}{c}{Hospital 1} & \multicolumn{1}{c}{Hospital 2} & \multicolumn{1}{c}{Hospital 3} \\ 

        \midrule
        
        \makecell{ Input \\ frames } & 
        
        \includegraphics[width=1.0\linewidth]{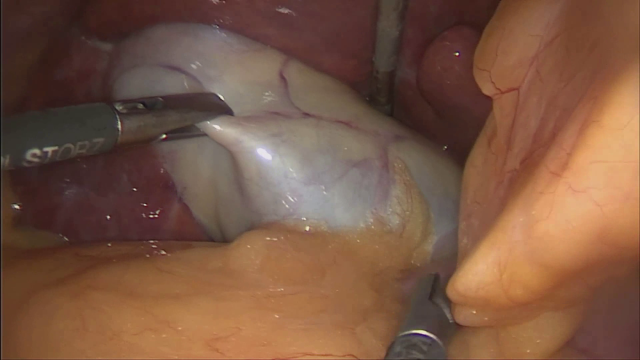} & 
        \includegraphics[width=1.0\linewidth]{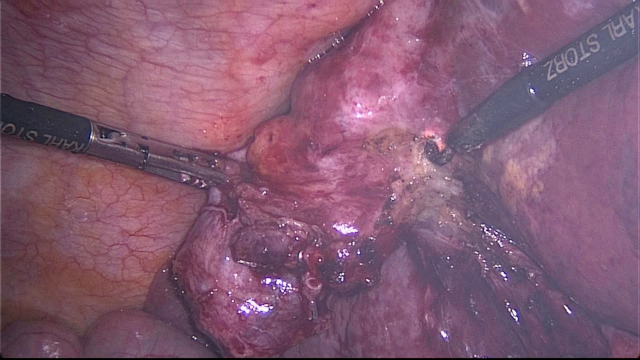} & 
       \includegraphics[width=1.0\linewidth]{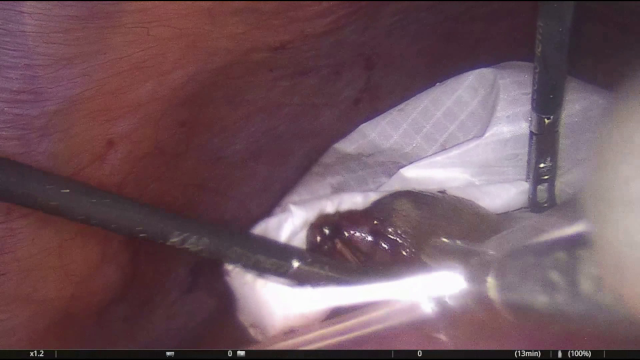} \\

        \rowcolor{gray!10}
        \makecell{ Surgical \\ phase \\ recognition } & 
        \hspace{-0.15cm}
        \input{00_dataset_hospital_1_phase}
        & 
        \hspace{-0.15cm}
        \input{00_dataset_hospital_2_phase}
        & 
        \hspace{-0.15cm}
        \input{00_dataset_hospital_3_phase} \\

        \rowcolor{gray!10}
        \makecell{ Instrument \\ instance \\ segmentation } &
        \includegraphics[width=1.0\linewidth]{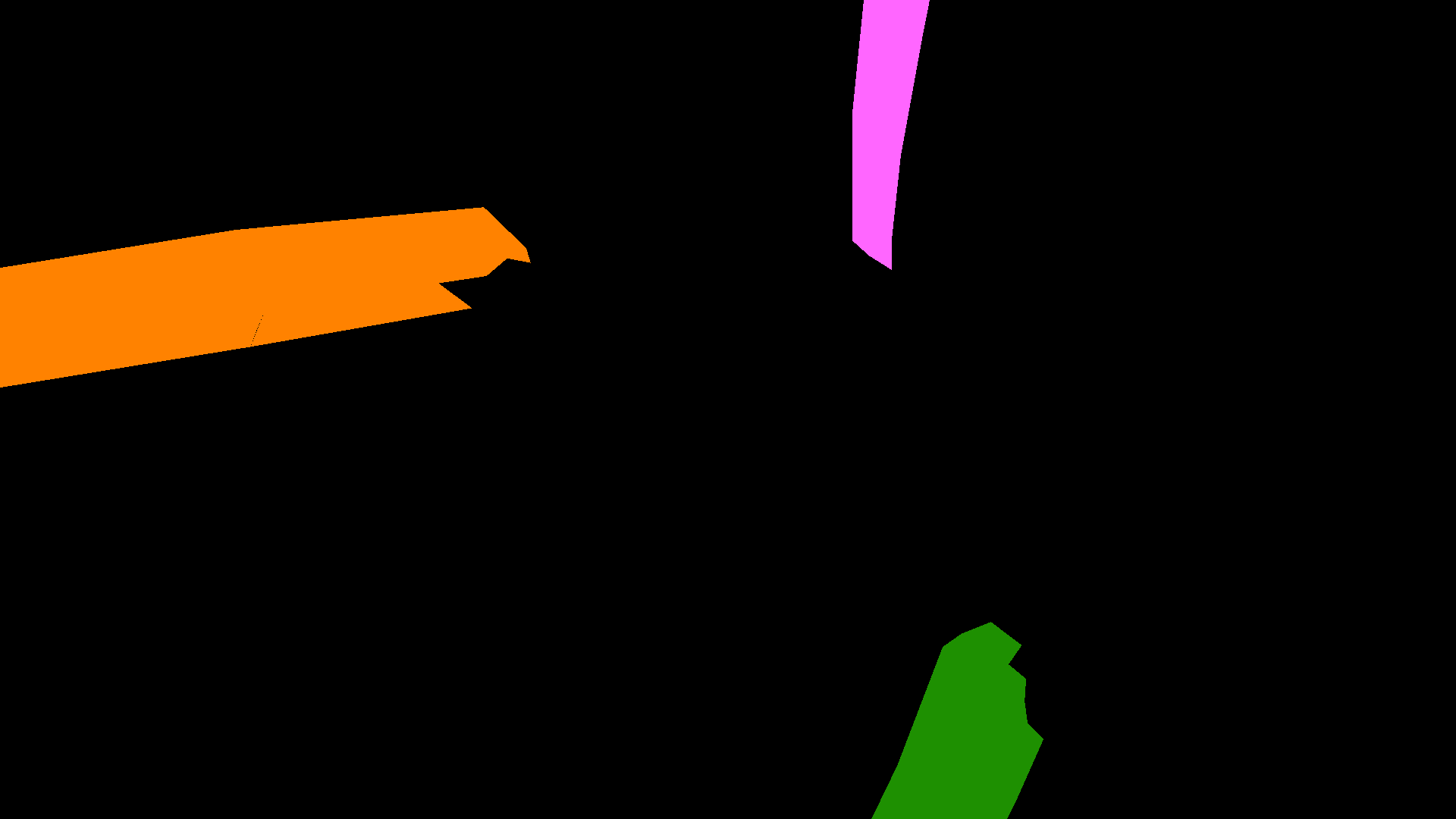} & 
        \includegraphics[width=1.0\linewidth]{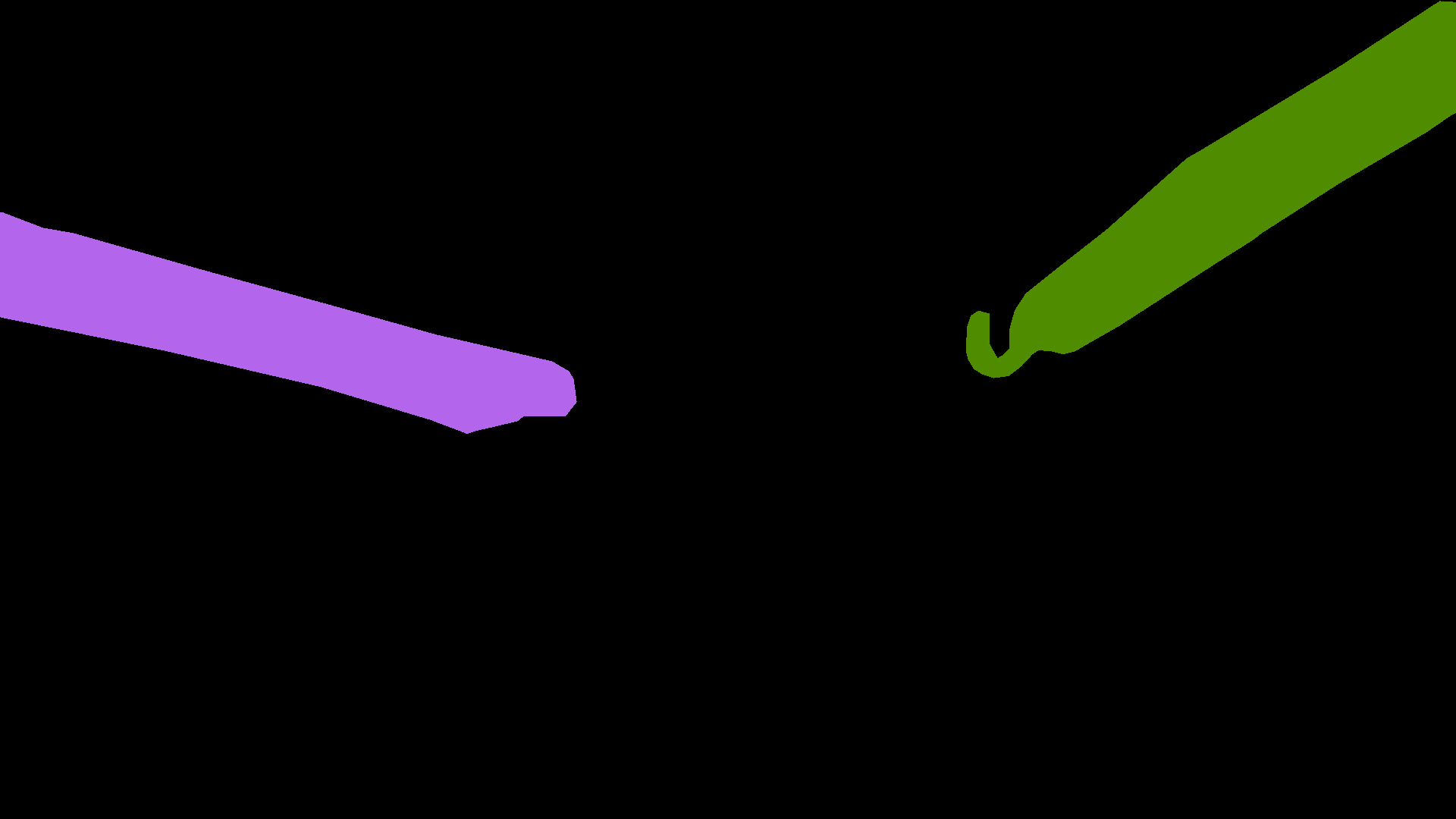} & 
        \includegraphics[width=1.0\linewidth]{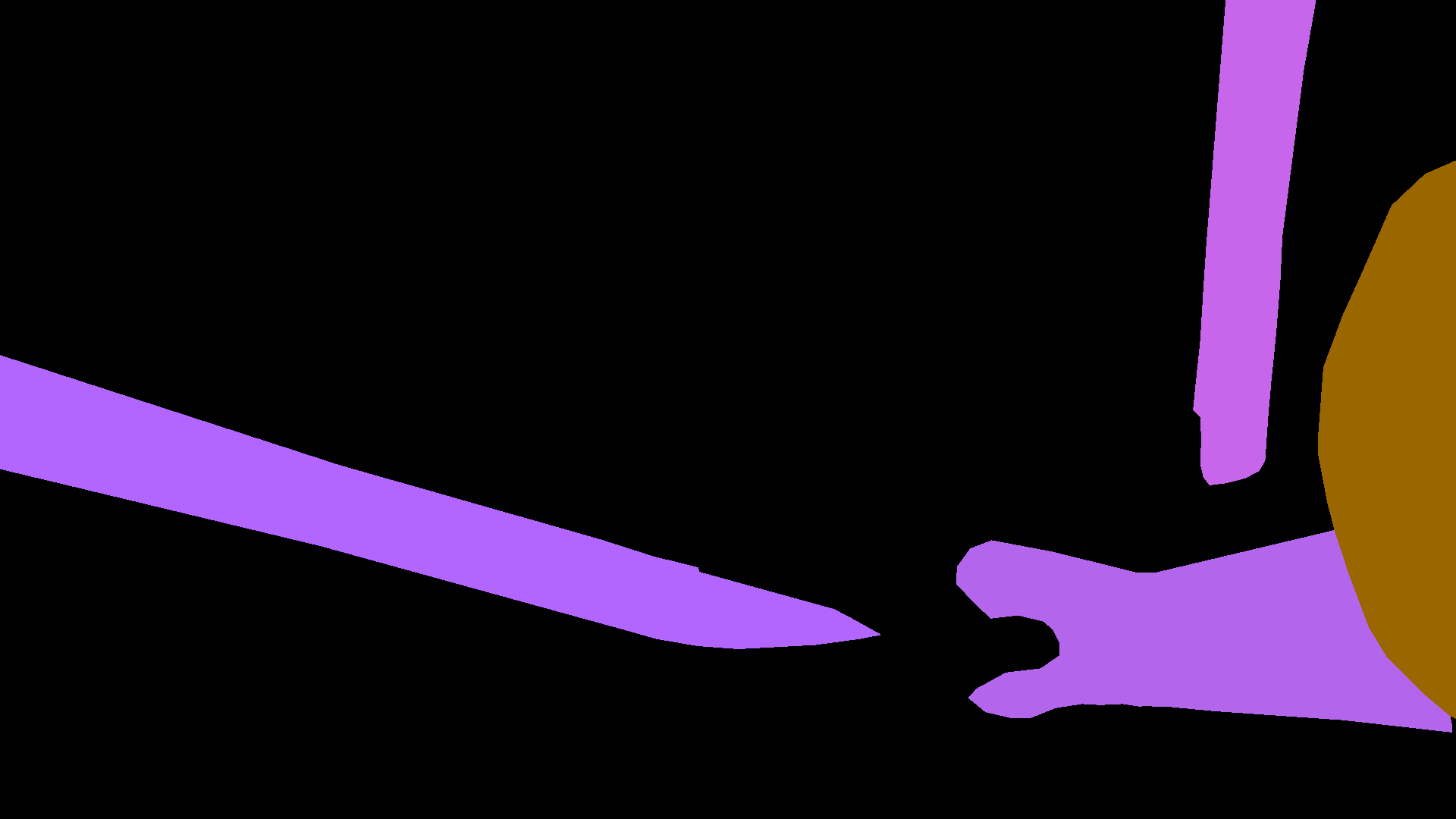} \\

        \rowcolor{gray!10}
        \makecell{ Instrument \\ keypoint \\ estimation } & 
        \includegraphics[width=1.0\linewidth]{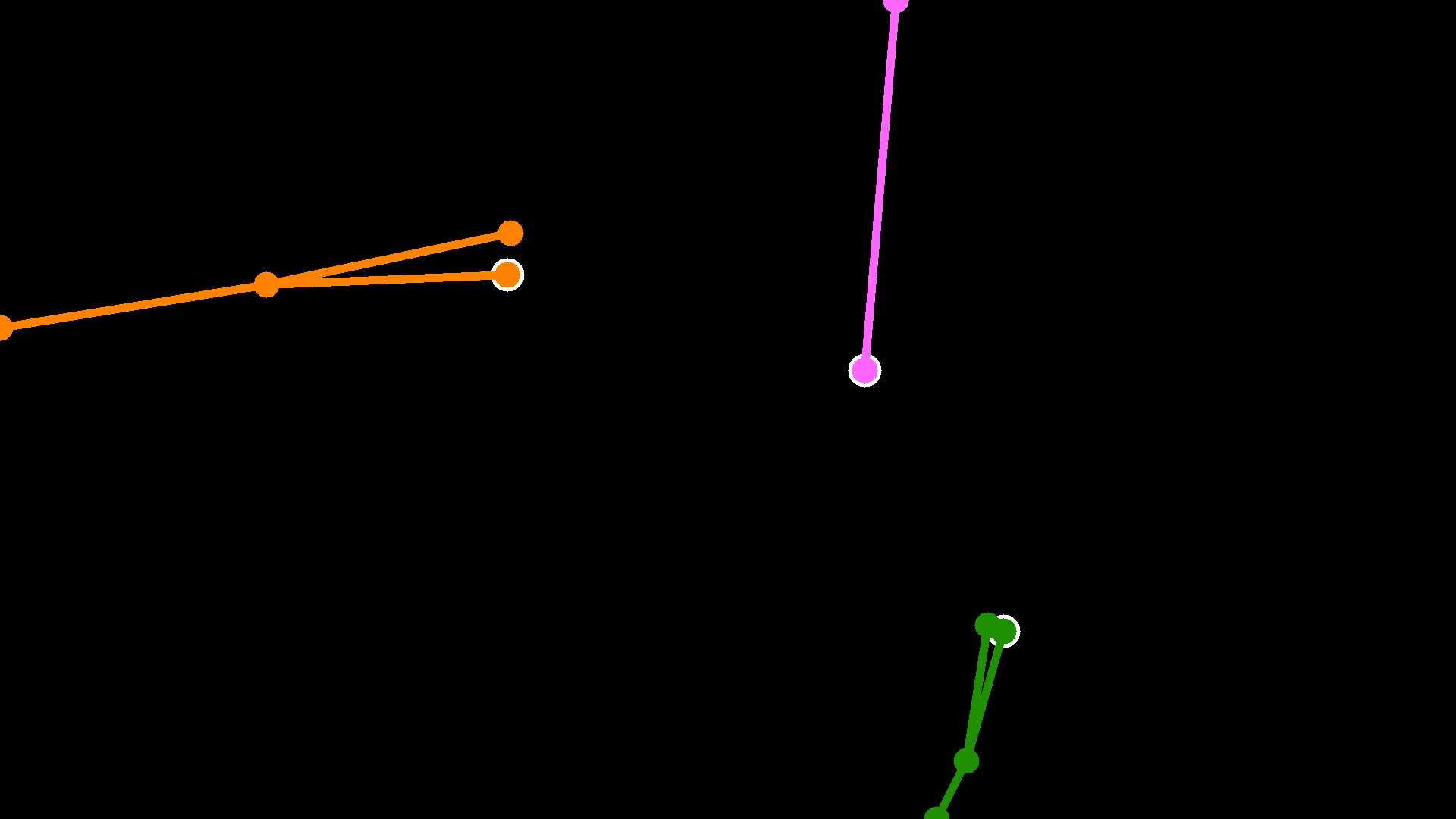} & 
        \includegraphics[width=1.0\linewidth]{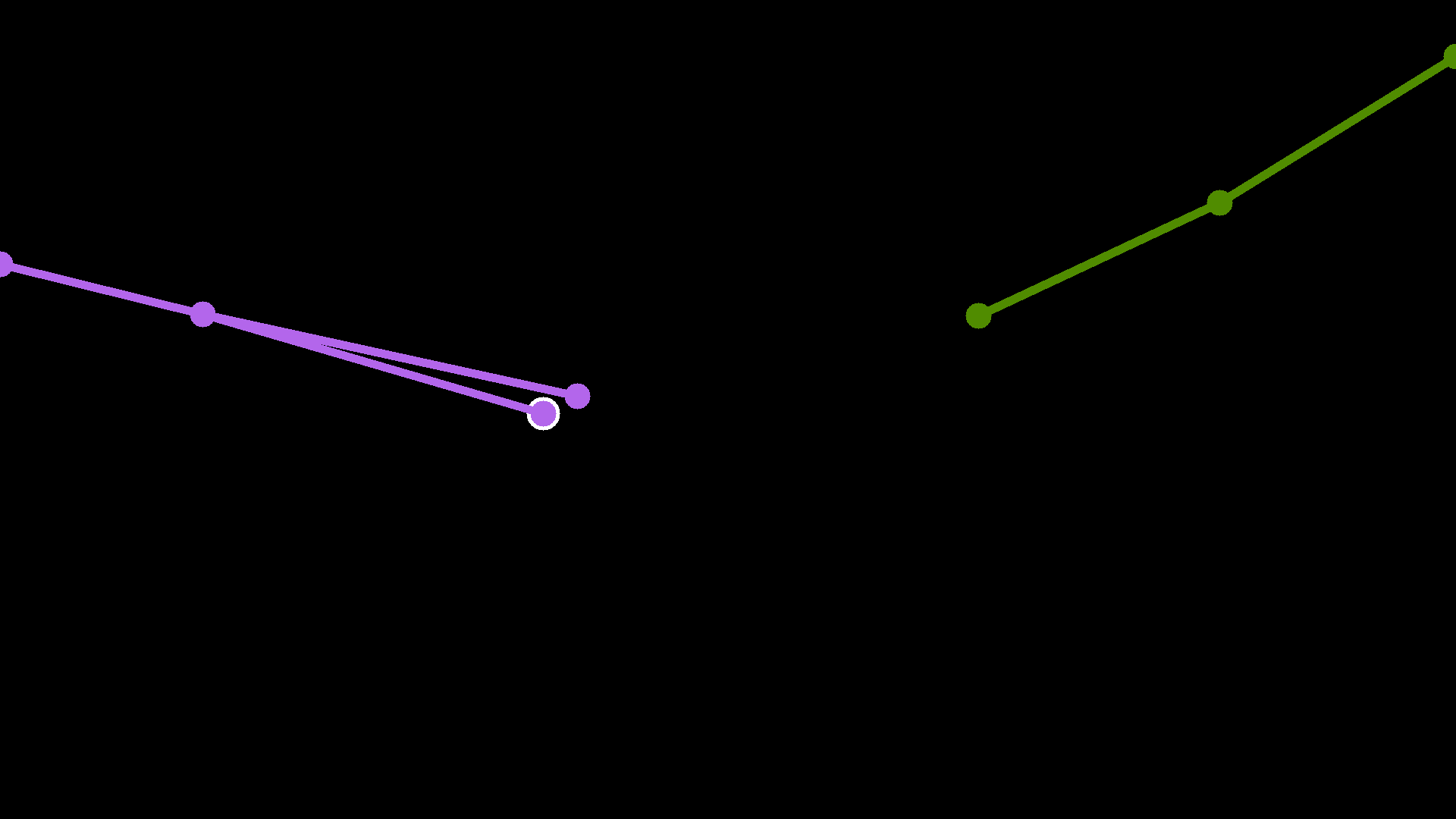} & 
        \includegraphics[width=1.0\linewidth]{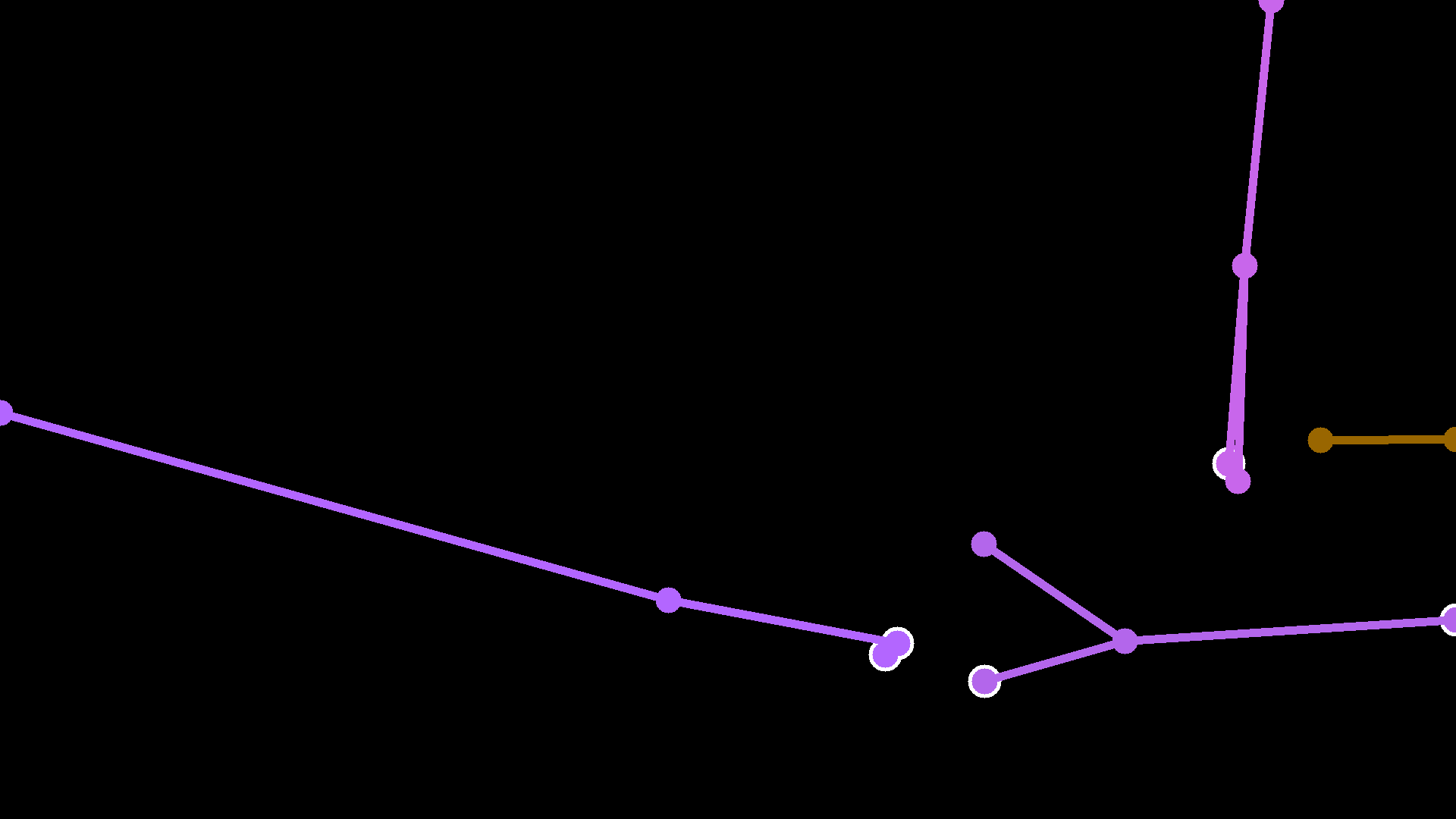} \\
        \bottomrule
    \end{tabularx}
    \caption{
    Visualization of the PhaKIR tasks and annotations for each of the three medical centers. For the phase recognition task, the phases preparation (P), calot triangle dissection (CTD), clipping and cutting (ClCu), gallbladder dissection (GD), gallbladder packaging (GP), cleaning and coagulation (ClCo), and gallbladder retraction (GR) are shown. For the instrument instance segmentation task, the color-encoded masks are presented. For the instrument keypoint estimation task, a visualization of the keypoint coordinates is depicted, including hidden keypoints surrounded by white.
    }
    \label{fig:annotations}
\end{figure*}

However, the datasets provided by these sub-challenges do not cover all the necessary aspects required for the development of real-world applications.
For the phase recognition task, only two out of three papers present a dataset that originates from multiple centers~\cite{bodenstedt2021heichole, wagner2019comparative}.
In the context of the instrument segmentation task, several challenges do not differentiate between distinct instrument classes~\cite{bodenstedt2015comparative,allan2018robotic,ross2019comparative,bodenstedt2021heichole,malpani2023synthetic} or between individual instances within the same class \cite{bodenstedt2015comparative,allan2017robotic,allan2018robotic,bodenstedt2021heichole,psychogyios2022sar,malpani2023synthetic}.
This lack of distinction hampers the ability to accurately identify overlapping instruments and complicates the task of tracking the tools over time, which relies on the consistent identification of each instrument across successive frames.
The recordings of some sub-challenges operate on porcine tissue~\hbox{\cite{allan2017robotic,allan2018robotic}}, which significantly simplifies the recognition of surgical instruments compared to human tissue~\cite{allan2018robotic}, or do not use any biological tissue at all~\cite{malpani2023synthetic}.
Some datasets only provide single frames~\cite{allan2017robotic,allan2018robotic,bodenstedt2021heichole,bodenstedt2023,malpani2023synthetic} or short clips \hbox{\cite{ross2019comparative}} instead of full video sequences, precluding the use of temporal information for the segmentation task.
Information on relevant instrument keypoints is available in only one sub-challenge~\hbox{\cite{bodenstedt2015comparative}}. However, for in-vivo recordings involving manual instruments, only the coordinates of a single keypoint and the instrument orientation are provided.
Meanwhile, sequences featuring robotic instruments are based on ex-vivo data from animal tissue, representing highly simplified movements that do not account for real-world surgical complexities.
Additionally, none of the sub-challenge datasets for the segmentation of surgical instruments or the keypoint estimation task include multi-center data, thereby limiting the ability to evaluate the participant methods across different clinical environments and surgical tools.
An overview regarding previous related EndoVis sub-challenges together with the conducted tasks, the number of participating medical centers, and key properties of the challenges and datasets is given in Table~\ref{introduction:challenge_comparison}.

To address these challenges, we provided a dataset comprising full endoscopic video sequences of thirteen human cholecystectomies from three medical centers, enabling the incorporation of temporal information in method development as well as the consideration of real-world conditions.
The annotations span three tasks: determining the surgical phase of an intervention, performing pixel-precise instance segmentation of surgical instruments, and localizing specific keypoints of these instruments. 
This setup supports both the recognition of surgical instruments and the understanding of surgical context in one dataset.
An overview of the tasks and the associated annotations is shown in Figure~\ref{fig:annotations}.

\begin{table*}[t]
    \small
    \centering
    \caption{Overview of related previous EndoVis sub-challenges including the conducted tasks phase recognition (PR), instrument segmentation (IS), and keypoint estimation (KE), the number of participating medical centers (\#Med. centers), and key properties of the challenges and datasets, organized according to the task sequence presented in Chapter~\ref{mission}.}
        \begin{tabularx}{1.0\textwidth}{p{3.9cm}P{0.25cm}P{0.25cm}P{0.25cm}P{0.65cm}p{10.5cm}} 
        \toprule
        \multirow{3}{*}{Challenge} & \multicolumn{3}{c}{Task} & \multirow{2}{*}{\#Med.} & \multirow{3}{*}{Challenge / dataset properties} \\
        \cmidrule{2-4}
         & PR & IS & KE & centers  &  \\ 
        \midrule 
        
        \rowcolor{gray!10}
        Speidel et al. (2017) \cite{speidel2017surgical} & \ding{51} &   &   & 1  & single-centered, single-task, methods no longer considered state of the art \\
        
        Wagner et al. (2023) \cite{wagner2019comparative} & \ding{51} &   &   & 3 & single-task, strong focus on one leading, more traditional architecture  \\
        
        \rowcolor{gray!10}
        Bodenstedt et al. (2021) \cite{bodenstedt2021heichole} & \ding{51} & \ding{51} &   & 3 & no instrument classification, no instance segmentation, single frames \\
        
        Allan et al. (2019) \cite{allan2017robotic} &   & \ding{51} &   & 1 &  single-centered, single-task, porcine tissue, no instance segmentation, single frames \\

        \rowcolor{gray!10}
        \multirow{2}{*}{Allan et al. (2020) \cite{allan2018robotic}} &   & \multirow{2}{*}{\ding{51}} &   & \multirow{2}{*}{1}  &  single-centered, single-task, porcine tissue, no instrument classification, no instance segmentation, single frames \\

        Roß et al. (2021) \cite{ross2019comparative} &   & \ding{51} &   & 1  &  single-centered, single-task, no instrument classification, short clips  \\

        \rowcolor{gray!10}
        Malpani and Glock (2023) \cite{malpani2023synthetic} &   & \ding{51} &   & -  &  single-centered, single-task, synthetic data, single frames \\

        Bodenstedt et al. (2023) \cite{bodenstedt2023} &   & \ding{51} &   & 1  & single-centered, single-task, single frames  \\

        \rowcolor{gray!10}
        Psychogyios et al. (2024) \cite{psychogyios2022sar} &   & \ding{51} &   & 1  & single-centered, single-task, no instance segmentation \\

        Bodenstedt et al. (2018) \cite{bodenstedt2015comparative} &   & \ding{51} & \ding{51} & 1  & single-centered, no instrument classification, no instance segmentation \\
        
        \midrule
        \rowcolor{gray!10}
        \multirow{2}{*}{PhaKIR (2025)} & \multirow{2}{*}{\ding{51}} & \multirow{2}{*}{\ding{51}} & \multirow{2}{*}{\ding{51}} & \multirow{2}{*}{3} & multi-centered, multi-task, instrument classification, instance segmentation, real-world surgeries, full sequences \\
        \bottomrule
        \end{tabularx}
    \label{introduction:challenge_comparison}
\end{table*}

Our paper is structured according to the transparent reporting of biomedical image analysis challenges (BIAS) guidelines published by~\cite{maier2020bias}.
As our challenge consists of three tasks, we refrain from using an overarching methods section and instead present the Chapters~\ref{challenge_organization}, \ref{mission}, \ref{data_and_annotations}, and~\ref{assessment} separately to ensure a clear structure.

The appendix contains information regarding the challenge organizers (see \ref{appendix:challenge_orga}), the data description and labeling instructions (see \ref{appendix:data_description}), the submission instructions for the participants (see \ref{appendix:submission_instructions}), as well as the challenge design document (see \ref{appendix:challenge_design}).

\section{Challenge organization}
\label{challenge_organization}

As part of the MICCAI 2024 conference, which took place in Marrakech, Morocco, from October 6 to 10, we conducted our sub-challenge entitled PhaKIR: Surgical Procedure \textbf{Pha}se, \textbf{K}eypoint, and \textbf{I}nstrument \textbf{R}ecognition as one of a total of eight EndoVis sub-challenges.
The number of registered participants and submissions for each individual sub-challenge is shown in Figure~\ref{organization:fig:endovis24}.

Three parties were involved in the challenge organization~(see~\ref{appendix:challenge_orga}).
The research group for Minimally Invasive Interdisciplinary Therapeutic Intervention (MITI)\footnote{\url{https://web.med.tum.de/en/miti/}} at the Technical University of Munich (TUM) obtained the source data from the three different medical centers and created the annotations (see~\ref{appendix:data_description}). 
The Regensburg Medical Image Computing (ReMIC)\footnote{\url{https://re-mic.de/}} laboratory at the Ost\-baye\-ri\-sche Technische Hochschule Regensburg (OTH Regensburg) carried out the technical implementation, which included the pre-processing and the provision of the data, the registration and administration of the participants, the submission process, and the evaluation and presentation of the results.
The company AKTORmed - Robotic Surgery\footnote{\url{https://aktormed.info/}} provided the prize money of €3,000 which was distributed equally across the three tasks.
The sub-challenge is designed as a one-time event with a fixed submission deadline, after which no submissions are possible.
As soon as the joint sub-challenge paper is published, the training dataset will be publicly available.
The sub-challenge platform can be accessed at~\url{https://phakir.re-mic.de/} and was hosted entirely on infrastructure set up and managed by the ReMIC laboratory~\cite{klausmann2025diy}, i.e., no established platforms such as Synapse or Grand Challenge were used.

With respect to the participation policies, only fully automated methods were permitted. 
Participants were restricted to using only the training data provided by the challenge organizers and publicly available datasets. 
In addition, the use of networks pre-trained on publicly available datasets was allowed.
While members of the organizing institutions were permitted to participate, they were not eligible to receive any awards.
The results of all teams were presented at the MICCAI 2024 EndoVis challenge event on October 10, 2024 and were subsequently published on the sub-challenge website.
All results and further analysis from participating teams are included in this joint challenge publication, with all contributing members of each team listed as co-authors.
Authors of submitted methods were not permitted to publish any results prior to the release of this joint challenge paper.

Information on the submission process (see~\ref{appendix:submission_instructions}) was provided to the participants both directly on the website\footnote{\url{https://phakir.re-mic.de/participation/}} and in a separate GitHub repository\footnote{\url{https://github.com/schnadoslin/PhaKIR_Submission_Template/}}.
While the instructions on the website are more concerned with organizational aspects, the repository explanations provide more technical details, such as the structure of the input and desired output data of the algorithms, information on the Docker submission, on the submission process in general, and on the metrics used. 
A Dockerfile, together with sample data and an exemplary network model, was also made available to the participants via this GitHub repository.
The submissions were provided as Docker containers, which were manually reviewed by the organizers to verify both technical functionality and compliance with the required output format.
Submissions with missing results for test cases were not permitted.
Participants were notified of the status of their submission by email and were allowed to update their models as often as needed until the final deadline.

\begin{figure*}
\centering
\begin{adjustbox}{width=\textwidth}
\begin{tikzpicture}[thick, node distance=1cm]

\fill[fill=timeline_bg] (-1.0, -1.5) rectangle (22.5, 3); 
\draw[arrow] (0, 0) -- (22, 0); 

\draw[fill=reg_phase, rounded corners=15pt] (0, 0) rectangle (3, 1.5);
\draw[fill=train_phase, rounded corners=15pt] (3, 0) rectangle (12, 1.5);
\draw[fill=sub_phase, rounded corners=15pt] (12, 0) rectangle (18, 1.5);
\draw[fill=eval_phase, rounded corners=15pt] (18, 0) rectangle (21, 1.5);

\node[phase] at (1.5, 0.75) {Registration \\ phase};
\node[phase] at (7.5, 0.75) {Training \\ phase};
\node[phase] at (14.63, 0.75) {Submission \> phase};
\node[phase] at (19.5, 0.75) {Evaluation \\ phase};

\foreach \x/\label in {
    0/{Registration \\ opens},
    3/{1st part of training \\ data released},
    6/{2nd part of training \\ data released},
    9/{Submission \\ instructions released},
    12/{Submission \\ opens},
    15/{Methodology report \\ deadline},
    18/{Submission \\ deadline},
    21/{Challenge \\ day}} {
    \draw[arrow] (\x, 0.1) -- ++(0, 1.5) node[event, above] {\label};
}

\foreach \x/\month/\day in {
    0/Apr/29, 3/May/27, 6/Jul/11, 9/Jul/25,
    12/Sep/8, 15/Sep/15, 18/Sep/24, 21/Oct/10} {
    \draw[thick] (\x, -0.1) -- (\x, 0.1);
    \node[below] at (\x, -0.3) {\footnotesize \month, \day};
}

\end{tikzpicture}
\end{adjustbox}
\caption{Timeline of our PhaKIR sub-challenge highlighting key milestones and dates. All dates refer to the year 2024.}
\label{fig:organization:timeline}
\end{figure*}
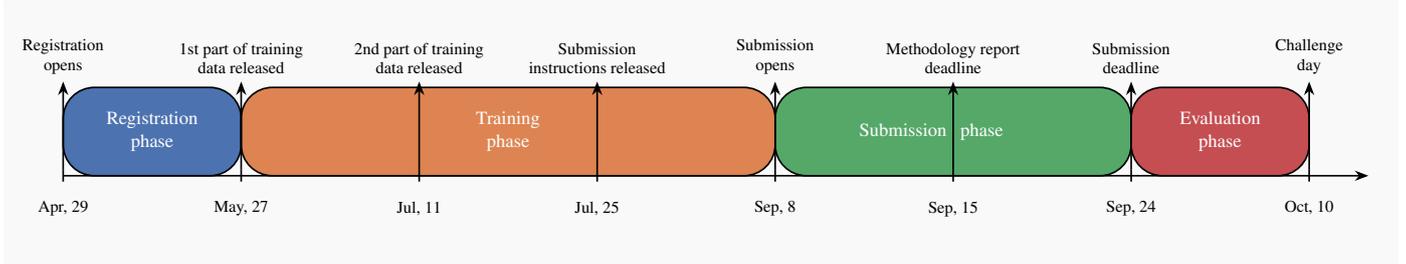

The sub-challenge was structured into four distinct phases: registration, training, submission, and evaluation.
The most important dates corresponding to these phases are shown in Figure~\ref{fig:organization:timeline}. 
Key milestones included the opening of the registration on April 29, the start of the training phase on May 27, the beginning of the submission phase on September 8, the submission deadline on September 24, and the publication of the results at the EndoVis event at MICCAI 2024 and online on the website on October 10.

\begin{figure*}[t]
    \centering
    \includegraphics[width=0.9\linewidth]{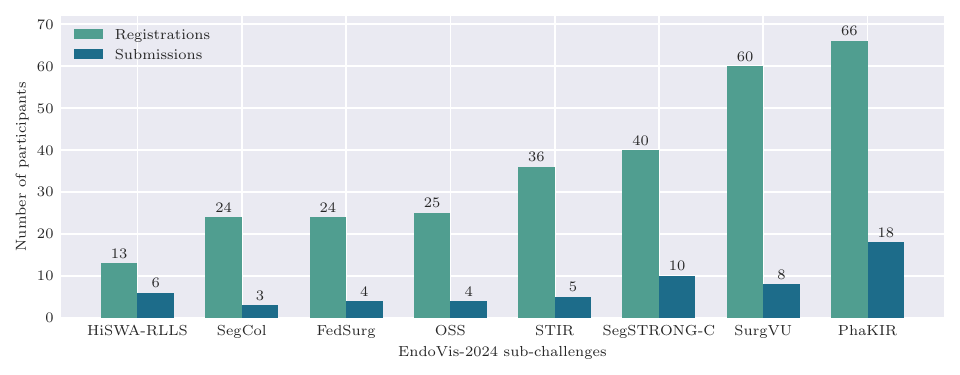}
    \caption{Number of participants that registered and submitted for each of the eight individual EndoVis-2024 sub-challenges, sorted in ascending order based on the number of registrations.}
    \label{organization:fig:endovis24}
\end{figure*}

Regarding the dataset, only anonymized data collected in an ethically approved research project approved by the local ethics committee of the TUM (approval code 337/21 S-EB) was used.
During the registration phase, each participant was required to sign the EndoVis rules document, which specifies that the provided data may only be used for the purposes of the challenge, must not be shared with third parties and cannot be used for commercial purposes.
Use of the data for individual publications by participants is permitted only after the release of this joint challenge publication, with notification provided via email and through the official website.
The dataset is published under a Creative Commons Attribution-NonCommercial-ShareAlike (CC BY-NC-SA)\footnote{\url{https://creativecommons.org/licenses/by-nc-sa/4.0/}} license, which means that it can be used for non-commercial purposes once the sub-challenge has been conducted and this paper has been released.
Regarding code availability, participants were provided with Docker container templates, including functional example code and example outputs.
While sharing the code for their submissions was not mandatory, participants were encouraged to do so.
Apart from the company AKTORmed, no additional sponsors or supporters were involved in the challenge.
Access to the test data was strictly limited: only a predefined group within the MITI research group responsible for acquiring the raw data and annotations, and a designated team at the ReMIC laboratory, tasked with data provisioning and evaluation, were granted access.

\section{Mission of the challenge}
\label{mission}

The objective of our PhaKIR sub-challenge was to make a contribution in the field of RAMIS by providing a high-quality, comprehensive real-world dataset of minimally invasively performed cholecystectomies, along with the benchmarking of various algorithms on this dataset.
We defined the target cohort as all individuals around the world with gallbladder-related problems that require cholecystectomy.
For this purpose, the challenge cohort consisted of adult humans of different ages and genders who had undergone cholecystectomy at three medical centres in Germany.
The data collected consists of endoscopic video recordings of the abdominal cavity during interventions, without any additional clinical patient data.
The challenge comprised the following three tasks, in which participants were allowed to take part either individually or as part of a team and could choose to participate in one or multiple tasks.
Each task was evaluated individually.

\begin{itemize}
    \item Surgical phase recognition: Classification of the surgical phases of a cholecystectomy as accurately as possible.
    \item Instrument instance segmentation: Segmentation of the contours of the surgical instruments as precisely as possible through pixel-accurate predictions and distinguishing different instrument classes and instances of the same instrument class.
    \item Instrument keypoint estimation: Location of certain keypoints on the surgical instruments as precisely as possible through pixel-accurate coordinates, and distinguishing different instrument classes and instances of the same instrument class, considering that the number of keypoints depends on the type of the individual instrument.
\end{itemize}

A central goal across all challenge tasks was the development of algorithms robust to real-world surgical conditions.
These include sudden smoke development, image blur, motion blur due to rapid instrument movement, occlusions from tissue and organ overlap, and reduced visibility caused by significant bleeding.
Another key objective was to leverage the temporal information inherent in the provided video sequences.
By moving beyond single-frame, participants were encouraged to improve both the accuracy and the robustness of their approaches in the face of these challenging conditions.

\section{Challenge dataset}
\label{data_and_annotations}

\subsection{Data collection}

A total of $n=13$ video sequences ranging between 23 and 60 minutes in duration and showing laparoscopic cholecystectomies were recorded with different endoscope systems from three surgical centers in Germany, namely the TUM University Hospital Rechts der Isar\footnote{\url{https://www.mri.tum.de/}} ($n = 9$), the Heidelberg University Hospital\footnote{\url{https://www.klinikum.uni-heidelberg.de/}} ($n = 2$), and the Weilheim-Schongau Hospital\footnote{\url{https://www.meinkrankenhaus2030.de/}} ($n = 2$).
For the sake of readability, we will refer to these three hospitals as hospital 1, hospital 2, and hospital 3 in the remainder of this article.
Heidelberg recordings have already been published as part of the Surgical Workflow and Skill Analysis Challenge (HeiChole Benchmark, \cite{wagner2019comparative}) and comply with the video files entitled \texttt{HeiChole2.mp4} and \texttt{HeiChole3.mp4}.
All sequences were recorded with monocular laparoscopic cameras, a resolution of $1920 \times 1080$ pixels, and a frame rate of 25 frames per second (FPS).
A video is recorded from the moment the camera is first inserted into the patient's abdomen and ends when the laparoscope is removed from the abdominal cavity for the last time.

\subsection{Training and test case definition}

The training and test cases each corresponded to a single video frame of a cholecystectomy recording and were structured as individual, independent samples.
Neither video frames nor labels were being provided for the test cases.
The participants were provided with the full-length sequences of the interventions.
Care was taken to ensure that full sequences were used either for training or for testing and that no subdivision was made within a video.
From hospital 1, $n=6$ videos were randomly selected for training and the remaining $n=3$ videos were used for testing.
The sequences from the other two surgical centers were divided equally into training and test data.
This approach was intended to ensure that each center was represented in the test data, thus requiring a high degree of generalizability of the participants' methods.
The recordings were named consistently, with \texttt{Video\_01}, \texttt{Video\_02}, ..., \texttt{Video\_09} originated from hospital 1, the sequences \texttt{Video\_11} and \texttt{Video\_12} were taken from hospital 2, and the recordings \texttt{Video\_13} and \texttt{Video\_14} were recorded at hospital 3.
It is important to note that the file entitled \texttt{Video\_10} was initially considered, but was excluded before the start of the challenge due to the significantly different instrumentation used, which varied greatly from that of the other videos.
The test dataset therefore consisted of the sequences \texttt{Video\_06}, \texttt{Video\_08}, \texttt{Video\_09}, \texttt{Video\_12}, \texttt{Video\_14}.
To ensure the anonymity of the patient and the surgical staff, all frames recorded outside the abdominal cavity (e.g., to clean the laparoscope) were removed from the videos.
For each video sequence, the participants were provided with a file containing the frame number after which such outside abdomen frames were extracted.
For the phase recognition task, every individual frame was annotated, whereas for the instrument instance segmentation and the instrument keypoint estimation tasks, annotations were provided for one frame per second.
The participants were therefore provided with 485,875 frames for training, with the same number of labels for the phase recognition task, as well as with instance segmentation and keypoint annotations for 19,435 frames.
The hidden test data contained 323,025 images, with an identical number of phase recognition labels and segmentation and keypoint labels for 12,921 frames.

\subsection{Data annotation}

The data was annotated by a team of four medical professionals, with the head of the group being an expert surgeon with many years of experience and the remaining members being students with a medical background.
The generation of the annotations followed specific rules which were defined and documented at the beginning of the process (see~\ref{appendix:data_description}).
A visualization of the annotations for each task, as well as for the three medical centers that provided the data, is shown in Figure~\ref{fig:annotations}.
To ensure high annotation quality and to avoid potential sources of error such as inaccurate timestamps or incorrectly classified instruments, a three-stage process was employed for each video.
This process involved the initial creation of the annotations, followed by two independent review stages conducted by different annotators.
The following sections describe the ground truth generation process in detail for each task.

\subsubsection{Task 1: Surgical phase recognition}

The ground truth data for the surgical phase recognition task was generated by setting timestamps during the visual inspection of the video recordings.
Each frame between a defined start and end timestamp was labeled with the corresponding surgical phase, based on the actions performed in that segment.
A total of seven distinct surgical phases was used, following the classification scheme introduced in the Cholec80 dataset by~\hbox{\cite{twinanda2016endonet}}.
Additionally, an eighth category -- an \texttt{undefined} phase -- was introduced to label transitional frames between two phases.
Although the video recordings from \cite{wagner2019comparative}, captured at the Heidelberg University, already included phase annotations, we re-annotated these sequences to ensure consistency with our annotation guidelines and alignment with the data from the other medical centers.
Examples of the surgical phase annotations are presented in the second row of Figure~\ref{fig:annotations}.

\subsubsection{Task 2: Instrument instance segmentation}

For the creation of ground truth data for the instrument instance segmentation task, the Computer Vision Annotation Tool (CVAT)\footnote{\url{https://www.cvat.ai/}} was used.
The surgical instruments were defined by manually drawn contours, where each contour defined the area of an instrument, from which segmentation masks were subsequently generated.
Following the approach by \hbox{\cite{ross2019comparative}} in the ROBUST-MIS 2019 challenge, this method ensured that each pixel in an image was assigned to exactly one instrument or to the background, enabling pixel-precise multi-instance segmentation. 
Annotators had access to entire video sequences -- including past and future frames -- for both the training and the test data, allowing for more accurate segmentation.
However, only visible portions of the instruments were annotated.
No assumptions were made regarding occluded or inferred parts.
A total of 19 different instrument classes was annotated.
Within each frame, individual instruments were distinguished by color-encoded masks, where the red (R) and green (G) channels encoded the instrument class, and the blue (B) channel indicated the instance of the object within that class. 
For example, two instruments of the same class in a single frame shared identical R and G values, while differing in B values.
To ensure consistency in manual segmentation, a set of annotation rules and visual examples was provided and strictly followed by all annotators.
A visualization of the color-coded instance segmentation masks for the three contributing medical centers is presented in the third row of Figure~\ref{fig:annotations}.

\subsubsection{Task 3: Instrument keypoint estimation}

For the instrument keypoint estimation task, annotations were also created using the CVAT tool, which allowed annotators to specify the coordinates of the relevant keypoints.
Depending on the type of the surgical instrument, between two and four keypoints were annotated per tool to cover the relevant locations.
The annotated keypoints included:
\begin{itemize}
    \item the endpoint, marking where the instrument intersects the image boundary,
    \item the shaft-tip transition, indicating the junction between the shaft and the tip, and
    \item the instrument tip, which may consist of two separate keypoints for opening instruments.
\end{itemize}
For each keypoint, its visibility status was documented -- categorized as visible, hidden or outside the image (and therefore unannotated).
As in the instrument instance segmentation task, annotators had access to entire video sequences for both training and test data, allowing them to consult preceding and following frames to achieve accurate localization. 
In contrast to segmentation, the annotation of keypoints did involve inference based on temporal continuity -- i.e., estimated instrument motion was considered to refine keypoint placement, particularly when visibility was limited.
A visualization of the annotated keypoints and joints between the instrument parts is depicted in the last row of Figure~\ref{fig:annotations}, following the same color-scheme used for the instrument instance segmentation masks.
Both visible and invisible keypoints displayed, with hidden keypoints indicated by a white circle around their respective coordinates.

After publication of this challenge paper and the corresponding dataset paper~\cite{rueckert2025video}, the training dataset will be publicly available~\cite{rueckert2025data}.

\section{Assessment method}
\label{assessment}

The following outlines the procedure for evaluation of the performance of the participants' methods. First, the aggregation of metric values is explained in Section~\ref{assessment:metric_aggregation}, then the ranking strategy is described in Section~\ref{assessment:ranking_strategy}, and finally, the selected metrics are presented in detail in Section~\ref{assessment:metric_description}.

\begin{figure*}[t]
    \centering
    \includegraphics[width=1.0\linewidth]{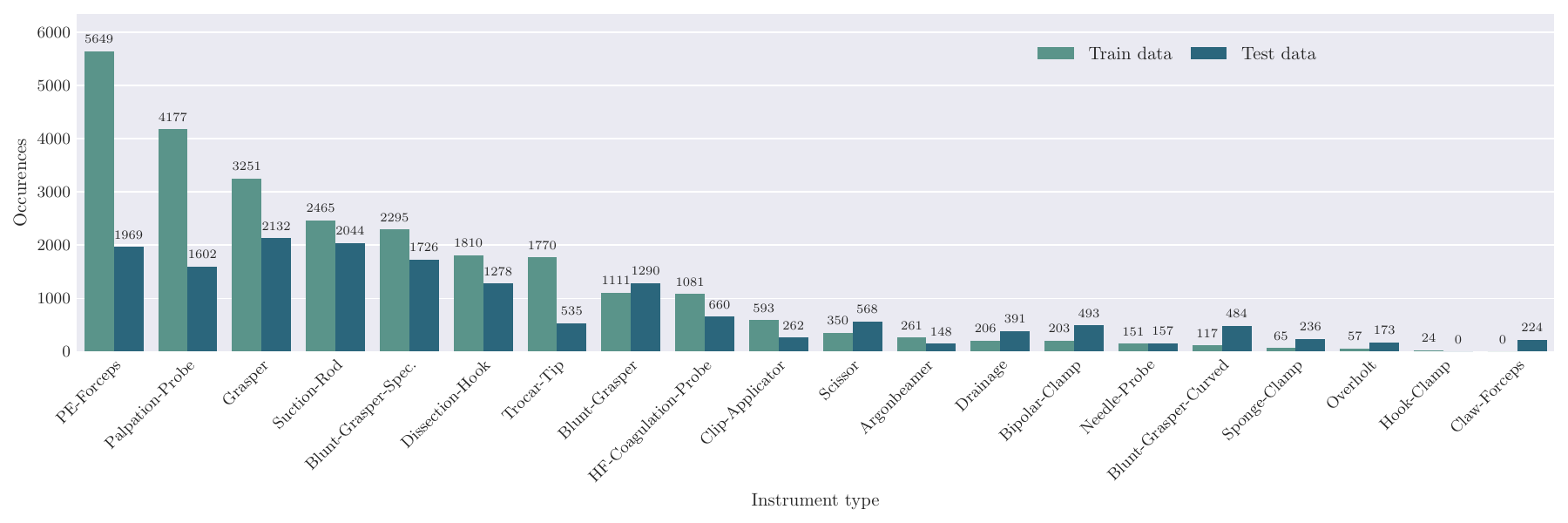}
    \caption{Number of occurrences for each instrument type, specified for training and test data and sorted in descending order according to the number of occurrences in the training data.}
    \label{fig:instrument_occurences}
\end{figure*}

\subsection{Metric aggregation}
\label{assessment:metric_aggregation}

For the selection of evaluation metrics, we considered the work of~\cite{reinke2024understanding} and followed the recommendations of the Metrics Reloaded Framework proposed \hbox{by~\cite{maier2024metrics}}.
Metrics were calculated per surgical phase and instrument class for each annotated frame in a video. 
As illustrated in Figure~\ref{fig:instrument_occurences}, the distribution of instrument classes is highly imbalanced, with some instruments appearing significantly more frequently than others.
To ensure a fair evaluation that does not favor common classes or phases, we first averaged the metric values across all frames of a video for each phase and instrument class individually.
These class-specific values were then averaged with equal weighting to compute a single overall score per video.
To aggregate across the test set, video-level scores were again averaged with equal weighting, ensuring that all videos contributed equally to the final evaluation -- regardless of their lengths.

To improve robustness and stability, we applied bootstrapping~\cite{efron1992bootstrap} with 10,000 iterations.
For each run, a number of samples equal to the number of annotated frames in the test set was drawn randomly with replacement.
The resulting metric values were averaged across all iterations to produce a final score per metric and participant, forming the ranking basis.

\subsection{Ranking strategy}
\label{assessment:ranking_strategy}

Since the number and type of evaluation metrics varied across tasks ($n = 2$ for surgical phase recognition, $n = 3$ for instrument instance segmentation, $n = 1$ for instrument keypoint estimation), direct aggregation was not feasible due to differences in scale and units. 
To address this, we calculated a ranking score for each individual metric.
These scores were then averaged with equal weighting to obtain an overall ranking value per participating group. 
The final challenge ranking was derived by comparing these aggregated ranking values across participating groups.

\subsection{Metric description}
\label{assessment:metric_description}

In the following, the employed metrics are described in detail for the tasks surgical phase recognition (Sec.~\ref{assessment:phase}), instrument instance segmentation (Sec.~\ref{assessment:segmentation}), and instrument keypoint estimation (Sec.~\ref{assessment:keypoint}). 

\subsubsection{Task 1: Surgical phase recognition}
\label{assessment:phase}

For the task of surgical phase recognition, the F1-score~\cite{van1979information,chinchor1992muc4} was used as the per-class performance metric, while Balanced Accuracy (BA,~\cite{tharwat2021classification}) served as the metric for evaluating the overall multi-class performance. 
The F1-score, defined as:
\begin{equation}
\text{F1-score} = \frac{2 \cdot \text{precision} \cdot \text{recall}}{\text{precision} + \text{recall}},
\label{eq:phase:f1_score}
\end{equation}
represents the harmonic mean of precision and recall, making it particularly suitable for the surgical phase recognition task as it accounts for both false positives and false negatives.
Similarly, the BA, defined as:
\begin{equation}
\text{BA} = \frac{\text{sensitivity} + \text{specificity}}{2},
\label{eq:phase:ba}
\end{equation}
corresponds to the unweighted average of sensitivity and specificity, ensuring that both shorter and longer phases contribute equally to the metric's value.
It is important to note that for frames labeled as \texttt{undefined} in the ground truth -- typically corresponding to phase transitions -- participants were free to handle them as they deemed appropriate.
However, these frames were excluded from the evaluation and were not considered in the calculation of the performance metrics for the test data.

\subsubsection{Task 2: Instrument instance segmentation}
\label{assessment:segmentation}

For the instrument instance segmentation task, three metrics were employed.
For localization, the Dice Similarity Coefficient (DSC, \cite{dice1945measures}) was used as the multi-instance multi-class overlap metric, the mean Average Precision \hbox{($\text{mAP}_{\text{IoU}}$, \cite{lin2014microsoft,everingham2015pascal})} was applied as the multi-threshold metric, and the 95\% Hausdorff-Distance (95\% HD, \cite{huttenlocher1993comparing}) served as the boundary-based metric. 
The DSC measures the overlap between two binary segmentation masks \(A\) and \(B\), and is computed as twice the area of overlap between \(A\) and \(B\) divided by the sum of the areas of the two masks.
Formally, for two sets \(A\) and \(B\), the DSC is given by:
\begin{equation}
\text{DSC}(A, B) = \frac{2 \cdot |A \cap B|}{|A| + |B|},
\label{eq:segmentation:dsc}
\end{equation}
where \( |A| \) and \( |B| \) denote the cardinalities of \(A\) and \(B\), respectively.
The DSC ranges from 0 (no overlap) to 1 (perfect overlap), and is commonly used as a measure of segmentation accuracy in medical imaging research and challenge design~\cite{taha2015metrics,muller2022towards,ross2019comparative}.

The mean Average Precision (mAP) is computed following the COCO evaluation protocol~\cite{lin2014microsoft} and the procedure described by~\cite{ross2019comparative}. 
It is defined as the mean of the Average Precision (AP) values across all \( C \) object classes and all Intersection-over-Union (IoU) thresholds \( T = \{0.50, 0.55, \ldots, 0.95\} \):
\begin{equation}
\text{mAP}_{\text{IoU}} = \frac{1}{|C|} \sum_{c=1}^C \left( \frac{1}{10} \sum_{t \in [0.5:0.05:0.95]} \left( \int_0^1 p_{c, t}(r) \, \mathrm{d}r \right) \right).
\label{eq:segmentation:map:iou}
\end{equation}
For each class \( c \in C \) and IoU threshold \( t \in T \), the Average Precision \( \text{AP}_{c,t} \) is defined by the area under the precision-recall curve:
\begin{equation}
\text{AP}_{c,t} = \int_{0}^{1} p_{c,t}(r) \, dr.
\end{equation}
Here, \( p_{c,t}(r) \) denotes the precision at a given recall level \( r \). 
The final mAP score is obtained by averaging \( \text{AP}_{c,t} \) across all classes and IoU thresholds.
Since participants' algorithms produced only binary segmentation masks without class probabilities, instance matching was based on the IoU between predicted instances \( M_1 \) and ground truth instances \( M_2 \), as defined by:
\begin{equation}
\text{IoU}(M_1, M_2) = \frac{|M_1 \cap M_2|}{|M_1 \cup M_2|},
\end{equation}
which was originally proposed by~\cite{jaccard1912distribution}.
The 95\% Hausdorff Distance (HD) quantifies the boundary agreement between two point sets \(A\) and \(B\), typically representing the contours of predicted and ground truth segmentations.
It is defined as:
\begin{equation}
\begin{split}
\text{95\% HD}(A, B) = \max \biggl\{ 
&\underset{a \in A}{\text{quant}_{95}} \left( \min_{b \in B} d(a, b) \right), \\
&\underset{b \in B}{\text{quant}_{95}} \left( \min_{a \in A} d(b, a) \right)
\biggr\}.
\end{split}
\label{eq:segmentation:hd}
\end{equation}
This metric computes the maximum of the $95^{\text{th}}$ percentile of the shortest distances from each point in one set to its closest point in the other. 
The point-to-point distance \(d(a, b)\) is measured using the Euclidean norm: \(d(a, b) = \|a - b\|_2\).
By using the $95^{\text{th}}$ percentile instead of the absolute maximum, this variant of the Hausdorff Distance is less sensitive to outliers and offers a more robust measure of segmentation accuracy.
To assign predicted segmentations to ground truth instances, we applied the Hungarian Maximum Matching Algorithm~\cite{kuhn1955hungarian} based on the IoU between all possible prediction-ground truth pairs.
Note that one instrument class present in the test data did not appear in the training set.
This class was therefore excluded from the evaluation to ensure fairness and metric consistency.

\subsubsection{Task 3: Instrument keypoint estimation}
\label{assessment:keypoint}

The instrument keypoint estimation task was evaluated using the \(\text{mAP}_{\text{OKS}}\) metric based on the Object Keypoint Similarity (OKS) following the COCO evaluation protocol~\hbox{\cite{cocoKeypointEval}.}
This approach is conceptually similar to the one used for instance segmentation, with the key difference being the substitution of the IoU with the OKS as the matching criterion between ground truth and predicted instances.
The matching strategy follows the procedure described in Section~\ref{assessment:segmentation}.
The OKS is defined as: 
\begin{equation}
\text{OKS} = \frac{\sum_i[\exp(-d_i^2 / 2s^2 k_i^2) \cdot \delta (v_i > 0)]}{\sum_i [\delta (v_i > 0)]},
\label{eq:keypoint:oks}
\end{equation}
where \(d_i\) represents the Euclidean distance between the \(i\)-th predicted and ground truth keypoint, \(v_i\) denotes the visibility flag of the ground truth keypoints, \(s\) is the object scale, and \(k_i\) is a constant controlling the falloff of the Gaussian distribution for each keypoint.
The function \(\delta(v_i > 0)\) is a conditional operator that checks whether the visibility flag of the ground truth keypoint is greater than zero.
The calculation of the OKS involves passing \(d_i\) through an unnormalized Gaussian distribution with standard deviation \(s k_i\), with only visible keypoints being considered; hidden keypoints, where \(v_i = 0\), are excluded from the calculation. 
In cases where predicted and ground truth instances contain different numbers of keypoints:
\begin{itemize}
    \item If the predicted instance has more keypoints, only those corresponding to the ground truth class are considered. 
    \item If fewer keypoints are predicted than present in the ground truth, placeholder coordinates \((0, 0)\) are appended for the missing ones.
\end{itemize}
For instruments with two tip keypoints, the optimal one-to-one correspondence is determined using the Hungarian Algorithm.
If the prediction includes more tip keypoints than expected, only the first predicted tip keypoint is considered for evaluation.

\section{Participating teams and methods}

\begin{table*}[t]
    \centering
    \small
    \caption{The available tasks of the PhaKIR challenge, surgical phase recognition, instrument instance segmentation, and instrument keypoint estimation, along with the participation of individual teams in each task.}
\begin{tabularx}{\textwidth}{
p{1.75cm}
wc{0.5cm}
wc{0.6cm}
wc{0.6cm}
wc{0.8cm}
wc{0.5cm}
wc{1.0cm}
c
wc{1.0cm}
wc{0.5cm}
c
c
c
wc{0.8cm}
c
}
\toprule
\multirow{2}{*}{Task} & 
\multirow{2}{*}{alvaro} & 
\multirow{2}{*}{augi}  &
\multirow{2}{*}{goncalo} &
\makecell[Xt]{hang- \\ lok} & 
\makecell[Xt]{jmees\_ \\ inc.} &
\makecell[Xt]{recog- \\ na} &
\makecell[Xt]{kist\_ \\ harilab} &
\multirow{2}{*}{floor9} &
\multirow{2}{*}{ryze} &
\makecell[Xt]{sds- \\ hd} &
\multirow{2}{*}{sk} &
\makecell[Xt]{smartlab\_ \\ hkust} &
\makecell[Xt]{unian- \\ des24} & 
\multirow{2}{*}{yipingli} 
\\

\midrule

\rowcolor{gray!10}
Phase & & & & & & & & & & & & & & \\
\rowcolor{gray!10}
Recognition & & \multirow{-2}{*}{\ding{51}} & & \multirow{-2}{*}{\ding{51}} & \multirow{-2}{*}{\ding{51}} & & & & \multirow{-2}{*}{\ding{51}} & & & \multirow{-2}{*}{\ding{51}} & \multirow{-2}{*}{\ding{51}} & \multirow{-2}{*}{\ding{51}} \\

Instance & & \multirow{2}{*}{\ding{51}} & \multirow{2}{*}{\ding{51}} & \multirow{2}{*}{\ding{51}} & \multirow{2}{*}{\ding{51}} & \multirow{2}{*}{\ding{51}} & \multirow{2}{*}{\ding{51}} & \multirow{2}{*}{\ding{51}} & & & \multirow{2}{*}{\ding{51}} & & \multirow{2}{*}{\ding{51}} & \\
Segmentation & & & & & & & & & & & & & & \\
\rowcolor{gray!10}
Keypoint & & & & & & & & & & & & & & \\
\rowcolor{gray!10}
Estimation & \multirow{-2}{*}{\ding{51}} & & & & & & & & & \multirow{-2}{*}{\ding{51}} & & & & \\

\bottomrule
\end{tabularx}
    \label{tab:participants_with_tasks}
\end{table*}

This section provides a detailed overview of the participating teams and their methodological contributions across the three challenge tasks. 
Table~\ref{tab:participants_with_tasks} summarizes the task-wise participation. 
A total of 14 unique teams took part in the challenge.
Of these, seven teams submitted solutions for the surgical phase recognition task, nine for instrument instance segmentation, and two for instrument keypoint estimation.
Notably, five teams participated in more than one task, demonstrating interdisciplinary interest and expertise. 
Within each task, the teams are listed in ascending alphabetical order.
The descriptions of their approaches are based on the methodological reports submitted by the participants and reflect the diversity of strategies employed to tackle the challenge problems.

\subsection{Task 1: Surgical phase recognition}

\begin{table*}[t]
\footnotesize
  \centering
  \caption{Overview of participating teams and their methodological choices for the surgical phase recognition task. The table summarizes the architectures used, incorporation of temporal information, utilization of external data, loss functions, optimizers, schedulers, learning rates (LR), weight decays (WD), number of training epochs, and batch sizes (BS). 
  Frequently used abbreviations and commonly referenced elements include: 
  stochastic gradient descent (SGD, \cite{kiefer1952stochastic}), adaptive moment estimation (Adam, \cite{kingma2015adam}), exponential moving average (EMA, \cite{klinker2011exponential}), long short-term memory (LSTM, \cite{Hochreiter1997LongSM}), cross-entropy (CE, \cite{good1956some}), weighted cross-entropy (WCE, \cite{ho2020real}), cosine annealing (CA, \cite{loshchilov2017sgdr}), reduce lr on plateau (RLROP, \cite{al2022scheduling}), and ImageNet \cite{russakovsky2015imagenet}.}
  
    \begin{tabularx}{\textwidth}{p{1.5cm}p{1.6cm}p{2.0cm}p{2.9cm}p{1.0cm}p{1.7cm}p{1.5cm}p{2.8cm}}
        \toprule
            \multirow{2}{*}{Team} & Basic & Video data & Additional & Loss & Optimizer / & LR / & Epochs / \\
            
            & architecture & used? & data & functions & Scheduler & WD & BS \\
            
             \midrule

            \rowcolor{gray!10}
            \textit{augi} & LoViT~\cite{liu2025lovit} & Yes, unidirectional ASMamba~\cite{chen2024video} & LoViT pre-trained on Cholec80~\cite{nwoye2022cholec} & WCE, phase transition penalties &  Adam / \newline RLROP & 1e-3 (LoViT), 5e-4 (ASMamba) / \newline No & 10 / \newline 1 \\
             
             \textit{hanglok} & TranSVnet~\cite{gao2021trans} & Yes, TeCNO~\cite{czempiel2020tecno} & ResNet-50 backbone pre-trained on Cholec80~\cite{nwoye2022cholec} & CE & Adam / \newline RLROP & 1e-4 (stage 1), 5e-4 (stage 2) / \newline 5e-4 & 50 (stage 1), \newline 150 (stage 2) / \newline 100 \\

            \rowcolor{gray!10}
             \textit{jmees\_inc.} & EVA-02~\cite{fang2024eva} & Yes, EMA + post-processing rules & Pre-trained on ImageNet & CE & SGD / \newline CA  & 1e-5 / \newline 2e-5 & 5 / \newline 32  \\
             
             \textit{ryze} & ResNet-50~\cite{he2016deep} & Yes, LSTM layer & Pre-trained on ImageNet & CE & N/A / \newline RLROP & 5e-4 / \newline 5e-4 & 25 / \newline 150 (train), \newline 320 (validation) \\

             \rowcolor{gray!10}
             \textit{smartlab\_hkust} & Surgformer~\cite{yang2024surgformer} & Yes, HTA~\cite{yang2024surgformer} & Weights of TimeSFormer~\cite{bertasius2021space} pre-trained on Kinetics-400 dataset~\cite{kay2017kinetics} & CE & AdamW / \newline CA & 5e-4 / \newline 5e-2 & 50 / \newline 32 \\
             
             \textit{uniandes24} & MuST~\cite{perez2024must} & Yes, MTFE + TCM~\cite{perez2024must} & MViT \cite{fan2021multiscale} backbone pre-trained on Kinetics-400~\cite{kay2017kinetics} and Cholec80~\cite{twinanda2016endonet} & CE & AdamW / \newline CA & 1e-4 (MViT, MuST), 1e-5 (TCM) / \newline 1e-4 & 20 (MViT pre-training, TCM), \newline 5 (MuST finetuning) \newline 24 (MViT pre-training, MuST finetuning), \newline 256 (TCM)  \\

             \rowcolor{gray!10}
             \textit{yipingli} & CAFormer~\cite{yu2023metaformer} & Yes, LSTM layer & Yes, pre-trained weights from SurgeNet~\cite{jaspers2024exploring} & CE & AdamW / \newline Not used & 1e-4 / \newline 1e-2 & 200 / \newline 1 sequence of 128 frames \\
        \bottomrule
    \end{tabularx}
  \label{tab:methods:phase}
\end{table*}

This section provides a detailed overview of the methods submitted for the surgical phase recognition task.
Table~\ref{tab:methods:phase} complements this discussion by summarizing key methodological components for each approach.
Specifically, it outlines the model architectures, strategies for temporal modeling, and whether external data sources were utilized.
Additionally, the table includes relevant training configurations such as loss functions, optimizers, learning rate schedules, initial learning rates, weight decay parameters, number of training epochs, and batch sizes.

\subsubsection{\textbf{augi} - Kyung Hee University, South Korea}

\noindent \textbf{Architecture} The augi team employed the LoViT architecture~\cite{liu2025lovit} as a feature extractor to capture rich spatial representations.\\
\textbf{Temporal context} To model temporal dependencies, they used ASMamba~\cite{chen2024video}, a modified version of ASFormer~\cite{yi2021asformer} originally designed for temporal action segmentation. 
ASMamba replaces the attention mechanism in ASFormer with the state-space model (SSM) operation from Vision Mamba~\cite{zhu2024vision}, improving efficiency of temporal interaction learning.
While ASMamba typically leverages bidirectional Vision Mamba blocks to predict the phase of the current frame, the authors modified the model to operate in a unidirectional, causal manner to ensure causal inference.
Additionally, they extended the loss function with a penalty term to discourage implausible phase transitions and encourage sequences that align with typical surgical workflows.\\
\textbf{Dataset split} An eight-fold cross-validation was conducted, selecting the median-performing video for validation.\\
\textbf{Augmentations} No data augmentation was applied. 
Input frames were resized to $224 \times 224$ and normalized using ImageNet statistics~\cite{russakovsky2015imagenet} before feature extraction.

\subsubsection{\textbf{hanglok} - Hanglok Tech, China}

\noindent \textbf{Architecture} The hanglok team based their approach on the TranSVnet architecture~\cite{gao2021trans}, which follows a two-stage design. 
In the first stage, spatial embeddings are extracted from each video frame using a ResNet-50 backbone~\hbox{\cite{he2016deep}.}\\
\textbf{Temporal context} In the second stage, these embeddings are passed to a temporal convolutional network (TeCNO, ~\cite{czempiel2020tecno}), which combines the spatial features of the current frame with temporal features from preceding frames. 
A subsequent transformer layer produces the final classification, where the spatial embeddings of the current frame serve as queries, and the temporal embeddings from preceding frames serve as keys and values.\\
\textbf{Dataset split} For evaluation, the authors used a fixed dataset split, allocating six videos for training and one video each for validation and testing.\\
\textbf{Augmentations} Input frames were resized to $250 \times 250$ pixels. 
During training, data augmentation consisted of random horizontal flipping and rotation within a range of 
$[-5 \degree, +5 \degree]$. 
Normalization was applied to both training and validation sets using channel-wise means of $[0.417, 0.261, 0.258]$ and standard deviations of $[0.219, 0.198, 0.193]$ for the RGB channels.

\subsubsection{\textbf{jmees\_inc.} - Jmees Inc., Japan}

\noindent \textbf{Architecture} The jmees\_inc.~team employed the EVA-02 architecture~\cite{fang2024eva} using default hyperparameter settings.\\ 
\textbf{Temporal context} Temporal information was incorporated via an exponential moving average (EMA) update of the model weights, combined with predefined post-processing rules. 
These rules stipulated that for frames with an ID greater than 30,000, the ``Preparation'' phase was disallowed, while for frames with an ID below 2,000, this phase was enforced.
For prediction, if the model's confidence score exceeded 0.9, the class with the highest adjusted probability was selected as the final output. 
Otherwise, a weighted sum of class probabilities over the previous 1,000 frames was calculated, with linearly increasing weights ranging from 1 to 1,000.\\ 
\textbf{Dataset split} To ensure label diversity, two videos from different hospitals were selected for validation. 
For the final submission, the model was trained on the full dataset.\\
\textbf{Augmentations} Input images were resized to $448 \times 448$ pixels. 
Data augmentation during training included random horizontal flipping, random resizing with a scale factor sampled from the range $[0.08, 1.0]$, random aspect ratio adjustments in the range $[0.75,1.33]$, and color jittering with brightness, contrast, saturation, and hue variations each sampled from the range $[-0.4, +0.4]$.

\subsubsection{\textbf{ryze} - University of Hong Kong, China}

\noindent \textbf{Architecture} The ryze team employed a ResNet-50 architecture~\cite{he2016deep} for feature extraction, followed by a fully connected layer for surgical phase classification.\\
\textbf{Temporal context} To model temporal dependencies, a long short-term memory (LSTM, \cite{Hochreiter1997LongSM}) layer was added after the feature extractor, configured with an input size of 2048, a hidden size of 512, and a sequence length of ten frames.\\
\textbf{Dataset split} For dataset partitioning, a five-fold cross-validation strategy was used.\\
\textbf{Augmentations} During training, input images were resized to $250 \times 250$ pixels; for validation, they were resized to 
$224 \times 224$. 
Data augmentation during training included random cropping to
$224 \times 224$ pixels, random horizontal flipping, random rotation in the range $[-5 \degree, +5 \degree]$, and color jittering.
The color jittering parameters allowed for brightness, contrast, and saturation variations within $[-0.1, +0.1]$ and hue adjustments within $[-0.05, +0.05]$. 
Normalization was performed using dataset-specific mean and standard deviation values.

\subsubsection{\textbf{smartlab\_hkust} - The Hong Kong University of Science and Technology, China}

\noindent \textbf{Architecture} The smartlab\_hkust team employed the Surgformer architecture~\cite{yang2024surgformer} to process untrimmed surgical videos by constructing frame volumes from temporally sampled sequences.\\
\textbf{Temporal context} Frames were extracted at predefined intervals and divided into non-overlapping patches, which were then transformed into spatial-temporal tokens. 
These tokens were passed through a series of stacked transformer blocks, each incorporating Hierarchical Temporal Attention (HTA), Spatial Attention (SA), and a Multilayer Perceptron (MLP).  
The SA mechanism aggregates pixel features at identical temporal positions across the spatial dimension, effectively capturing spatial representations within individual frames. 
In contrast, the HTA mechanism aggregates features at identical spatial locations over the temporal dimension, leveraging multiple temporal windows at varying resolutions to enhance temporal dependency modeling.\\
\textbf{Dataset split} The dataset was partitioned into a fixed split, allocating six videos for training and two for validation.\\
\textbf{Augmentations} During training, the RandAugment augmentation strategy~\cite{cubuk2020randaugment} was applied, producing input images with a resolution of $224 \times 224$ pixels.
The configuration used was rand-m7-n4-mstd0.5-inc1, which indicates that four random transformations (n4) were applied to each image, each with a base magnitude of 7 (m7), a magnitude standard deviation of 0.5 (mstd0.5) for variability, and inclusion of the identity operation (inc1) to probabilistically retain the original image.
To further enhance data diversity, random erasing~\cite{zhong2020random} was employed, with a probability of $0.25$.
Additionally, mixup~\cite{zhang2017mixup} was incorporated as a regularization strategy.

\subsubsection{\textbf{uniandes24} - Los Andes University, Colombia}

\noindent \textbf{Architecture} The uniandes24 team utilized the Multi-Scale Transformers for Surgical Phase Recognition (MuST) architecture~\cite{perez2024must}, which follows a two-stage pipeline.\\
\textbf{Temporal context} In the first stage, a Multi-Term Frame Encoder (MTFE) is employed to adapt to diverse surgical contexts by capturing dependencies across different temporal scales.
This module constructs temporal windows of varying lengths around the target frame to generate embeddings that encode both short- and mid-term temporal information. 
The second stage consists of a Temporal Consistency Module (TCM), which enhances long-term temporal understanding by modeling relationships among the frame embeddings generated by the MTFE within an extended temporal window.  
The overall approach builds a temporal multi-scale pyramid, where lower levels provide fine-grained temporal detail and higher offer broader but sparser temporal context.  
The method employs Multiscale Vision Transformers (MViT, \cite{fan2021multiscale}) as the video backbone, leveraging its Multi-Head Pooling Attention mechanism to progressively reduce spatio-temporal resolution while increasing feature dimensionality. 

This generates multi-scale spatio-temporal features at each pyramid level, consisting of a class token and patch embeddings. 
At each level, the features are processed through a Multi-Temporal Attention Module, which combines cross-attention with residual multi-temporal self-attention units, thereby enriching the representation with information from other temporal scales.
The class tokens from all levels are then extracted and aggregated to form a comprehensive multi-term embedding of the keyframe, which is further refined through the TCM for the final phase classification. 
To enhance model predictions, the authors applied a post-processing technique that smooths output probabilities using a sliding spanning two frames.
This smoothing considers both the predicted class probabilities and their associated class labels.\\
\textbf{Dataset split} A fixed dataset split was used, assigning six videos for training and two for validation.  
During training, all images were resized to a resolution of $250 \times 250$ pixels.\\
\textbf{Augmentations} Data augmentation included random cropping to $224 \times 224$ pixels, random horizontal flipping, and AlexNet-style PCA color augmentation using the same eigenvalues and eigenvectors as in MViT.
For validation, images were resized to $224 \times 224$ pixels.  
Normalization for both training and validation followed the standard ImageNet values~\hbox{\cite{russakovsky2015imagenet}.}

\subsubsection{\textbf{yipingli} - Eindhoven University of Technology, Netherlands}

\noindent \textbf{Architecture} The yipingli team developed their approach using a backbone composed of a pre-trained CAFormer model~\cite{yu2023metaformer} to extract spatial features from input images.\\
\textbf{Temporal context} This feature extractor was coupled with a temporal head incorporating an LSTM network~\cite{Hochreiter1997LongSM} configured with a sequence length of 128 frames to capture temporal dependencies.
The LSTM module propagates hidden states across batches to maintain temporal continuity and enable a more comprehensive understanding of the video content.  
During fine-tuning, only the final block of the CAFormer backbone was unfrozen, allowing effective end-to-end supervised training while preserving the rich feature representations learned during pre-training.\\
\textbf{Dataset split} A cross-validation strategy was employed, with the dataset partitioned into six videos for training and two for validation.\\  
\textbf{Augmentations} To enhance model robustness, several data augmentation techniques were applied, including random shifting and scaling with a factor in the range $[-0.05, +0.05]$, random rotation between $[-15 \degree, +15 \degree]$, color shifts applied individually to each RGB channel within the range $[-15, +15]$, and random brightness and contrast adjustments with factors in the range $[-0.2, +0.2]$.
All images were resized to $224 \times 224$ pixels and normalized using the standard ImageNet values~\hbox{\cite{russakovsky2015imagenet}.}  

\subsection{Task 2: Instrument instance segmentation}

This section provides a detailed overview of the methods submitted by the participating teams for the instrument instance segmentation task. 
Table~\ref{tab:methods:segmentation} is structured  analogously to Table~\ref{tab:methods:phase} and summarizes the key methodological components of each approach, including the architectures employed, temporal modeling strategies, data utilization, and training configurations. 
The table serves as a concise reference for comparing the core characteristics and design decisions of the submitted solutions.

\begin{table*}[htbp]
\footnotesize
  \centering
    \caption{Overview regarding teams and methods participated in the task of surgical instrument instance segmentation. Shown are the used architectures, the usage of the temporal information, the utilization of external data, the employed loss functions, optimizers, schedulers, learning rates (LR), weight decays (WD), epochs, and batch sizes (BS). Common terms are abbreviated as follows, together with their citations: 
  stochastic gradient descent (SGD, \cite{kiefer1952stochastic}), adaptive moment estimation (Adam, \cite{kingma2015adam}), cross-entropy (CE, \cite{good1956some}), step decay (SD, \cite{johnson2023learning}), cosine annealing (CA, \cite{loshchilov2017sgdr}), flat cosine annealing (FCA, \cite{lyu2022rtmdet}), reduce lr on plateau (RLROP, \cite{al2022scheduling}), and ImageNet \cite{russakovsky2015imagenet}.}
  
    \begin{tabularx}{\textwidth}{p{1.4cm}p{1.8cm}p{1.2cm}p{3.3cm}p{1.8cm}p{2.1cm}p{1.7cm}p{1.7cm}}
        \toprule
            \multirow{2}{*}{Team} & Basic & Video data & Additional & Loss & Optimizer / & LR / & Epochs / \\
            
            & architecture & used? & data & functions & Scheduler & WD & BS \\
            
             \midrule

            \rowcolor{gray!10}
             \textit{augi} & Mask2Former \cite{cheng2022masked}, \newline PINS module & No & Swin-L~\cite{liu2021swin} backbone pre-trained on ImageNet & Binary CE + dice loss (segmentation), focal loss (classification) & AdamW / \newline SD & 1e-4 (Mask2Former, PINS) / \newline 5e-2 (Mask2Former) & 4775 (Mask2Former), 20 (PINS) / \newline 8 (Mask2Former), 16 (PINS) \\

             \textit{goncalo} & \hbox{Mask~R-CNN}~\cite{he2017mask} & No & Microsoft COCO~\cite{lin2014microsoft} & Binary CE (mask head, objectness), CE (classification) & Adam / \newline SD & 1e-3 / \newline 1e-2 & 30 / \newline 8 \\

            \rowcolor{gray!10}
             \textit{hanglok} & SAM~\cite{kirillov2023segment}, \newline DETR~\cite{carion2020end} & No & SAM pre-trained ViT-H weights \cite{kirillov2023segment}, DETR~R50 weights \cite{carion2020end}, DETR backbone pre-trained on ImageNet & Dice loss + NTXentLoss~\cite{he2020momentum, chen2020simple} (SAM), CE + IoU loss (DETR) & Adam (SAM), AdamW (DETR) / \newline SD (DETR) & 1e-3 (SAM), 1e-4 (DETR), 1e-5 (DETR backbone) / \newline 1e-4 (SAM), 1e-4 (DETR, DETR backbone) & 300 (SAM), 200 (DETR, DETR backbone) / \newline 32 (SAM), 10 (DETR, DETR backbone)  \\

             \textit{jmees\_inc.} & Mask2Former \cite{cheng2022masked}, \newline SAM 2 \cite{ravi2024sam} & No & No external data, pseudo labels generated from unlabeled PhaKIR video frames & CE (classification), dice loss + CE (segmentation) & AdamW / \newline CA & 1e-4 / \newline 5e-2 & 10 / \newline 4 \\

             \rowcolor{gray!10}
             \textit{kist\_harilab} & TP-SIS~\cite{zhou2023text}, \newline BERT~\cite{kenton2019bert}, \newline Swin transformer \cite{liu2021swin} & No & Pre-trained TP-SIS based on CLIP, EV-17, EV-18 \cite{radford2021learning, allan2017robotic, allan2018robotic}, BERT pre-trained on BooksCorpus~\cite{zhu2015aligning} and English wikipedia, Swin backbone pre-trained on ImageNet & CE & AdamW / \newline SD & 3e-5 / \newline 1e-2 & 50 / \newline 32 \\ 

             \textit{floor9} & RTMDet-Ins-l \cite{lyu2022rtmdet}, \newline SAM 2 \cite{ravi2024sam} & Yes, match instances along frames & RTMDet-Ins-l pre-trained on CholecInstanceSeg \cite{alabi2024cholecinstanceseg} & Dice-loss & AdamW / \newline FCA & 4e-3 / \newline 5e-2 & 300 / \newline 16 \\

            \rowcolor{gray!10}
            \textit{recogna} & U-Net~\cite{ronneberger2015u}, \newline SAM~\cite{kirillov2023segment} & No & U-Net trained from scratch, \newline SAM pre-trained on SA-1B \cite{kirillov2023segment} & CE & Adam / \newline RLROP & 1e-3 / \newline 1e-4 & 100 / \newline 20 \\

             \textit{sk} & U-Net \cite{ronneberger2015u} & No & ConvNext-Base \cite{liu2022convnet} backbone pre-trained on ImageNet & Dice focal loss, Hausdorff distance loss & AdamW / \newline CA & 1e-4 / \newline 1e-5 & 50 / \newline 24 \\

            \rowcolor{gray!10}
            \textit{uniandes24} & MATIS~\cite{ayobi2023matis} & Yes, TAPIS \cite{ayobi2024pixel}, SAM2 \cite{ravi2024sam} & Pre-trained ViT on ImageNet, Mask2Former on Microsoft COCO \cite{lin2014microsoft}, MViT on Kinetics-400 \cite{kay2017kinetics}, SAM2 on SA-V \cite{ravi2024sam} & CE & AdamW (ViT, Mask2Former), Adam (MViT, Transformer) / \newline SD (Mask2Former), CA (MViT, Transformer) & 1e-5 (ViT, MViT, Transformer), 1e-4 (Mask2Former) / \newline 1e-4 (MViT, Transformer) & 30 / \newline 24 (Mask2Former), 64 (MViT, Transformer) \\
        \bottomrule
    \end{tabularx}
  \label{tab:methods:segmentation}
\end{table*}

\subsubsection{\textbf{augi} - Kyung Hee University, South Korea}

\noindent \textbf{Architecture} The augi team employed the Mask2Former architecture~\cite{cheng2022masked} in combination with a Presence-aware Instrumental Segmentation (PINS) module, specifically designed to refine segmentation predictions in frames devoid of surgical instruments.  
In the first stage, feature extraction is performed at multiple scales using a Swin-Large (\mbox{Swin-L} backbone~\cite{liu2021swin}, allowing the model to accurately segment both small and large surgical tools within complex surgical scenes. 
Leveraging the transformer-based architecture of Mask2Former, the model generated multiple object queries per frame, each corresponding to a detected instance of a surgical instrument.
To reduce false-positive predictions, the authors introduced the PINS module as a second stage of their approach. 
This module, build upon a Swin-Base (Swin-B) backbone~\cite{liu2021swin}, performs binary classification to determine whether any surgical instrument is present in a given frame.  
If Mask2Former produces a segmentation mask for a frame classified by PINS as instrument-free, the PINS module suppresses the erroneous mask, thereby enhancing the overall segmentation quality.
During training, deep supervision~\hbox{\cite{lee2015deeply}} was employed to guide learning across the network's multiple stages.\\
\textbf{Temporal context} No temporal information from the video sequences was utilized.\\
\textbf{Dataset split} A fixed dataset split was used, with seven videos designated for training and one for validation.  
For Mask2Former training, only frames containing at least one instrument were included, while the PINS module was trained on all available frames.\\  
\textbf{Augmentations} Input images for Mask2Former were resized to $1024 \times 1024$ pixels with random scaling applied using a factor in the range $[-0.9, 1.0]$.   
For the PINS module, augmentation techniques included random horizontal and vertical flipping, random rotation within $[-30 \degree, +30 \degree]$, random resized cropping with scale range $[0.8, 1.0]$, color jittering (brightness, contrast, saturation, and hue variation between $[0.5, 1.5]$), and the addition of Gaussian noise with a standard deviation of $\sigma=0.05$.  
All training and validation images were resized to $224 \times 224$ pixels.  

\subsubsection{\textbf{goncalo} - University of Minho, Portugal}

\noindent \textbf{Architecture} The goncalo team developed their method based on the Mask R-CNN architecture~\cite{he2017mask}, employing a ResNet-50 backbone~\cite{he2016deep}.  
To leverage pre-trained weights effectively, the first three layers of the backbone were frozen during training, while only the subsequent layers were fine-tuned.\\
\textbf{Temporal context} Temporal information of the video sequences was not incorporated into the approach.\\
\textbf{Dataset split} A fixed dataset split was used, with six videos allocated for training and two for validation.\\
\textbf{Augmentations} No data augmentation techniques were applied during either training or validation phases.
All input images were resized to a uniform resolution of $512 \times 512$ pixels.  

\subsubsection{\textbf{hanglok} - Hanglok Tech, China}

\noindent \textbf{Architecture} The hanglok team employed the Segment Anything Model (SAM, \cite{kirillov2023segment}), utilizing the ViT-H variant of the Vision Transformer \hbox{(ViT, \cite{alexey2020image})} as the backbone.
The segmentation prompt of their model was derived from the outputs of the DEtection TRansformer (DETR, \cite{carion2020end}), which itself was based on a Swin-Transformer~\cite{liu2021swin} backbone.  
The SAM model comprises three core components: an image encoder, a prototype-based class prompt encoder, and a mask decoder.  
The image encoder extracts embeddings from the input image, which are subsequently processed by the prompt encoder.  
Following the methodology of SurgicalSAM~\cite{yue2024surgicalsam}, the authors replaced the original SAM prompt encoder with a class prompt encoder to enhance performance.  
The class prompt encoder calculates the spatial similarity between the image embedding and a class embedding bank, activating class-specific regions within the image.  
The class predicted by the DETR model is then used to generate prompt embeddings for the mask decoder.   
To improve class discrimination and facilitate more accurate class prompting, contrastive learning was applied to the class prompt encoder.\\
\textbf{Temporal context} Temporal information from the video sequences was not utilized.\\
\textbf{Dataset split} For dataset partitioning, the authors adopted a class-based splitting strategy rather than a video-based one, assigning individual frames to training and validation sets in a 9:1 ratio.\\
\textbf{Augmentations} During training, data augmentation techniques included horizontal and vertical flipping, random rotation in the range $[-30 \degree, +30 \degree]$, and random batch sampling.  

\subsubsection{\textbf{jmees\_inc.} - Jmees Inc., Japan}

\noindent \textbf{Architecture} The jmees\_inc.~team utilized the Mask2Former architecture~\cite{cheng2022masked} with a Swin-Transformer-Base~\cite{liu2021swin} backbone for the instance segmentation task.  
Pseudo labels were generated using the Segment Anything Model 2 (SAM 2, \cite{ravi2024sam}) and subsequently employed as supervision targets for training the Mask2Former model.  
To convert instance segmentation outputs into semantic segmentation masks, a post-processing step was applied, in which each predicted instance was binarized using a threshold of 0.5.\\
\textbf{Temporal context} The authors experimented with incorporation of temporal information by considering the frequency of instances across multiple frames for determining the instrument class.  
However, this approach was found to degrade validation performance and was therefore excluded from the final method.\\
\textbf{Dataset split} A fixed dataset split was used, allocating six videos for training and the remaining two for validation, with careful attention to ensuring all instrument classes were present in the validation set.  
For the final submission, the model was trained on the entire dataset.\\  
\textbf{Augmentations} During training, several data augmentation techniques were applied, including random horizontal flipping, large-scale jittering \cite{ghiasi2021simple} with a resize factor randomly sampled from the range $[0.1, 2.0]$, target resolution of $750 \times 1333$ pixels, and random cropping to $750 \times 1333$, allowing for negative crops.
For validation, no data augmentation was performed beyond resizing to the fixed resolution of $750 \times 1333$ pixels. 

\subsubsection{\textbf{kist\_harilab} - Korea Institute of Science and Technology, South Korea}

\noindent \textbf{Architecture} The kist\_harilab team employed a text-promptable surgical instrument segmentation model (TP-SIS, \cite{zhou2023text}), which leverages text embeddings generated by Bidirectional Encoder Representations from Transformers (BERT, \cite{kenton2019bert}).
For image encoding, a Swin Transformer~\cite{liu2021swin} was used as the backbone.  
The model operates based on text prompts constructed from instrument class names and additional prompts derived from object location information, specifically the centroid of each object.  
To mitigate potential bias from only positive prompts (e.g., prompts for objects present in the image), the authors also generated negative prompts for instruments not present in the frame.
This strategy enhanced the model's ability to distinguish object existence more reliably.  
To ensure training stability, the number of positive and negative prompts was balanced.\\
\textbf{Temporal context} The temporal context from the video sequences was not utilized.\\
\textbf{Dataset split} Additionally, the team did not define a separate validation dataset, opting to use the entire available dataset for model training.\\
\textbf{Augmentations} All input images were resized to a resolution of $480 \times 480$ pixels, and the network outputs were upsampled to match the original image resolutions for the final segmentation mask generation.  
No data augmentation techniques were applied during training.  

\subsubsection{\textbf{floor9} - King’s College London, UK}

\noindent \textbf{Architecture} The floor9 team utilized a model from the Real-Time Models for Object Detection (RTMDet) framework, specifically the RTMDet-Ins-l architecture~\cite{lyu2022rtmdet}, which was initialized with weights pre-trained on the CholecInstanceSeg dataset~\cite{alabi2024cholecinstanceseg} -- a large-scale tool instance segmentation benchmark specifically curated for cholecystectomy procedures.
This model was chosen for its optimal balance between high performance and efficient training time compared to other architectures in its category.
As part of the post-processing pipeline, the authors applied non-maximum suppression across classes to ensure that, for instruments with multiple predicted instances across different classes, only the instance with the highest confidence score was retained.\\
\textbf{Temporal context} In a subsequent step, temporal information was incorporated by comparing the predicted class labels of each instance in the keyframe with those of neighbouring frames.  
To facilitate instance matching across frames, the SAM~2 model~\cite{ravi2024sam} was employed to propagate labels from adjacent predictions.  
The instrument class in the keyframe was updated only when sufficient supporting evidence from surrounding frames was present.\\
\textbf{Dataset split} The authors did not report any specific details regarding a training-validation dataset split.\\
\textbf{Augmentations} For training, the default augmentation strategies from the RTMDet framework were employed, including Cached Mosaic, Cached MixUp, and large scale jittering~\cite{lyu2022rtmdet,bochkovskiy2020yolov4,zhang2017mixup,ghiasi2021simple}.
However, the parameter ranges used for these transformations were not specified.  

\subsubsection{\textbf{recogna} - São Paulo State University, Brazil}

\noindent \textbf{Architecture} The recogna team developed their method based on a two-stage coarse-to-fine procedure.  
In the first stage, a U-Net architecture~\cite{ronneberger2015u} was employed for object prediction and initial coarse segmentation.  
The second stage involved segmentation refinement using the Segment Anything Model (SAM, \cite{kirillov2023segment}), where bounding boxes extracted from the U-Net outputs were used as prompt inputs, enabling more precise segmentations during the refinement phase.\\
\textbf{Temporal context} Temporal information from the video sequences was not utilized.\\
\textbf{Dataset split} A fixed data split was applied, with 80\% of the images assigned to training and 20\% to validation. 
The images for both training and validation were randomly selected.\\  
\textbf{Augmentations} During training, the authors employed several data augmentation techniques, including horizontal and vertical flipping, random rotation within the range of $[-30 \degree, +30 \degree]$, and random color jittering with individual adjustment factors for hue, saturation, brightness, and contrast, all sampled from the interval $[-0.1, +0.1]$.  
All training and validation images were resized to a resolution of $256 \times 256$ pixels.

\subsubsection{\textbf{sk} - Muroran Institute of Technology, Japan}

\noindent \textbf{Architecture} The sk team utilized a U-Net architecture~\cite{ronneberger2015u} with a ConvNext-Base~\cite{liu2022convnet} backbone for feature extraction.  
The encoder was initialized with pre-trained weights from the ImageNet dataset~\cite{russakovsky2015imagenet}, while the decoder weights were initialized randomly.\\
\textbf{Temporal context} The authors did not incorporate any temporal information into their method.\\
\textbf{Dataset split} A fixed data split for training and validation was applied.
Specifically, three complete video sequences, along with the first halves of two additional videos, were used for training, while two complete videos and the second halves of the aforementioned videos were assigned to the validation set.\\
\textbf{Augmentations} During training, random horizontal and vertical flipping were applied as data augmentation techniques, and all input images were resized to a resolution of $960 \times 544$.

\begin{table*}[t]
\footnotesize
  \centering
  \caption{Overview of participating teams and their methods for the surgical instrument keypoint estimation task. The table summarizes the architectures employed, the use of temporal information, the utilization of external data, as well as key training parameters including loss functions, optimizers, schedulers, learning rates (LR), weight decays (WD), epochs, and batch sizes (BS). Common abbreviations and frequently used terms together with their associated references are as follows: 
  stochastic gradient descent (SGD, \cite{kiefer1952stochastic}), cross-entropy (CE, \cite{good1956some}), cosine annealing (CA, \cite{loshchilov2017sgdr}), one cycling lr (OCLR, \cite{smith2019super}).}
        \begin{tabularx}{\textwidth}{p{1.2cm}p{2.5cm}p{1.2cm}p{1.3cm}p{4.5cm}p{1.5cm}p{1.4cm}p{1.4cm}}
        \toprule
            \multirow{2}{*}{Team} & Basic & Video data & Additional & Loss & Optimizer / & LR / & Epochs / \\
            
            & architecture & used? & data & functions & Scheduler & WD & BS \\
            
             \midrule

            \rowcolor{gray!10}
             \textit{alvaro} & YOLOv8 (yolo-v8-x variant) \cite{varghese2024yolov8} & No & No & CE (classification), complete IoU loss (bounding box regression), distributed focal loss (bounding box identification), OKS loss (keypoint localization)  & SGD / \newline  OCLR with CA & 1e-2 / \newline  5e-4 & 300 / \newline 32 \\

             \textit{sds-hd} & YOLOv8 (yolo-v8-x-pose-p6 variant) \cite{varghese2024yolov8} & No & No & CE (classification, keypoint locations and visibility), complete IoU loss (bounding box regression), distributed focal loss (bounding box identification) & SGD / \newline No & 1e-2 / \newline  0.0 & 500 / \newline  104 \\
        \bottomrule
    \end{tabularx}
  \label{tab:methods:keypoints}
\end{table*}

\subsubsection{\textbf{uniandes24} - Los Andes University, Colombia}

\noindent \textbf{Architecture} The uniandes24 team proposed a fully supervised approach based on a two-stage mask classification pipeline using Masked-Attention Transformers for Surgical Instrument Segmentation (MATIS, \cite{ayobi2023matis}), extended to enhance temporal consistency in mask predictions.  
The core method relies on an adapted version of Mask2Former~\cite{cheng2022masked}, which predicts region candidates for each surgical instrument within a frame and classifies them into specific instrument types.\\  
\textbf{Temporal context} To improve classification consistency across frames, the team integrated a Temporal Consistency Module (TCM) inspired by the Transformers for Actions, Phases, Steps, and Instrument Segmentation (TAPIS) model~\cite{ayobi2024pixel}, employing the MViT~\cite{fan2021multiscale} backbone for video analysis.  
This module aggregates spatio-temporal features over time to provide essential contextual information for classification.

To maintain temporal coherence in instance predictions, SAM2~\cite{ravi2024sam} was employed to track individual instrument instances across frames, generating a temporal mask sequence for each instance.  
These tracked instances were then associated with semantic features from Mask2Former’s predictions using the IoU metric. 
Each semantic feature window was processed independently by a transformer encoder, capturing local temporal dependencies for each instance. 

Further improvements in cross frame mask classification were achieved by incorporating a cross-attention module, again inspired by TAPIS~\cite{ayobi2024pixel} and leveraging MViT~\cite{fan2021multiscale} to aggregate global spatio-temporal features.
These were fused with the local features from the transformer encoder, providing rich contextual information to support accurate classification. 
To reduce false positives in frames without instruments, a ViT \cite{lin2014microsoft} model was trained for binary classification to distinguish background from non-background frames. 
If a frame was classified as background, the mask threshold for all classes was increased to 0.9, thereby reducing the likelihood of erroneous predictions.\\
\textbf{Dataset split} The dataset was split into six videos for training and two for validation.\\
\textbf{Augmentations} For Mask2Former, input frames were resized to $750 \times 1333$ pixels, and random horizontal flipping was used as data augmentation.  
For the MViT and transformer pipeline, training images were resized to $250 \times 250$, followed by a random crop to $224 \times 224$.  
Additional augmentations included horizontal flipping, color jittering, and AlexNet-style PCA color augmentation (using the same eigenvalues and eigenvectors as in MViT), followed by ImageNet normalization~\cite{russakovsky2015imagenet} values.  
During validation, images were resized to $224 \times 224$ pixels, with only the normalization step applied.

\subsection{Task 3: Instrument keypoint estimation}

The following sections outline the methodological approaches employed by the participants in the instrument keypoint estimation task. 
Table~\ref{tab:methods:keypoints} follows the same structure as the Tables~\ref{tab:methods:phase} and \ref{tab:methods:segmentation} providing a concise overview of each team’s approach. 
It summarizes the key characteristics of the submitted methods, including the architectures used, the incorporation of temporal information, the utilization of external data, as well as essential training parameters.

\subsubsection{\textbf{alvaro} - Medtronic, UK, and \newline University College London, UK}

\noindent \textbf{Architecture} The alvaro team addressed the task of surgical instrument keypoint localization by incorporating uncertainty estimation into a state-of-the-art object detection framework.
Specifically, they utilized the YOLOv8 architecture, selecting the yolo-v8-x variant~\cite{varghese2024yolov8} and extended it to jointly predict keypoint localizations, bounding box coordinates, and class labels.  
Their objective was to leverage recent developments in deterministic uncertainty quantification methods. particularly the Deep Deterministic Uncertainty (DDU) approach~\cite{mukhoti2021deep}, building upon the foundation established by Deterministic Uncertainty Methods (DUM, \cite{postels2021practicality}).
To realize this, the authors introduced spectral normalization layers~\cite{miyato2018spectral} into each convolutional block of the YOLOv8 model.  
The rationale behind this modification is that spectral normalization enforces Lipschitz continuity, effectively constraining both the upper and lower Lipschitz bounds.  
This regularization is intended to prevent the functional collapse and enhance training stability.\\
\textbf{Temporal context} The authors did not incorporate any temporal information from the video sequences.\\
\textbf{Dataset split} The dataset was partitioned into 65\% for training, 10\% for validation, and 25\% for testing, ensuring that each instrument class was represented in all subsets.\\ 
\textbf{Augmentations} During training, a range of data augmentation techniques were applied.
Initially, images were resized to $640 \times 500$ while preserving the aspect ratio through padding, followed by horizontal and vertical flipping.  
Further augmentations included brightness reduction, contrast enhancement, and various blurring transformations -- such as motion blur, median blur, and random fog generation. 
The specific parameter ranges for these transformations were not reported.
Following augmentation, images were resized again to the final resolution of $640 \times 384$.

\subsubsection{\textbf{sds-hd} - German Cancer Research Center, Germany}

\noindent \textbf{Architecture} The sds-hd team utilized the yolov8x-pose-p6 variant of the YOLOv8 architecture~\hbox{\cite{varghese2024yolov8}.}  
Given that the provided annotations featured a variable number of keypoints per object depending on the instrument class, a dedicated padding strategy was implemented to standardize the keypoint count.
This ensured comparability with YOLOv8-pose, which requires a consistent number of keypoints across all object classes.  
After pre-processing, the standardized annotations and corresponding images were fed into the YOLOv8 model.  
During post-processing, the predictions were filtered to retain only the relevant keypoints for each instrument, while padded keypoints were excluded from the final output.\\
\textbf{Temporal context} Temporal context of the video sequences was not employed.\\
\textbf{Dataset split} For model development, a leave-one-out cross-validation strategy was applied across the eight available videos, with each video iteratively excluded for validation.  
For the final submission, the model was trained on all eight videos to fully leverage the available data.\\
\textbf{Augmentations} A variety of data augmentation techniques were employed during training.  
These included color-based augmentations such as random hue shifts within the range $[-0.015, +0.015]$, saturation adjustments in the range $[-0.7, +0.7]$, and value modifications in the range $[-0.4, +0.4]$, aimed at addressing variations in lighting and color conditions.  
Geometric augmentations included random translation in the range $[-0.1, +0.1]$, random scaling within $[-0.5, +0.5]$, and both vertical and horizontal flipping.  
In addition, the mosaic augmentation technique \cite{kaur2021data} was employed with a factor of 1.0.
This augmentation was disabled after the 50th epoch to allow for more refined training in the later stages of model convergence.

\section{Results}
\label{results}

\begin{table*}[t]
    \small
    \centering
    \caption{Results and rankings for each individual team and metric regarding the surgical phase recognition task, together with the averaged and overall rank for each team. For each metric, the mean value over the 10,000 bootstrap runs ($\text{Mean}_\text{BT}$), the 95\% confidence interval (95\% CI), and the metric-specific rank are shown, and it is indicated whether a higher or lower value is desired.}
        \begin{tabularx}{\textwidth}{p{1.8cm}P{1.6cm}YP{1.3cm}P{1.6cm}YP{1.3cm}YY} 
        \toprule
        \multirow{3}{*}{Team} & \multicolumn{3}{c}{F1-score (in \%) $\uparrow$} & \multicolumn{3}{c}{BA (in \%) $\uparrow$} & \multirow{2}{*}{Avg.} & \multirow{2}{*}{\textbf{Global}} \\
        \cmidrule(lr){2-4}
        \cmidrule(lr){5-7}
         & $\text{Mean}_{\text{BT}}$ & 95\% CI & Rank & $\text{Mean}_{\text{BT}}$ & 95\% CI & Rank & rank & \textbf{rank}  \\ 
        \midrule 
        \rowcolor{gray!10}
        uniandes24 & 69.12 & $[68.90, 69.35]$ & 1 & 84.20 & $[84.09, 84.31]$ & 1 & 1.0 & \textbf{1st} \\
        jmees\_inc. & 65.46 & $[65.23, 65.70]$ & 2 & 82.20 & $[82.07, 82.33]$ & 2 & 2.0 & \textbf{2nd} \\
        \rowcolor{gray!10}
        yipingli & 58.01 & $[57.80, 58.23]$ & 3 & 79.02 & $[78.90, 79.15]$ & 4 & 3.5 & \textbf{3rd} \\
        smartlab\_hkust & 57.34 & $[57.13, 57.55]$ & 4 & 80.31 & $[80.21, 80.42]$ & 3 & 3.5 & \textbf{3rd} \\
        \rowcolor{gray!10}
        ryze & 48.32 & $[48.09, 48.55]$ & 5 & 73.78 & $[73.63, 73.93]$ & 5 & 5.0 & \textbf{5th} \\
        hanglok & 28.83 & $[28.62, 29.03]$ & 6 & 62.21 & $[62.06, 62.35]$ & 6 & 6.0 & \textbf{6th} \\
        \rowcolor{gray!10}
        augi & 17.46 & $[17.34, 17.58]$ & 7 & 57.97 & $[57.86, 58.08]$ & 7 & 7.0 & \textbf{7th} \\
        \bottomrule
        \end{tabularx}
    \label{results:phase_recognition:results_tab}
\end{table*}

\begin{figure*}[b!]
\centering
  \begin{subfigure}{0.33\textwidth}
    \includegraphics[width=\textwidth]{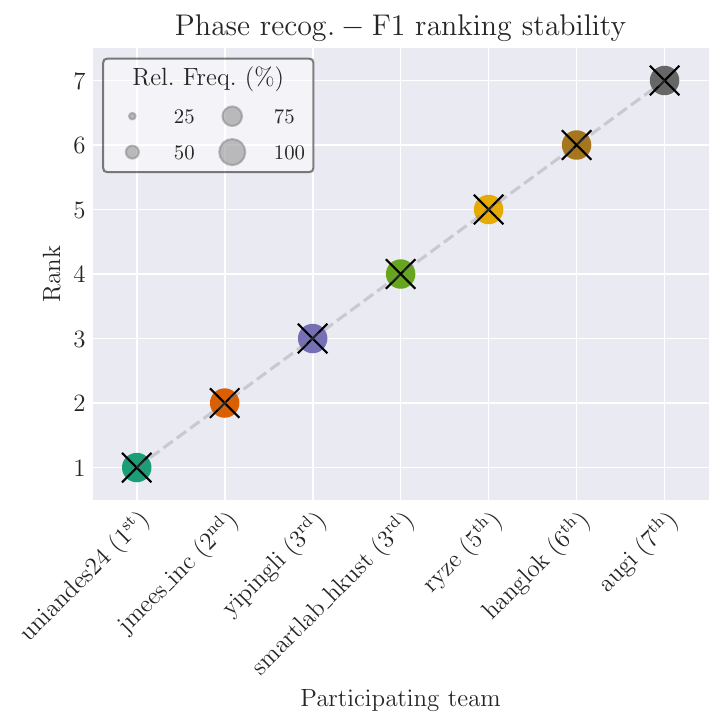}
      \caption{}
      \label{fig:results:phase_recognition:ranking_robustness:blobs:f1}
  \end{subfigure}
  \hfill
  \begin{subfigure}{0.33\textwidth}
    \includegraphics[width=\textwidth]{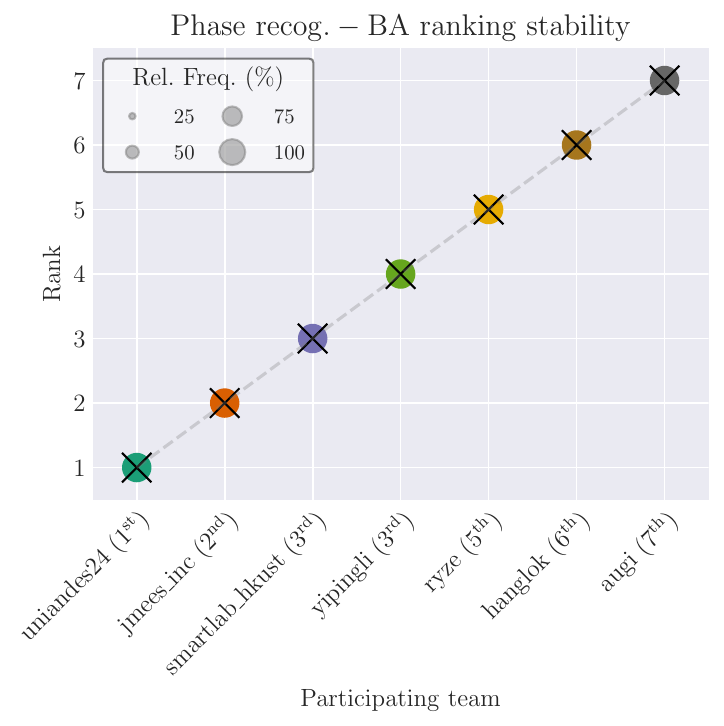}
      \caption{}
      \label{fig:results:phase_recognition:ranking_robustness:blobs:ba}
  \end{subfigure}
    \hfill
  \begin{subfigure}{0.33\textwidth}
    \includegraphics[width=\textwidth]{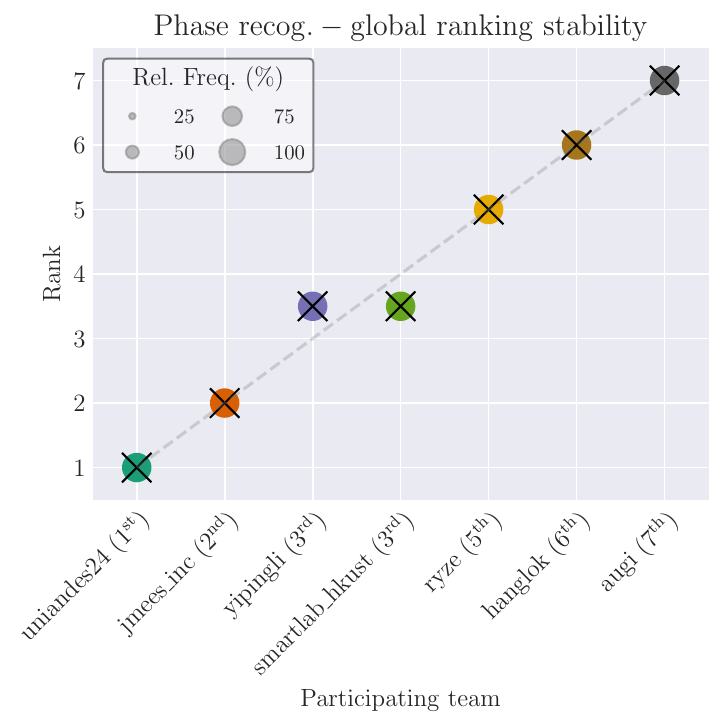}
      \caption{}
      \label{fig:results:phase_recognition:ranking_robustness:blobs:global}
  \end{subfigure}
\caption{
\label{fig:resuts:phase_recognition:ranking_robustness:blobs}%
Visualization of ranking stability based on bootstrapping for the surgical phase recognition task for the metrics F1-score~(\subref{fig:results:phase_recognition:ranking_robustness:blobs:f1}), BA~(\subref{fig:results:phase_recognition:ranking_robustness:blobs:f1}), and global~(\subref{fig:results:phase_recognition:ranking_robustness:blobs:global}). For each metric, the individual ranking stability is shown and the global rank is indicated after each team name.}
\end{figure*}

\begin{figure*}[t!]
\centering
  \begin{subfigure}{0.33\textwidth}
    \includegraphics[width=\textwidth]{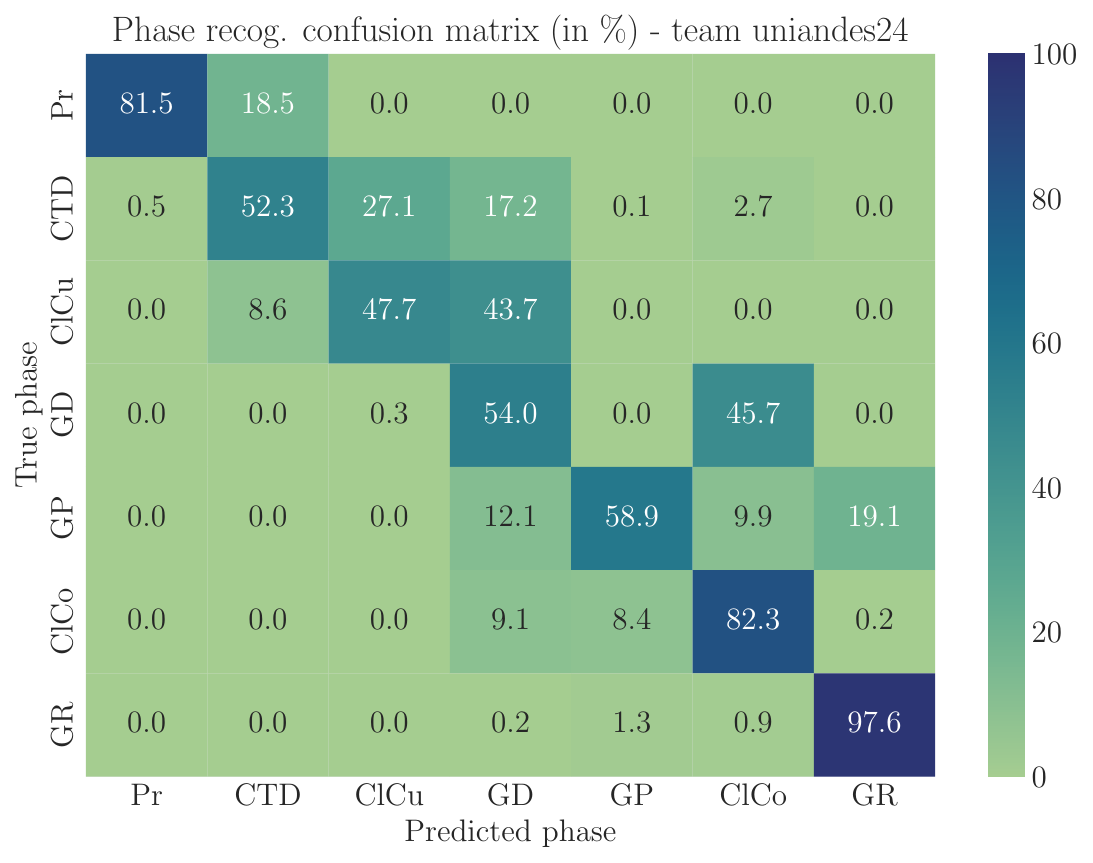}
      \caption{}
      \label{fig:results:phase_recognition_confusion_matrices:uniandes24}
  \end{subfigure}
  \hfill
  \begin{subfigure}{0.33\textwidth}
    \includegraphics[width=\textwidth]{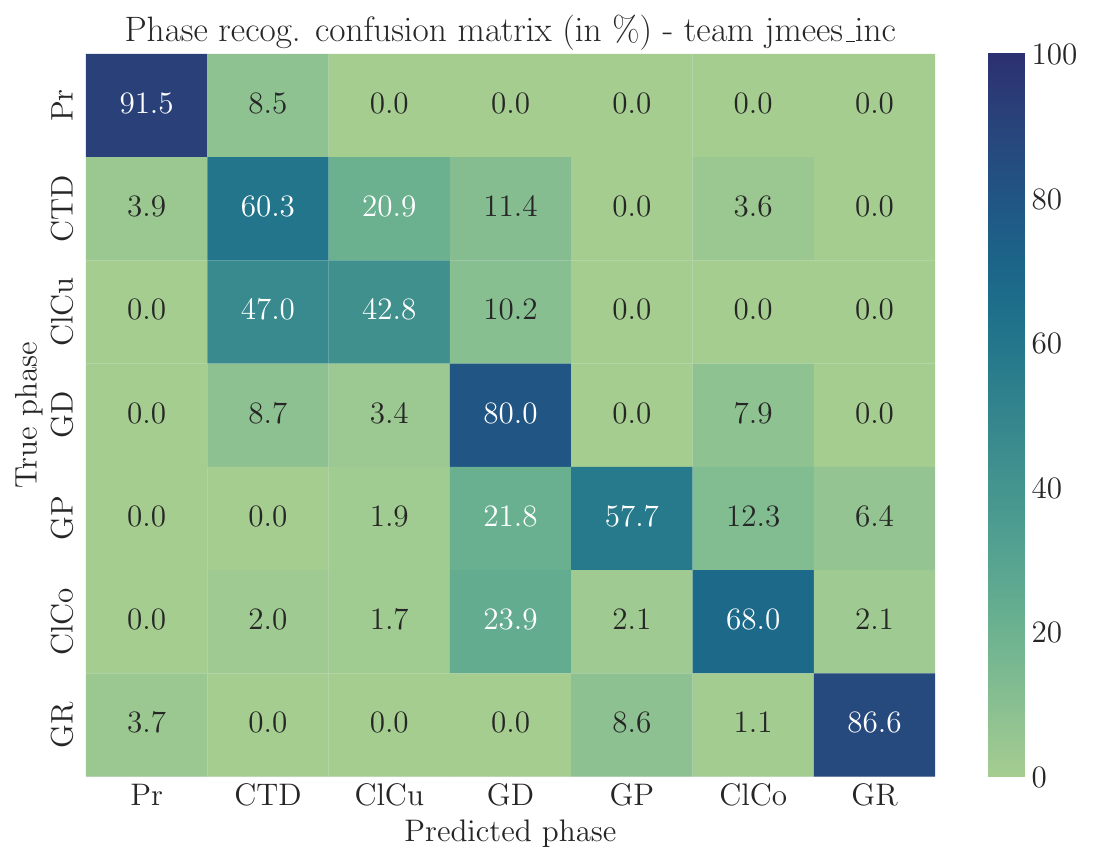}
      \caption{}
      \label{fig:results:phase_recognition_confusion_matrices:jmees_inc}
  \end{subfigure}
    \hfill
  \begin{subfigure}{0.33\textwidth}
    \includegraphics[width=\textwidth]{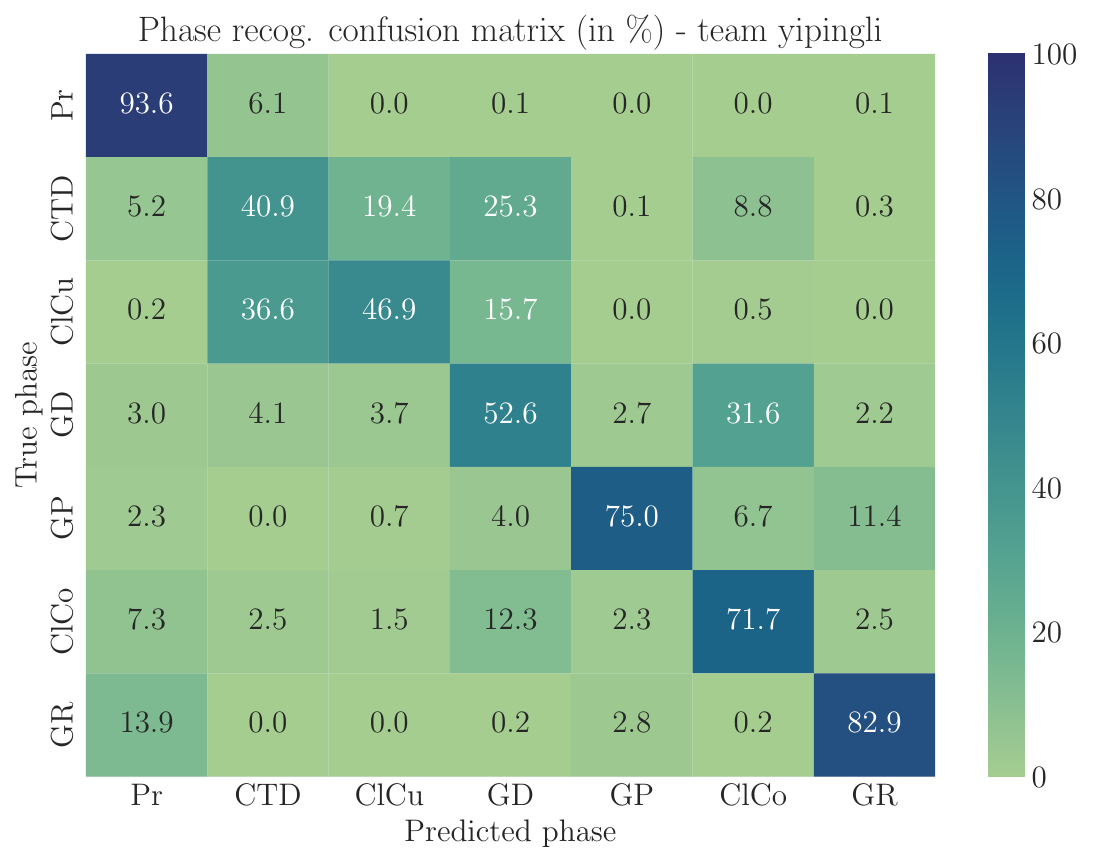}
      \caption{}
      \label{fig:results:phase_recognition_confusion_matrices:yipingli}
  \end{subfigure}
      \hfill
  \begin{subfigure}{0.33\textwidth}
    \includegraphics[width=\textwidth]{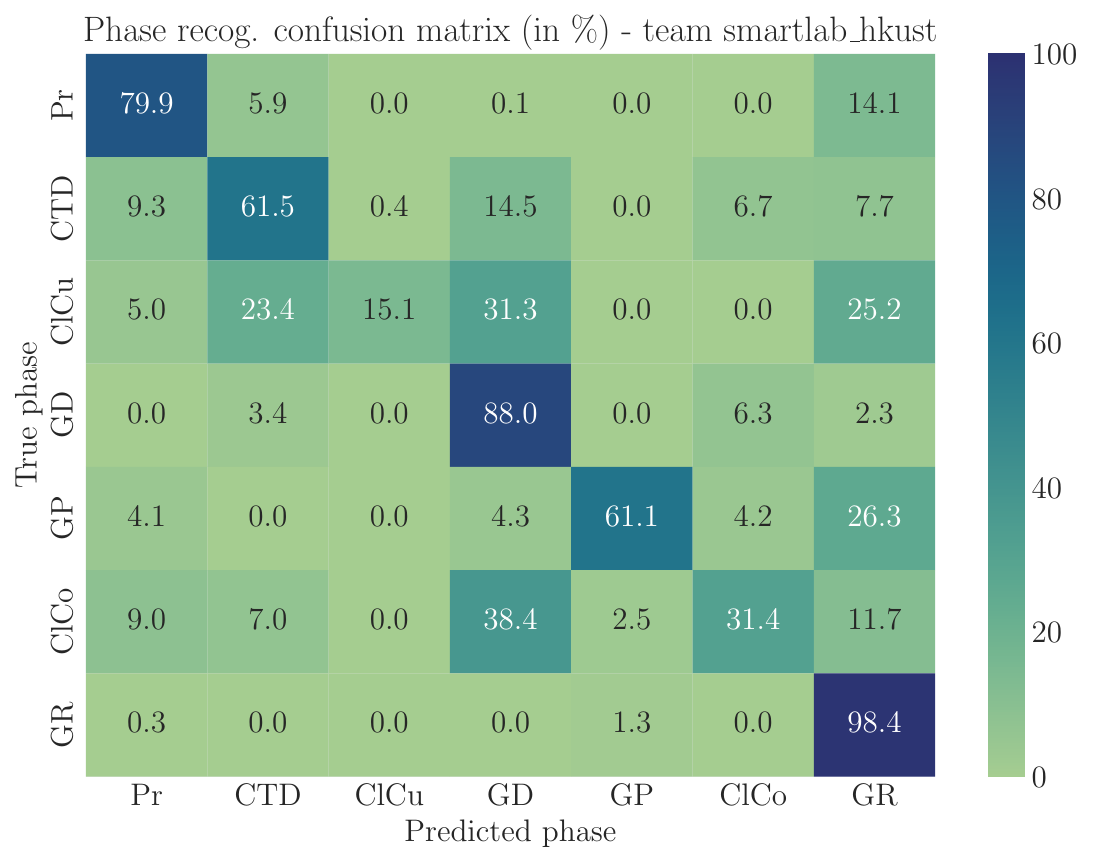}
      \caption{}
      \label{fig:results:phase_recognition_confusion_matrices:smartlab_hkust}
  \end{subfigure}
      \hfill
  \begin{subfigure}{0.33\textwidth}
    \includegraphics[width=\textwidth]{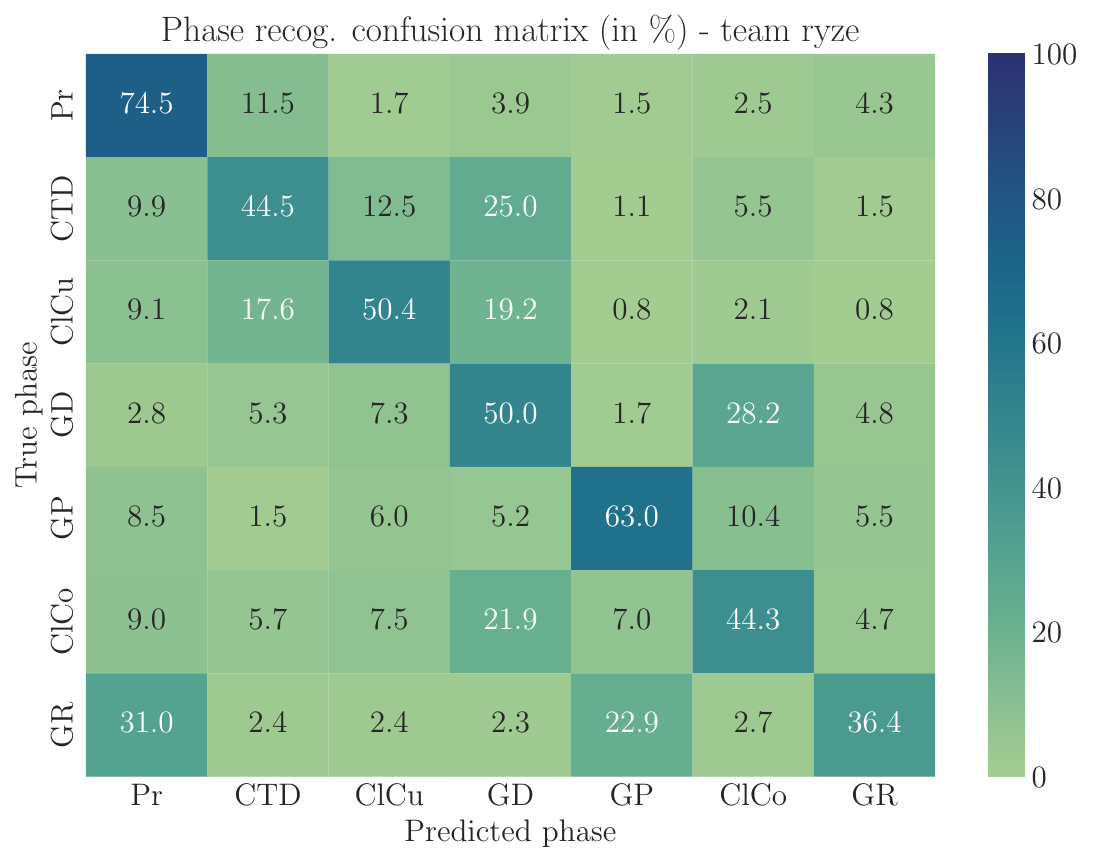}
      \caption{}
      \label{fig:results:phase_recognition_confusion_matrices:ryze}
  \end{subfigure}
      \hfill
  \begin{subfigure}{0.33\textwidth}
    \includegraphics[width=\textwidth]{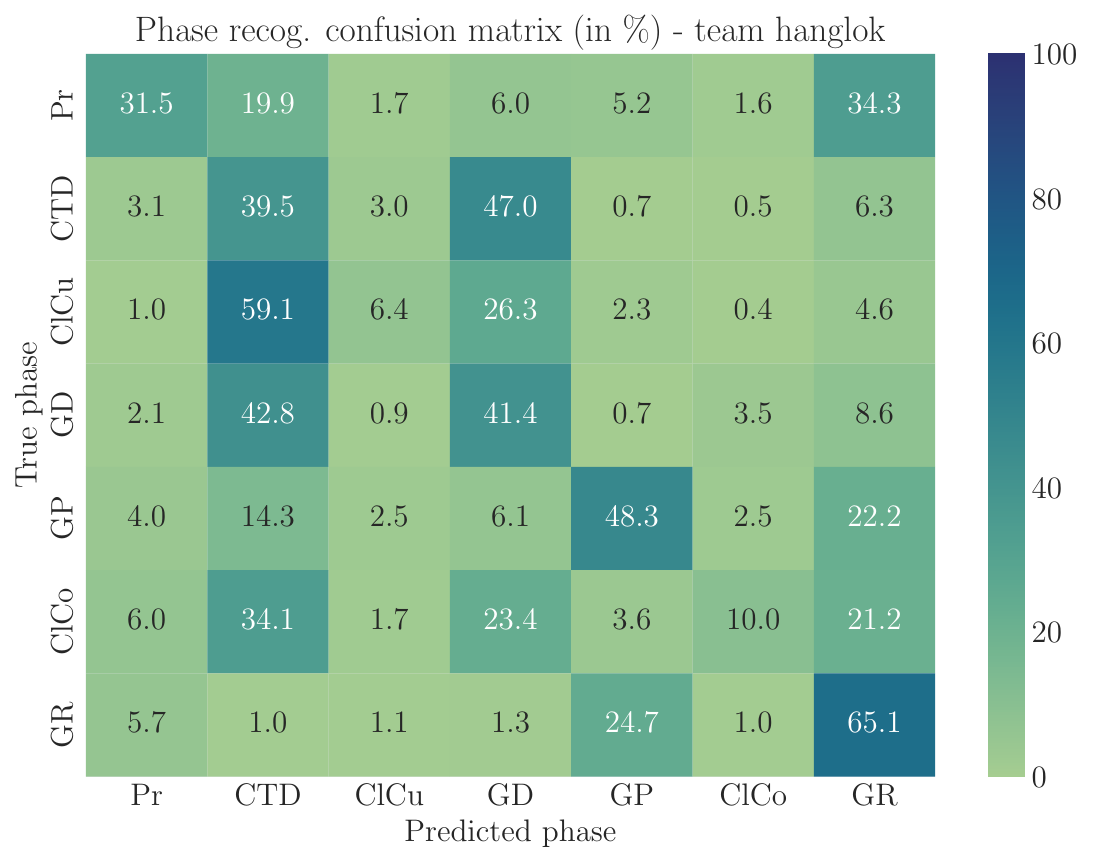}
      \caption{}
      \label{ffig:results:phase_recognition_confusion_matrices:hanglok}
  \end{subfigure}
      \hfill
  \begin{subfigure}{0.33\textwidth}
    \includegraphics[width=\textwidth]{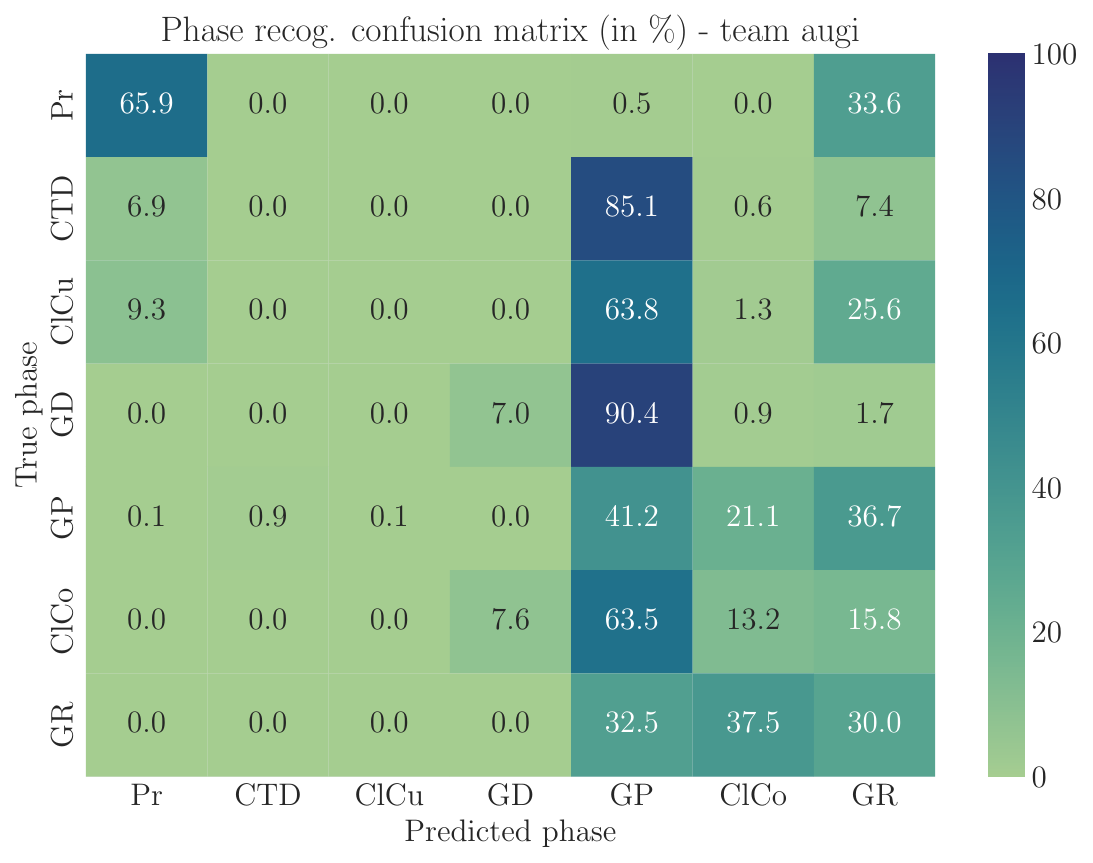}
      \caption{}
      \label{fig:results:phase_recognition_confusion_matrices:augi}
  \end{subfigure}
\caption{Confusion matrices for the surgical phase recognition task across all test videos, given in row-wise relative values, for teams uniandes24
(\subref{fig:results:phase_recognition_confusion_matrices:uniandes24}),
jmees\_inc. 
(\subref{fig:results:phase_recognition_confusion_matrices:jmees_inc}),
yipingli 
(\subref{fig:results:phase_recognition_confusion_matrices:yipingli}),
smartlab\_hkust 
(\subref{fig:results:phase_recognition_confusion_matrices:smartlab_hkust}),
ryze 
(\subref{fig:results:phase_recognition_confusion_matrices:ryze}),
hanglok
(\subref{ffig:results:phase_recognition_confusion_matrices:hanglok}),
augi
(\subref{fig:results:phase_recognition_confusion_matrices:augi}), sorted according to the results and rankings shown in Table~\ref{results:phase_recognition:results_tab}. For clearer visualization, the phase names are abbreviated as follows: preparation (Pr), calot triangle dissection (CTD), clipping and cutting (ClCu), gallbladder dissection (GD), gallbladder packaging (GP), cleaning and coagulation (ClCo), and gallbladder retraction (GR).
}
\label{fig:resuts:phase_recognition_confusion_matrices}
\end{figure*}

The following sections present the evaluation results, beginning with an overview of the two statistical analysis methods used~(Sec.~\ref{results:statistical_analysis}).
This is followed by the detailed results for the three challenge tasks: surgical phase recognition~(Sec.~\ref{results:phase_recognition}), instrument instance segmentation~(Sec.~\ref{results:instrument_segmentation}), and instrument keypoint estimation~(Sec.~\ref{results:keypoint_estimation}).  
For each task, we first report the overall performance results (Tab.~\ref{results:phase_recognition:results_tab}, \ref{results:instance_seg:results_tab}, \ref{results:keypoint_estimation:results_tab}), followed by a statistical analysis of ranking stability (Fig.~\ref{fig:resuts:phase_recognition:ranking_robustness:blobs}, \ref{fig:results:instrument_segmentation:ranking_robustness:blobs}, \ref{fig:results:keypoint_estimation:ranking_robustness:blobs}) and task-specific supplementary analyses.  
For the phase recognition task, we include a team-specific confusion matrix visualization (Fig.~\ref{fig:resuts:phase_recognition_confusion_matrices}).  
For the segmentation and keypoint estimation tasks, we present qualitative results  (Tab.~\ref{results:instrument_seg:tab:qualitative_results_page_01}, \ref{results:instrument_seg:tab:qualitative_results_page_02}, \ref{results:keypoint_estimation:tab:qualitative_results_page_01}).  
In addition, the instrument segmentation task includes a comparison of the methodological approaches used by the participating teams (Fig.~\ref{results:instance_seg:method_comparison}).  
Finally, for all three tasks, we analyze team performance at the level of individual test videos, enabling a more fine-grained assessment of robustness and consistency (Fig.~\ref{results:phase_recognition:density_plot:f1}, \ref{results:phase_recognition:density_plot:ba}, \ref{results:instance_seg:density_plot:dsc}, \ref{results:instance_seg:density_plot:map}, \ref{results:instance_seg:density_plot:hd}, \ref{results:keypoint_estimation:density_plot:map}; Tab.~\ref{results:phase_recognition:results_tab_individual_videos}, \ref{results:instance_seg:results_tab_individual_videos}, \ref{results:keypoint_estimation:results_tab_individual_videos}).

\subsection{Statistical analysis}
\label{results:statistical_analysis}

This section introduces two statistical analysis methods used in the subsequent chapters: ranking stability plots (Fig.~\ref{fig:resuts:phase_recognition:ranking_robustness:blobs}, \ref{fig:results:instrument_segmentation:ranking_robustness:blobs}, and \ref{fig:results:keypoint_estimation:ranking_robustness:blobs}) and kernel density estimation plots \hbox{(KDE, \cite{silverman2018density},} Fig.~\ref{results:phase_recognition:density_plot:f1}, \ref{results:phase_recognition:density_plot:ba}, \ref{results:instance_seg:density_plot:dsc}, \ref{results:instance_seg:density_plot:map}, \ref{results:instance_seg:density_plot:hd}, \ref{results:keypoint_estimation:density_plot:map}). 

Ranking stability plots illustrate the robustness of metric-based rankings by repeating the evaluation across multiple bootstrap iterations.  
In these plots, the size of each color-coded blob at position ($\text{A}_i$, rank $j$) reflects the relative frequency with which algorithm $\text{A}_i$ achieved rank $j$ across $b = 10{,}000$ bootstrap samples.  
The median rank of each algorithm is indicated by a black cross, and the corresponding 95\% bootstrap interval, ranging from the 2.5$^{\text{th}}$ to the 97.5$^{\text{th}}$ percentile, is shown as a black line.  
This visualization technique is adapted from the blob plot concept in~\cite{wiesenfarth2021methods}.

Kernel density estimation with a Gaussian kernel is a non-parametric technique used to estimate the probability density function of a continuous variable from sample data.  
Each KDE is constructed by centering a Gaussian kernel at each data point \( x_i \), and summing the contributions to form a smooth approximation of the underlying distribution.
The KDE is defined as:
\begin{equation}
f(x) = \frac{1}{n} \sum_{i=1}^{n} \frac{1}{\sqrt{2\pi}h} \exp\left(-\frac{(x - x_i)^2}{2h^2}\right),
\end{equation}
where \( x \) is the evaluation point and \( x_i \) are the observed data points and \( h \) is the bandwidth parameter controlling the smoothness of the estimate.
High-density regions of the bootstrap distribution appear as sharp peaks in the KDE, while low-density regions produce flatter shapes.  
To select an appropriate bandwidth \( h \), we apply Scott's Rule~\cite{scott2015multivariate}, which balances the sample size and data dimensionality:
\begin{equation}
h = n^{-\frac{1}{d+4}},
\end{equation}
where \( n = 10{,}000 \) denotes the number of bootstrap samples and \( d = 1 \) represents the dimensionality of the data.

KDE plots are valuable for visualizing bootstrap results because they provide an intuitive and smooth representation of an algorithm's performance across repeated evaluation runs. 
Rather than focusing solely on single summary statistics such as the mean or median, KDE plots reveal the full distribution of performance.
This makes it easier to compare methods, understand variability, and assess consistency in a more comprehensive and interpretable way.

\subsection{Task 1: Surgical phase recognition}
\label{results:phase_recognition}

\begin{figure*}[h!]
    \small
    \hfill
    \begin{overpic}[width=0.9\textwidth]{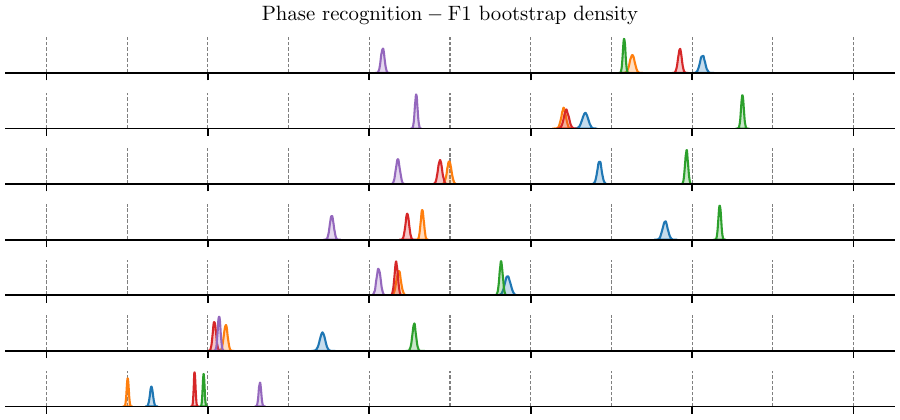}
    \put(-9, 40.6){\color{black} uniandes24}
    \put(-9, 34.5){\color{black} jmees\_inc.}
    \put(-9, 28.4){\color{black} yipingli}
    \put(-9, 22.3){\color{black} smartlab\_hkust}
    \put(-9, 16.2){\color{black} ryze}
    \put(-9, 10.1){\color{black} hanglok}
    \put(-9, 4){\color{black} augi}
    \put(3.8, -1.5){\color{black} 0.0}
    \put(21.5, -1.5){\color{black} 0.2}
    \put(39.6, -1.5){\color{black} 0.4}
    \put(57.5, -1.5){\color{black} 0.6}
    \put(75.5, -1.5){\color{black} 0.8}
    \put(93.5, -1.5){\color{black} 1.0}
    
\end{overpic}
    \caption{KDE plot based on the bootstrap results for the participating teams of the surgical phase recognition task with regard to the F1-score values for the test set videos 6 \legendsquare{darkblue}, 8 \legendsquare{darkorange}, 9 \legendsquare{darkgreen}, 12 \legendsquare{darkred}, and 14 \legendsquare{darkpurple}.}
    \label{results:phase_recognition:density_plot:f1}
\end{figure*}

\begin{figure*}[h!]
    \small
    \hfill
    \begin{overpic}[width=0.9\textwidth]{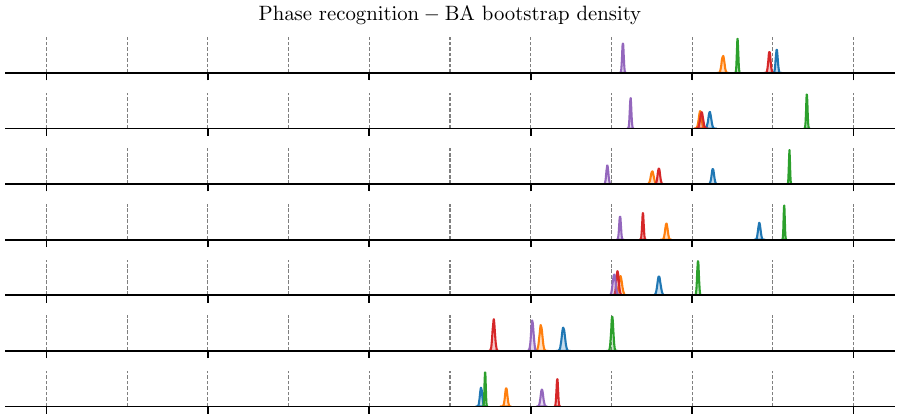}
    \put(-9, 40.6){\color{black} uniandes24}
    \put(-9, 34.5){\color{black} jmees\_inc.}
    \put(-9, 28.4){\color{black} yipingli}
    \put(-9, 22.3){\color{black} smartlab\_hkust}
    \put(-9, 16.2){\color{black} ryze}
    \put(-9, 10.1){\color{black} hanglok}
    \put(-9, 4){\color{black} augi}
    \put(3.8, -1.5){\color{black} 0.0}
    \put(21.5, -1.5){\color{black} 0.2}
    \put(39.6, -1.5){\color{black} 0.4}
    \put(57.5, -1.5){\color{black} 0.6}
    \put(75.5, -1.5){\color{black} 0.8}
    \put(93.5, -1.5){\color{black} 1.0}
    
\end{overpic}
    \caption{KDE plot based on the bootstrap results for the participating teams of the surgical phase recognition task with regard to the BA values for the test set videos 6 \legendsquare{darkblue}, 8 \legendsquare{darkorange}, 9 \legendsquare{darkgreen}, 12 \legendsquare{darkred}, and 14 \legendsquare{darkpurple}.}
    \label{results:phase_recognition:density_plot:ba}
\end{figure*}

The results for the surgical phase recognition task are summarized in Table~\ref{results:phase_recognition:results_tab}.  
The team uniandes24 achieved the highest scores in both metrics (69.12\% F1-score, 84.20\% BA), securing first place in the overall ranking and thereby winning the task.  
The individual rankings across both metrics are largely consistent, with the exception of a swap between third and fourth place: yipingli ranked higher in terms of F1-score, while smartlab\_hkust outperformed in BA.  
Since both teams attained the same average rank of 3.5, they shared third place in the final ranking.  

The ranking stability, assessed via bootstrapping, is visualized in Figure~\ref{fig:resuts:phase_recognition:ranking_robustness:blobs}.  
Figures~\ref{fig:results:phase_recognition:ranking_robustness:blobs:f1} and \ref{fig:results:phase_recognition:ranking_robustness:blobs:ba} depict the ranking stability for the F1-score and the BA metric, respectively, while Figure~\ref{fig:results:phase_recognition:ranking_robustness:blobs:global} presents the stability of the overall ranking based on the average of both metrics.  
The results demonstrate high robustness across all rankings and confirm the stability of yipingli's third place in F1-score and smartlab\_hkust's third place in BA.  

Figure~\ref{fig:resuts:phase_recognition_confusion_matrices} presents the confusion matrices for all participating teams, illustrating the relationship between the actual and predicted surgical phases and highlighting the proportion of correctly and incorrectly classified categories.  
Among all phases, the preparation phase (Pr) achieved the highest average classification accuracy, with 74.05\% across all participants.  
In contrast, the clipping and cutting phase (ClCu) proved to be the most challenging, with the lowest average accuracy of 29.89\%.  

\begin{table*}[t]
    \small
    \centering
    \caption{Results of the surgical phase recognition task for each team and each individual video in the test dataset, together with the mean metric values for each video.}
    \begin{tabularx}{\textwidth}{p{1.7cm}P{1.2cm}P{1.2cm}P{1.2cm}P{1.2cm}P{1.2cm} 
                              P{1.2cm}P{1.2cm}P{1.2cm}P{1.2cm}P{1.2cm}}  
        \toprule
        \multirow{2}{*}{Team} 
        & \multicolumn{5}{c}{F1-score (in \%) $\uparrow$} 
        & \multicolumn{5}{c}{BA (in \%) $\uparrow$} \\
        \cmidrule(lr){2-6} \cmidrule(l){7-11}
        & V\_06 & V\_08 & V\_09 & V\_12 & V\_14
        & V\_06 & V\_08 & V\_09 & V\_12 & V\_14 \\
        \midrule
        \rowcolor{gray!10}
        uniandes24 & 81.30 & 72.59 & 71.58 & 78.49 & 41.66 & 90.50 & 83.84 & 85.64 & 89.59 & 71.43 \\
        jmees\_inc. & 66.76 & 64.08 & 86.24 & 64.42 & 45.81 & 82.19 & 81.01 & 94.23 & 81.19 & 72.38 \\
        \rowcolor{gray!10}
        yipingli & 68.53 & 49.91 & 79.32 & 48.77 & 43.54 & 82.58 & 75.07 & 92.07 & 75.89 & 69.50 \\
        smartlab\_hkust & 76.67 & 46.57 & 83.43 & 44.70 & 35.34 & 88.34 & 76.82 & 91.42 & 73.91 & 71.08 \\
        \rowcolor{gray!10}
        ryze & 57.15 & 43.64 & 56.33 & 43.33 & 41.14 & 75.90 & 71.13 & 80.74 & 70.77 & 70.35 \\
        hanglok & 34.19 & 22.20 & 45.56 & 20.80 & 21.39 & 64.05 & 61.27 & 70.11 & 55.42 & 60.20 \\
        \rowcolor{gray!10}
        augi & 12.99 & 10.05 & 19.46 & 18.35 & 26.44 & 53.85 & 56.97 & 54.35 & 63.30 & 61.39 \\
        \midrule
        \textbf{mean} & \textbf{56.80} & \textbf{44.15} & \textbf{63.13} & \textbf{45.55} & \textbf{36.47} & \textbf{76.77} & \textbf{72.30} & \textbf{81.22} & \textbf{72.87} & \textbf{68.05} \\
        \bottomrule
    \end{tabularx}
    \label{results:phase_recognition:results_tab_individual_videos}
\end{table*}

\begin{table*}[b]
    \small
    \centering
    \caption{Results and rankings for each individual team and metric regarding the instrument instance segmentation task, together with the averaged and overall rank for each team. For each metric, the mean value over the 10,000 bootstrap runs ($\text{Mean}_\text{BT}$), the 95\% confidence interval (95\% CI), and the metric-specific rank are shown, and it is indicated whether a higher or lower value is desired.}
        \begin{tabularx}{\textwidth}{p{1.3cm}P{1cm}P{1.6cm}P{0.7cm}P{1cm}P{1.6cm}P{0.7cm}Z{1cm}P{2.0cm}P{0.7cm}YY} 
        \toprule
        \multirow{3}{*}{Team} 
        & \multicolumn{3}{c}{DSC (in \%) $\uparrow$} 
        & \multicolumn{3}{c}{$\text{mAP}_{\text{IoU}}$ (in \%) $\uparrow$} 
        & \multicolumn{3}{c}{95\% HD $\downarrow$} 
        & \multirow{2}{*}{Avg.} 
        & \multirow{2}{*}{\textbf{Global}} \\
        \cmidrule(lr){2-4}
        \cmidrule(lr){5-7}
        \cmidrule(lr){8-10}
         & $\text{Mean}_{\text{BT}}$ & 95\% CI & Rank 
         & $\text{Mean}_{\text{BT}}$ & 95\% CI & Rank 
         & $\text{Mean}_{\text{BT}}$ & 95\% CI & Rank 
         & rank & \textbf{rank} \\
        \midrule
        \rowcolor{gray!10}
        jmees\_inc. & 35.48 & $[34.12, 36.96]$ & 1 & 36.05 & $[34.51, 37.70]$ & 1 & 75.48 & $[68.98, 82.41]$ & 3 & 1.67 & \textbf{1st} \\
        uniandes24 & 33.10 & $[31.43, 35.20]$ & 2 & 33.39 & $[31.52, 35.66]$ & 3 & 71.47 & $[56.32, 85.03]$ & 1 & 2.00 & \textbf{2nd} \\
        \rowcolor{gray!10}
        augi & 32.95 & $[31.61, 34.56]$ & 3 & 33.53 & $[31.99, 35.31]$ & 2 & 73.10 & $[62.15, 82.24]$ & 2 & 2.33 & \textbf{3rd} \\
        kist\_harilab & 26.72 & $[26.01, 27.54]$ & 4 & 26.40 & $[25.63, 27.29]$ & 4 & 163.91 & $[145.75, 182.09]$ & 6 & 4.67 & \textbf{4th} \\
        \rowcolor{gray!10}
        floor9 & 21.64 & $[20.47, 23.07]$ & 6 & 20.63 & $[19.36, 22.16]$ & 6 & 82.91 & $[76.99, 89.24]$ & 4 & 5.33 & \textbf{5th} \\
        sk & 24.24 & $[23.59, 24.90]$ & 5 & 23.73 & $[22.96, 24.50]$ & 5 & 243.56 & $[222.71, 259.75]$ & 7 & 5.67 & \textbf{6th} \\
        \rowcolor{gray!10}
        goncalo & 18.57 & $[17.60, 19.73]$ & 8 & 18.34 & $[17.19, 19.62]$ & 8 & 155.12 & $[142.39, 166.73]$ & 5 & 7.00 & \textbf{7th} \\
        hanglok & 18.92 & $[17.80, 20.15]$ & 7 & 18.84 & $[17.45, 20.33]$ & 7 & 246.13 & $[232.84, 259.82]$ & 8 & 7.33 & \textbf{8th} \\
        \rowcolor{gray!10}
        recogna & 13.46 & $[12.73, 14.24]$ & 9 & 12.39 & $[11.57, 13.26]$ & 9 & 300.12 & $[286.60, 312.53]$ & 9 & 9.00 & \textbf{9th} \\
        \bottomrule
        \end{tabularx}
    \label{results:instance_seg:results_tab}
\end{table*}

The average classification accuracies for the remaining phases were as follows:
\begin{itemize}
    \item gallbladder retraction (GR): 71.00\%
    \item gallbladder packaging (GP): 57.89\%
    \item gallbladder dissection (GD): 53.28\%
    \item cleaning and coagulation (ClCo): 45.85\%
    \item calot triangle dissection (CTD): 42.71\%.
\end{itemize}
A closer inspection of misclassification patterns reveals consistent trends:
\begin{itemize}
    \item The Pr phase was most frequently confused with GR (12.35\%) and CTD (10.07\%).  
    \item The CTD phase was most often misclassified as GD (20.05\%) and GP (12.43\%).  
    \item The ClCu phase was commonly mistaken for CTD (27.46\%) and GD (20.92\%).  
    \item The GD phase was misclassified as ClCo (17.72\%) and GP (13.65\%).  
    \item The GP phase was most frequently confused with GR (18.22\%) and ClCo (9.57\%).  
    \item The ClCo phase was often mistaken for GD (19.52\%) and GP (12.77\%).  
    \item The GR phase was most commonly misclassified as GP (13.44\%) and Pr (7.81\%).  
\end{itemize}

Figures~\ref{results:phase_recognition:density_plot:f1} and \ref{results:phase_recognition:density_plot:ba} display the KDE plots for the F1-score and BA, respectively, derived from bootstrap evaluation across individual test videos.  
In the KDE plot for the F1-score (Fig.~\ref{results:phase_recognition:density_plot:f1}), video nine yielded the highest metric values for four out of seven teams, indicating it was the most accurately predicted video overall.  
In contrast, video 14 produced the lowest F1-scores for five out of seven teams, making it the most challenging sequence in the dataset.  
A similar pattern emerges in the KDE plot for BA (Fig.~\ref{results:phase_recognition:density_plot:ba}): video nine achieved the highest scores for five teams, while video 14 registered the lowest values for five teams.

Table~\ref{results:phase_recognition:results_tab_individual_videos}.  
presents quantitative results summarizing the mean metric values for each individual video in the test dataset. These numerical results align well with the KDE visualizations shown in Figures~\ref{results:phase_recognition:density_plot:f1} and \ref{results:phase_recognition:density_plot:ba}, reinforcing the observed performance trends. 
Among all test sequences, video nine was the most successfully recognized across teams, achieving an average F1-score of 63.13\% and BA of 81.22\%.
In contrast, video 14 proved most challenging, with the lowest average scores (36.47\% F1-score, 68.05\% BA).  
The second-highest performance was observed for video six (56.80\% F1-score, 76.77\% BA), followed by video 12 (45.55\% F1-score, 72.87\% BA), and video eight (44.15\% F1-score, 72.30\% BA).

\subsection{Task 2: Instrument instance segmentation}
\label{results:instrument_segmentation}

Table~\ref{results:instance_seg:results_tab} summarizes the detailed results of the instrument instance segmentation task.
The team jmees\_inc.~achieved the highest scores in two out of three evaluation metrics (35.48\% DSC and 36.05\% $\text{mAP}_{\text{IoU}}$) and ranked third in the 95\% HD metric (75.48\%), securing the best overall performance in the task.
A largely consistent ranking pattern emerges between the DSC and $\text{mAP}_{\text{IoU}}$ metrics, with only minor rank fluctuations caused by closely aligned performance values among competing teams.
In contrast, the ranking based on the 95\% HD metric diverges considerably from those observed in the other two metrics.
Notably, only the teams recogna and augi exhibit stable rankings that align with their performance in the $\text{mAP}_{\text{IoU}}$ metric, highlighting differences in boundary-level versus region-based segmentation performance.

\begin{figure*}[h!]
\centering
  \begin{subfigure}{0.49\textwidth}
    \includegraphics[width=\textwidth]{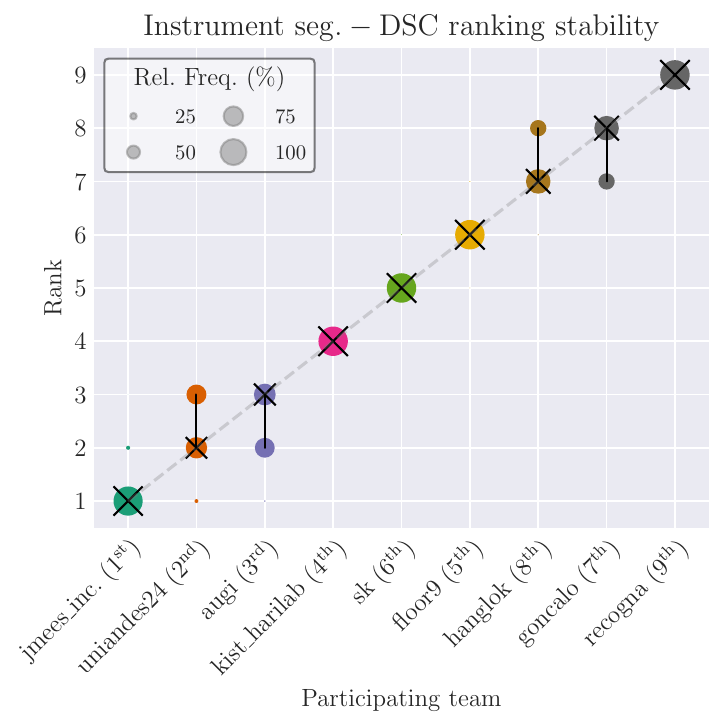}
      \caption{}
      \label{fig:results:instrument_segmentation:ranking_robustness:blobs:dsc}
  \end{subfigure}
  \hfill
  \begin{subfigure}{0.49\textwidth}
    \includegraphics[width=\textwidth]{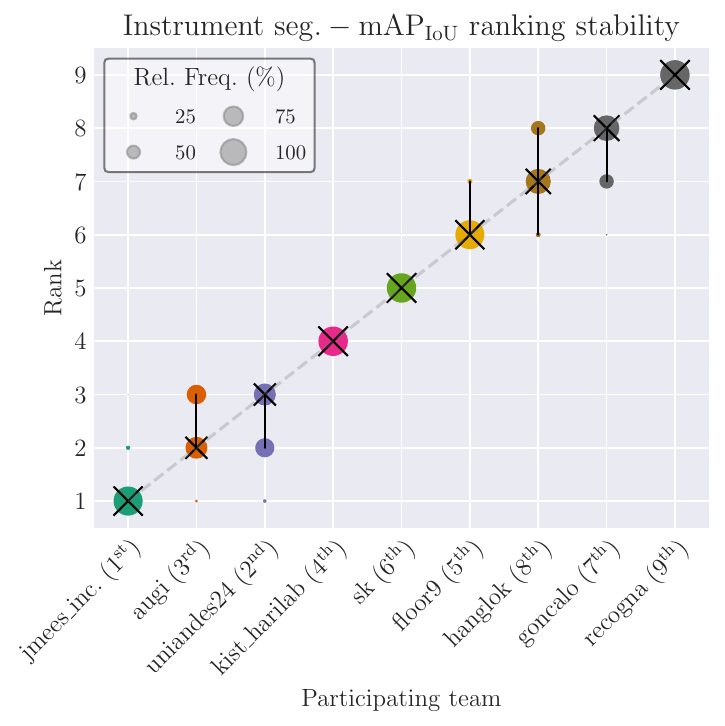}
      \caption{}
      \label{fig:results:instrument_segmentation:ranking_robustness:blobs:auc}
  \end{subfigure}
    \hfill
  \begin{subfigure}{0.49\textwidth}
    \includegraphics[width=\textwidth]{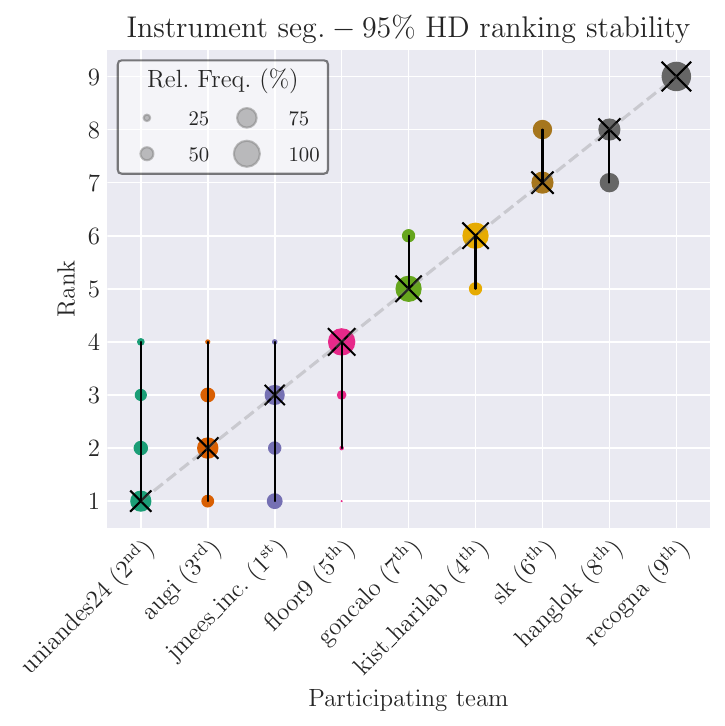}
      \caption{}
      \label{fig:results:instrument_segmentation:ranking_robustness:blobs:hd}
  \end{subfigure} 
      \hfill
  \begin{subfigure}{0.49\textwidth}
    \includegraphics[width=\textwidth]{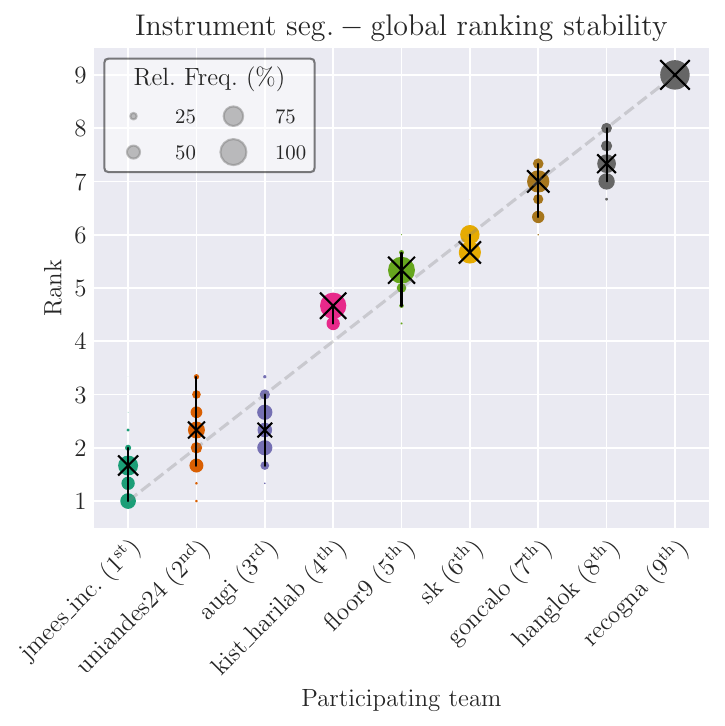}
      \caption{}
      \label{fig:results:instrument_segmentation:ranking_robustness:blobs:global}
  \end{subfigure} 
\caption{
\label{fig:results:instrument_segmentation:ranking_robustness:blobs}%
Visualization of ranking stability based on bootstrapping for the instrument instance segmentation task for the metrics DSC~(\subref{fig:results:instrument_segmentation:ranking_robustness:blobs:dsc}), $\text{mAP}_\text{IoU}$~(\subref{fig:results:instrument_segmentation:ranking_robustness:blobs:auc}), 95\% HD~(\subref{fig:results:instrument_segmentation:ranking_robustness:blobs:hd}), and global~(\subref{fig:results:instrument_segmentation:ranking_robustness:blobs:global}). For each metric, the individual ranking stability is shown and the global rank is indicated after each team name.}
\end{figure*}

Figure~\ref{fig:results:instrument_segmentation:ranking_robustness:blobs} depicts the ranking stability of each participating team across the three evaluation metrics DSC~(\ref{fig:results:instrument_segmentation:ranking_robustness:blobs:dsc}), $\text{mAP}_{\text{IoU}}$~(\ref{fig:results:instrument_segmentation:ranking_robustness:blobs:auc}), 95\% HD~(\ref{fig:results:instrument_segmentation:ranking_robustness:blobs:hd}), as well as the global ranking based on aggregated metrics~(\ref{fig:results:instrument_segmentation:ranking_robustness:blobs:global}).
The results reveal that team rankings are considerably more stable for both the DSC and $\text{mAP}_{\text{IoU}}$ metrics, indicating consistent performance across bootstrap iterations.
In contrast, the rankings for the 95\% HD metric exhibit greater variability, with a higher incidence of outliers and rank fluctuations, reflecting the increased sensitivity of this metric to boundary-level discrepancies.
In the global ranking plot (Figure~\ref{fig:results:instrument_segmentation:ranking_robustness:blobs:global}), a clear ordering of teams is observed.
Notably, the teams, uniandes24 and augi, which ranked second and third overall, demonstrate nearly indistinguishable performance.

\begin{figure*}[tbph!]
\footnotesize
\centering
\renewcommand{\arraystretch}{1.5}
\captionof{table}{Visualization of example participant predictions, together with quantitative results for DSC (in \%) $\uparrow$, $\text{mAP}_{\text{IoU}}$ (in \%) $\uparrow$, and 95\% HD $\downarrow$ below each image. An assignment of the color codings to the respective instrument categories is provided in Section 4.1 in \ref{appendix:data_description}.}
\begin{tabularx}{\textwidth}{m{1.5cm} X X X X }
\toprule
\multicolumn{5}{c}{\textbf{Challenge data}} \\
\midrule
\textbf{Team} & \textbf{Sample 1} & \textbf{Sample 2} & \textbf{Sample 3} & \textbf{Sample 4} \\
\midrule
\multirow{2}{*}{Input} &
\begin{minipage}{0.165\textwidth}
    \centering
    \includegraphics[width=\linewidth]{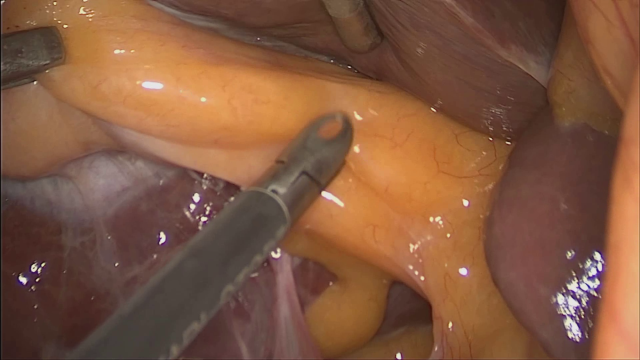}\\
\end{minipage}
&
\begin{minipage}{0.165\textwidth}
    \centering
    \includegraphics[width=\linewidth]{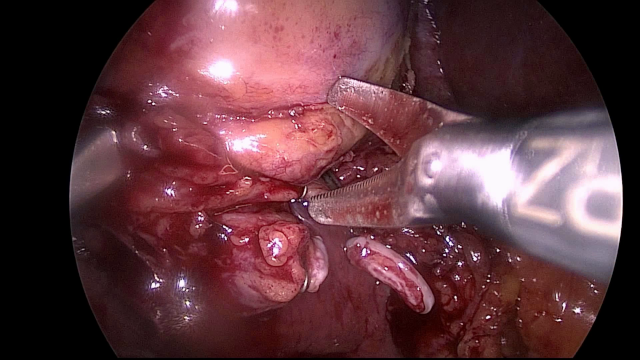}\\
\end{minipage}
&
\begin{minipage}{0.165\textwidth}
    \centering
    \includegraphics[width=\linewidth]{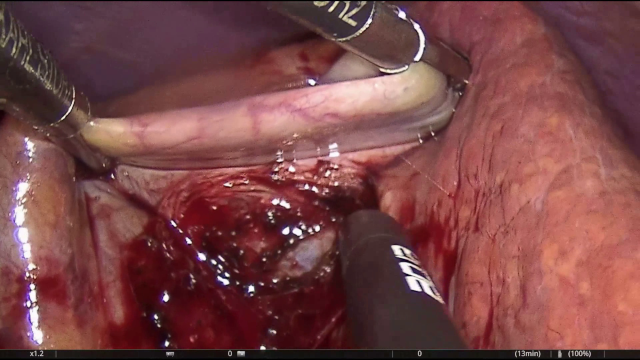}\\
\end{minipage}
&
\begin{minipage}{0.165\textwidth}
    \centering
    \includegraphics[width=\linewidth]{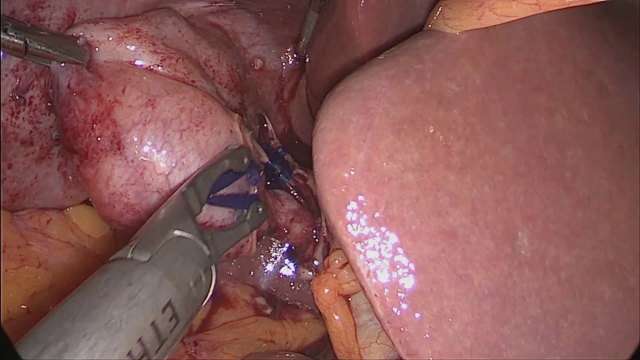}\\
\end{minipage}
\\[1cm]
\multirow{2}{*}{\shortstack[l]{Ground \\ truth}} &
\begin{minipage}{0.165\textwidth}
    \centering
    \includegraphics[width=\linewidth]{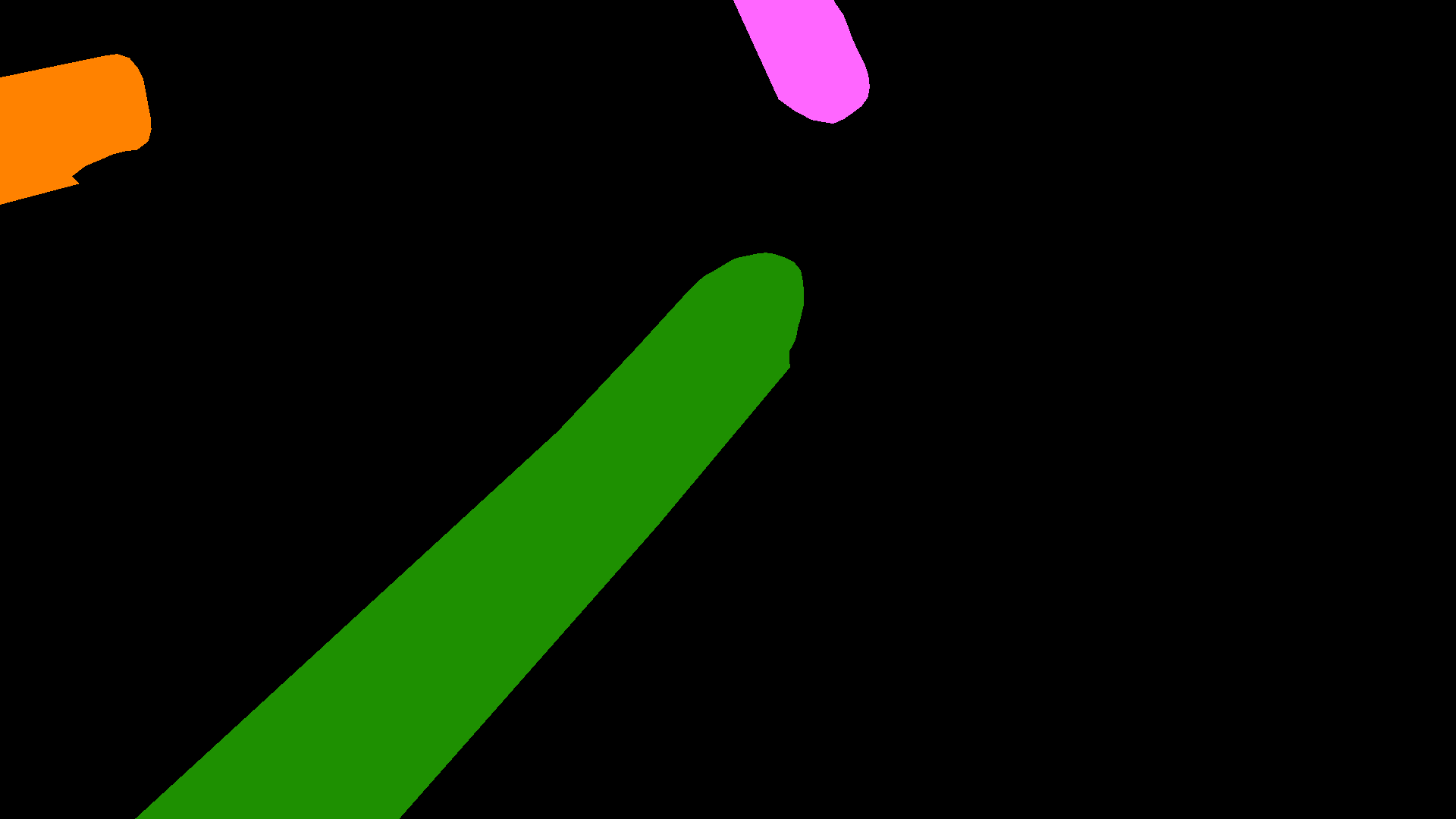}\\
\end{minipage}
&
\begin{minipage}{0.165\textwidth}
    \centering
    \includegraphics[width=\linewidth]{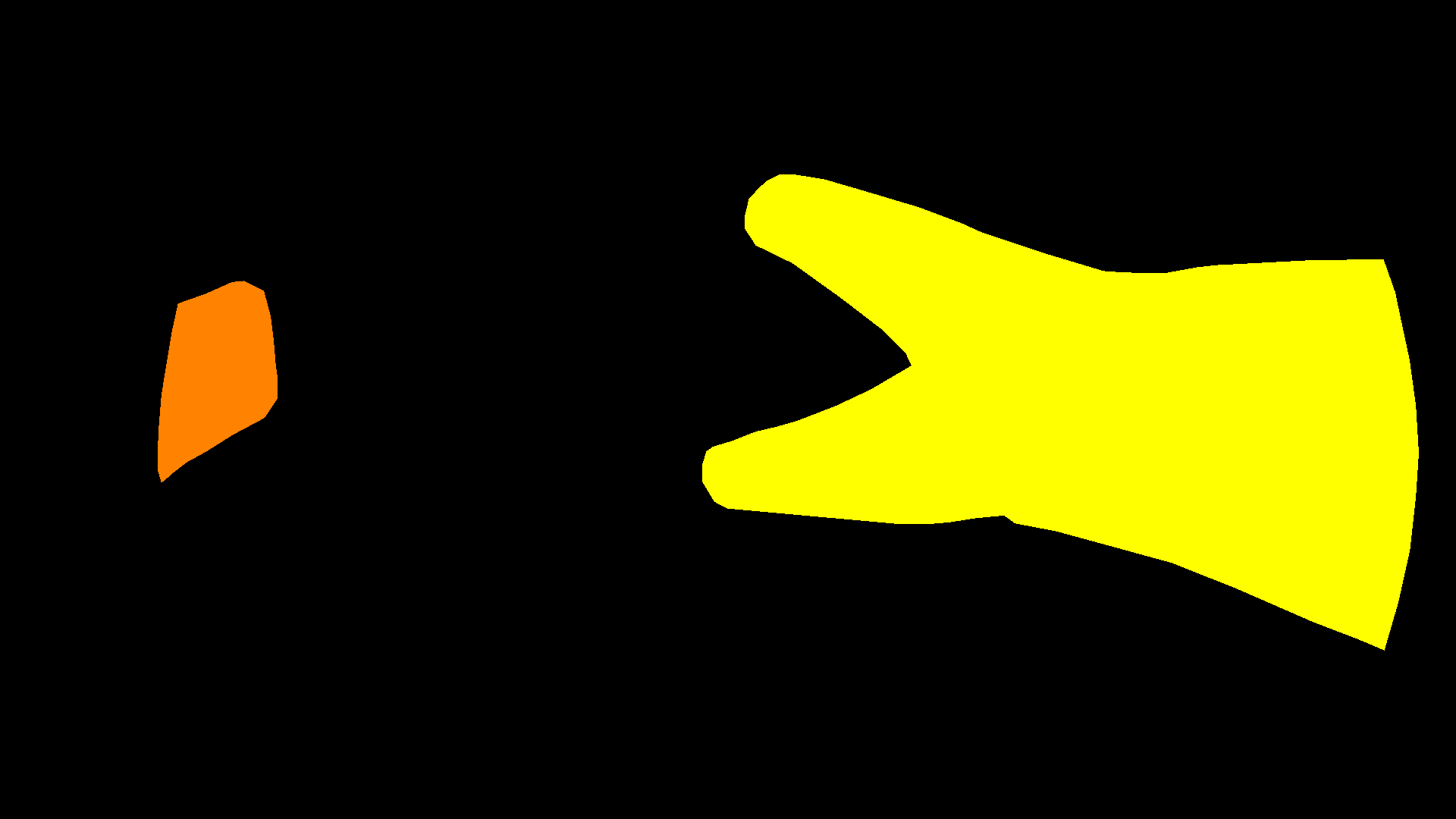}\\
\end{minipage}
&
\begin{minipage}{0.165\textwidth}
    \centering
    \includegraphics[width=\linewidth]{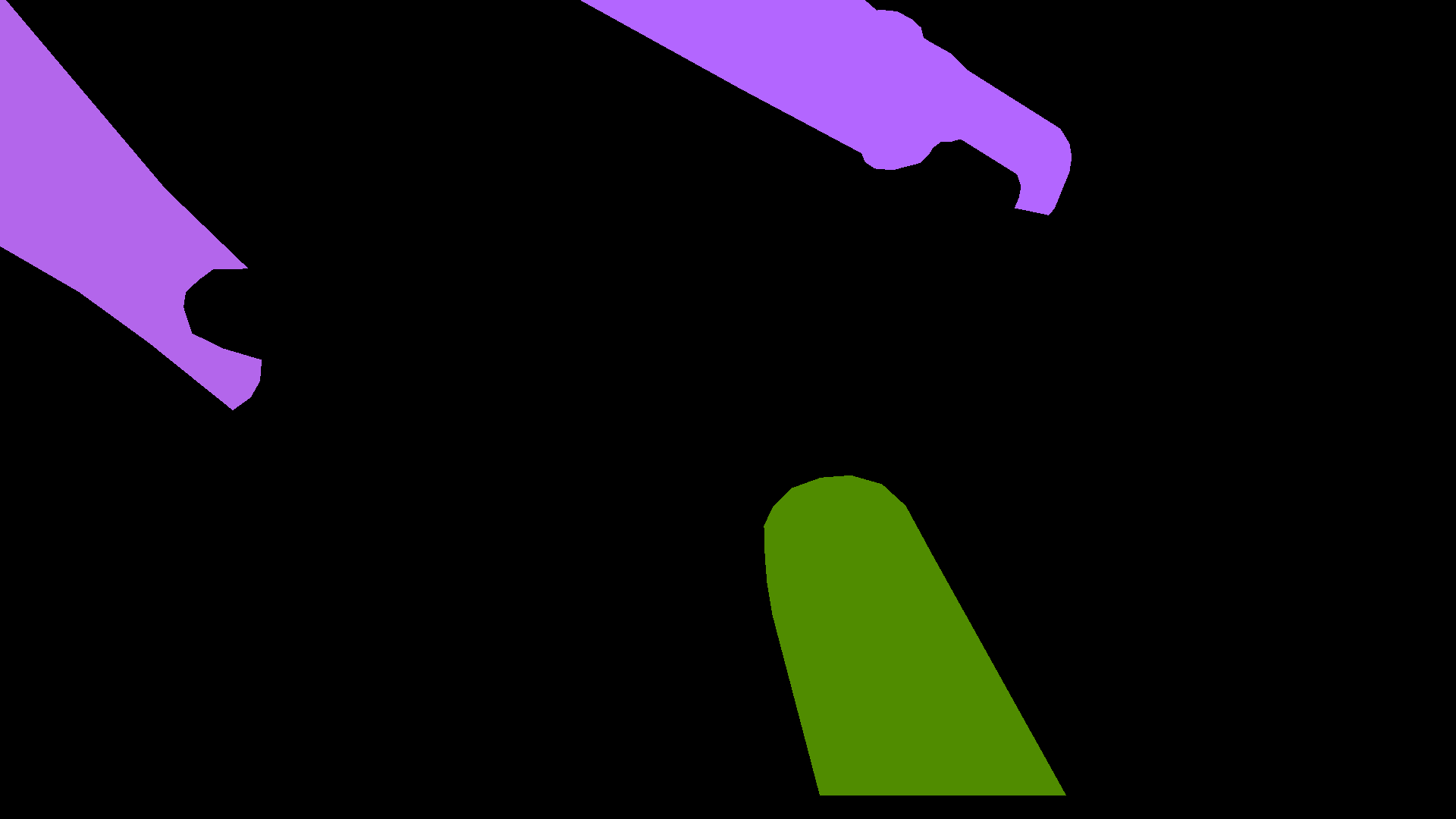}\\
\end{minipage}
&
\begin{minipage}{0.165\textwidth}
    \centering
    \includegraphics[width=\linewidth]{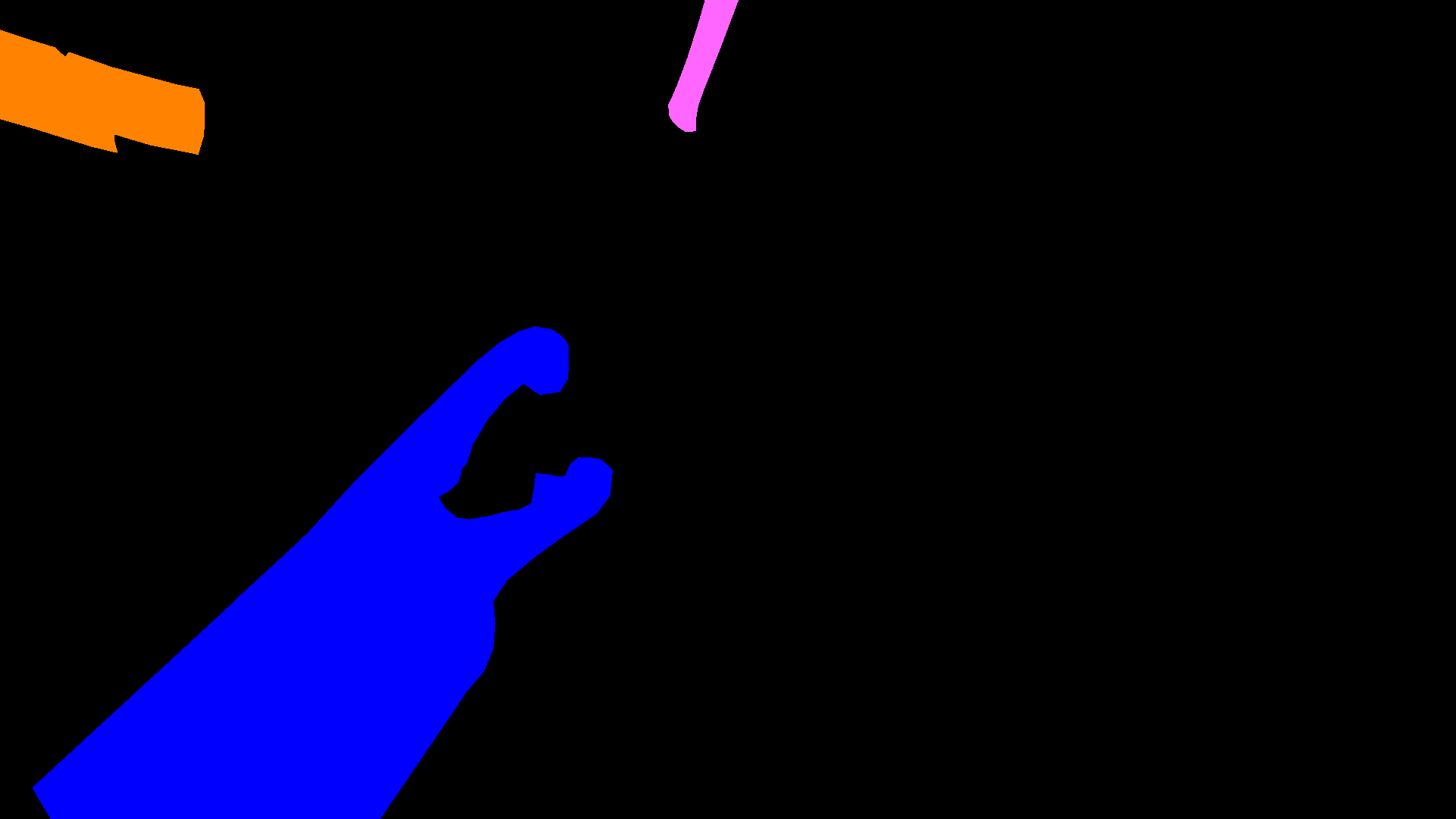}\\
\end{minipage}
\\
\midrule
\multicolumn{5}{c}{\textbf{Participant predictions}} \\
\midrule
\multirow{2}{*}{jmees\_inc.} &
\begin{minipage}{0.165\textwidth}
    \centering
    \includegraphics[width=\linewidth]{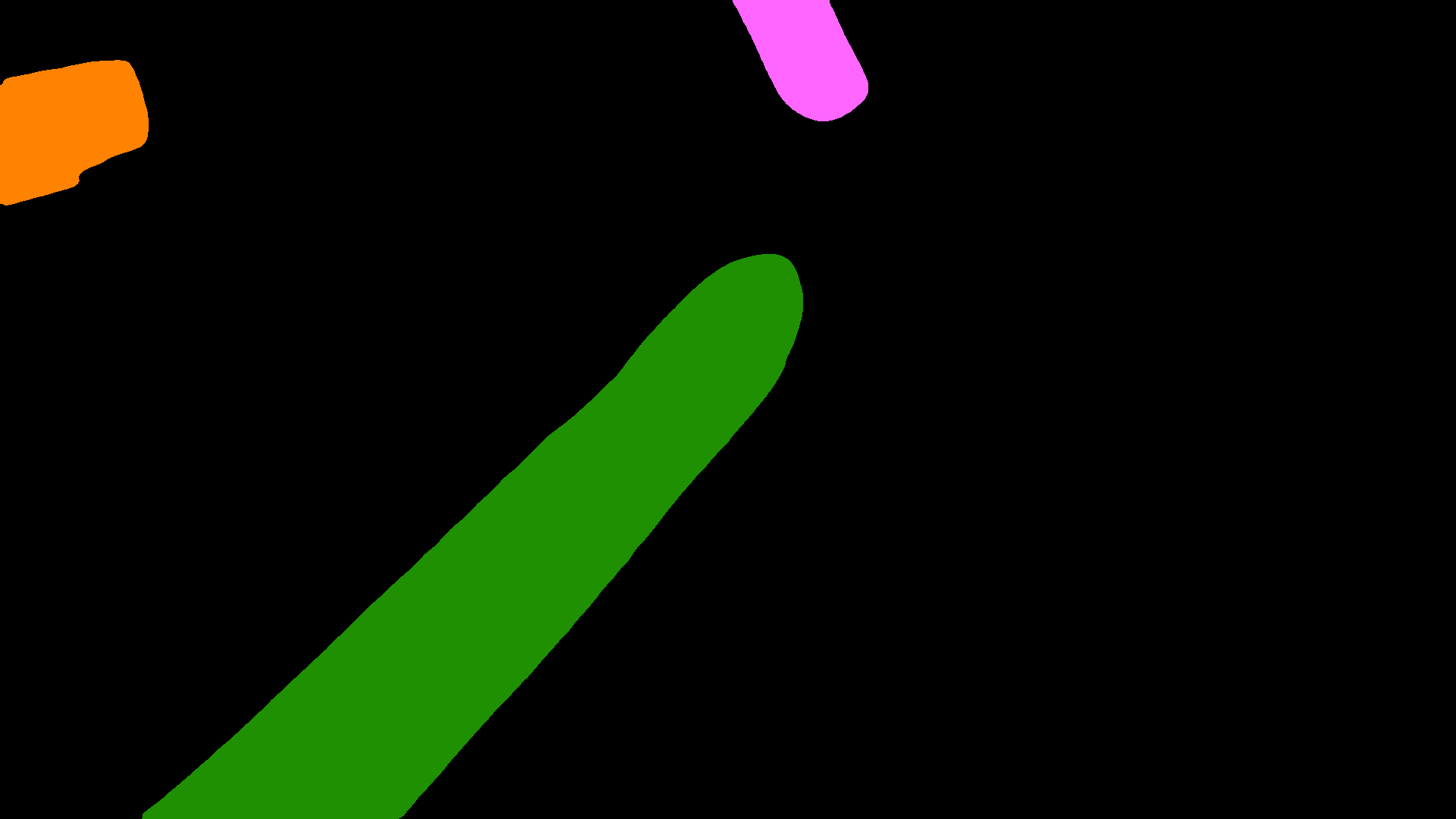}\\[-2pt]
    {\footnotesize 97.30 \(|\) 100.00 \(|\) 7.06}
\end{minipage}
&
\begin{minipage}{0.165\textwidth}
    \centering
    \includegraphics[width=\linewidth]{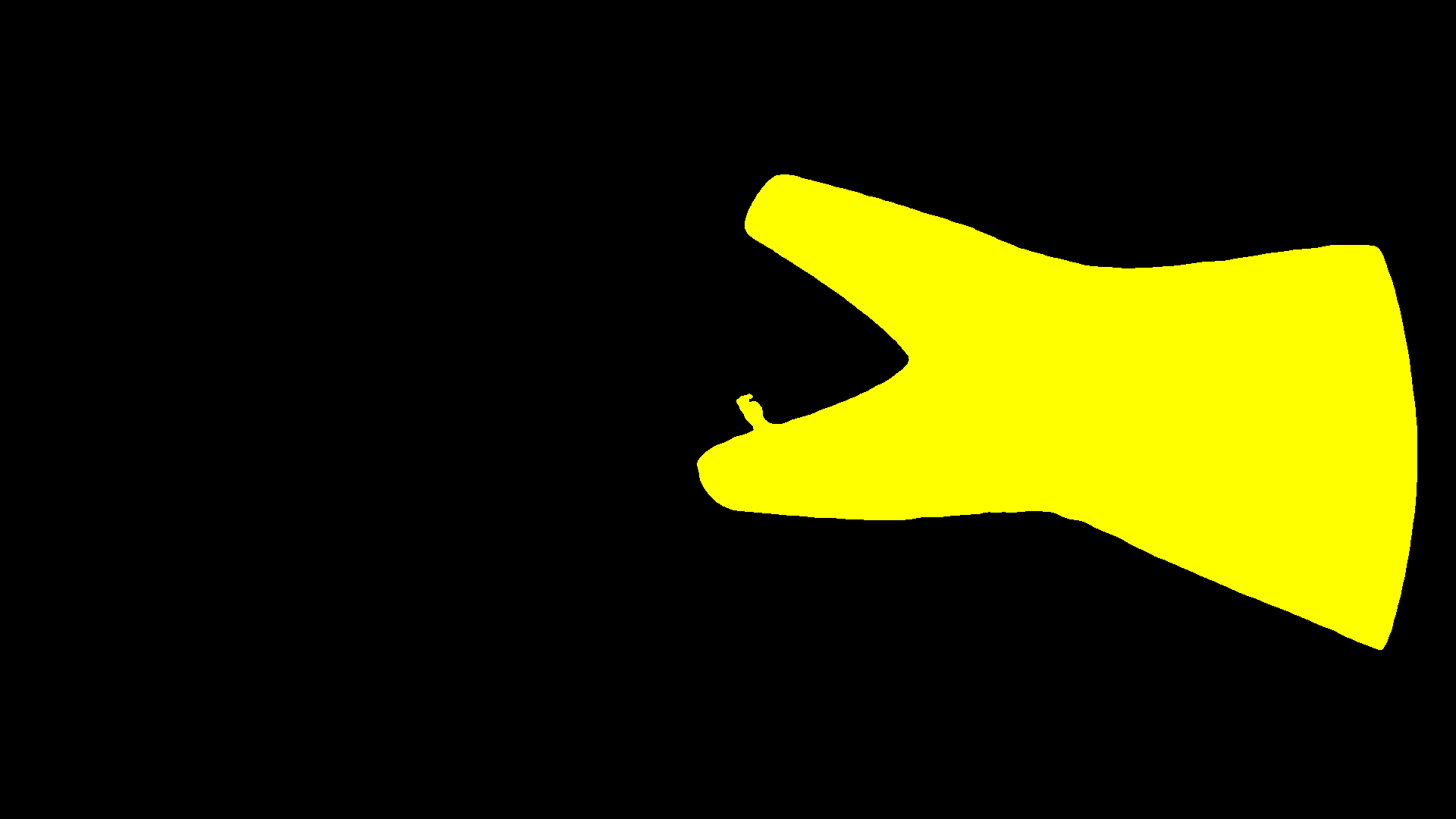}\\[-2pt]
    {\footnotesize 49.01 \(|\) 50.00 \(|\) 19.00}
\end{minipage}
&
\begin{minipage}{0.165\textwidth}
    \centering
    \includegraphics[width=\linewidth]{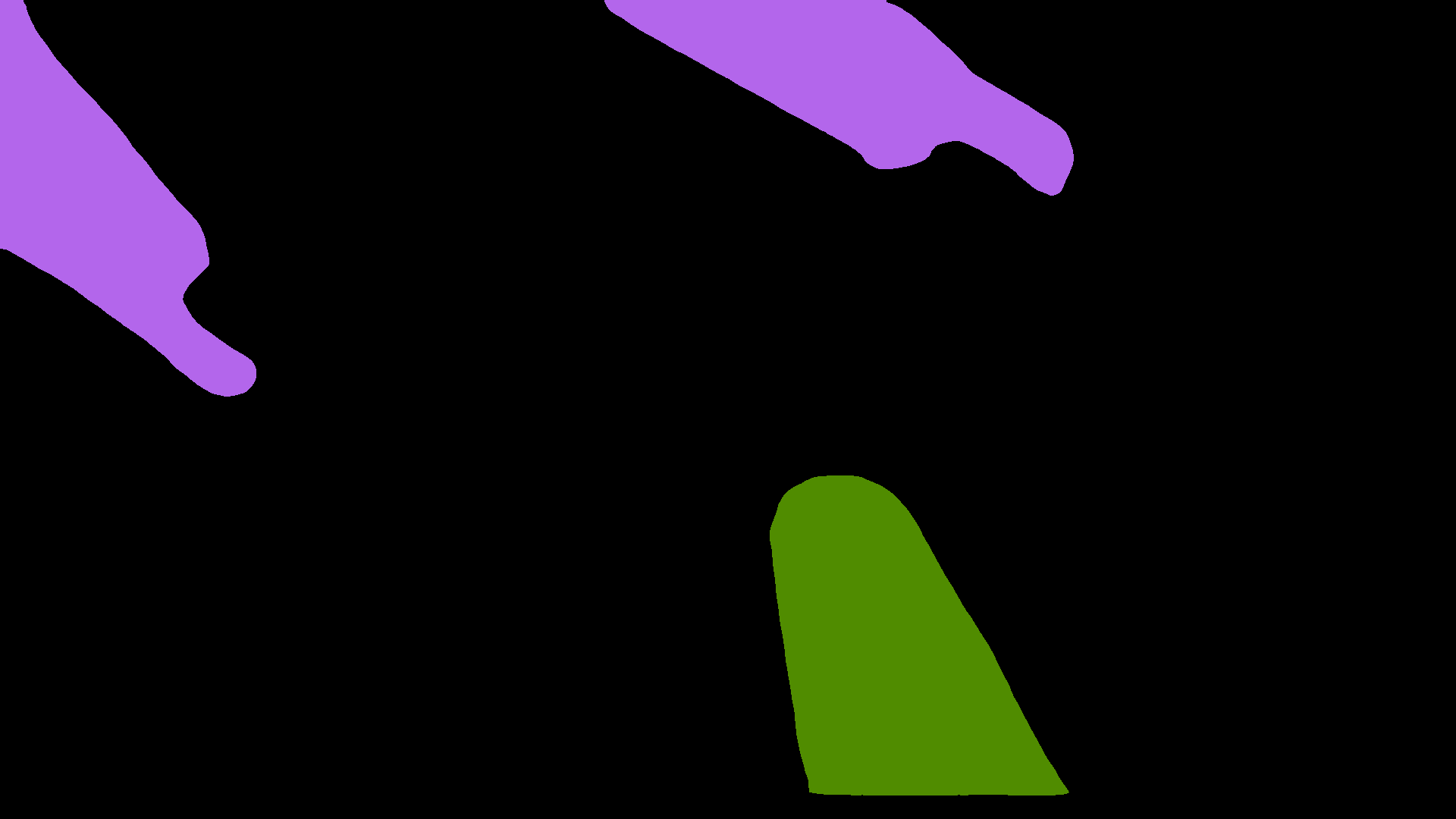}\\[-2pt]
    {\footnotesize 65.06 \(|\) 50.00 \(|\) 530.66}
\end{minipage}
&
\begin{minipage}{0.165\textwidth}
    \centering
    \includegraphics[width=\linewidth]{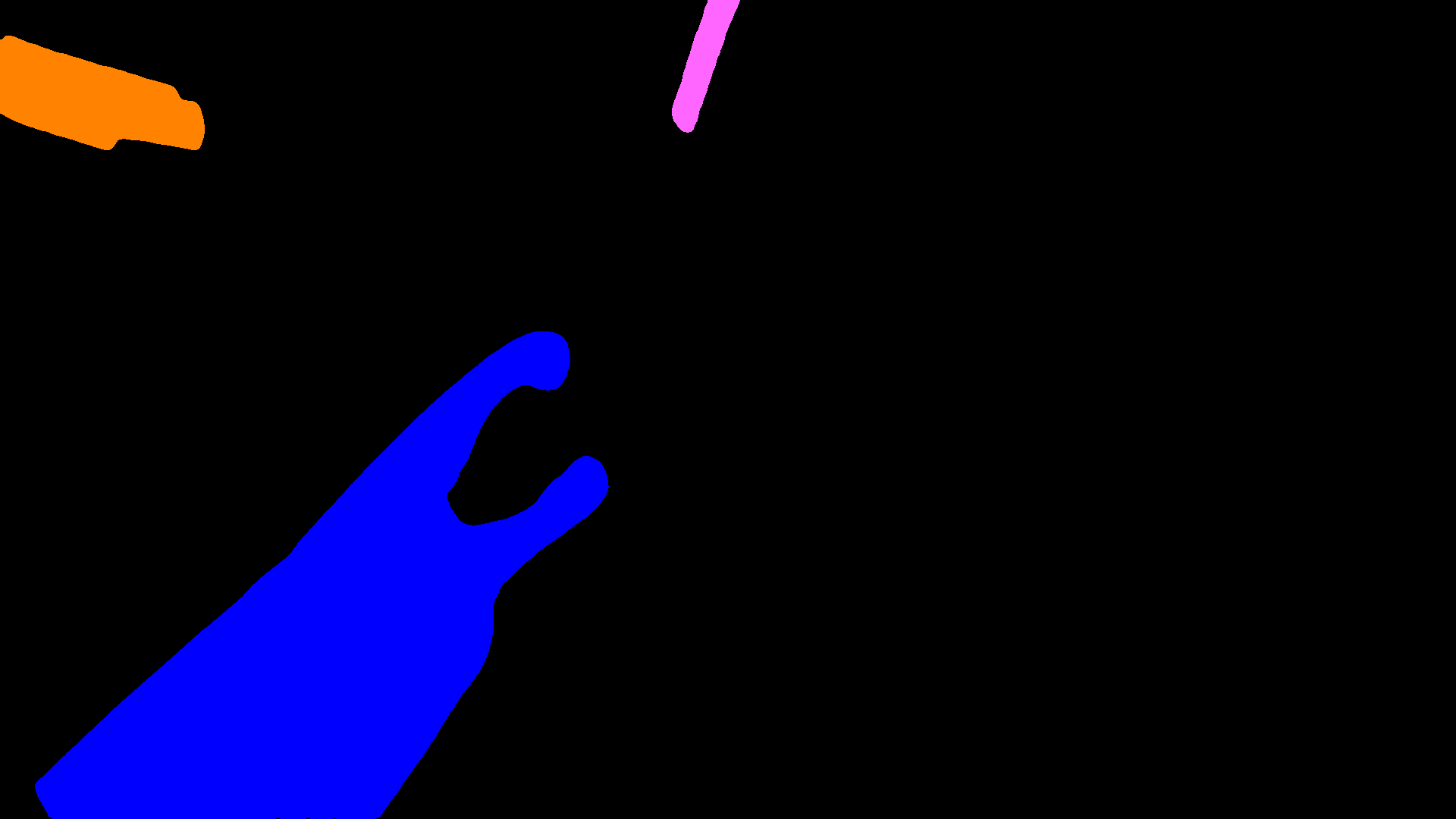}\\[-2pt]
    {\footnotesize 95.87 \(|\) 100.00 \(|\) 8.37}
\end{minipage}
\\
\multirow{2}{*}{uniandes24} &
\begin{minipage}{0.165\textwidth}
    \centering
    \includegraphics[width=\linewidth]{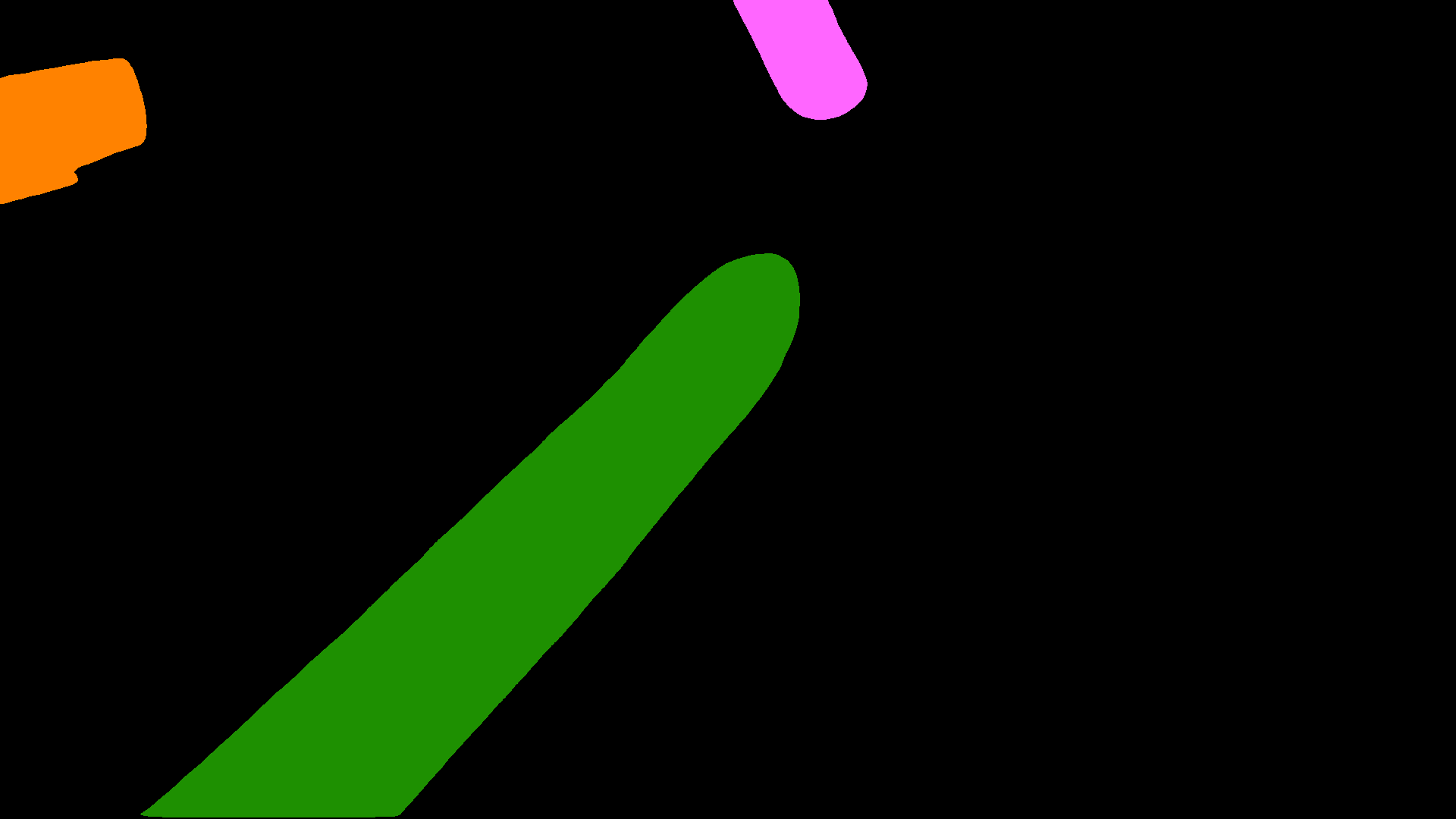}\\[-2pt]
    {\footnotesize 97.17 \(|\) 100.00 \(|\) 10.38}
\end{minipage}
&
\begin{minipage}{0.165\textwidth}
    \centering
    \includegraphics[width=\linewidth]{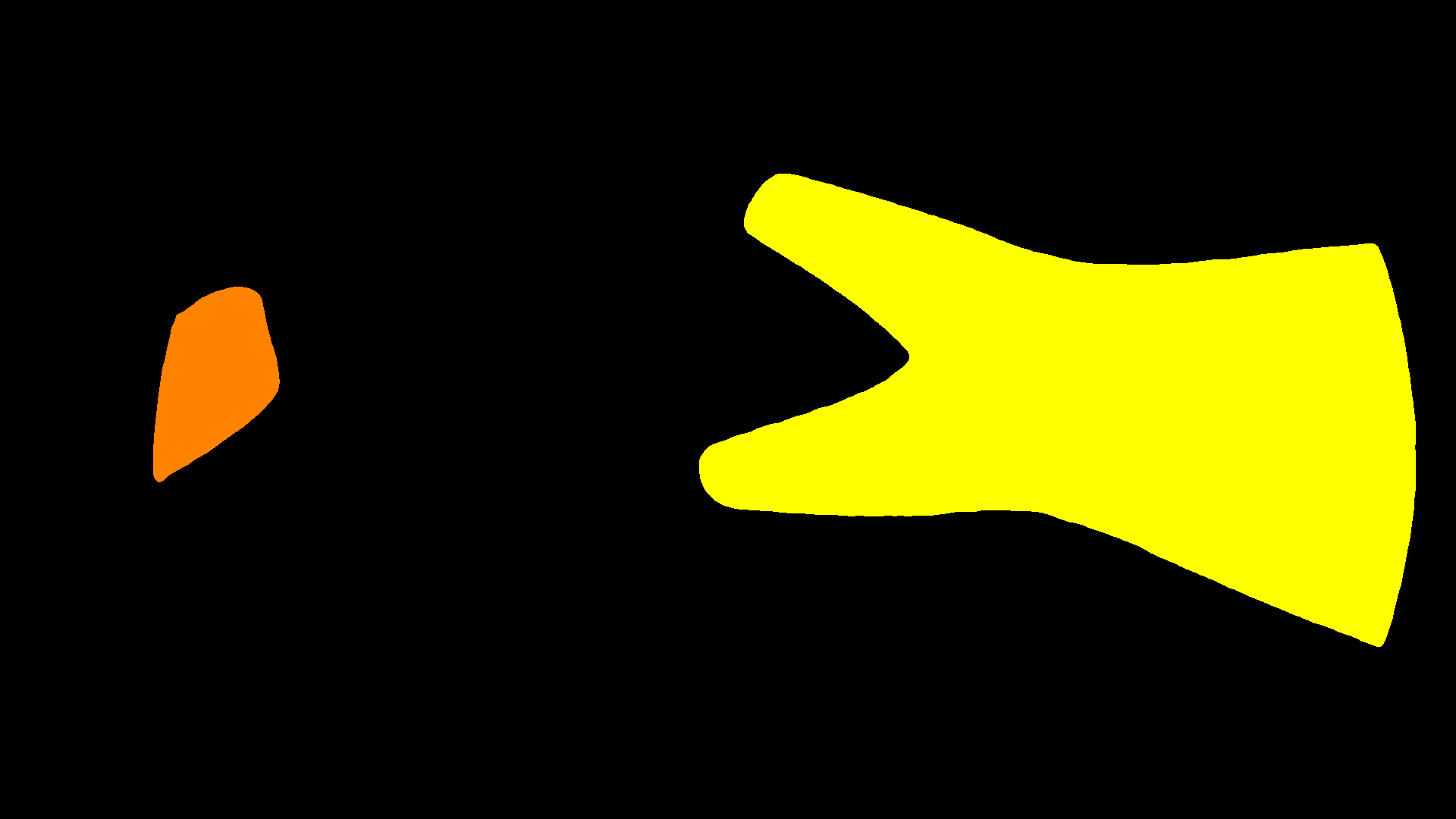}\\[-2pt]
    {\footnotesize 96.51 \(|\) 100.00 \(|\) 11.61}
\end{minipage}
&
\begin{minipage}{0.165\textwidth}
    \centering
    \includegraphics[width=\linewidth]{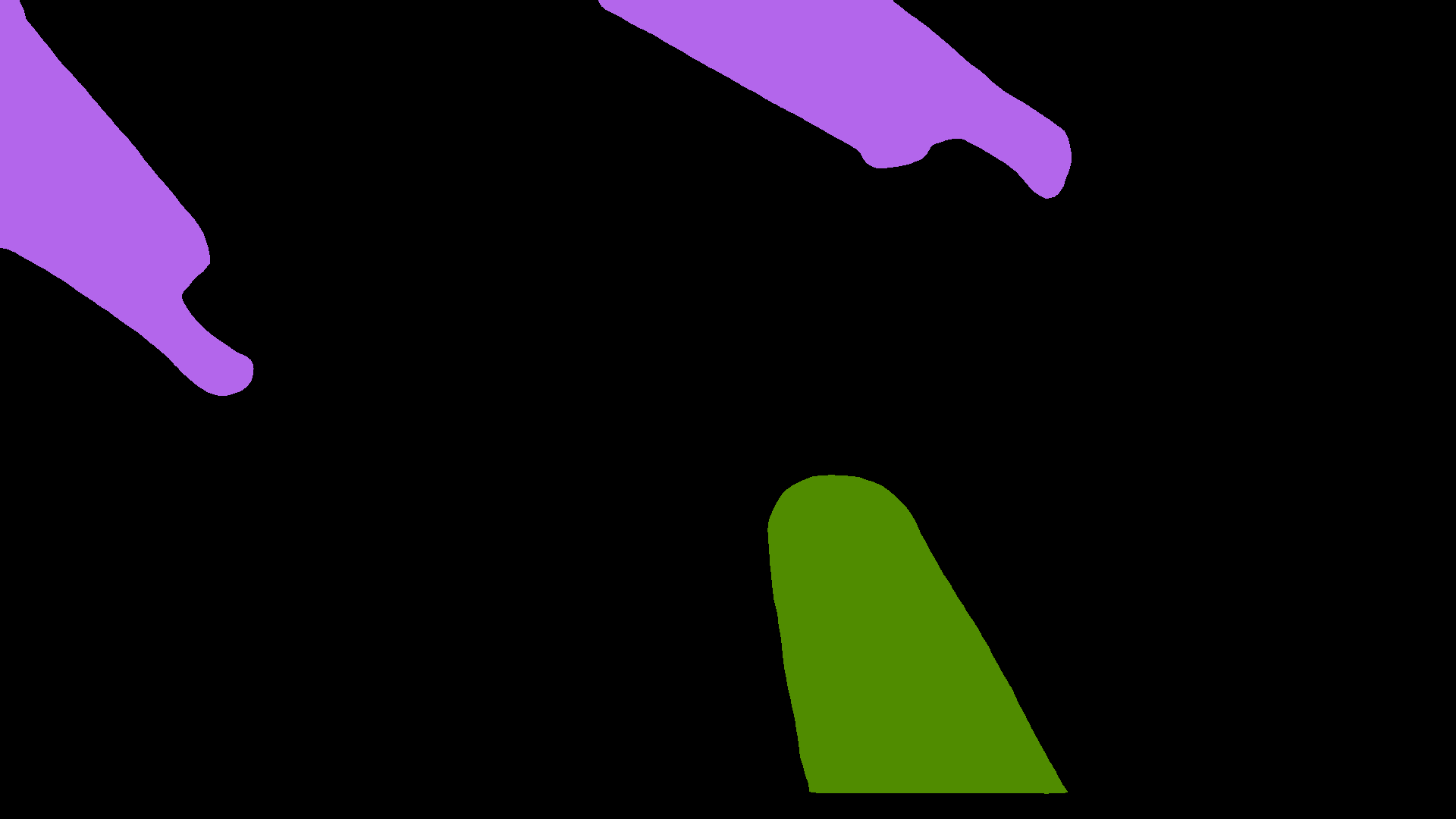}\\[-2pt]
    {\footnotesize 64.90 \(|\) 50.00 \(|\) 529.53}
\end{minipage}
&
\begin{minipage}{0.165\textwidth}
    \centering
    \includegraphics[width=\linewidth]{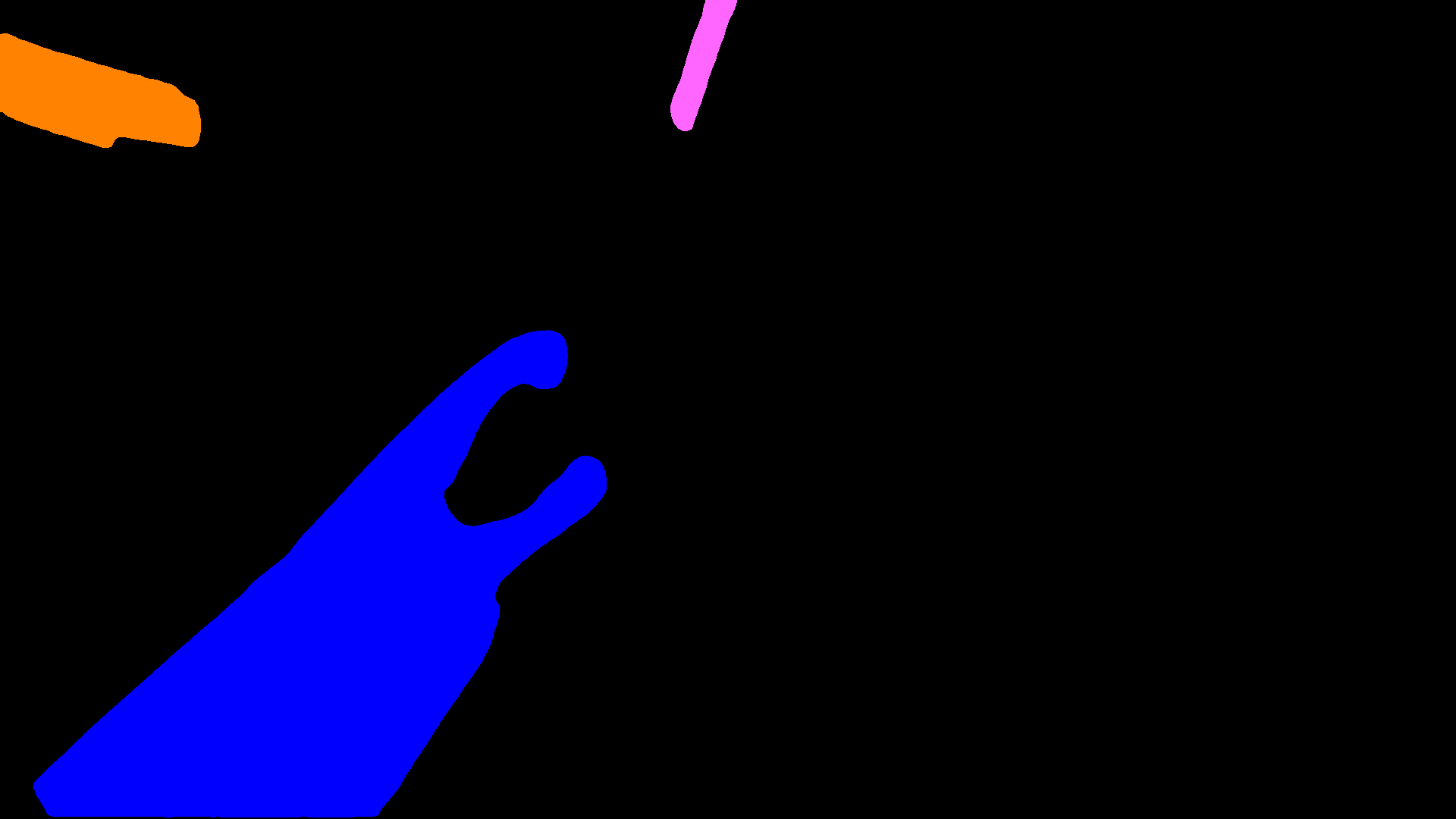}\\[-2pt]
    {\footnotesize 96.27 \(|\) 100.00 \(|\) 31.58}
\end{minipage}
\\
\multirow{2}{*}{augi} &
\begin{minipage}{0.165\textwidth}
    \centering
    \includegraphics[width=\linewidth]{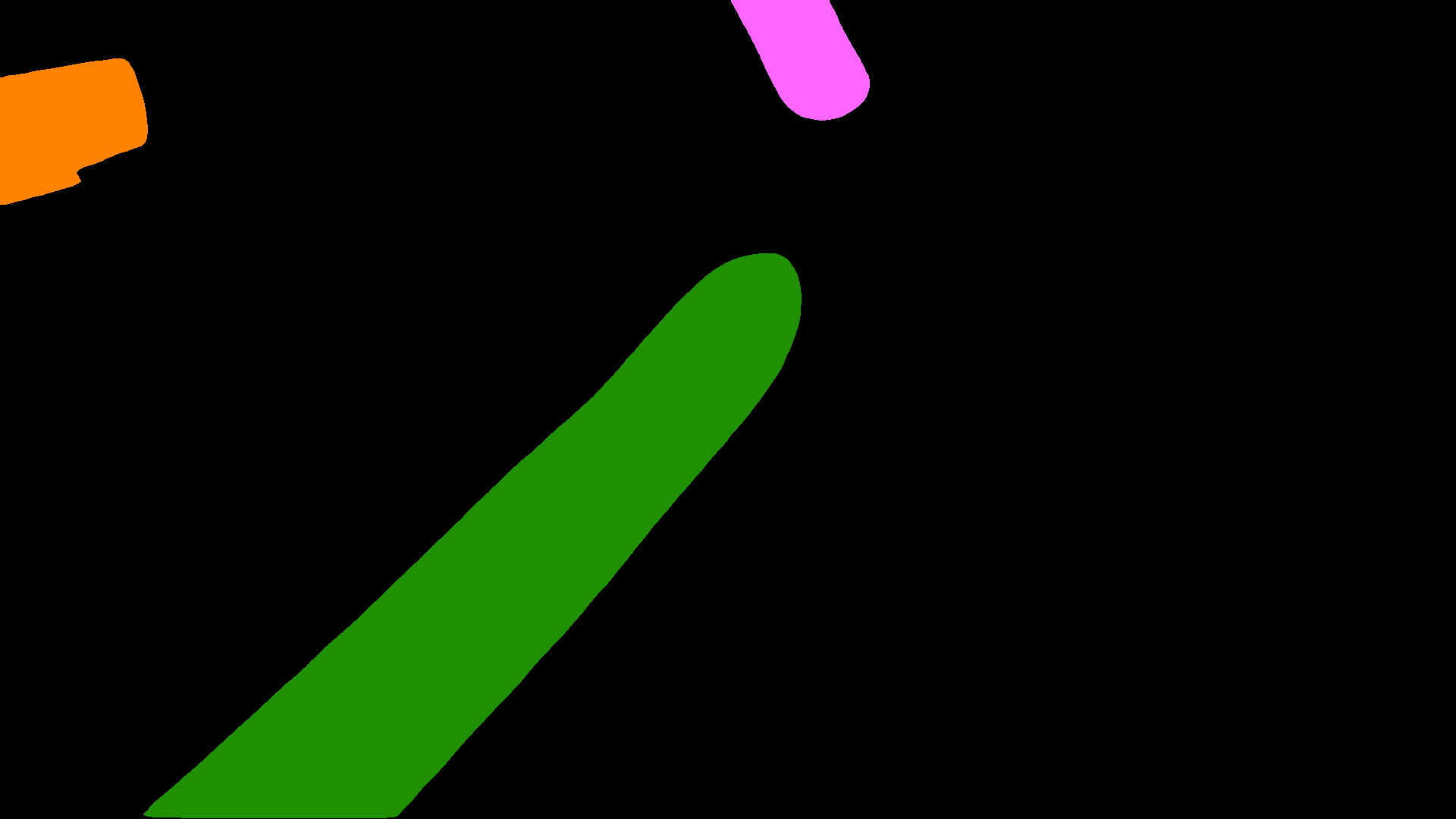}\\[-2pt]
    {\footnotesize 97.76 \(|\) 100.00 \(|\) 9.64}
\end{minipage}
&
\begin{minipage}{0.165\textwidth}
    \centering
    \includegraphics[width=\linewidth]{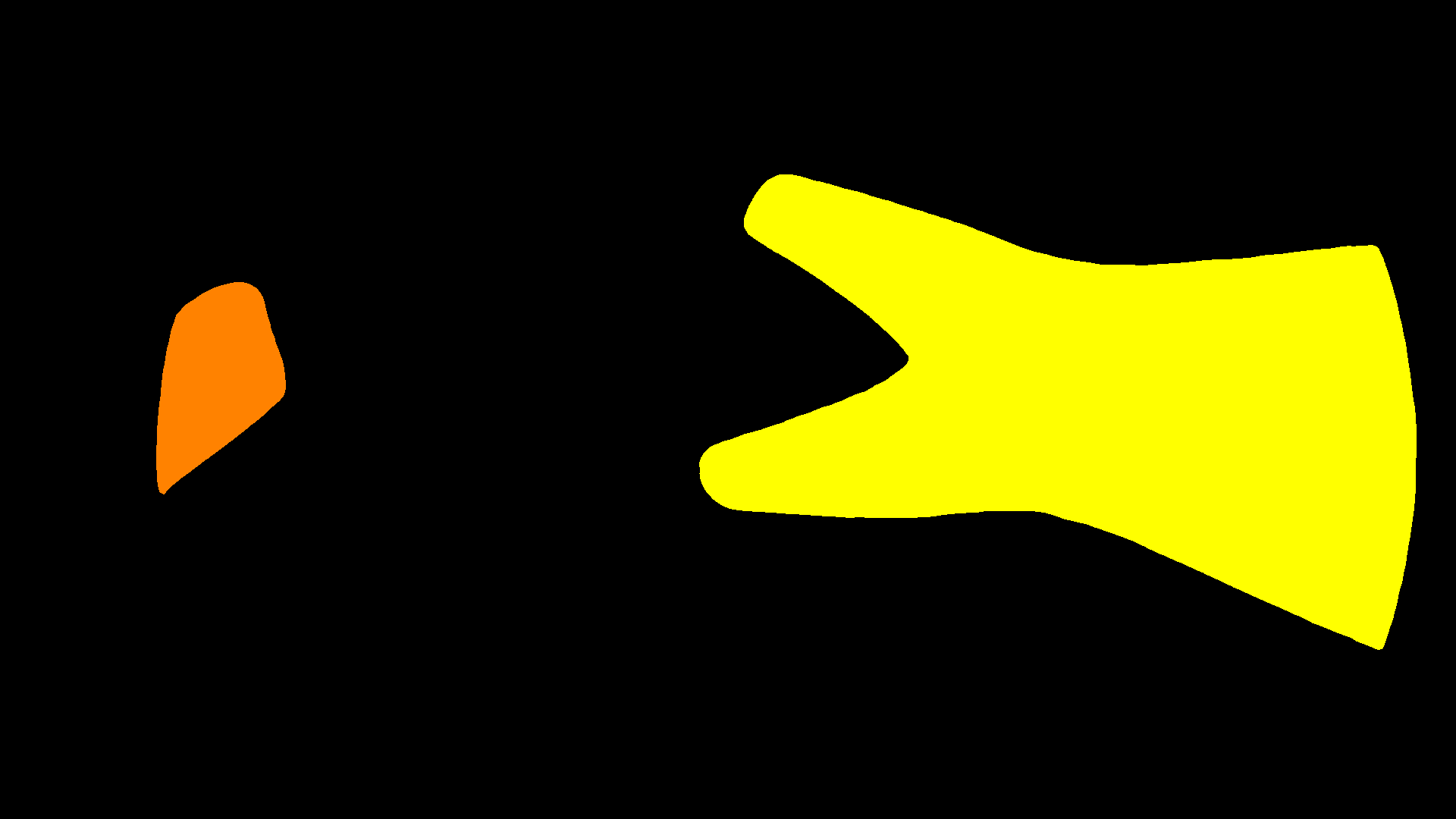}\\[-2pt]
    {\footnotesize 96.50 \(|\) 100.00 \(|\) 14.52}
\end{minipage}
&
\begin{minipage}{0.165\textwidth}
    \centering
    \includegraphics[width=\linewidth]{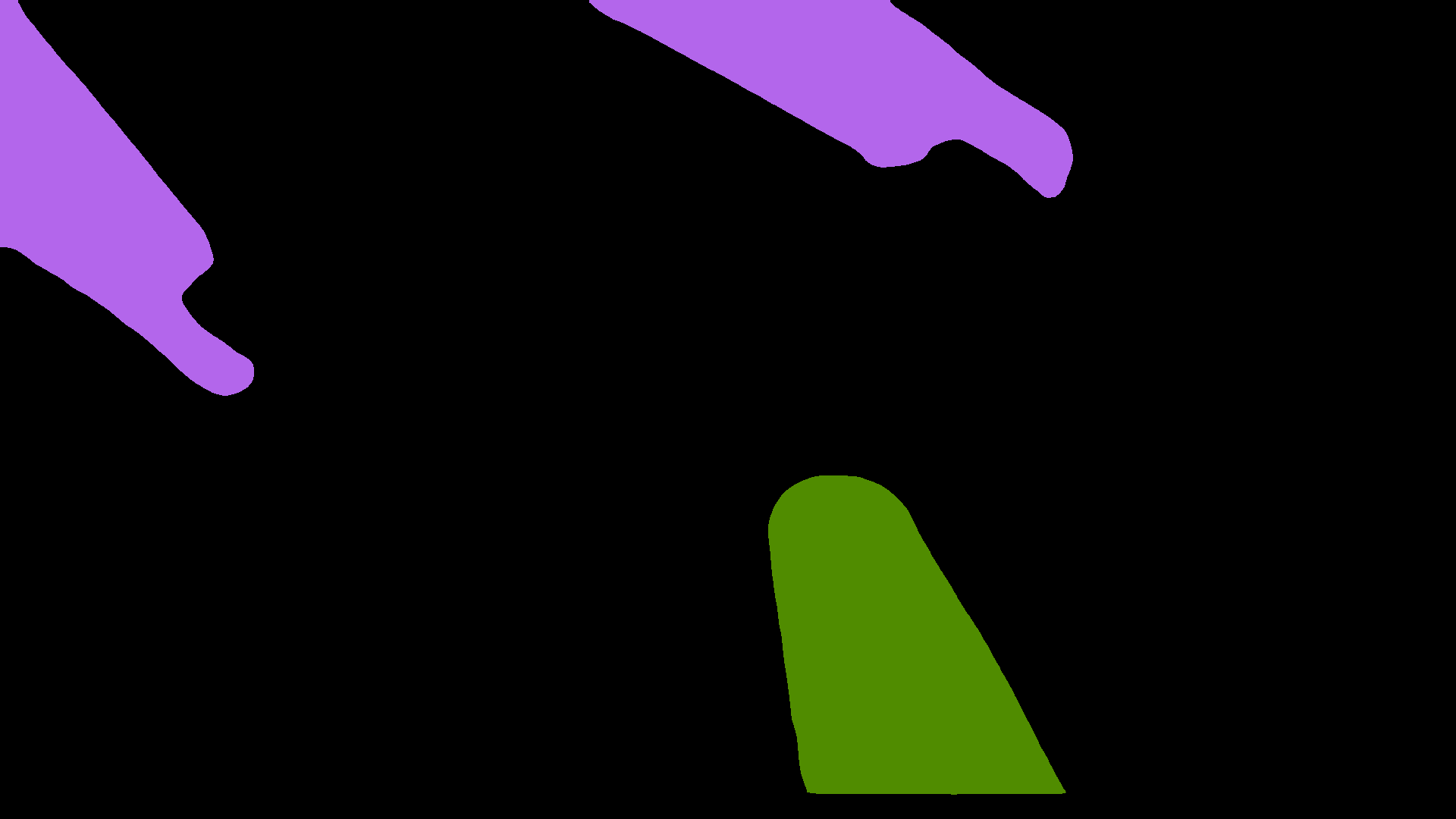}\\[-2pt]
    {\footnotesize 64.97 \(|\) 50.00 \(|\) 377.58}
\end{minipage}
&
\begin{minipage}{0.165\textwidth}
    \centering
    \includegraphics[width=\linewidth]{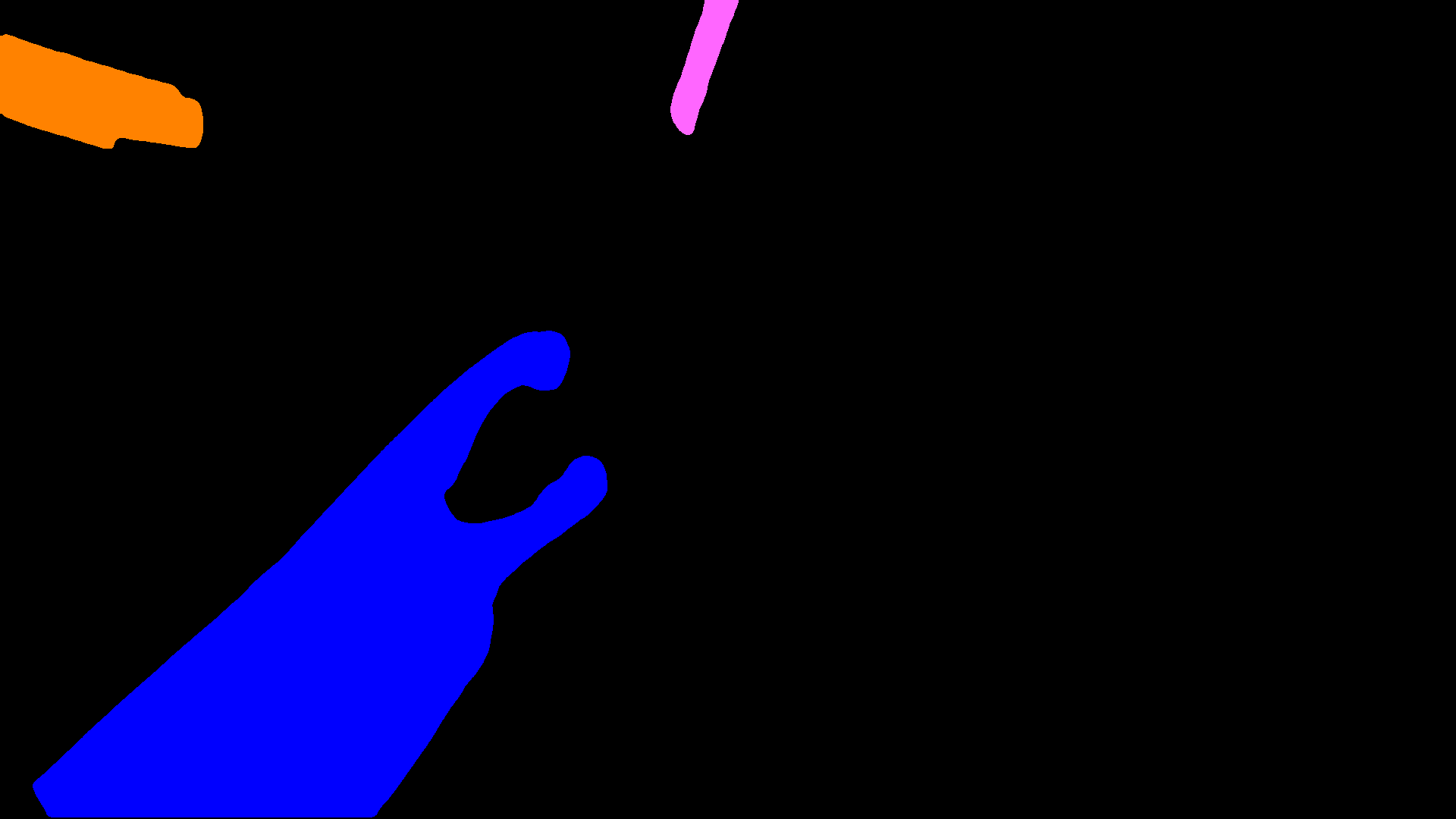}\\[-2pt]
    {\footnotesize 96.61 \(|\) 100.00 \(|\) 31.37}
\end{minipage}
\\
\multirow{2}{*}{kist\_harilab} &
\begin{minipage}{0.165\textwidth}
    \centering
    \includegraphics[width=\linewidth]{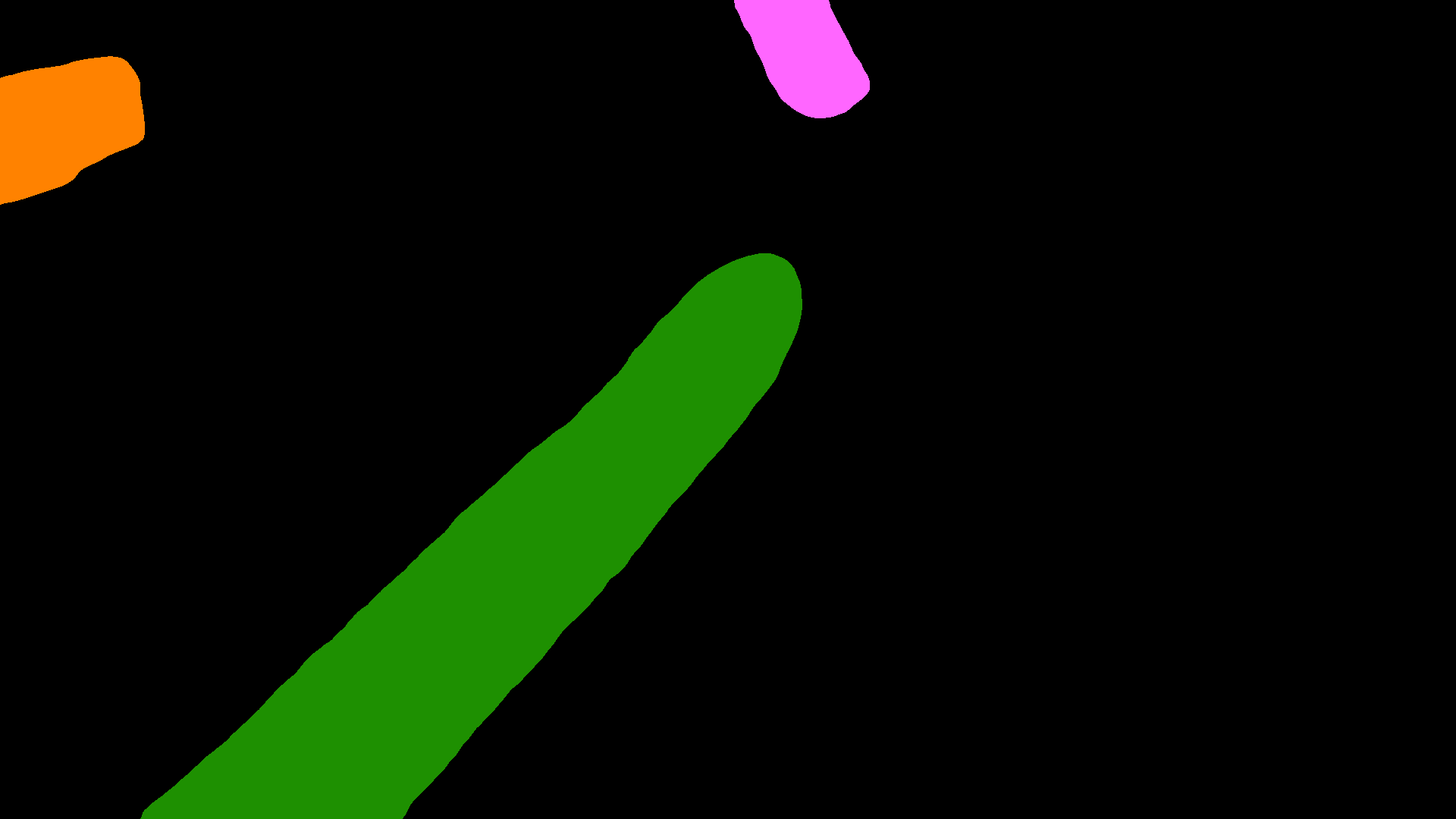}\\[-2pt]
    {\footnotesize 97.44 \(|\) 100.00 \(|\) 7.31}
\end{minipage}
&
\begin{minipage}{0.165\textwidth}
    \centering
    \includegraphics[width=\linewidth]{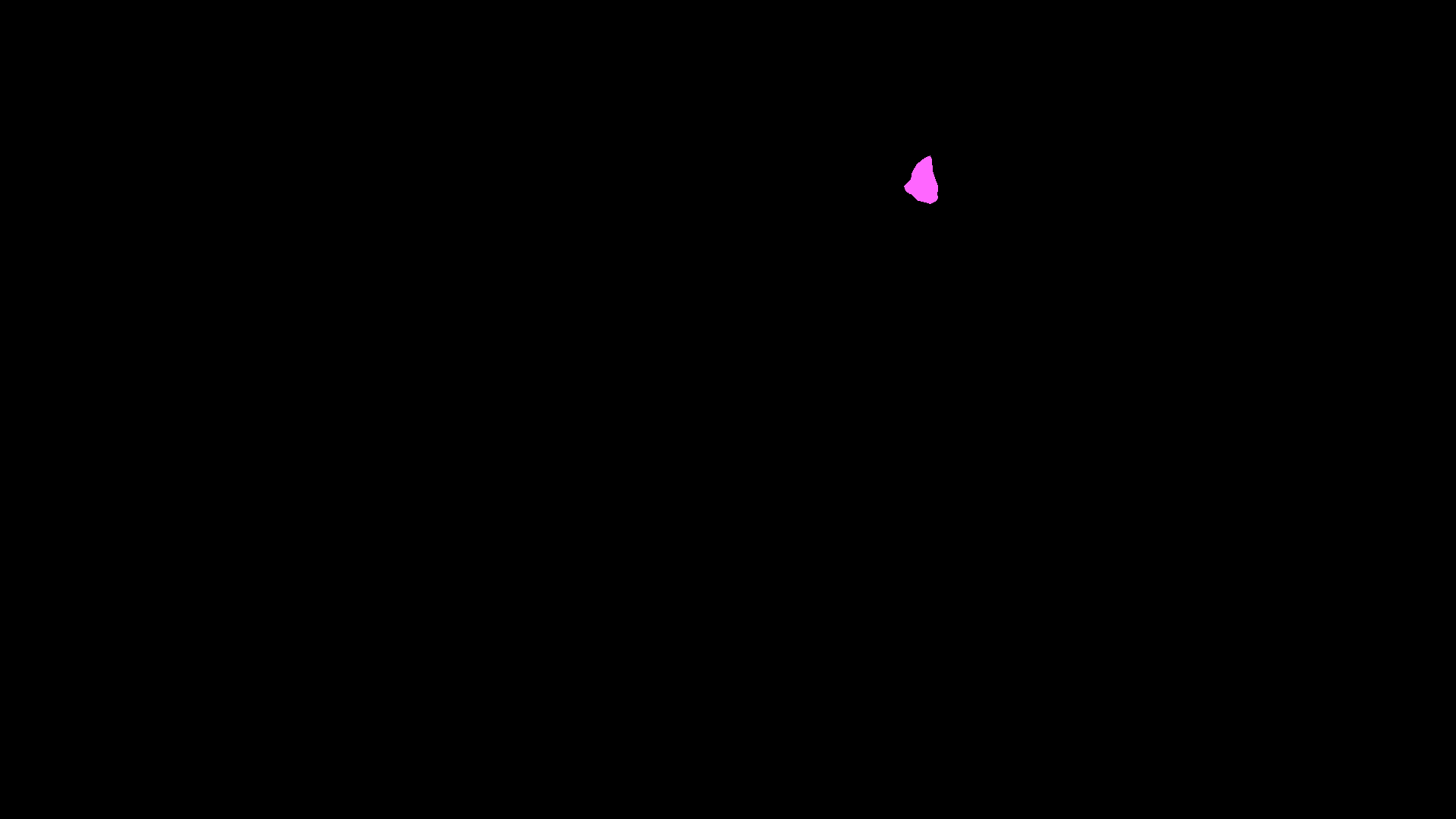}\\[-2pt]
    {\footnotesize 0.00 \(|\) 0.00 \(|\) 1045.36}
\end{minipage}
&
\begin{minipage}{0.165\textwidth}
    \centering
    \includegraphics[width=\linewidth]{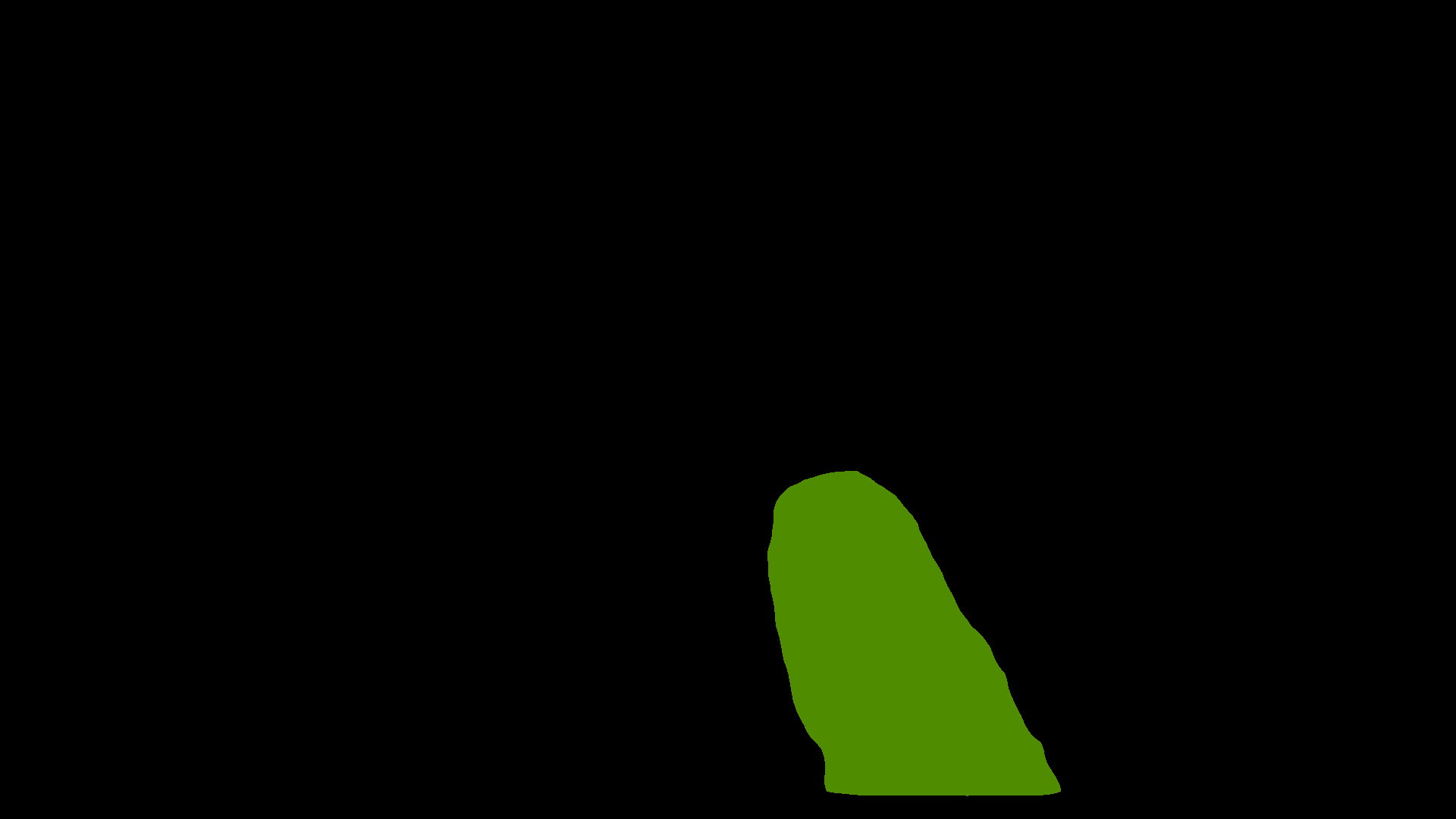}\\[-2pt]
    {\footnotesize 49.07 \(|\) 50.00 \(|\) 10.00}
\end{minipage}
&
\begin{minipage}{0.165\textwidth}
    \centering
    \includegraphics[width=\linewidth]{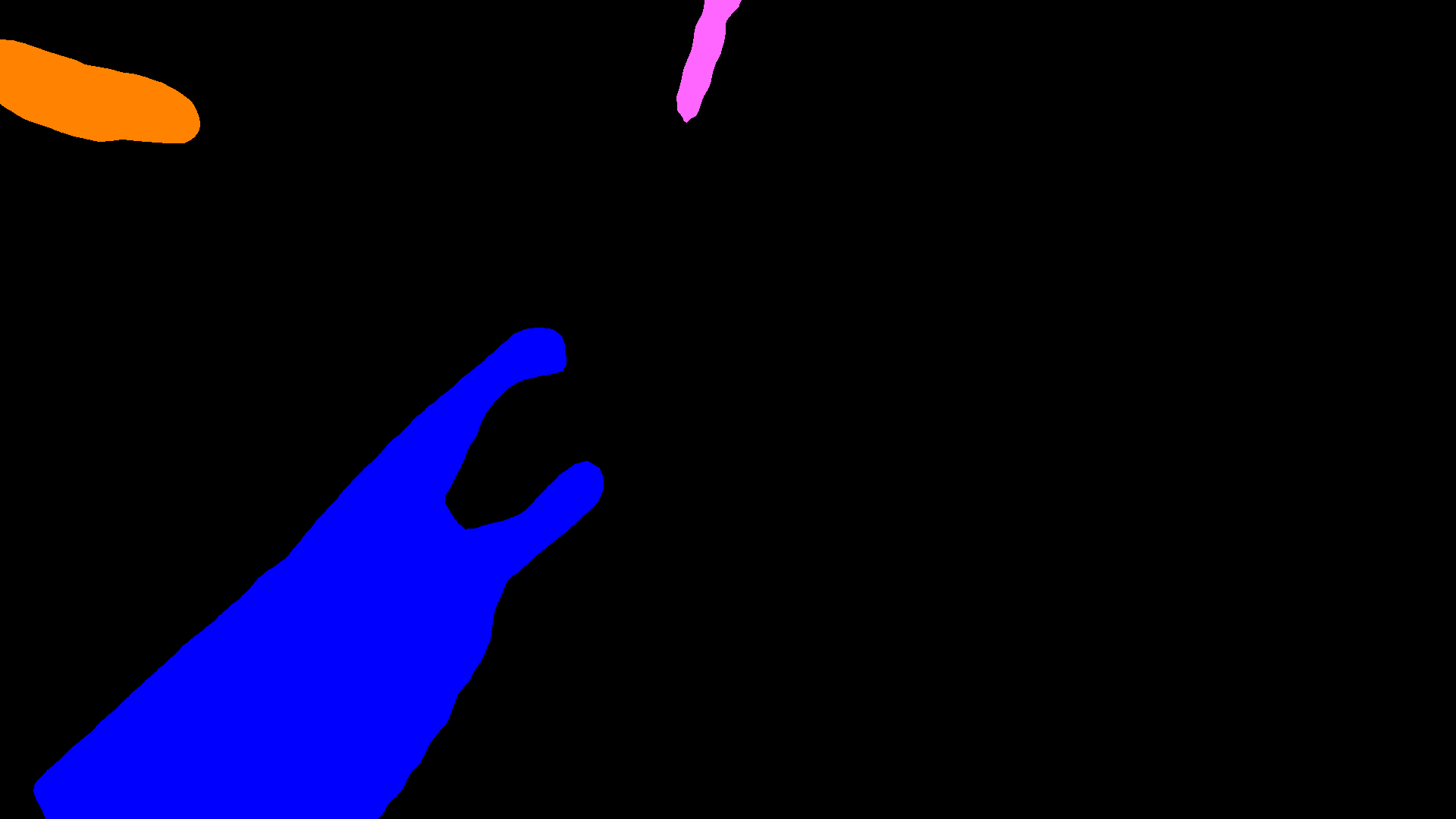}\\[-2pt]
    {\footnotesize 92.49 \(|\) 100.00 \(|\) 13.82}
\end{minipage}
\\
\multirow{2}{*}{floor9} &
\begin{minipage}{0.165\textwidth}
    \centering
    \includegraphics[width=\linewidth]{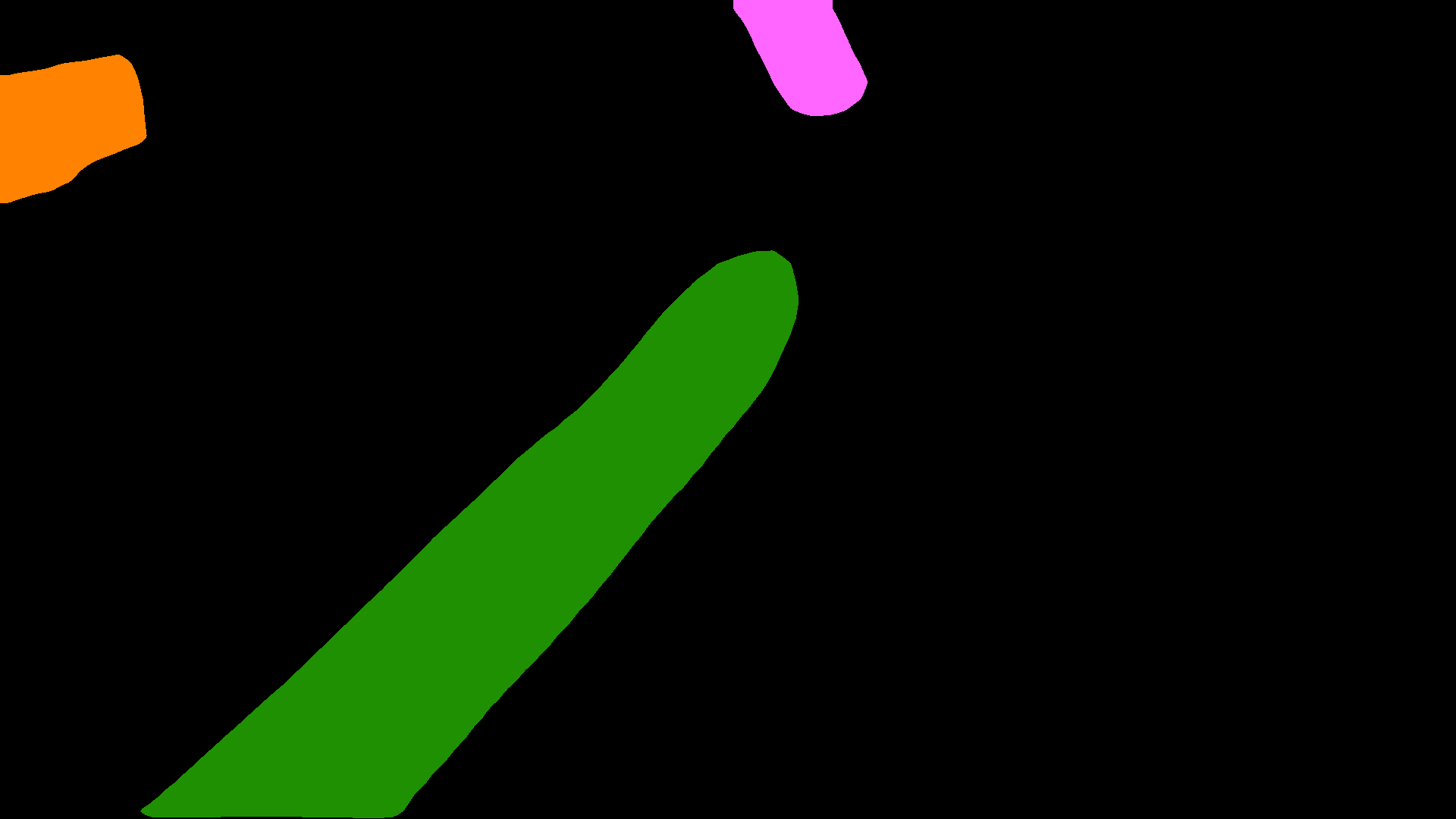}\\[-2pt]
    {\footnotesize 96.70 \(|\) 100.00 \(|\) 10.91}
\end{minipage}
&
\begin{minipage}{0.165\textwidth}
    \centering
    \includegraphics[width=\linewidth]{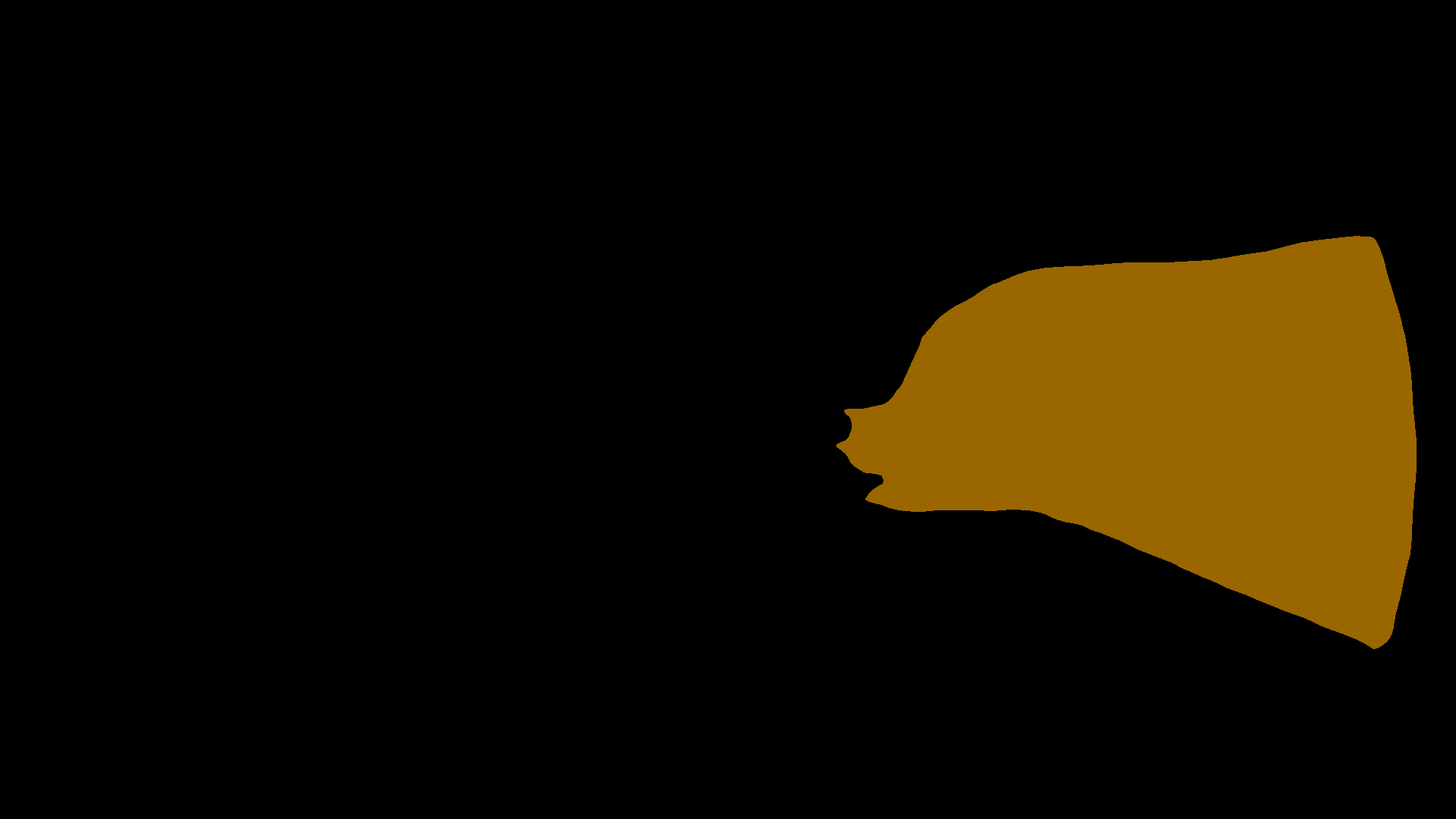}\\[-2pt]
    {\footnotesize 0.00 \(|\) 0.00 \(|\) 208.81}
\end{minipage}
&
\begin{minipage}{0.165\textwidth}
    \centering
    \includegraphics[width=\linewidth]{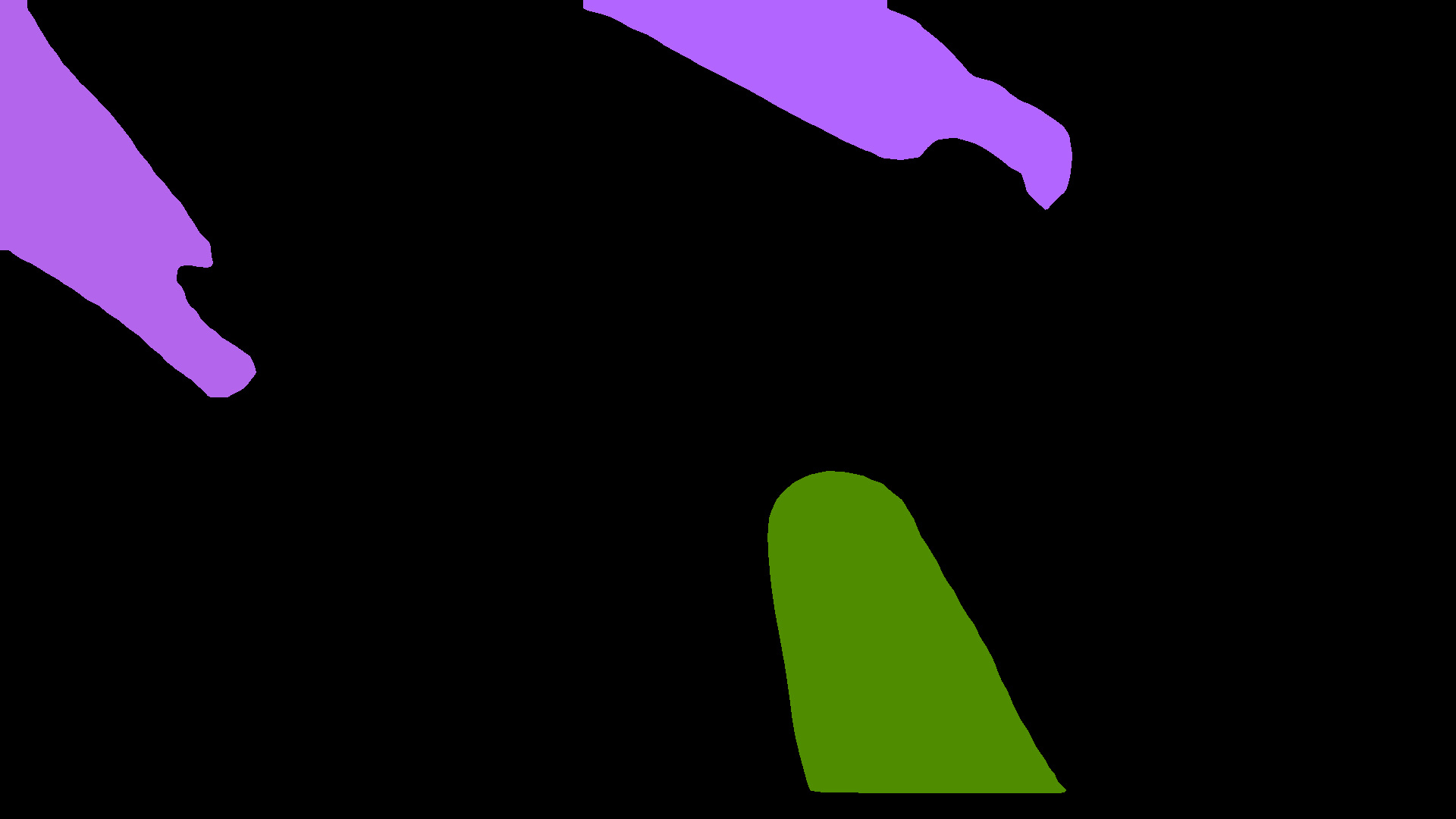}\\[-2pt]
    {\footnotesize 96.65 \(|\) 100.00 \(|\) 12.52}
\end{minipage}
&
\begin{minipage}{0.165\textwidth}
    \centering
    \includegraphics[width=\linewidth]{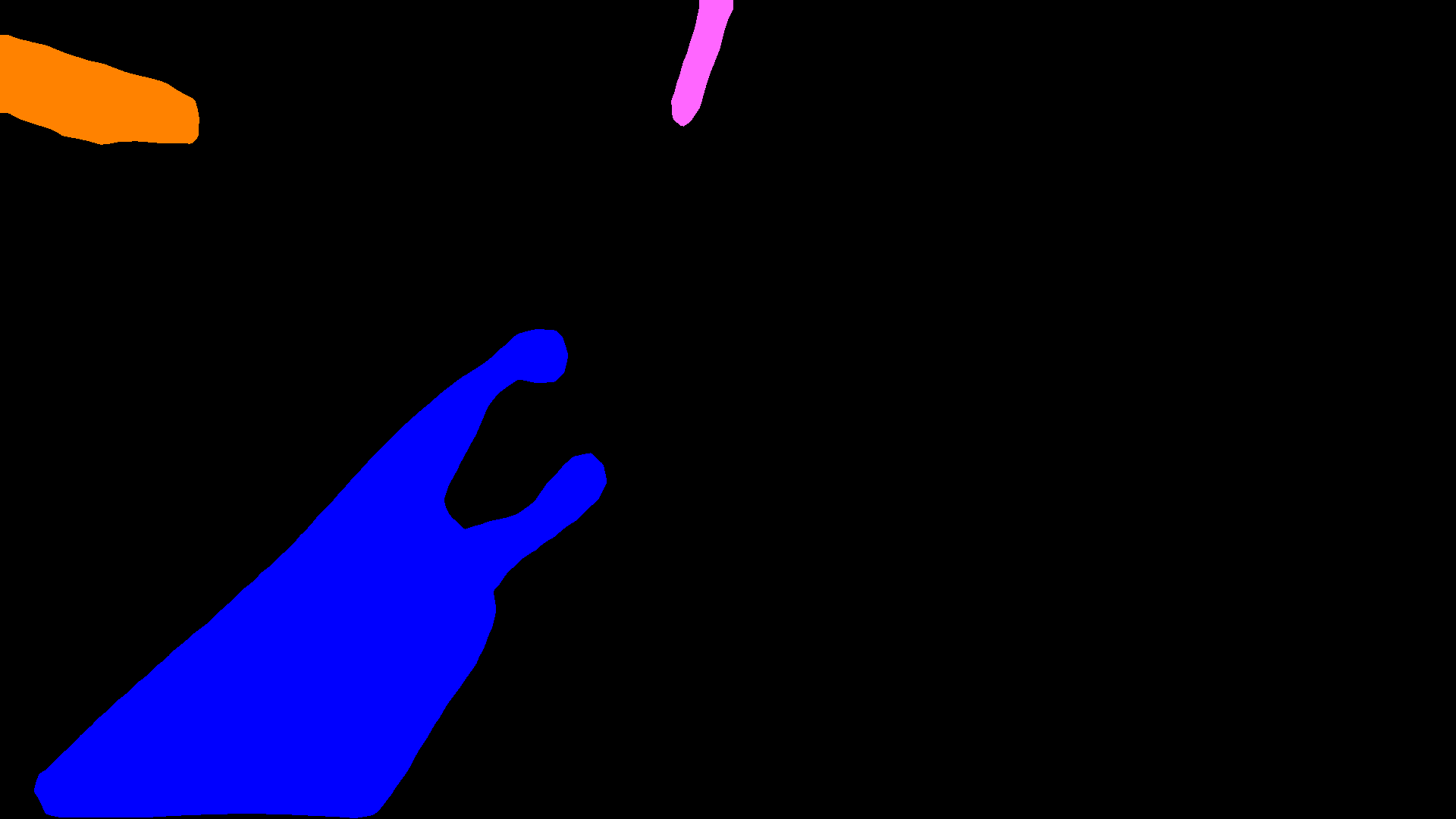}\\[-2pt]
    {\footnotesize 94.03 \(|\) 100.00 \(|\) 33.84}
\end{minipage}
\\
\multirow{2}{*}{sk} &
\begin{minipage}{0.165\textwidth}
    \centering
    \includegraphics[width=\linewidth]{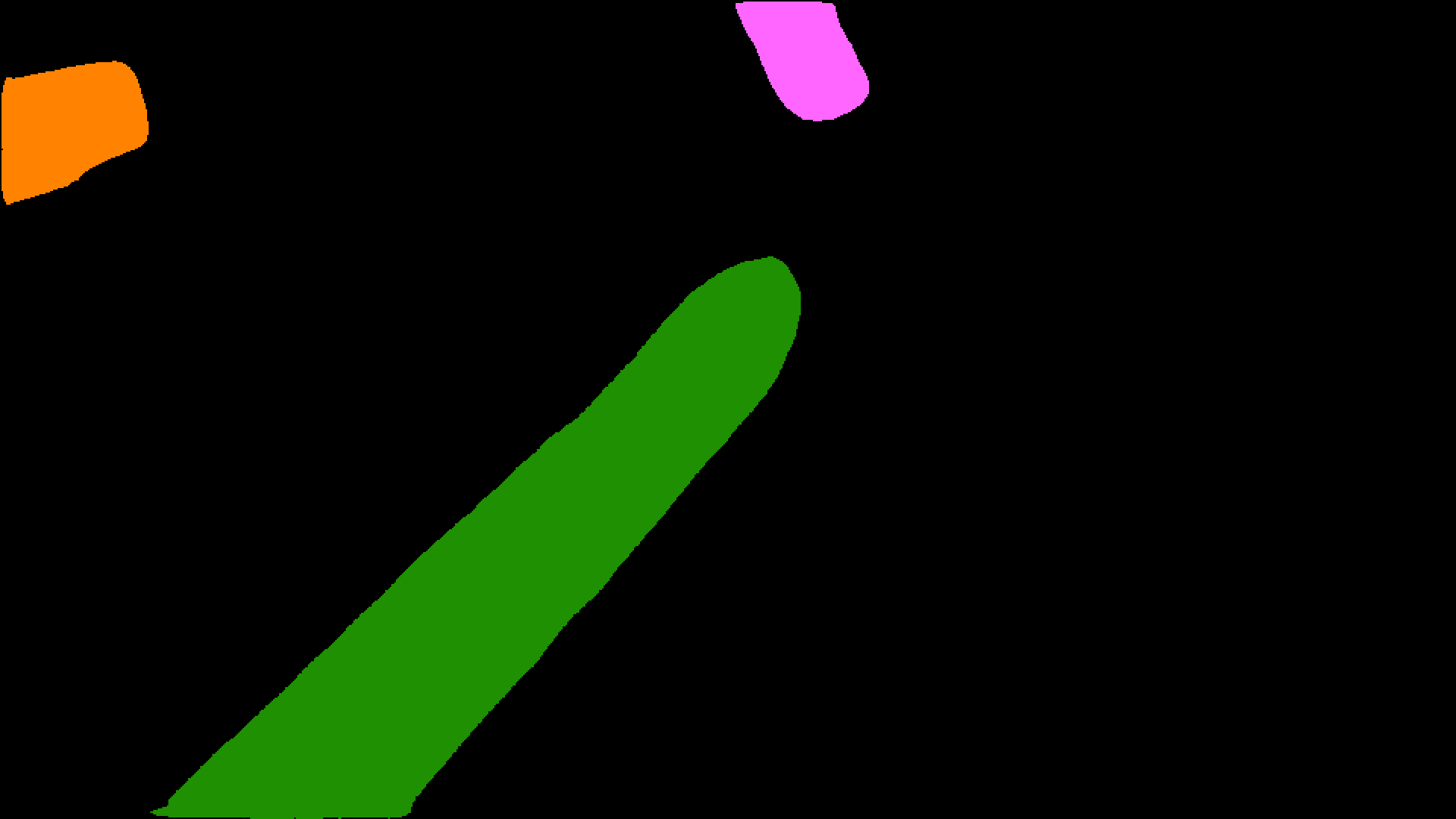}\\[-2pt]
    {\footnotesize 96.34 \(|\) 100.00 \(|\) 30.88}
\end{minipage}
&
\begin{minipage}{0.165\textwidth}
    \centering
    \includegraphics[width=\linewidth]{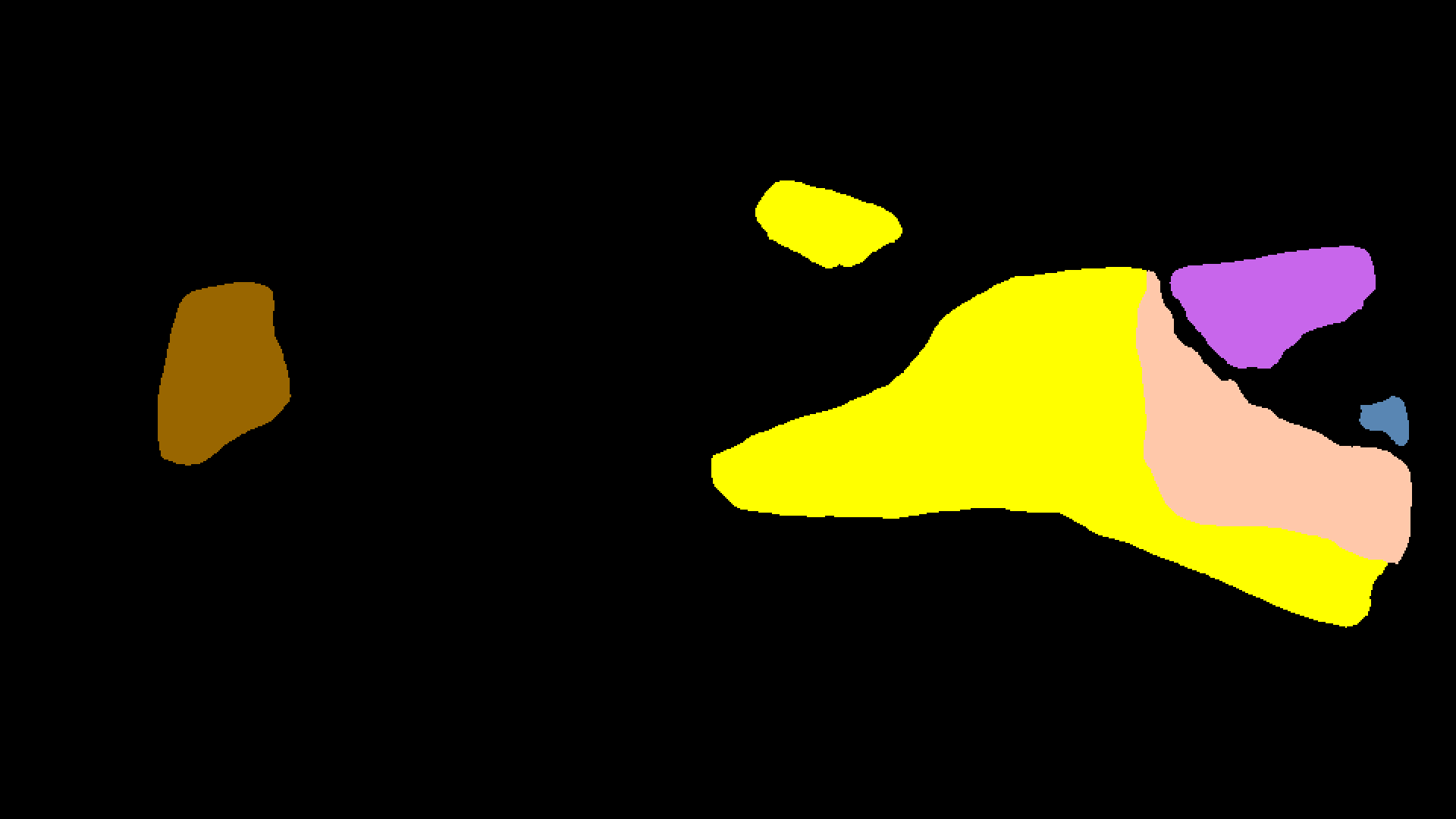}\\[-2pt]
    {\footnotesize 11.08 \(|\) 0.00 \(|\) 109.27}
\end{minipage}
&
\begin{minipage}{0.165\textwidth}
    \centering
    \includegraphics[width=\linewidth]{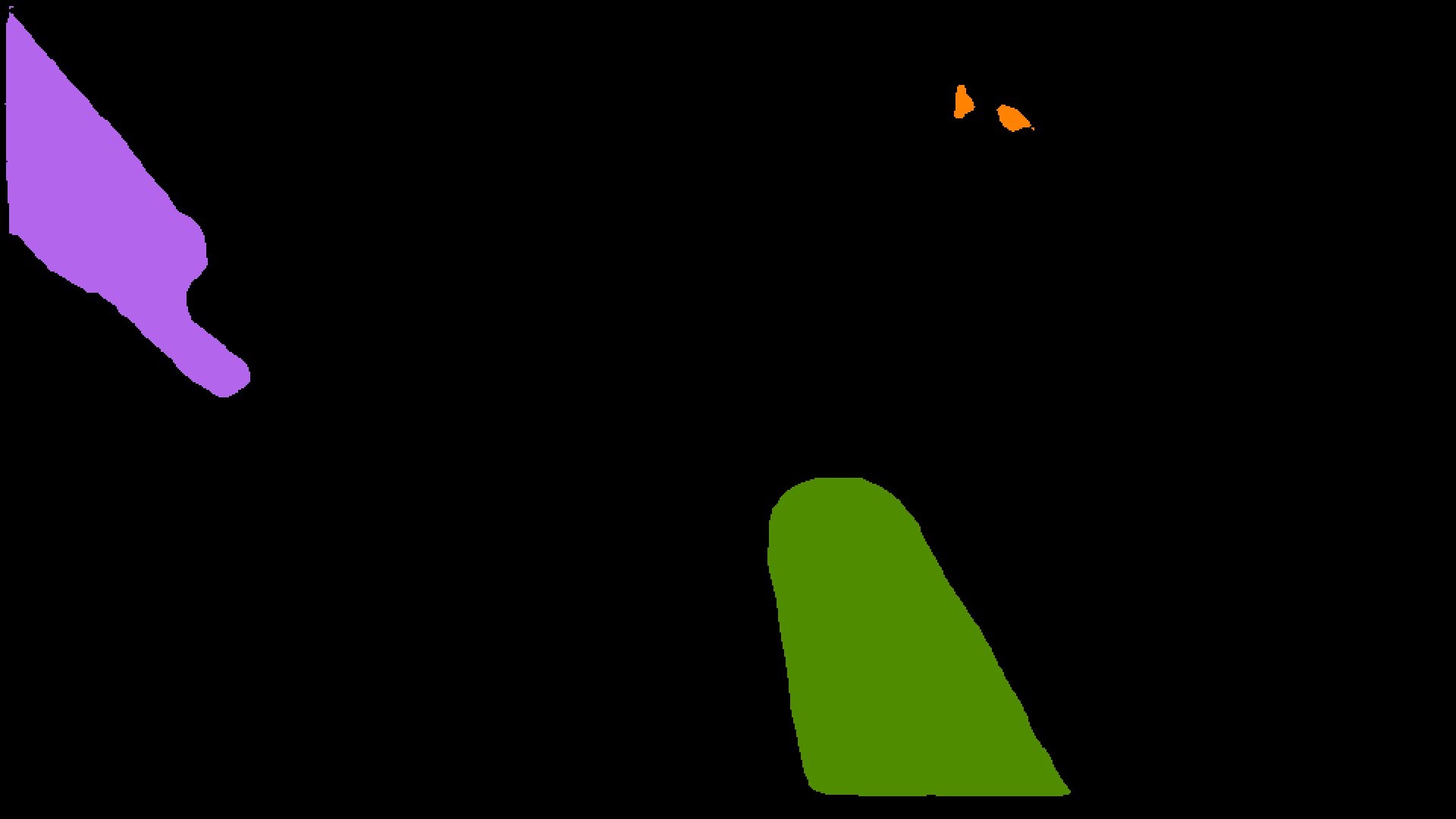}\\[-2pt]
    {\footnotesize 47.95 \(|\) 33.33 \(|\) 225.46}
\end{minipage}
&
\begin{minipage}{0.165\textwidth}
    \centering
    \includegraphics[width=\linewidth]{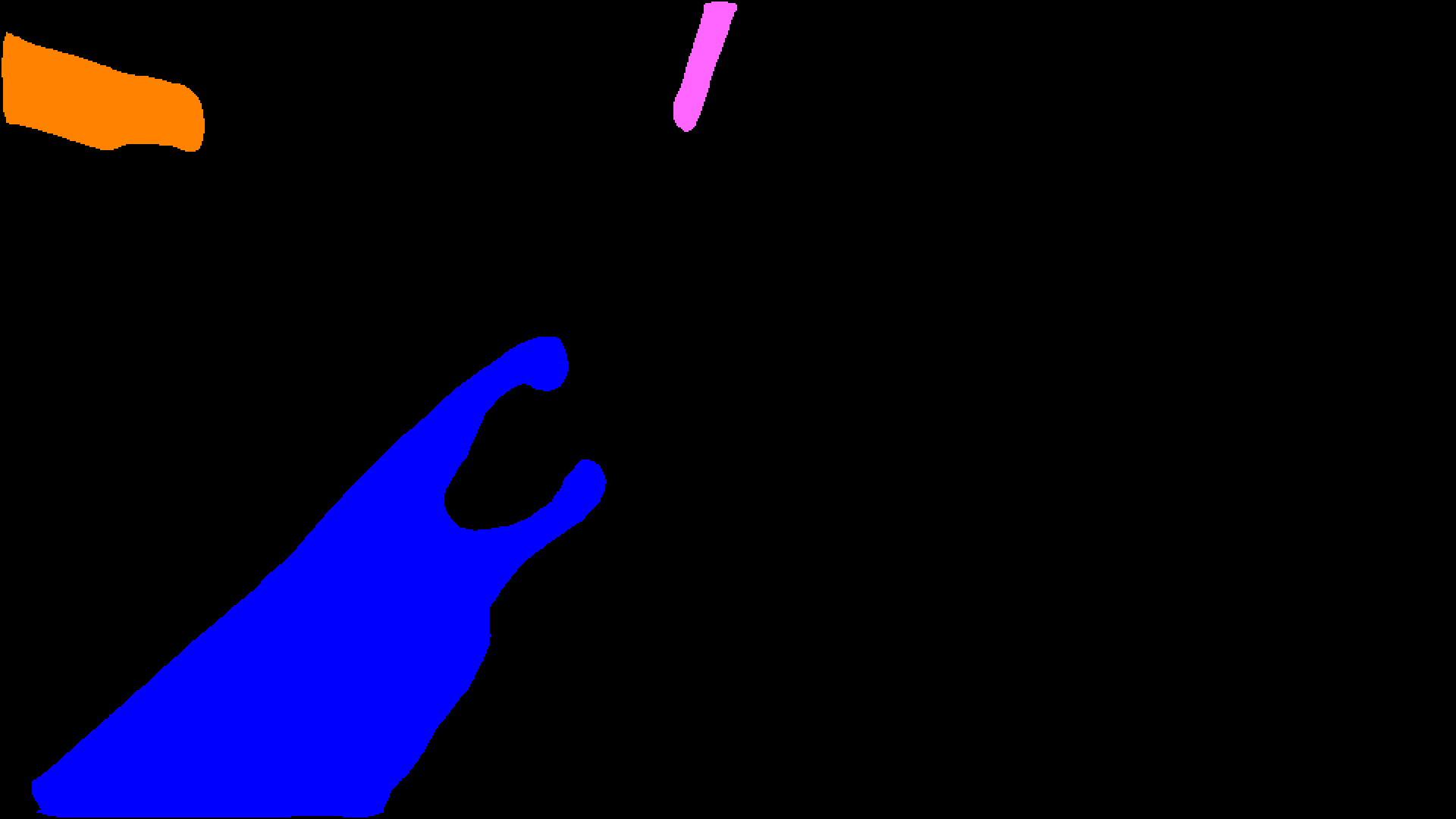}\\[-2pt]
    {\footnotesize 94.53 \(|\) 100.00 \(|\) 33.28}
\end{minipage}
\\
\multirow{2}{*}{goncalo} &
\begin{minipage}{0.165\textwidth}
    \centering
    \includegraphics[width=\linewidth]{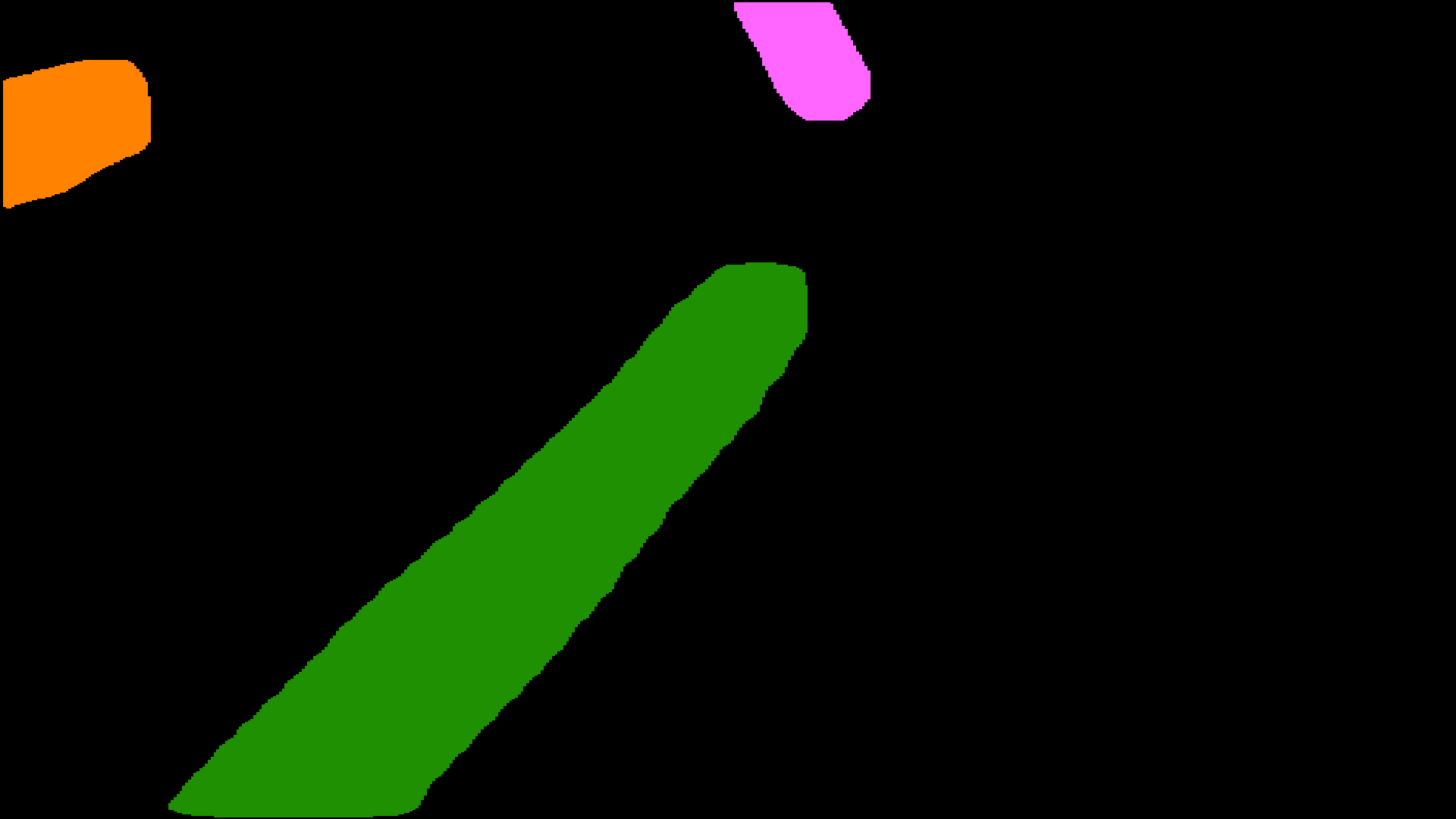}\\[-2pt]
    {\footnotesize 96.12 \(|\) 100.00 \(|\) 35.79}
\end{minipage}
&
\begin{minipage}{0.165\textwidth}
    \centering
    \includegraphics[width=\linewidth]{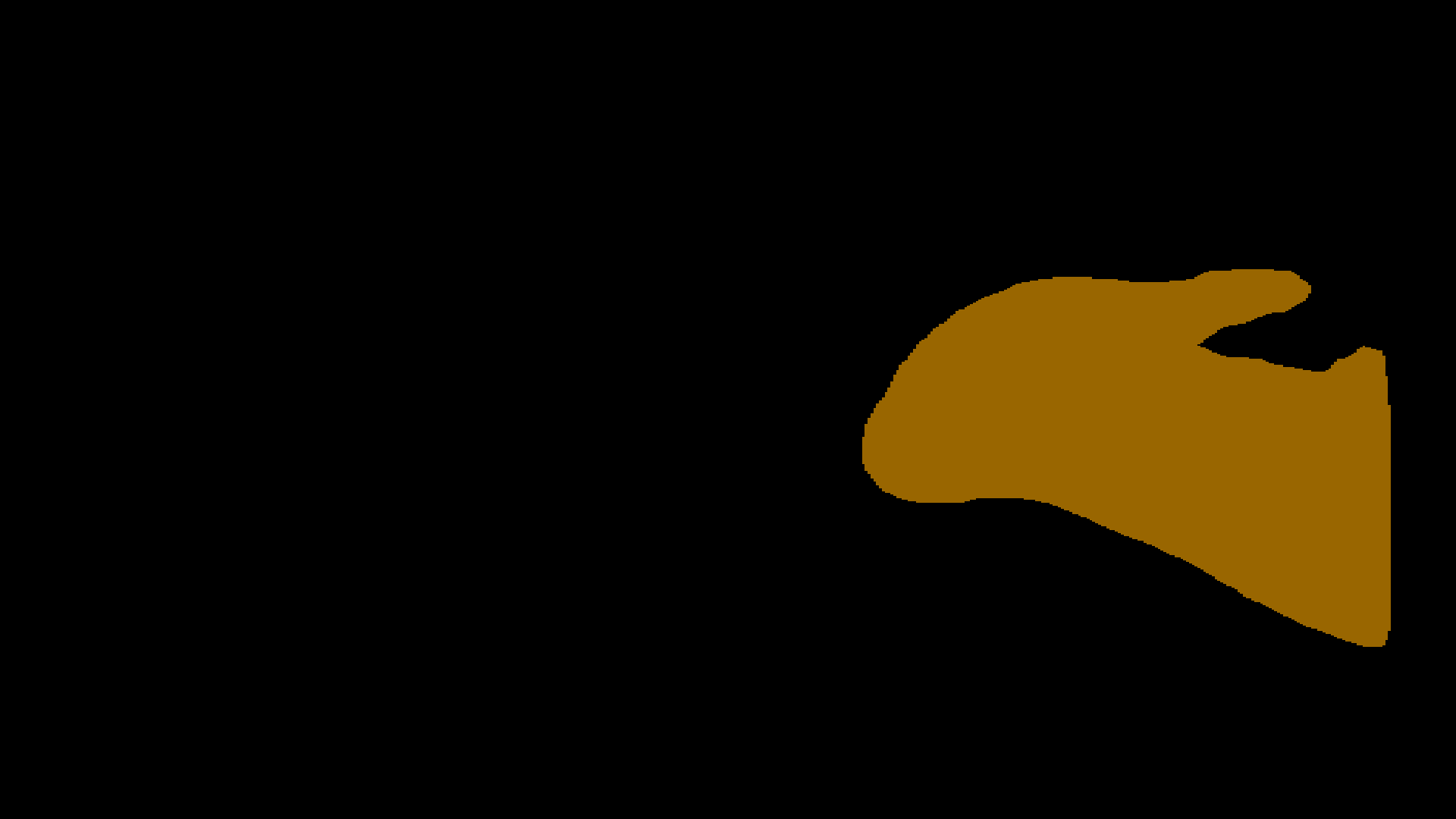}\\[-2pt]
    {\footnotesize 0.00 \(|\) 0.00 \(|\) 211.69}
\end{minipage}
&
\begin{minipage}{0.165\textwidth}
    \centering
    \includegraphics[width=\linewidth]{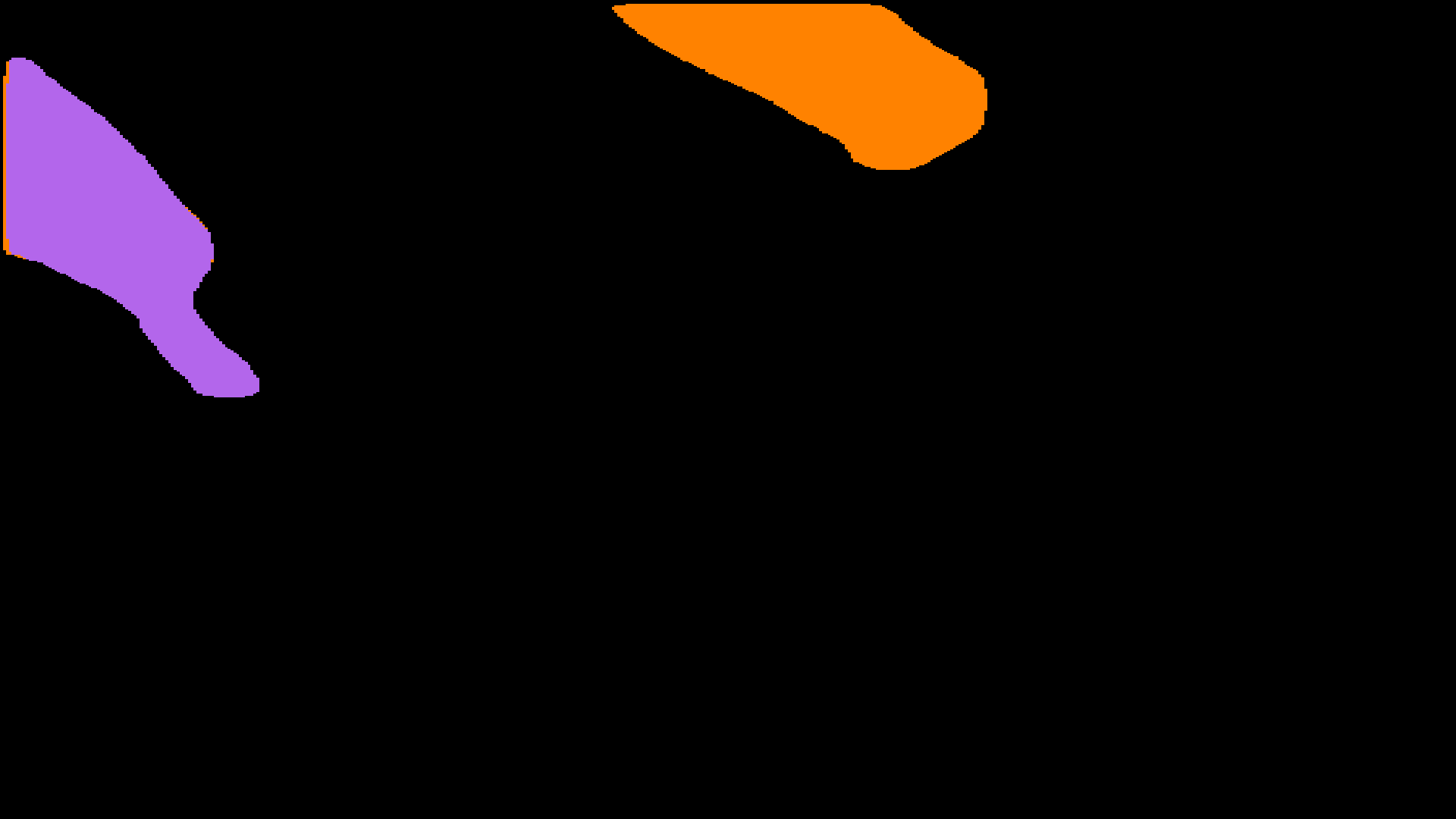}\\[-2pt]
    {\footnotesize 15.01 \(|\) 0.00 \(|\) 806.70}
\end{minipage}
&
\begin{minipage}{0.165\textwidth}
    \centering
    \includegraphics[width=\linewidth]{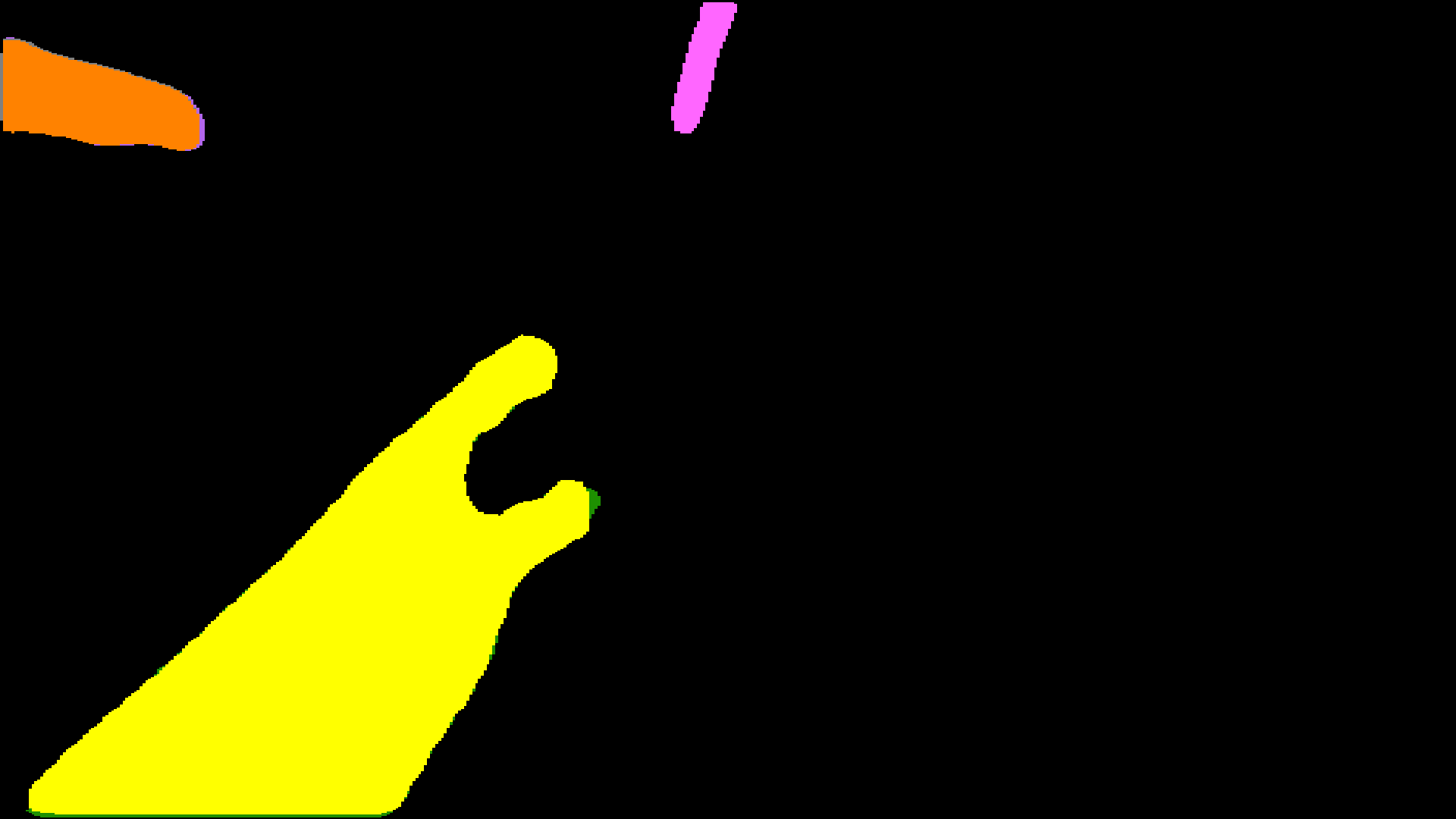}\\[-2pt]
    {\footnotesize 26.12 \(|\) 28.57 \(|\) 35.67}
\end{minipage}
\\
\multirow{2}{*}{hanglok} &
\begin{minipage}{0.165\textwidth}
    \centering
    \includegraphics[width=\linewidth]{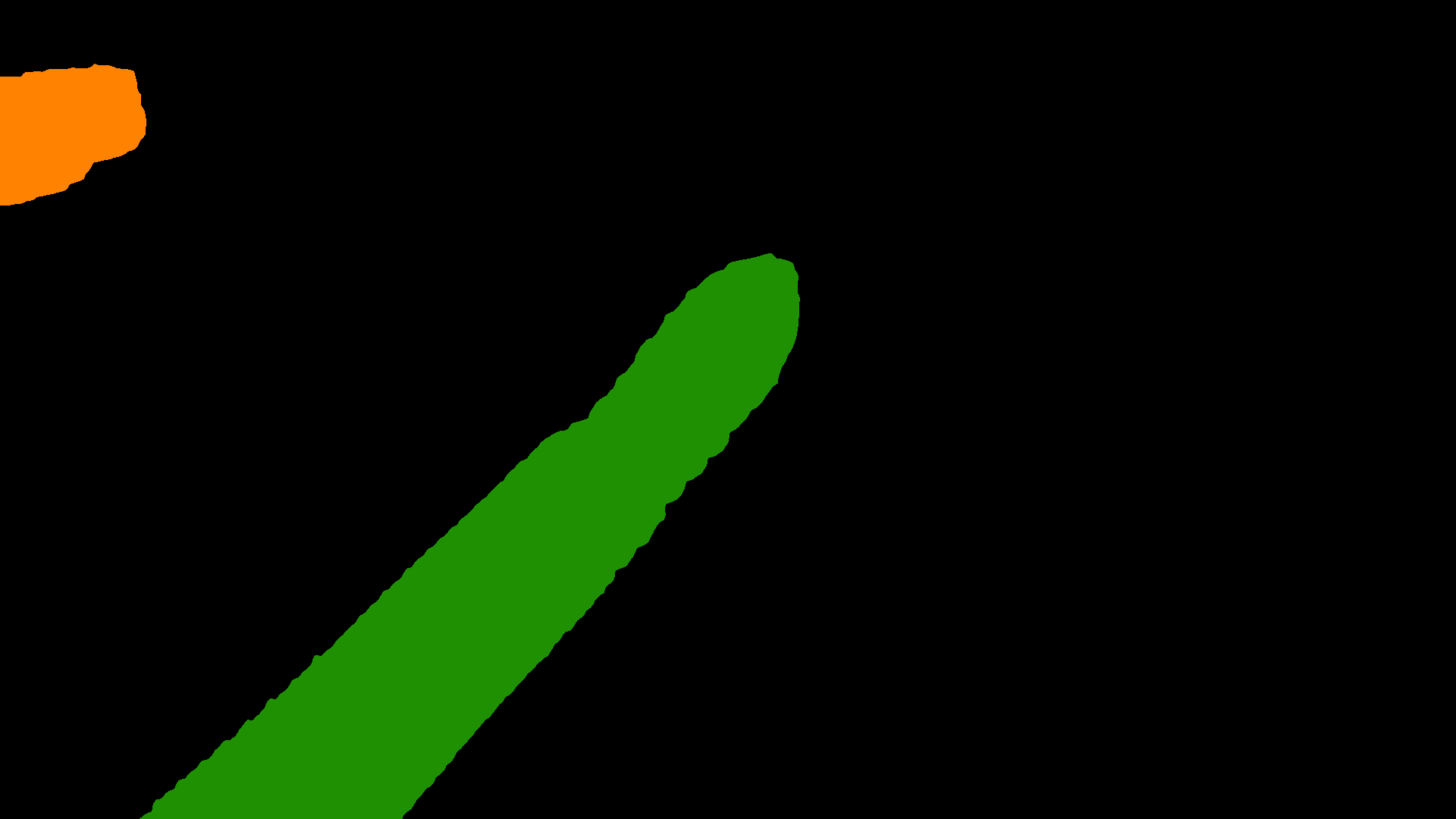}\\[-2pt]
    {\footnotesize 64.23 \(|\) 66.67 \(|\) 11.11}
\end{minipage}
&
\begin{minipage}{0.165\textwidth}
    \centering
    \includegraphics[width=\linewidth]{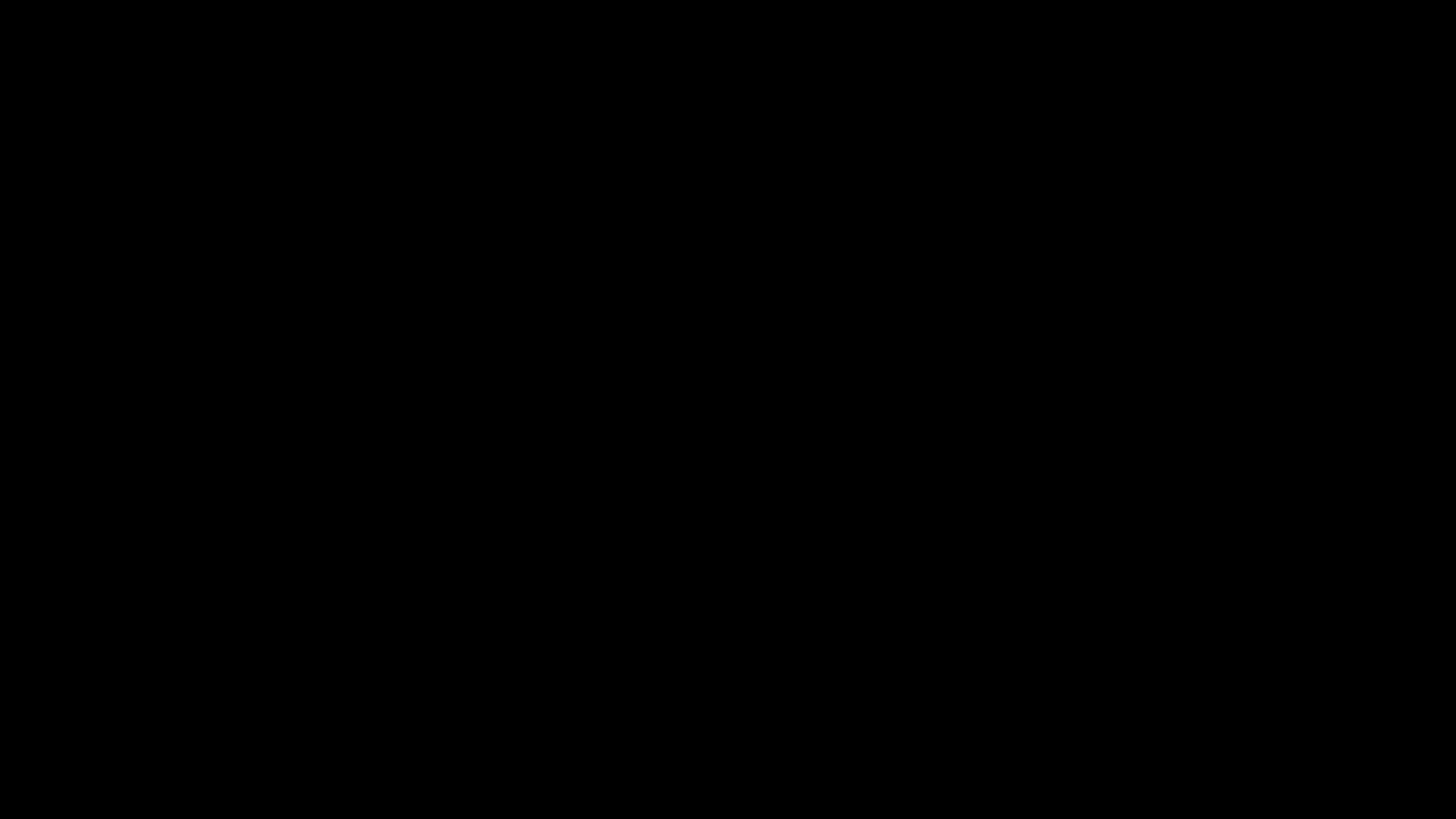}\\[-2pt]
    {\footnotesize 0.00 \(|\) 0.00 \(|\) nan}
\end{minipage}
&
\begin{minipage}{0.165\textwidth}
    \centering
    \includegraphics[width=\linewidth]{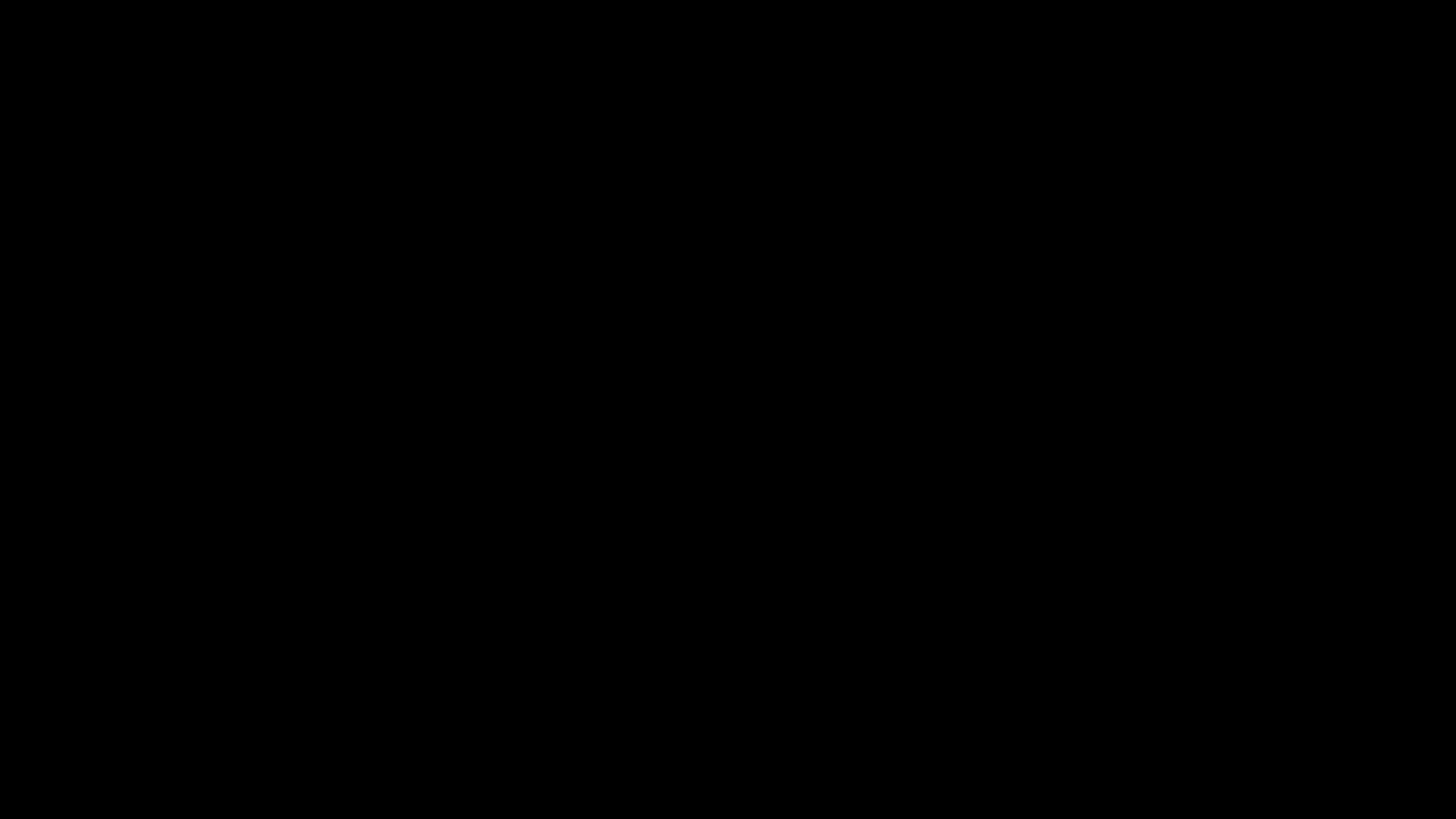}\\[-2pt]
    {\footnotesize 0.00 \(|\) 0.00 \(|\) nan}
\end{minipage}
&
\begin{minipage}{0.165\textwidth}
    \centering
    \includegraphics[width=\linewidth]{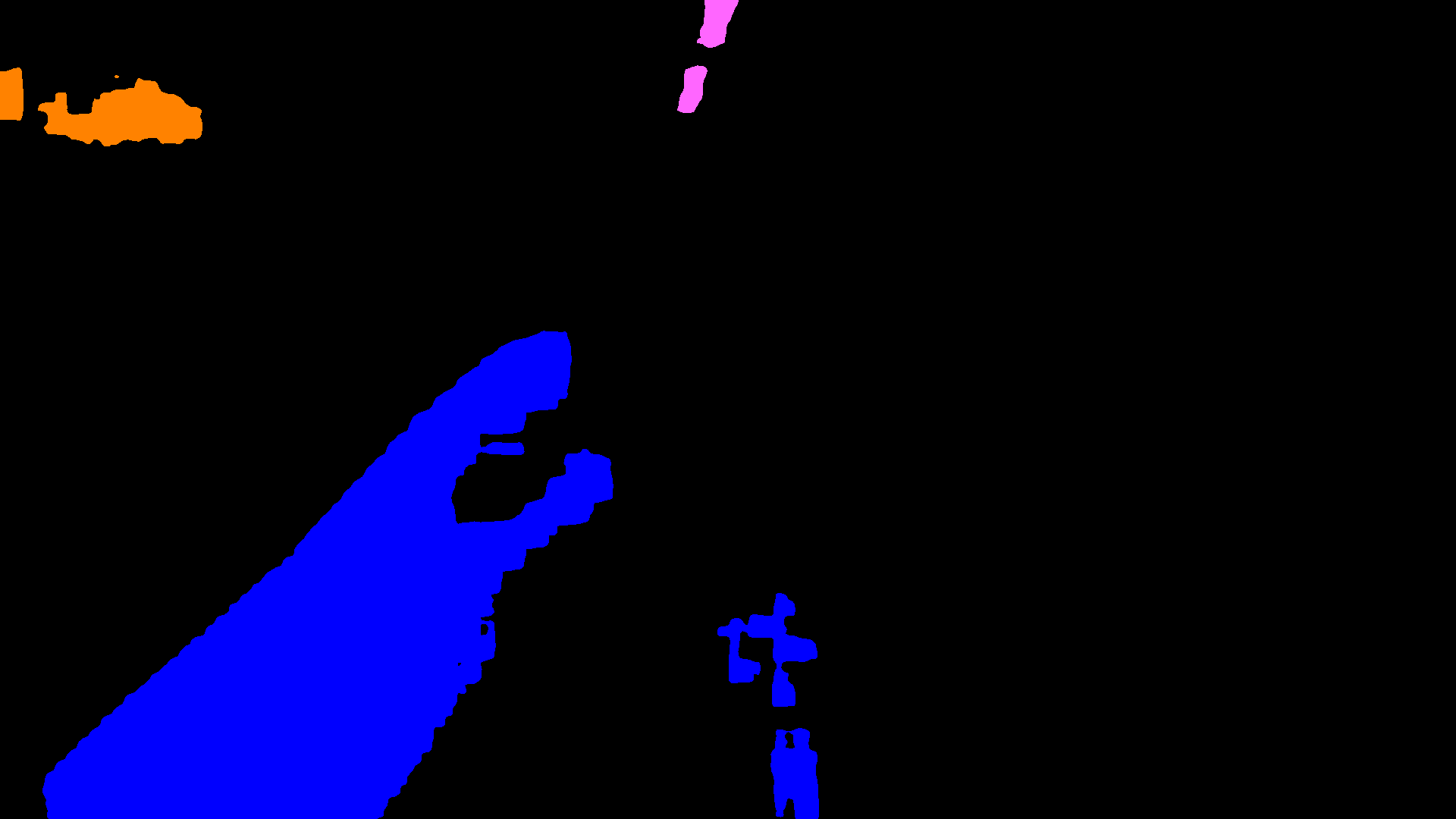}\\[-2pt]
    {\footnotesize 77.41 \(|\) 66.67 \(|\) 152.42}
\end{minipage}
\\
\multirow{2}{*}{recogna} &
\begin{minipage}{0.165\textwidth}
    \centering
    \includegraphics[width=\linewidth]{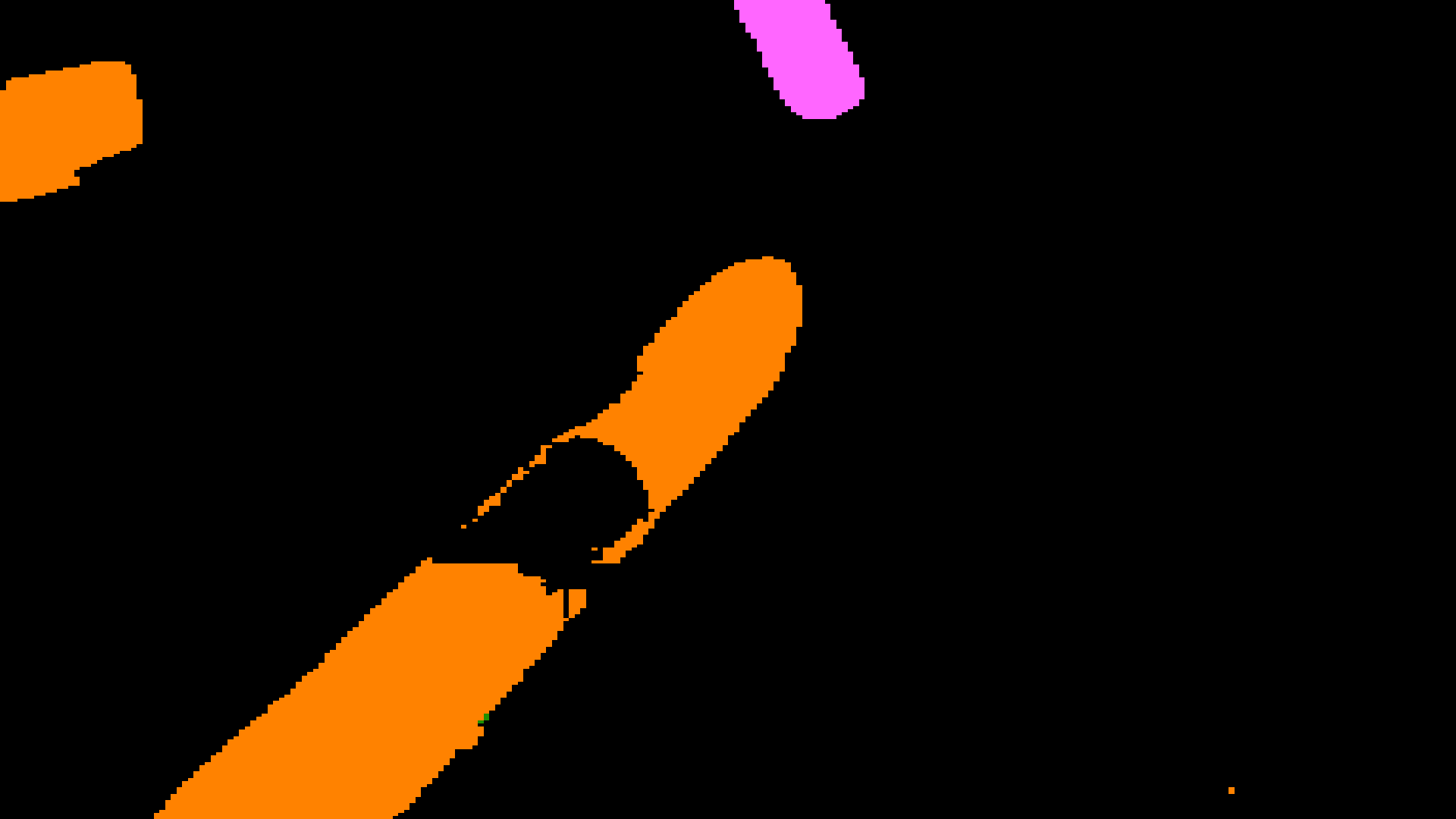}\\[-2pt]
    {\footnotesize 39.93 \(|\) 33.33 \(|\) 582.06}
\end{minipage}
&
\begin{minipage}{0.165\textwidth}
    \centering
    \includegraphics[width=\linewidth]{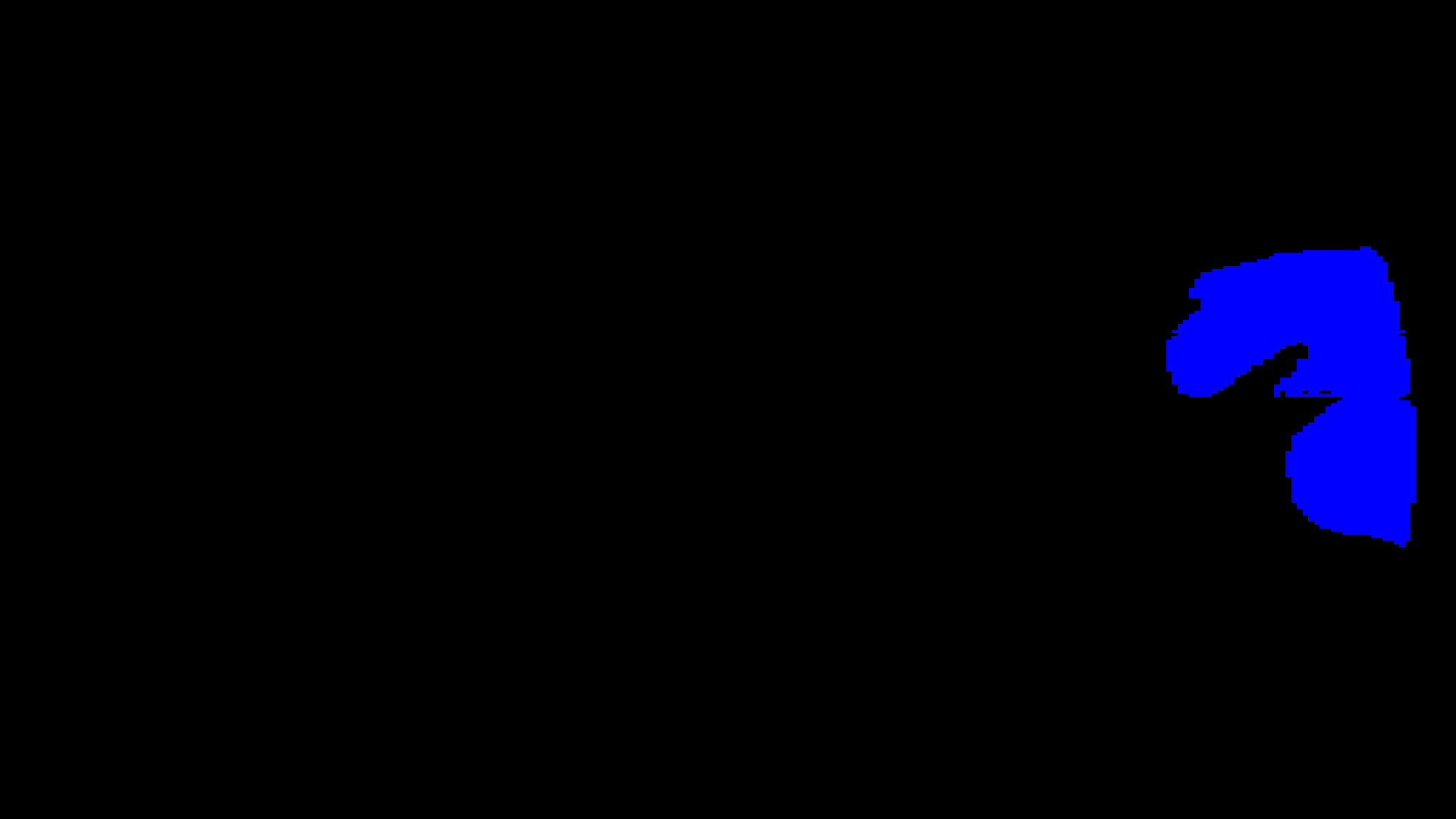}\\[-2pt]
    {\footnotesize 0.00 \(|\) 0.00 \(|\) 573.50}
\end{minipage}
&
\begin{minipage}{0.165\textwidth}
    \centering
    \includegraphics[width=\linewidth]{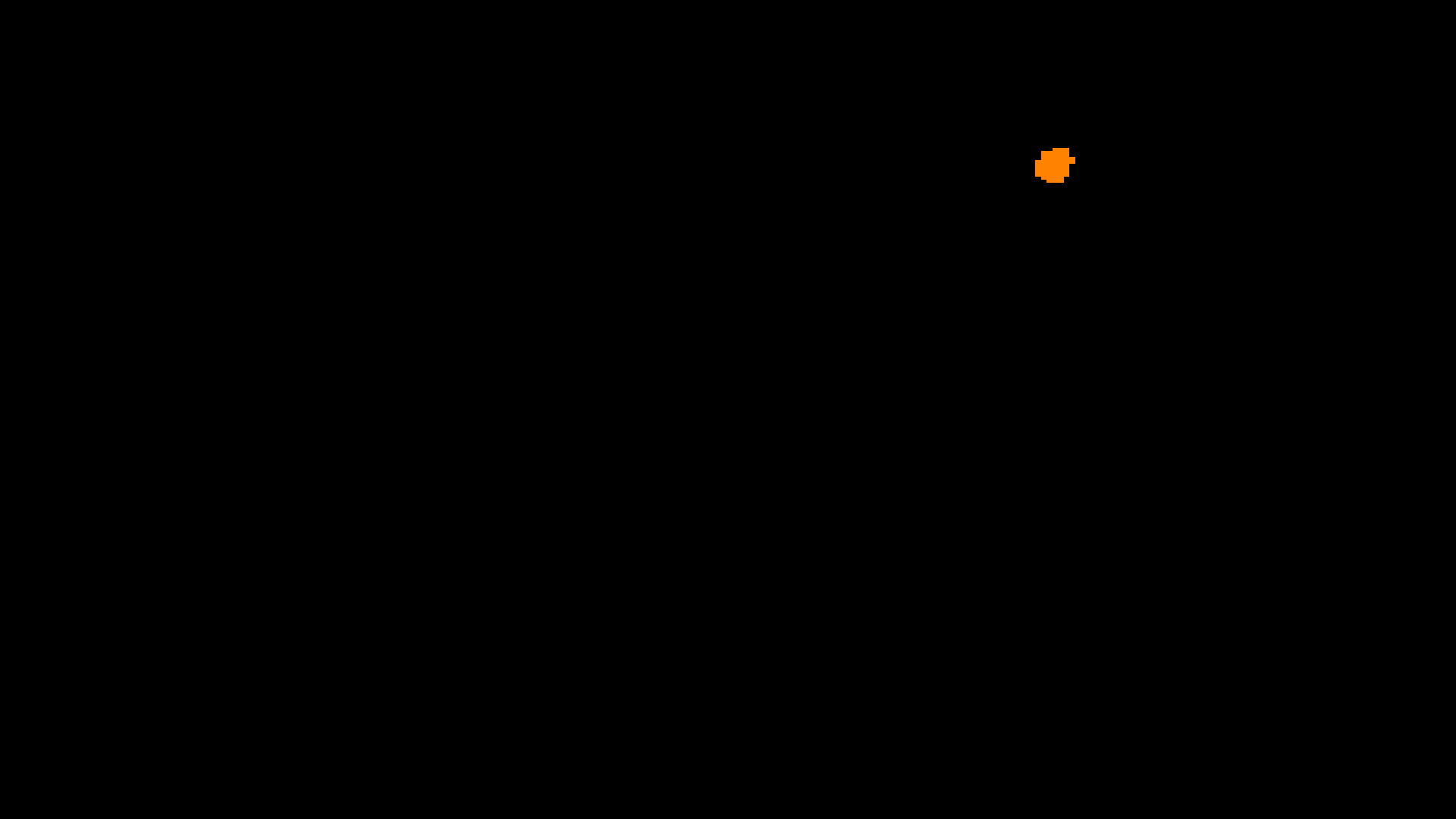}\\[-2pt]
    {\footnotesize 0.00 \(|\) 0.00 \(|\) 570.86}
\end{minipage}
&
\begin{minipage}{0.165\textwidth}
    \centering
    \includegraphics[width=\linewidth]{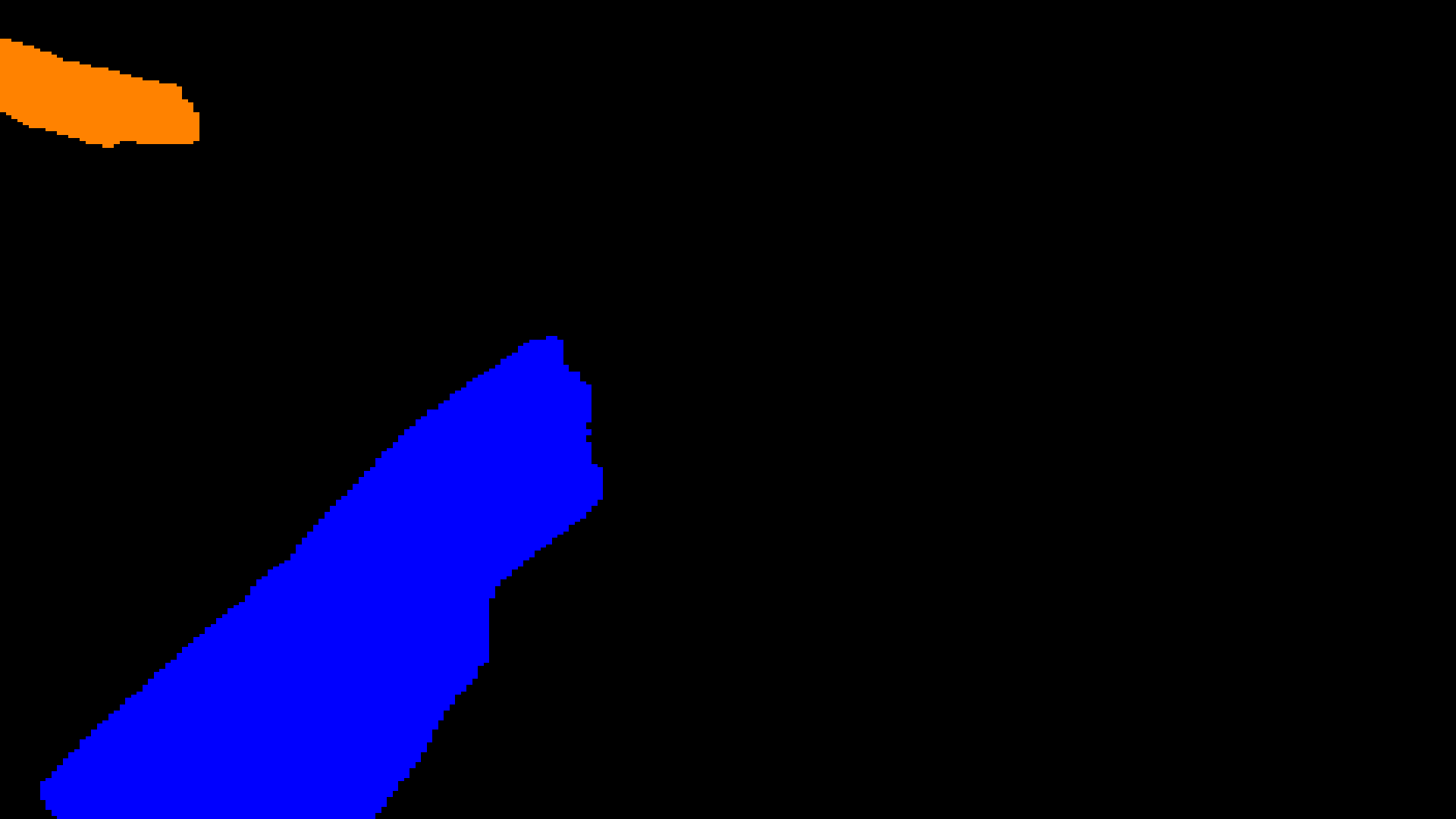}\\[-2pt]
    {\footnotesize 61.69 \(|\) 66.67 \(|\) 43.59}
\end{minipage}
\\

\bottomrule
\end{tabularx}
\label{results:instrument_seg:tab:qualitative_results_page_01}
\end{figure*}

\begin{figure*}[tbph!]
\footnotesize
\centering
\renewcommand{\arraystretch}{1.5}
\captionof{table}{Visualization of example participant predictions, together with quantitative results for DSC (in \%) $\uparrow$, $\text{mAP}_{\text{IoU}}$ (in \%) $\uparrow$, and 95\% HD $\downarrow$ below each image. An assignment of the color codings to the respective instrument categories is provided in Section 4.1 in \ref{appendix:data_description}.}
\begin{tabularx}{\textwidth}{m{1.5cm} X X X X }
\toprule
\multicolumn{5}{c}{\textbf{Challenge data}} \\
\midrule
\textbf{Team} & \textbf{Sample 5} & \textbf{Sample 6} & \textbf{Sample 7} & \textbf{Sample 8} \\
\midrule
\multirow{2}{*}{Input} &
\begin{minipage}{0.165\textwidth}
    \centering
    \includegraphics[width=\linewidth]{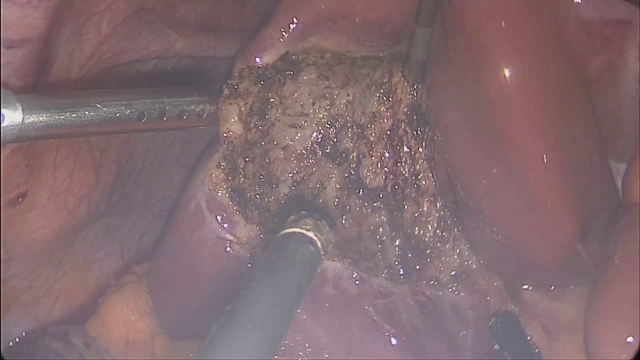}\\
\end{minipage}
&
\begin{minipage}{0.165\textwidth}
    \centering
    \includegraphics[width=\linewidth]{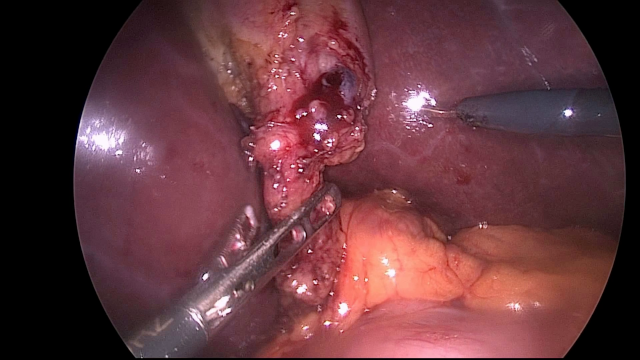}\\
\end{minipage}
&
\begin{minipage}{0.165\textwidth}
    \centering
    \includegraphics[width=\linewidth]{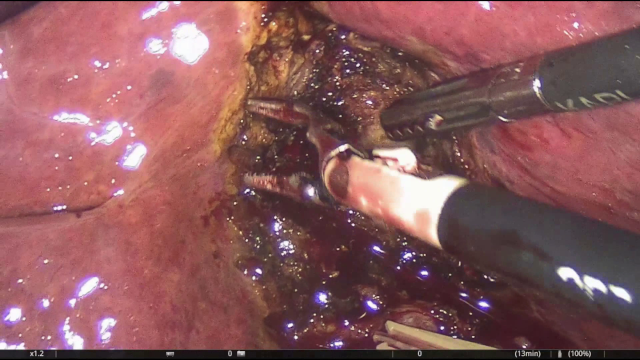}\\
\end{minipage}
&
\begin{minipage}{0.165\textwidth}
    \centering
    \includegraphics[width=\linewidth]{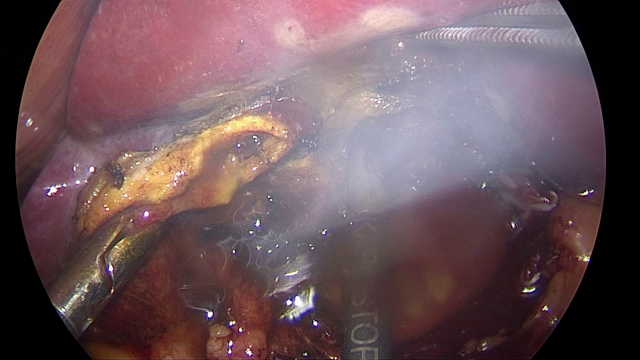}\\
\end{minipage}
\\[1cm]
\multirow{2}{*}{\shortstack[l]{Ground \\ truth}} &
\begin{minipage}{0.165\textwidth}
    \centering
    \includegraphics[width=\linewidth]{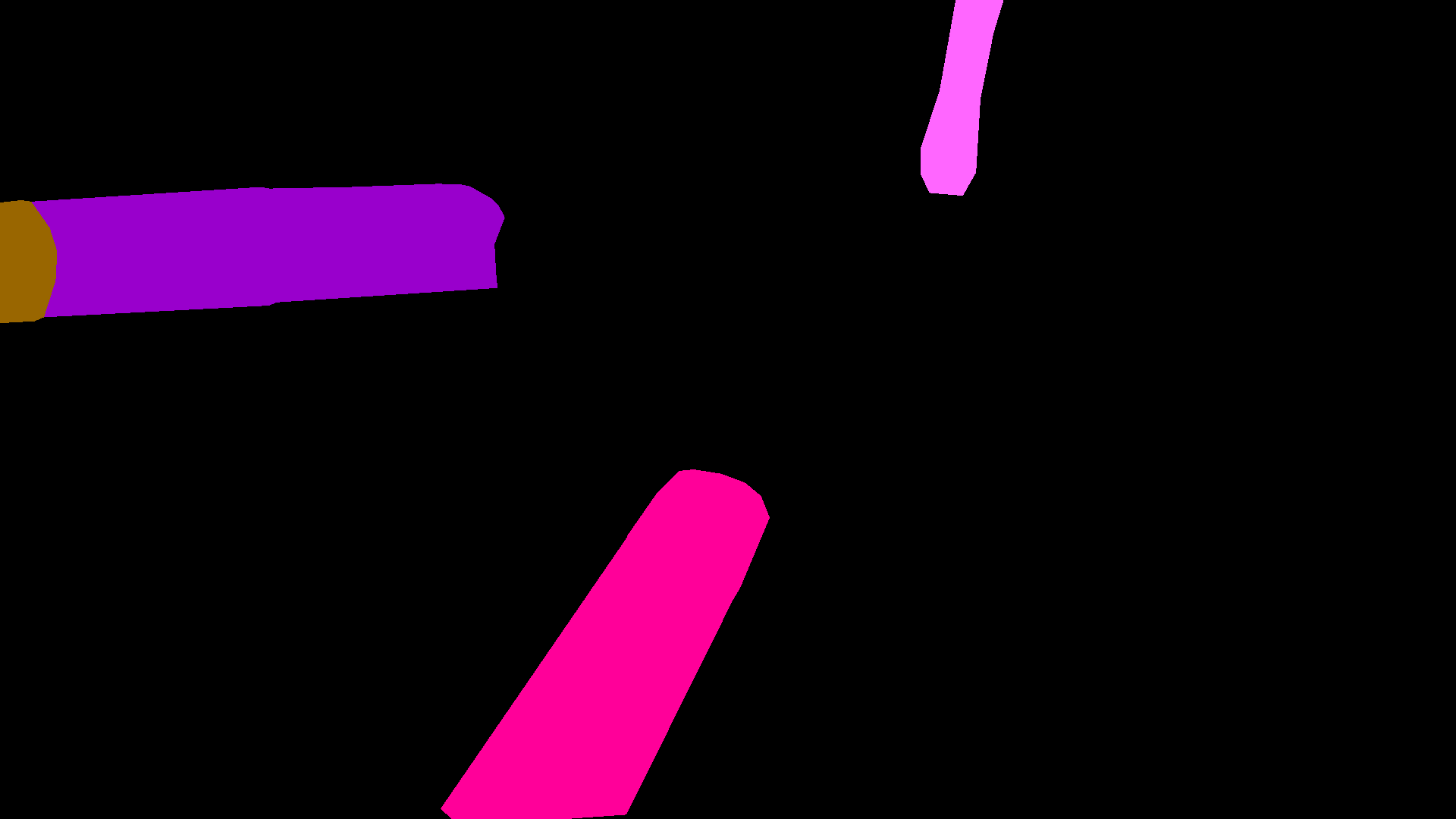}\\
\end{minipage}
&
\begin{minipage}{0.165\textwidth}
    \centering
    \includegraphics[width=\linewidth]{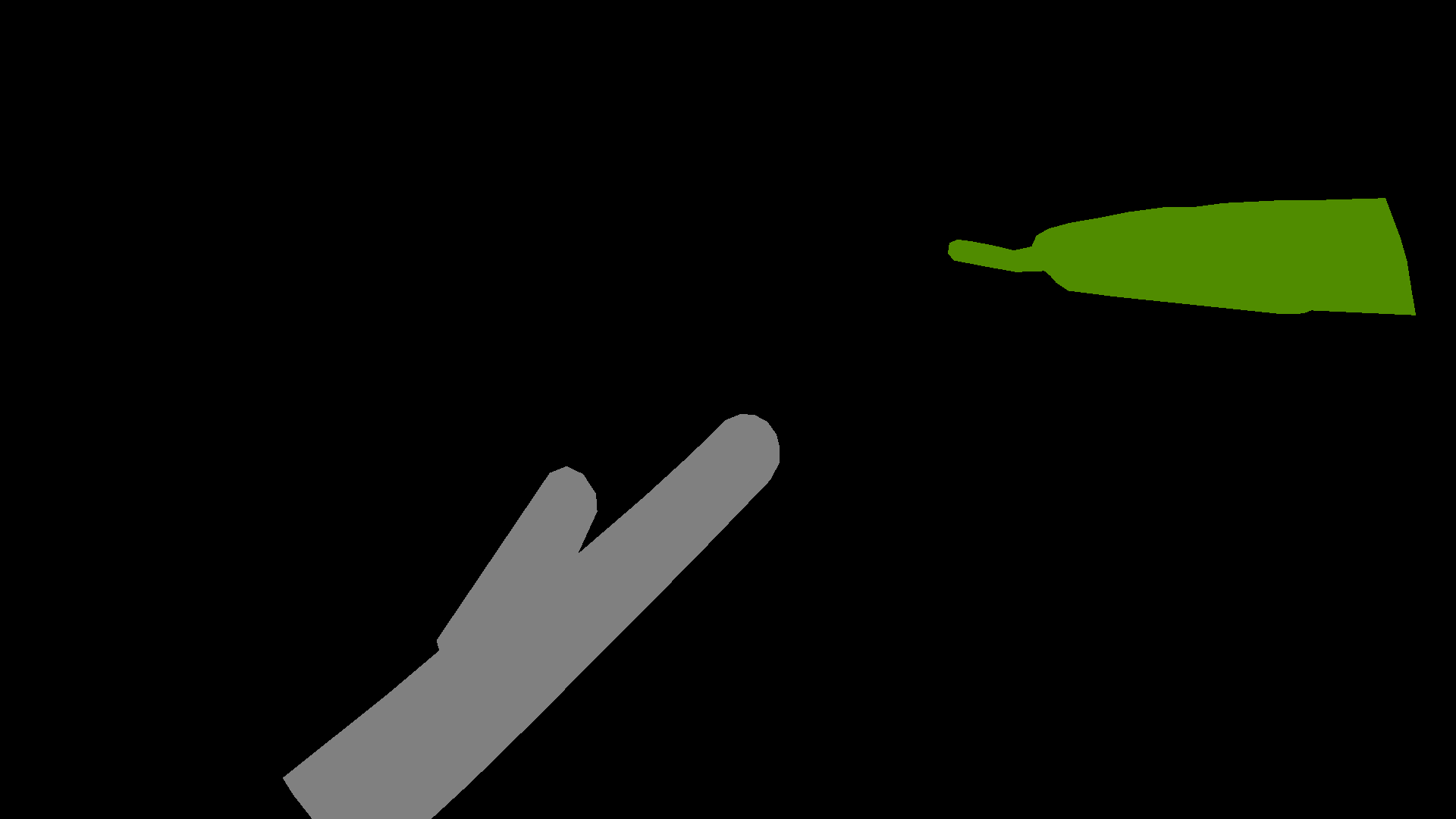}\\
\end{minipage}
&
\begin{minipage}{0.165\textwidth}
    \centering
    \includegraphics[width=\linewidth]{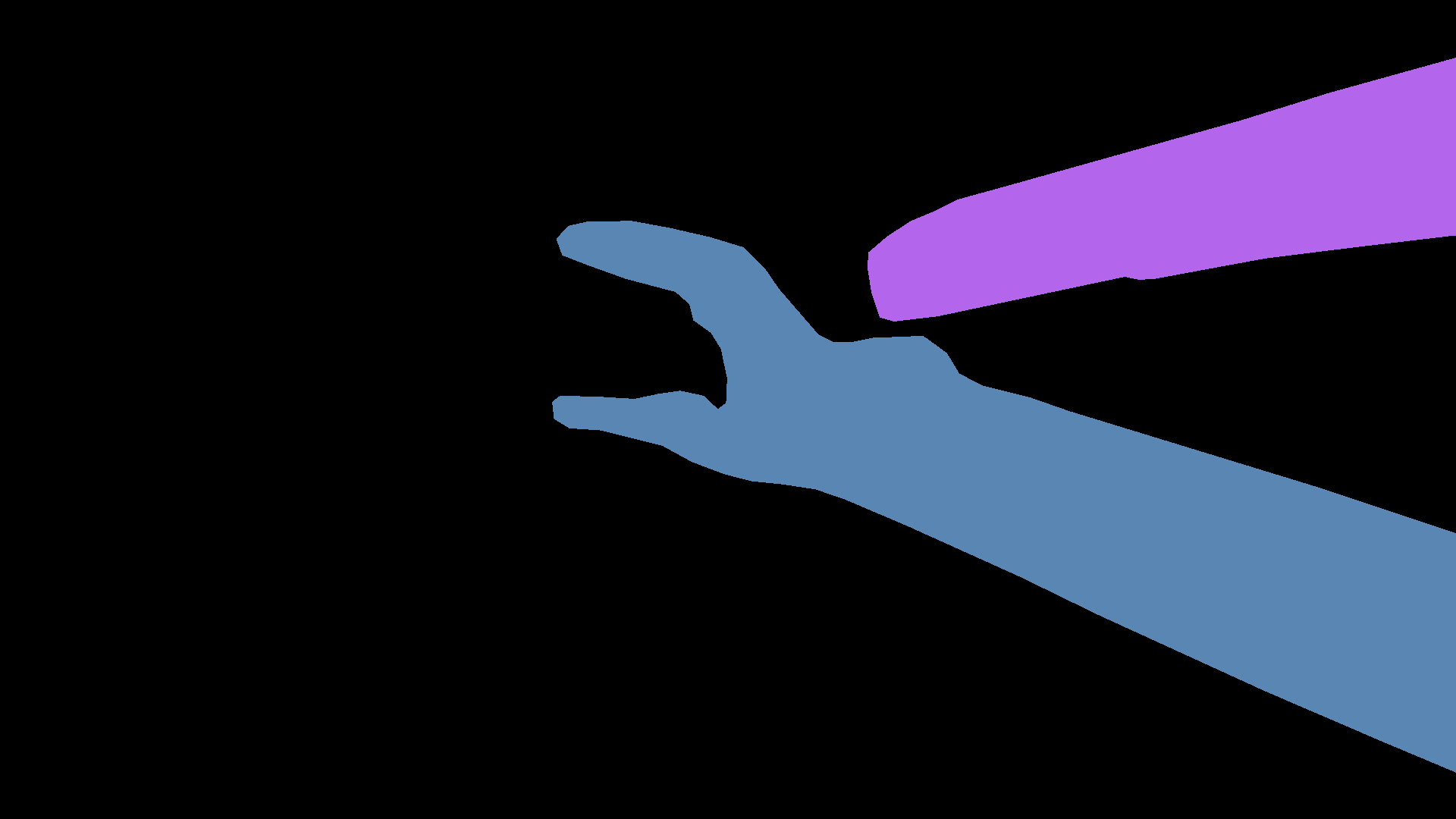}\\
\end{minipage}
&
\begin{minipage}{0.165\textwidth}
    \centering
    \includegraphics[width=\linewidth]{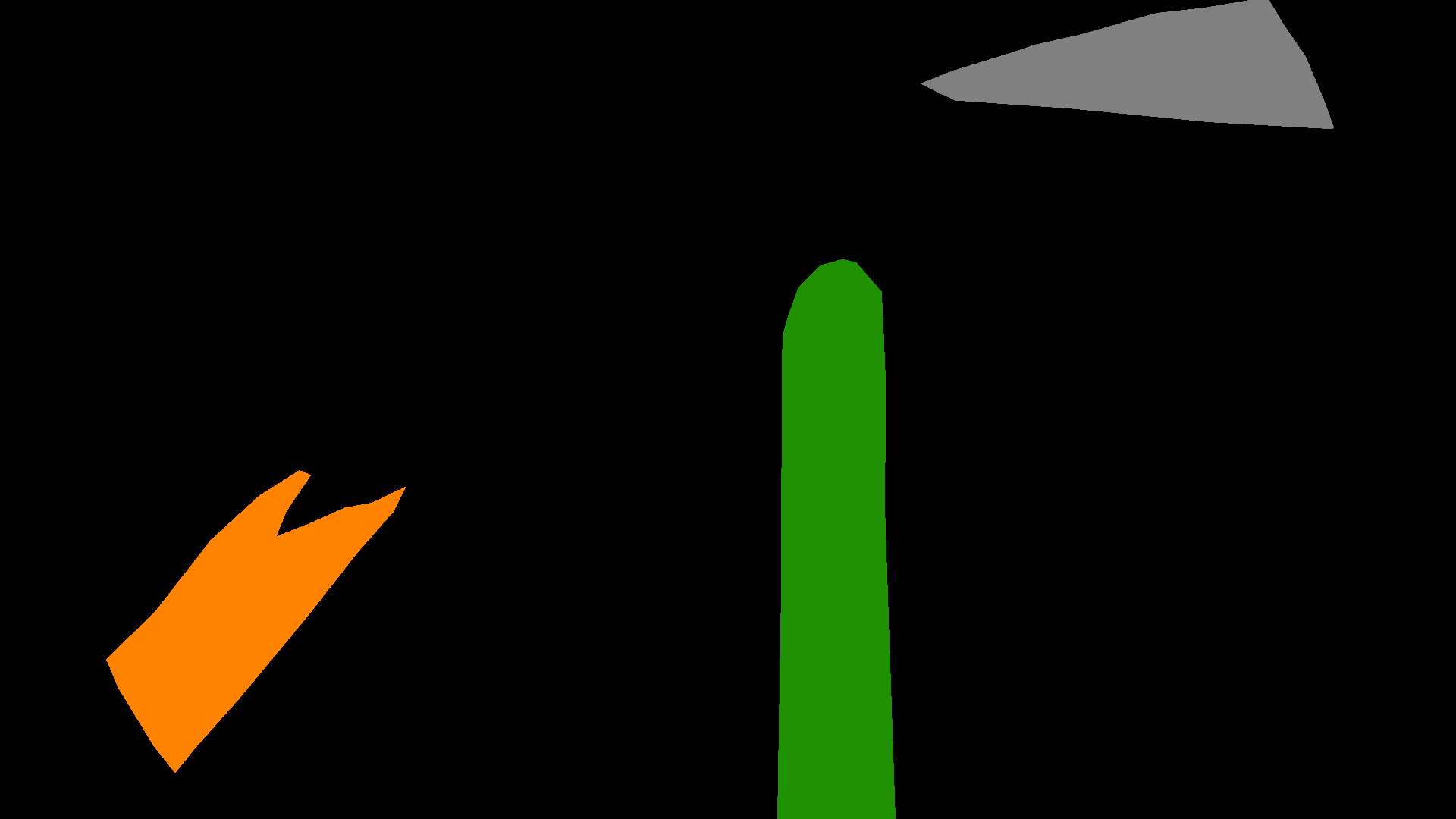}\\
\end{minipage}
\\
\midrule
\multicolumn{5}{c}{\textbf{Participant predictions}} \\
\midrule
\multirow{2}{*}{jmees\_inc.} &
\begin{minipage}{0.165\textwidth}
    \centering
    \includegraphics[width=\linewidth]{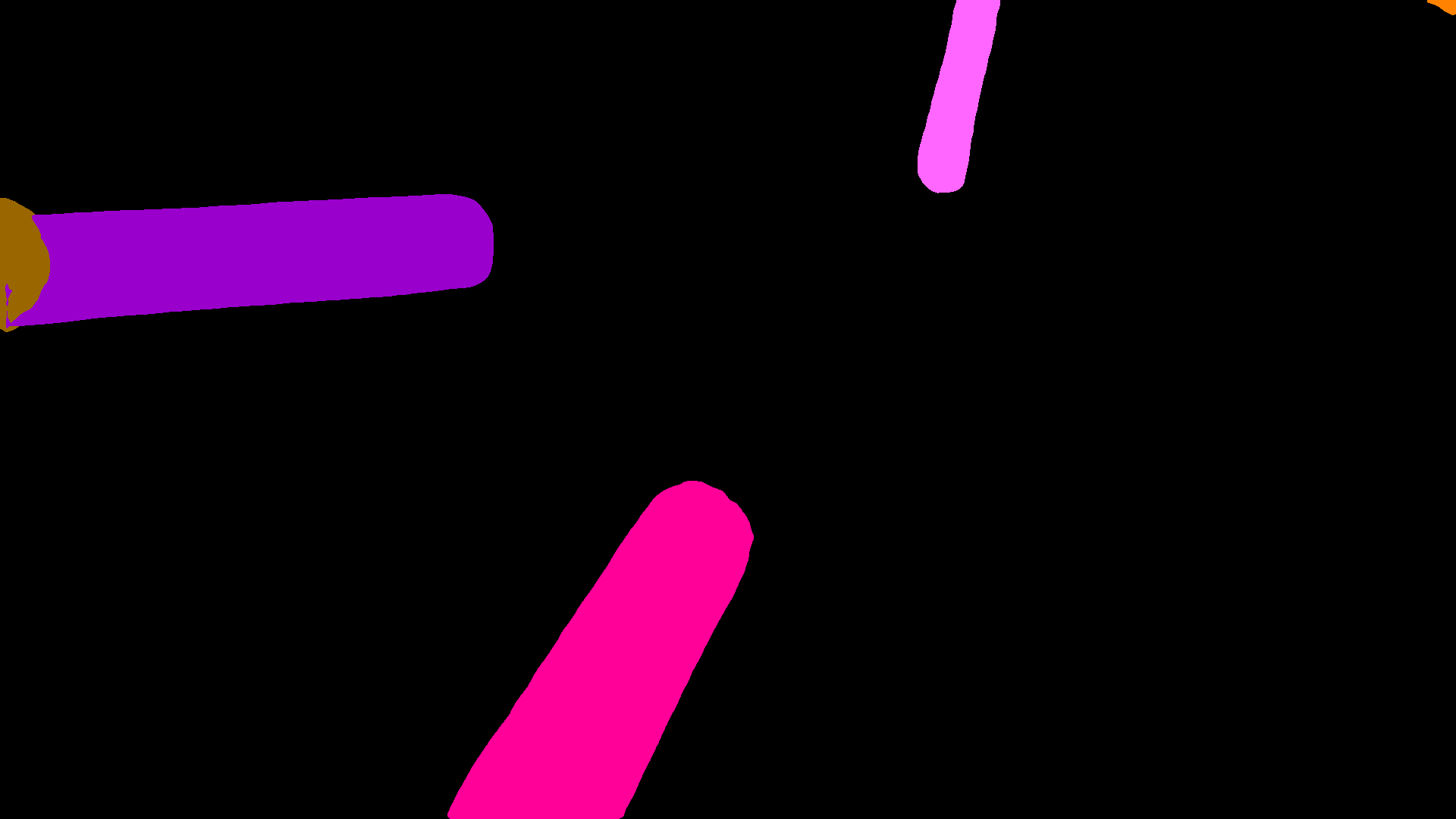}\\[-2pt]
    {\footnotesize 73.10 \(|\) 80.00 \(|\) 25.02}
\end{minipage}
&
\begin{minipage}{0.165\textwidth}
    \centering
    \includegraphics[width=\linewidth]{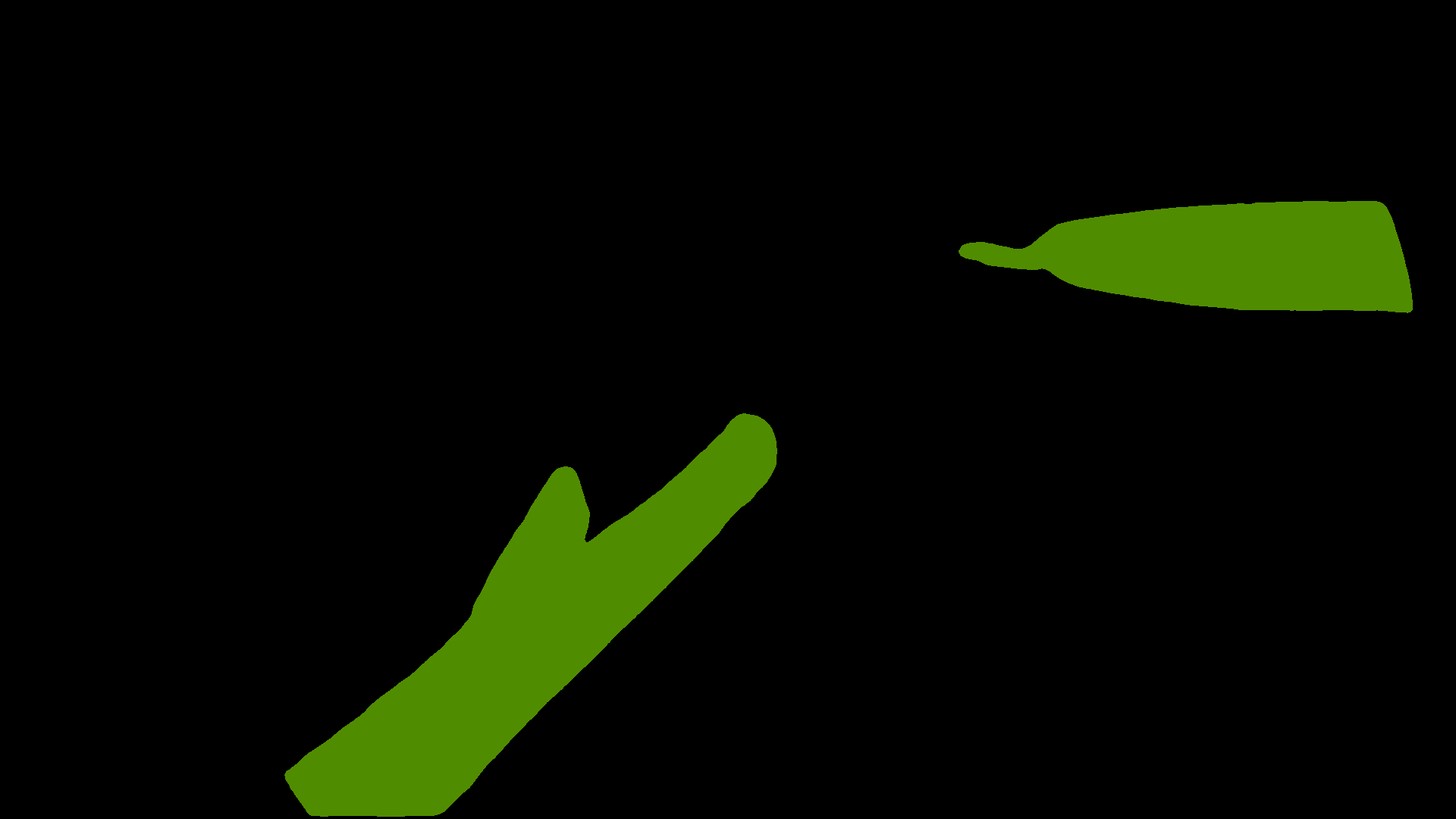}\\[-2pt]
    {\footnotesize 26.12 \(|\) 0.00 \(|\) 789.72}
\end{minipage}
&
\begin{minipage}{0.165\textwidth}
    \centering
    \includegraphics[width=\linewidth]{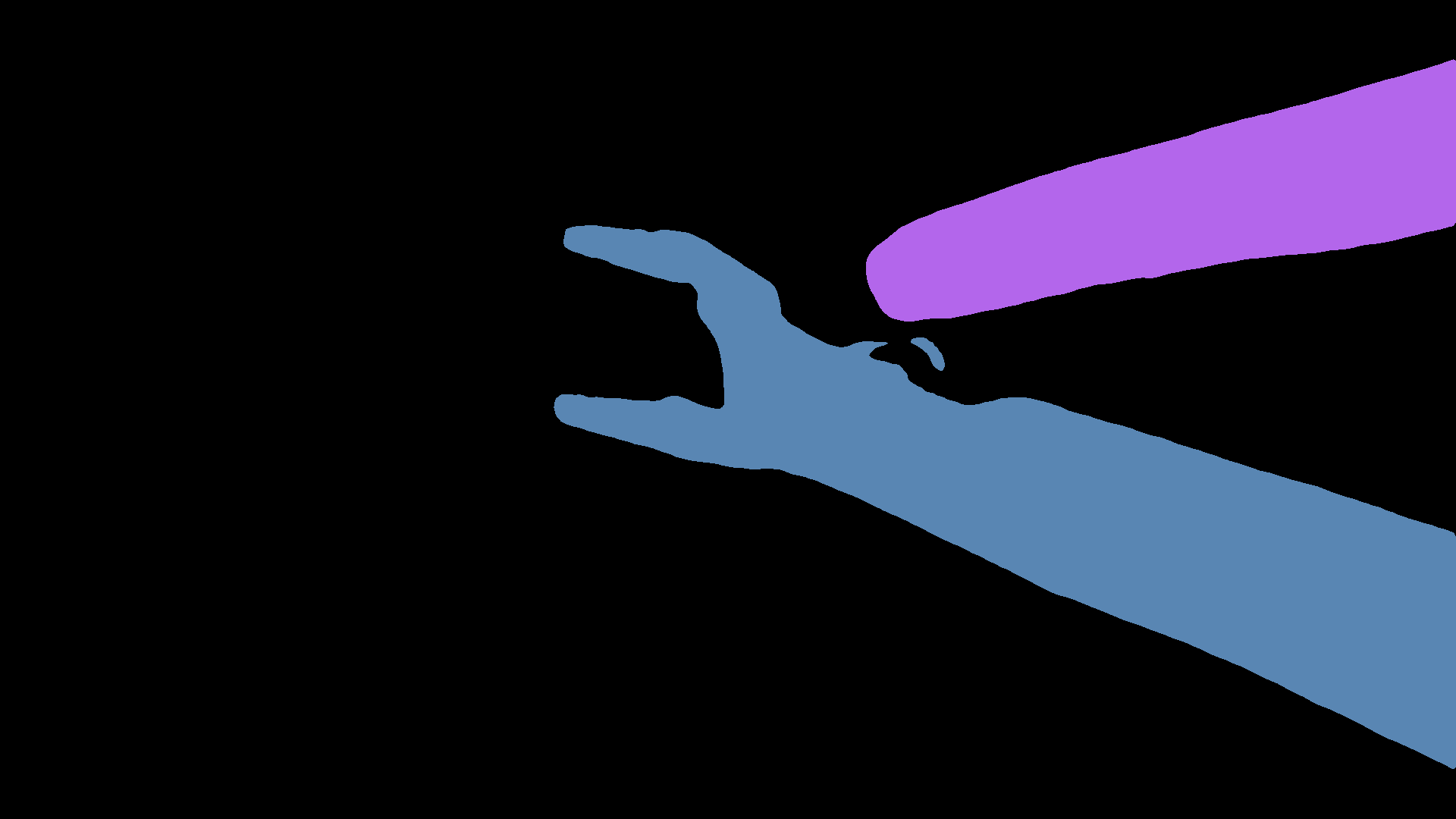}\\[-2pt]
    {\footnotesize 96.81 \(|\) 100.00 \(|\) 12.39}
\end{minipage}
&
\begin{minipage}{0.165\textwidth}
    \centering
    \includegraphics[width=\linewidth]{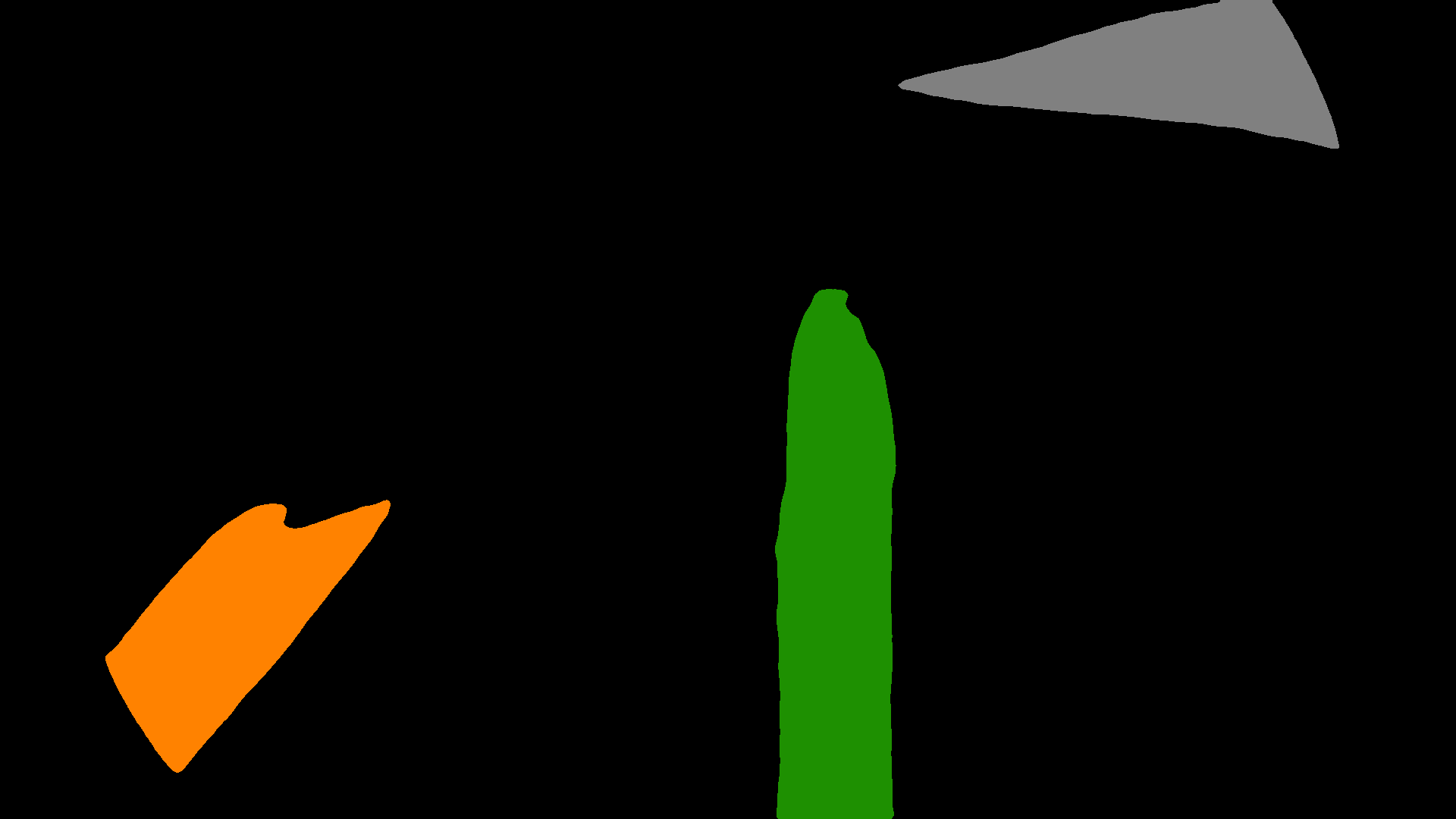}\\[-2pt]
    {\footnotesize 94.36 \(|\) 100.00 \(|\) 24.06}
\end{minipage}
\\
\multirow{2}{*}{uniandes24} &
\begin{minipage}{0.165\textwidth}
    \centering
    \includegraphics[width=\linewidth]{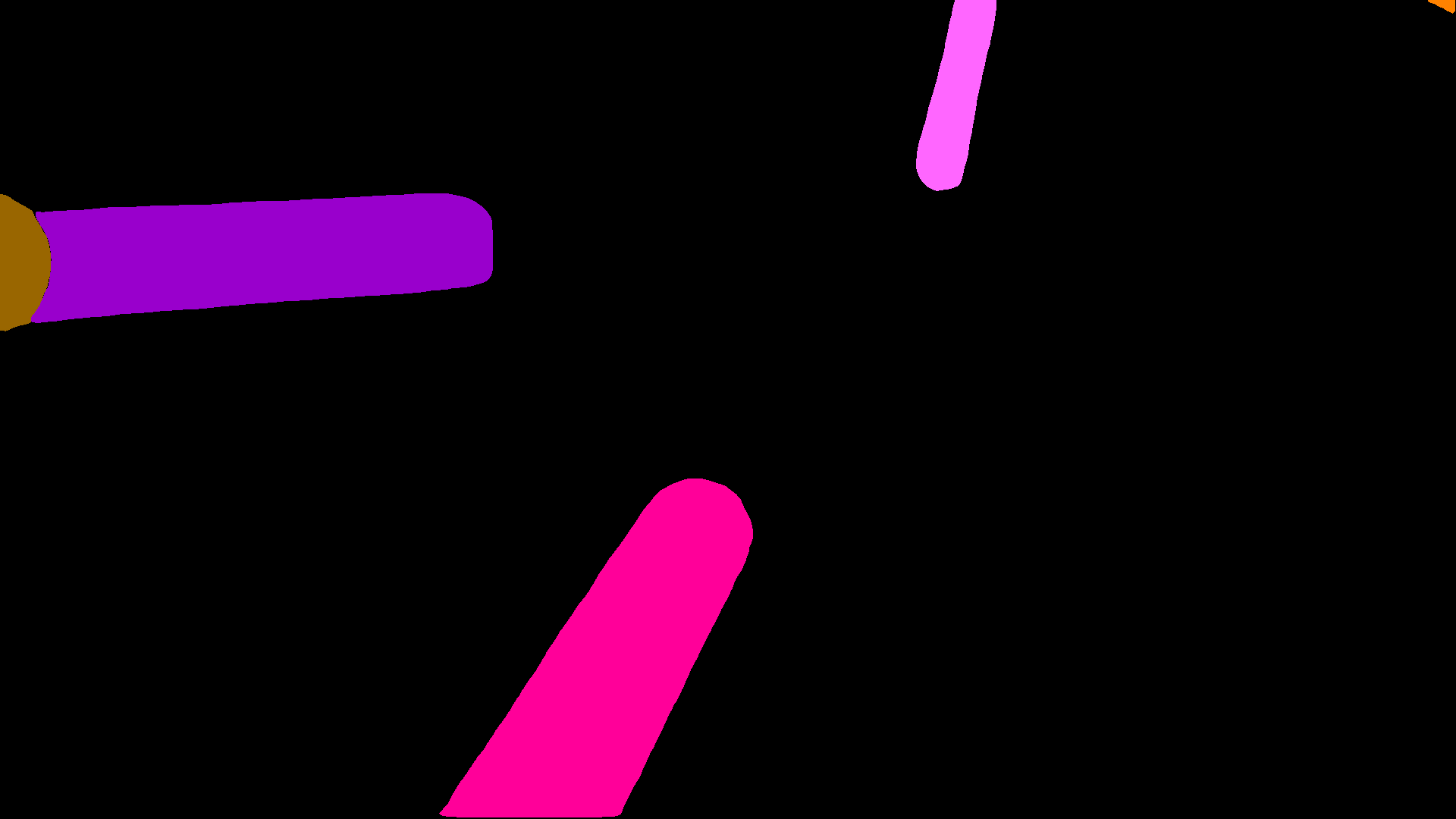}\\[-2pt]
    {\footnotesize 73.62 \(|\) 80.00 \(|\) 16.34}
\end{minipage}
&
\begin{minipage}{0.165\textwidth}
    \centering
    \includegraphics[width=\linewidth]{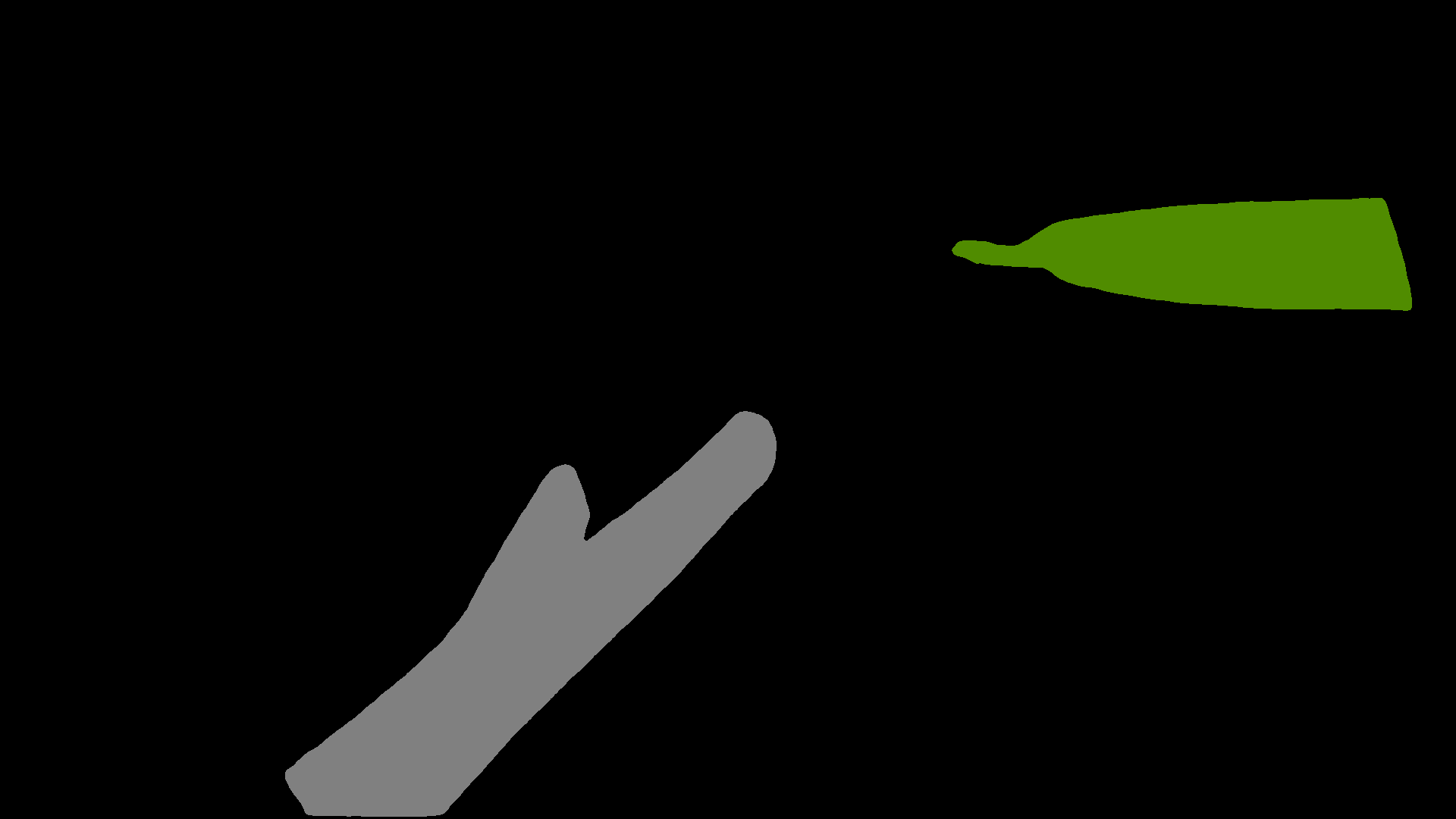}\\[-2pt]
    {\footnotesize 96.77 \(|\) 100.00 \(|\) 8.50}
\end{minipage}
&
\begin{minipage}{0.165\textwidth}
    \centering
    \includegraphics[width=\linewidth]{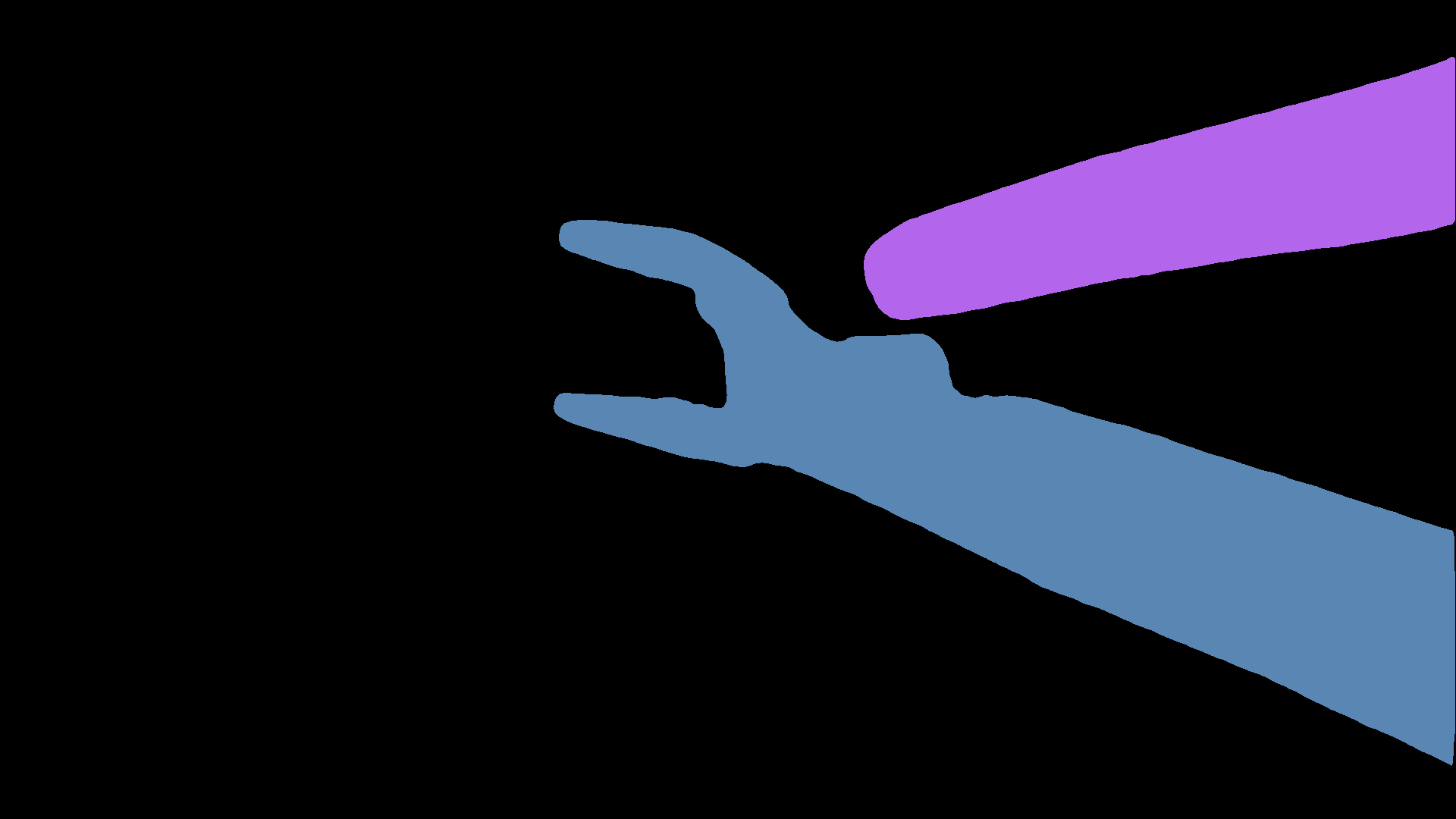}\\[-2pt]
    {\footnotesize 97.09 \(|\) 100.00 \(|\) 17.66}
\end{minipage}
&
\begin{minipage}{0.165\textwidth}
    \centering
    \includegraphics[width=\linewidth]{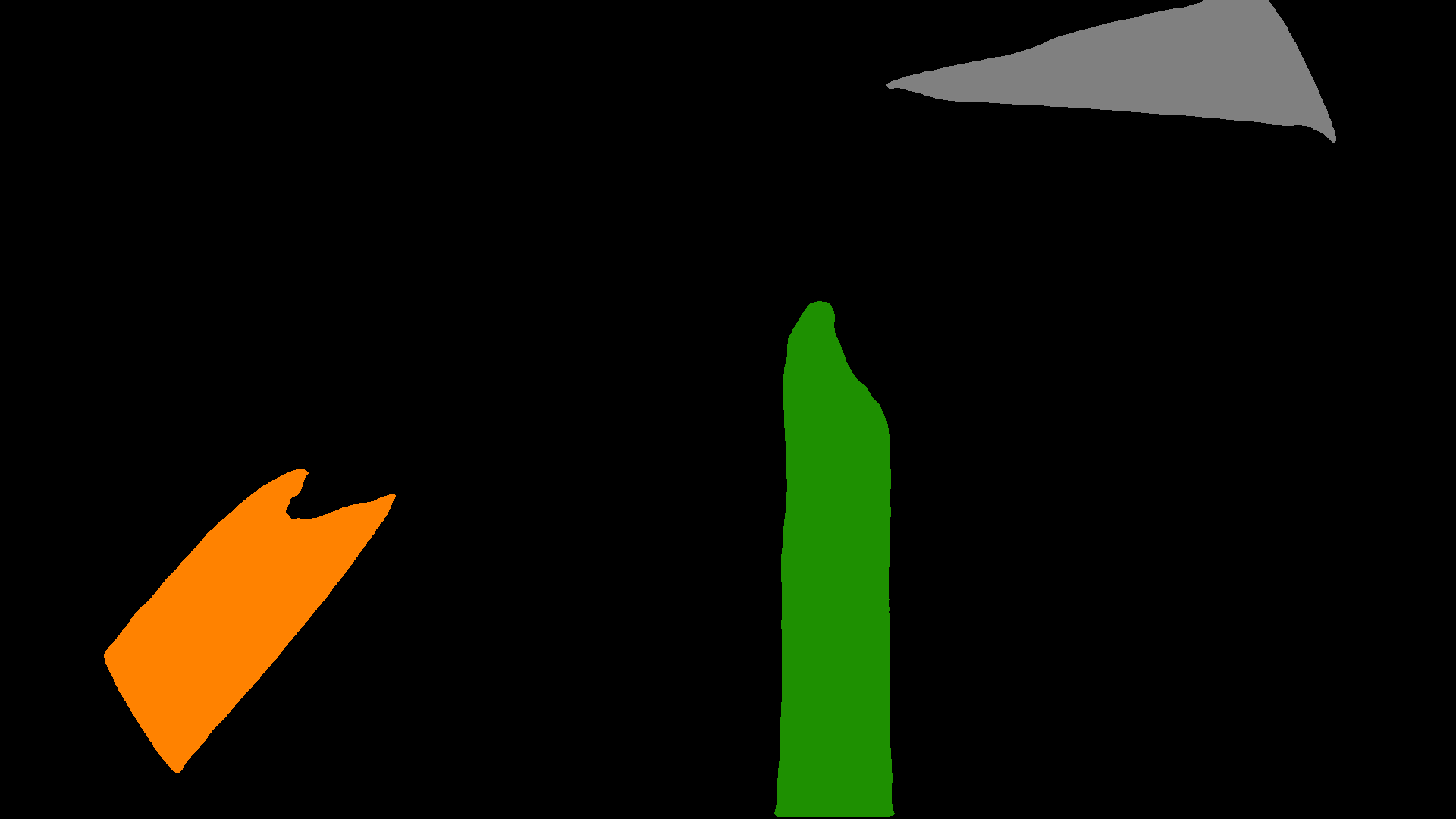}\\[-2pt]
    {\footnotesize 94.02 \(|\) 100.00 \(|\) 31.20}
\end{minipage}
\\
\multirow{2}{*}{augi} &
\begin{minipage}{0.165\textwidth}
    \centering
    \includegraphics[width=\linewidth]{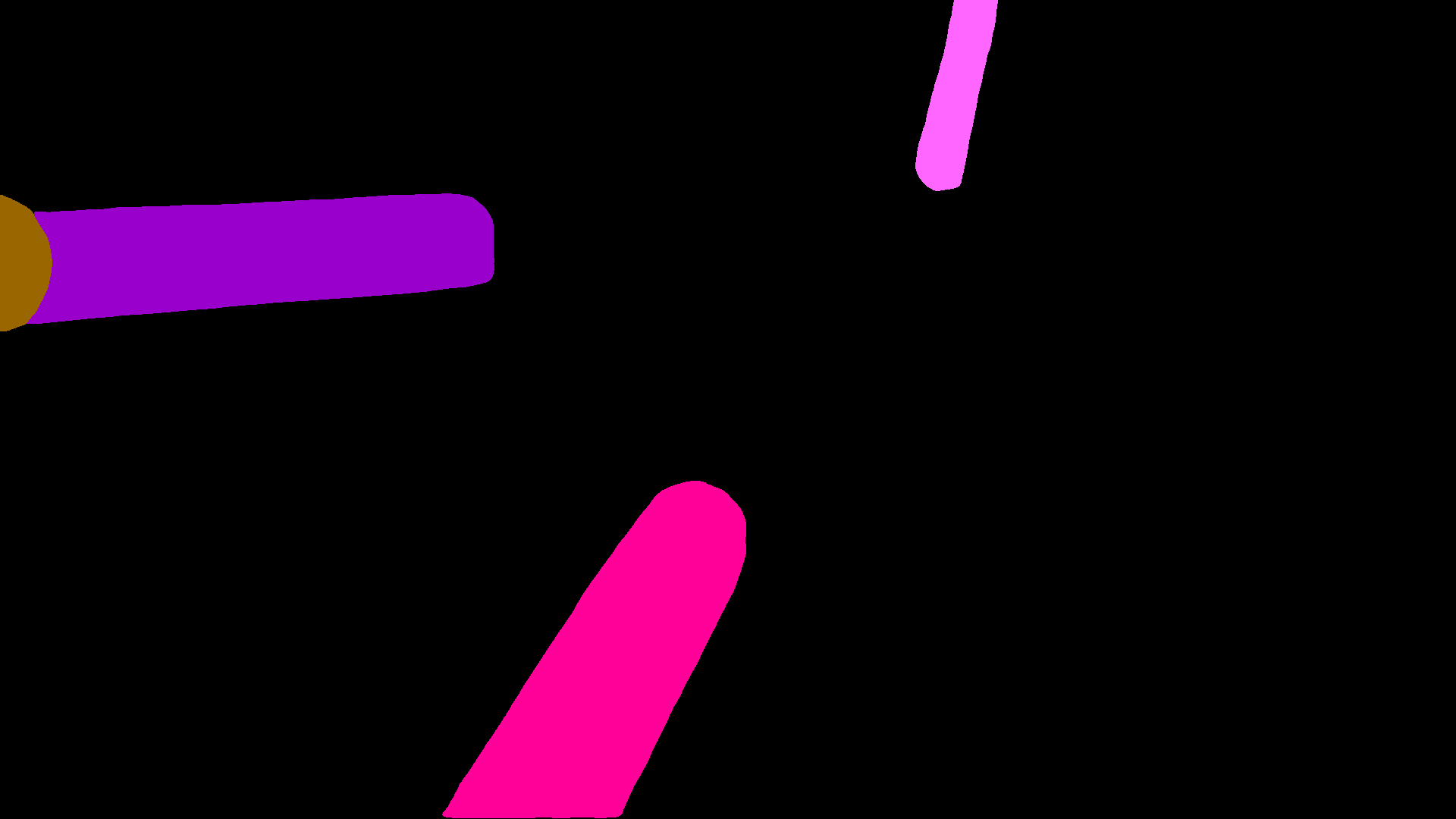}\\[-2pt]
    {\footnotesize 92.12 \(|\) 100.00 \(|\) 18.09}
\end{minipage}
&
\begin{minipage}{0.165\textwidth}
    \centering
    \includegraphics[width=\linewidth]{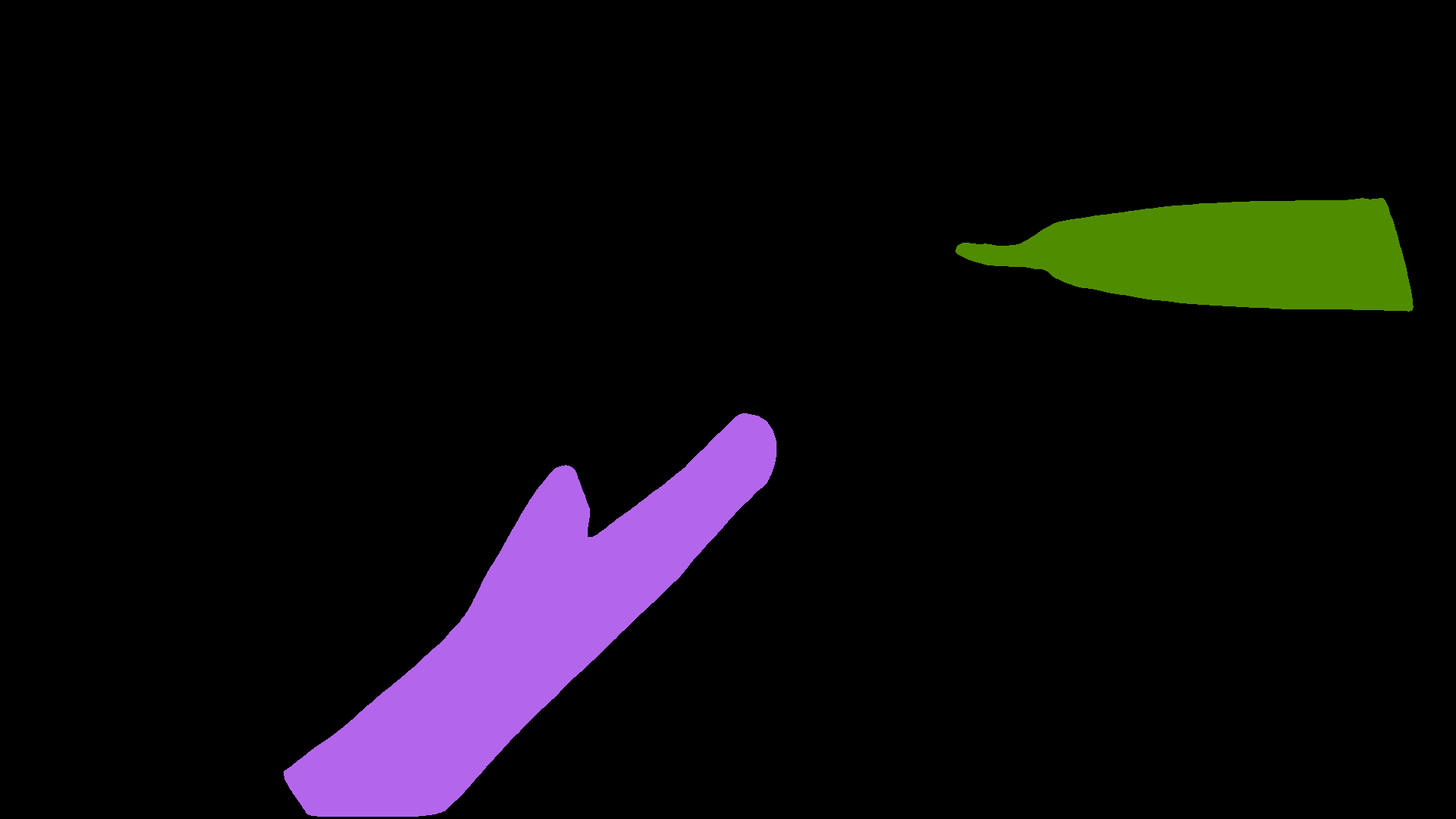}\\[-2pt]
    {\footnotesize 32.28 \(|\) 33.33 \(|\) 9.33}
\end{minipage}
&
\begin{minipage}{0.165\textwidth}
    \centering
    \includegraphics[width=\linewidth]{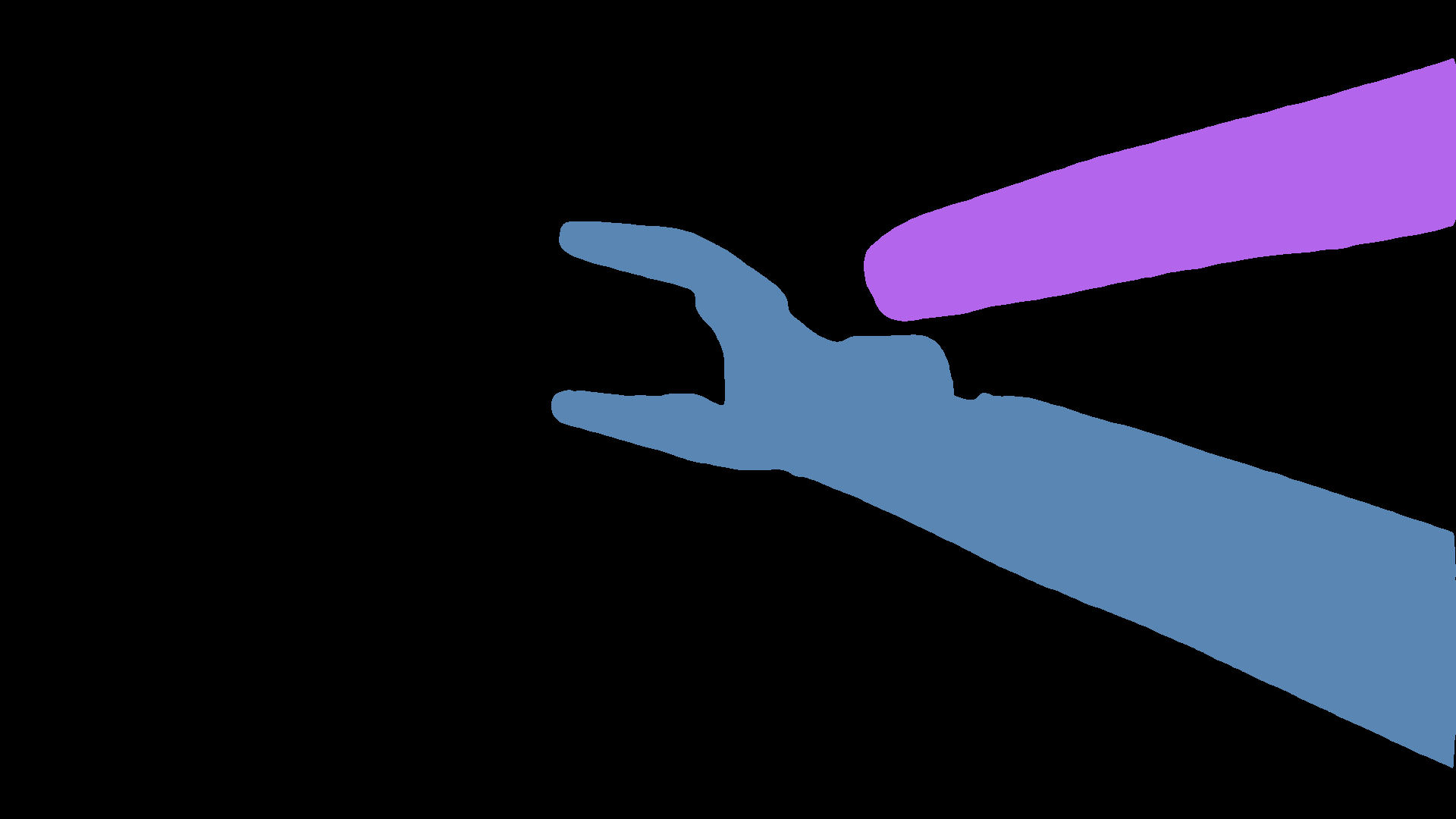}\\[-2pt]
    {\footnotesize 97.50 \(|\) 100.00 \(|\) 9.66}
\end{minipage}
&
\begin{minipage}{0.165\textwidth}
    \centering
    \includegraphics[width=\linewidth]{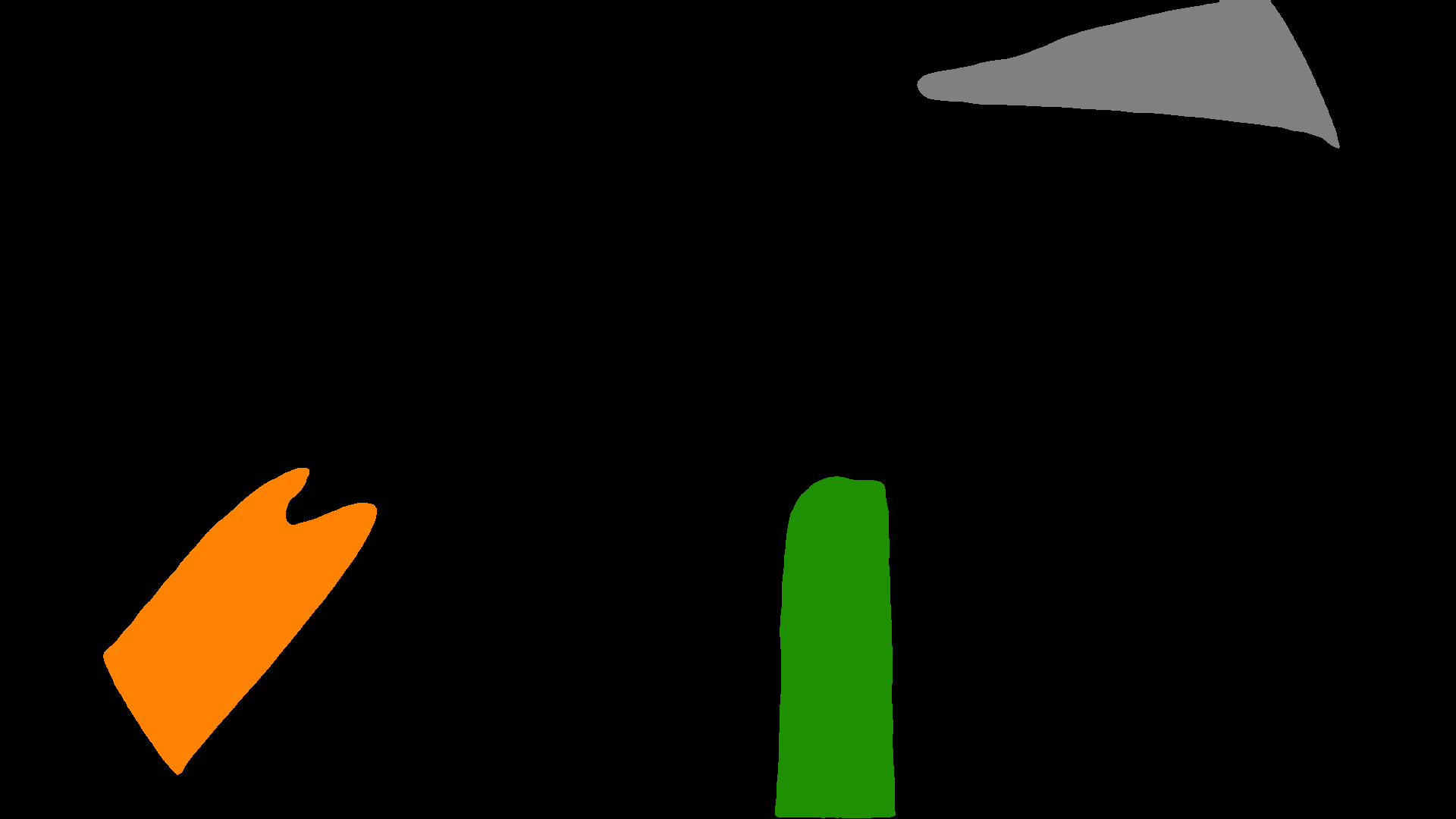}\\[-2pt]
    {\footnotesize 88.97 \(|\) 100.00 \(|\) 95.51}
\end{minipage}
\\
\multirow{2}{*}{kist\_harilab} &
\begin{minipage}{0.165\textwidth}
    \centering
    \includegraphics[width=\linewidth]{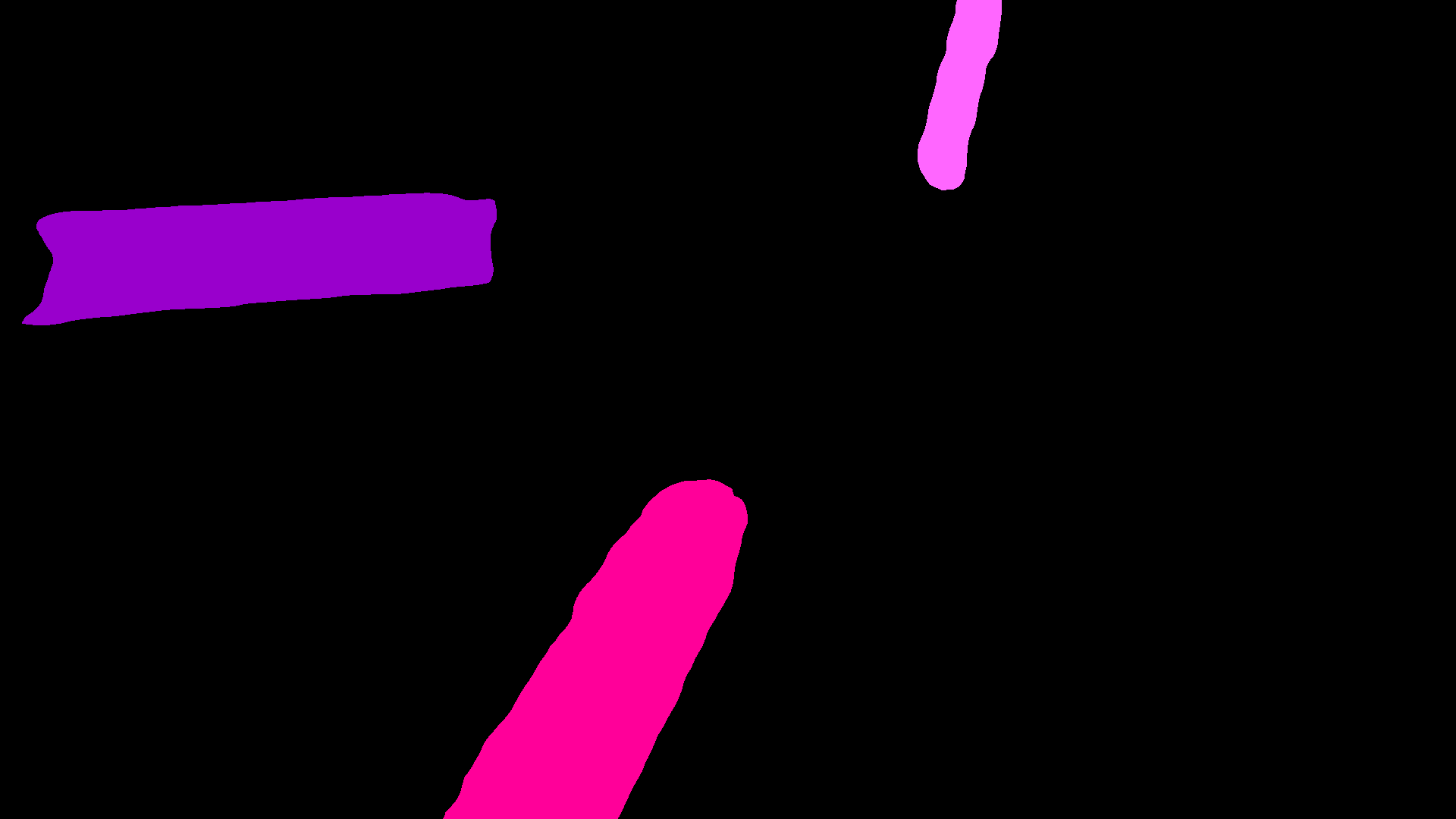}\\[-2pt]
    {\footnotesize 69.38 \(|\) 75.00 \(|\) 18.36}
\end{minipage}
&
\begin{minipage}{0.165\textwidth}
    \centering
    \includegraphics[width=\linewidth]{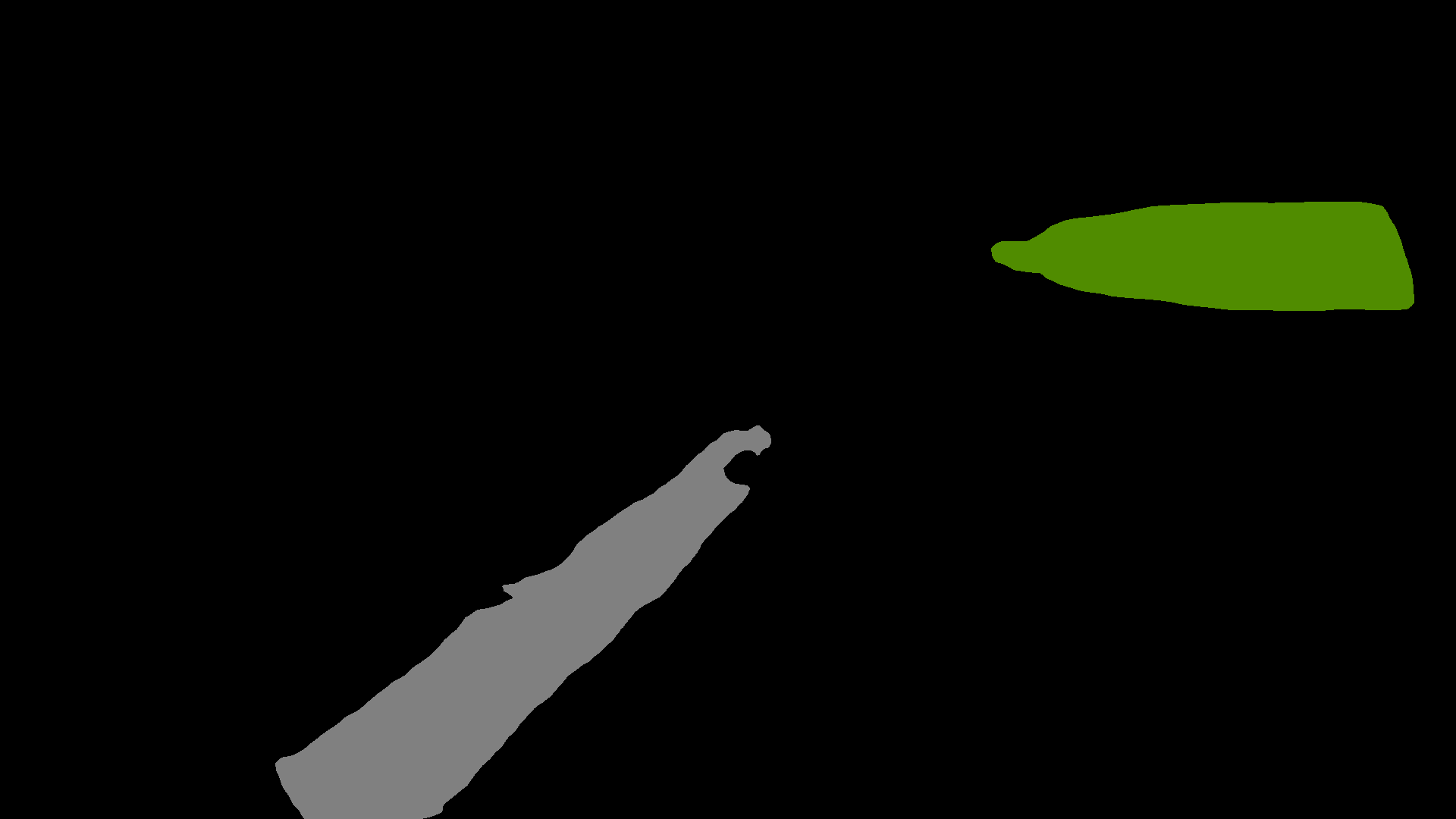}\\[-2pt]
    {\footnotesize 91.57 \(|\) 100.00 \(|\) 34.18}
\end{minipage}
&
\begin{minipage}{0.165\textwidth}
    \centering
    \includegraphics[width=\linewidth]{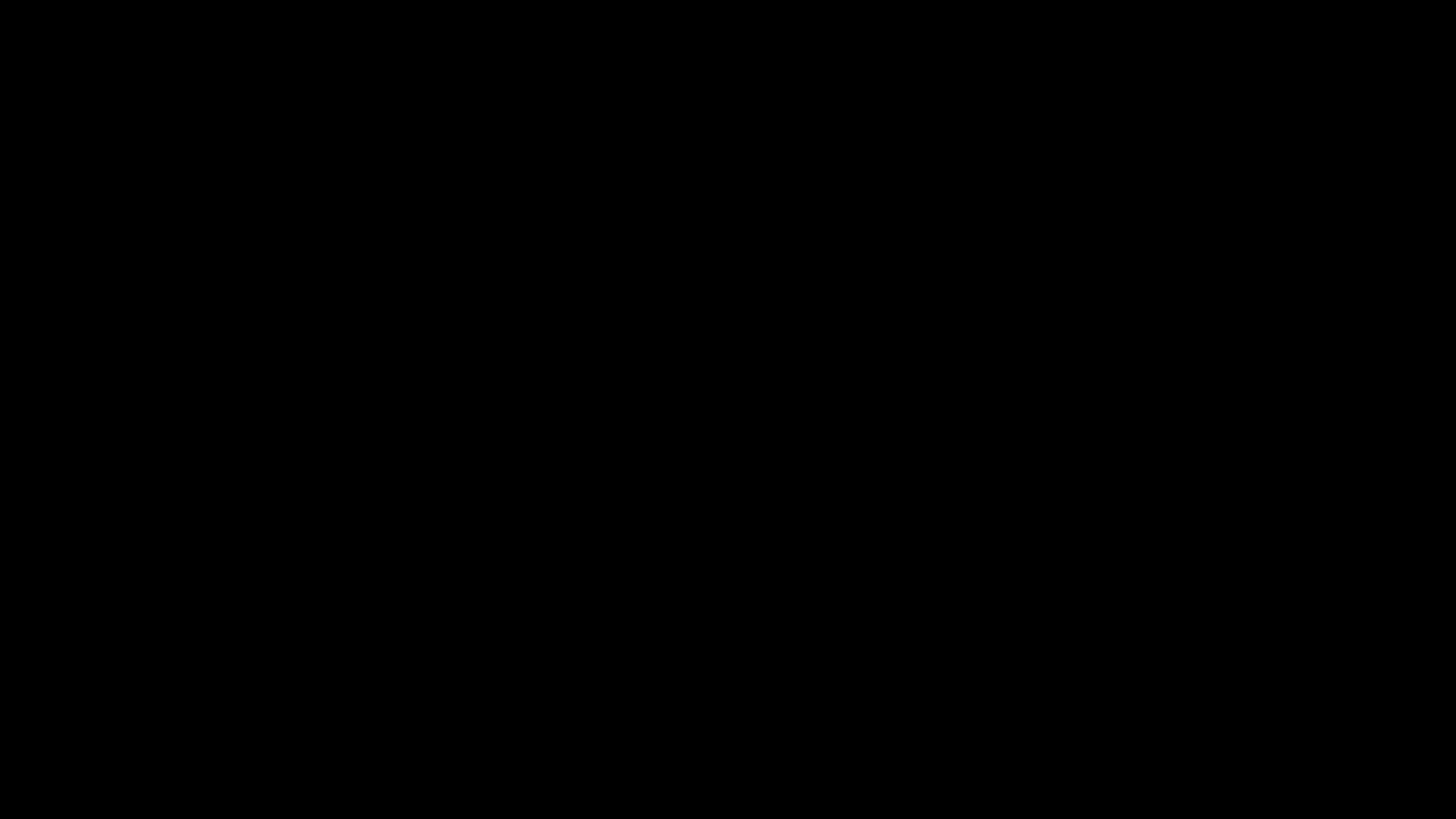}\\[-2pt]
    {\footnotesize 0.00 \(|\) 0.00 \(|\) nan}
\end{minipage}
&
\begin{minipage}{0.165\textwidth}
    \centering
    \includegraphics[width=\linewidth]{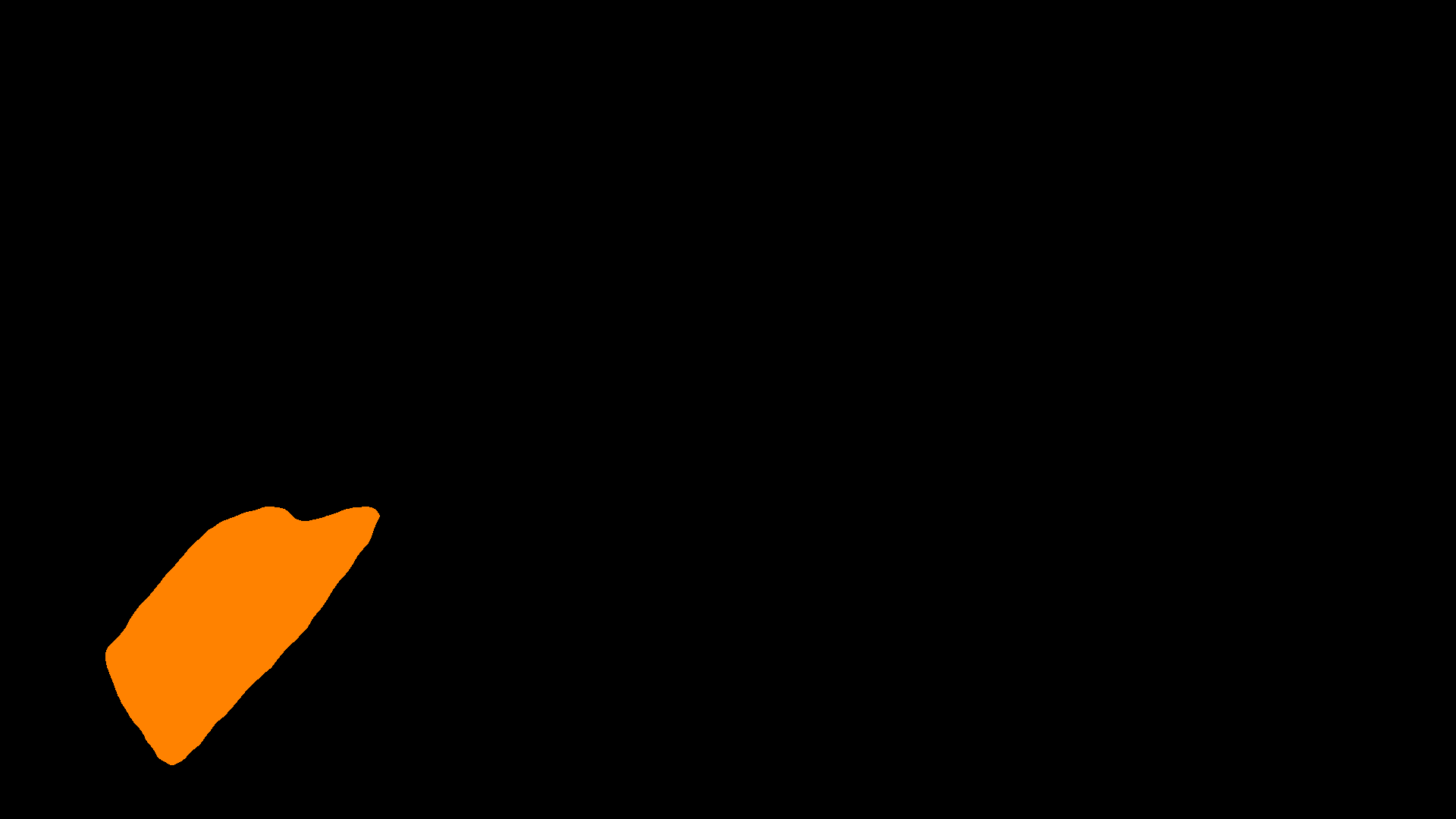}\\[-2pt]
    {\footnotesize 30.72 \(|\) 33.33 \(|\) 30.53}
\end{minipage}
\\
\multirow{2}{*}{floor9} &
\begin{minipage}{0.165\textwidth}
    \centering
    \includegraphics[width=\linewidth]{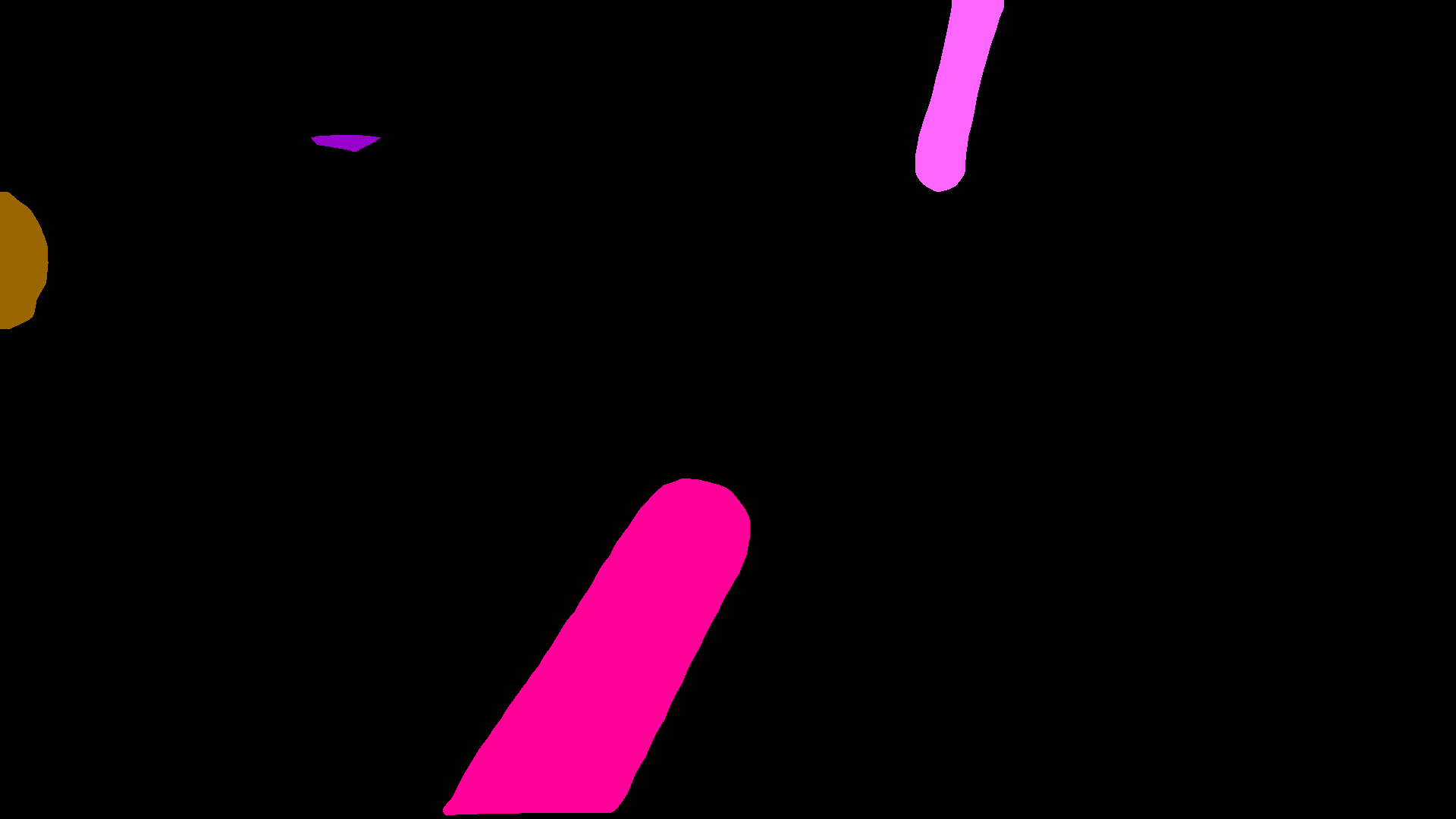}\\[-2pt]
    {\footnotesize 67.61 \(|\) 75.00 \(|\) 111.06}
\end{minipage}
&
\begin{minipage}{0.165\textwidth}
    \centering
    \includegraphics[width=\linewidth]{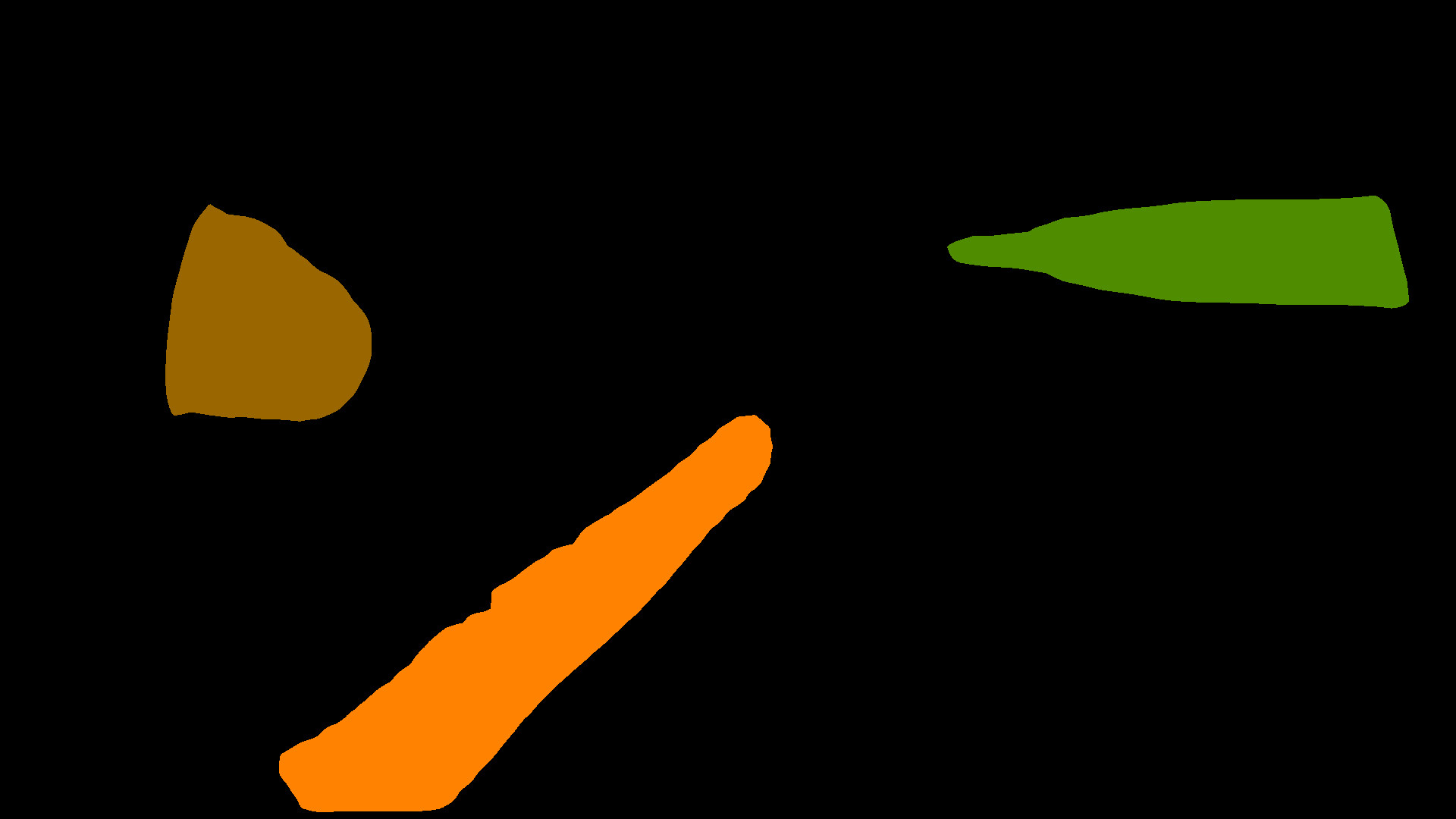}\\[-2pt]
    {\footnotesize 23.46 \(|\) 25.00 \(|\) 33.02}
\end{minipage}
&
\begin{minipage}{0.165\textwidth}
    \centering
    \includegraphics[width=\linewidth]{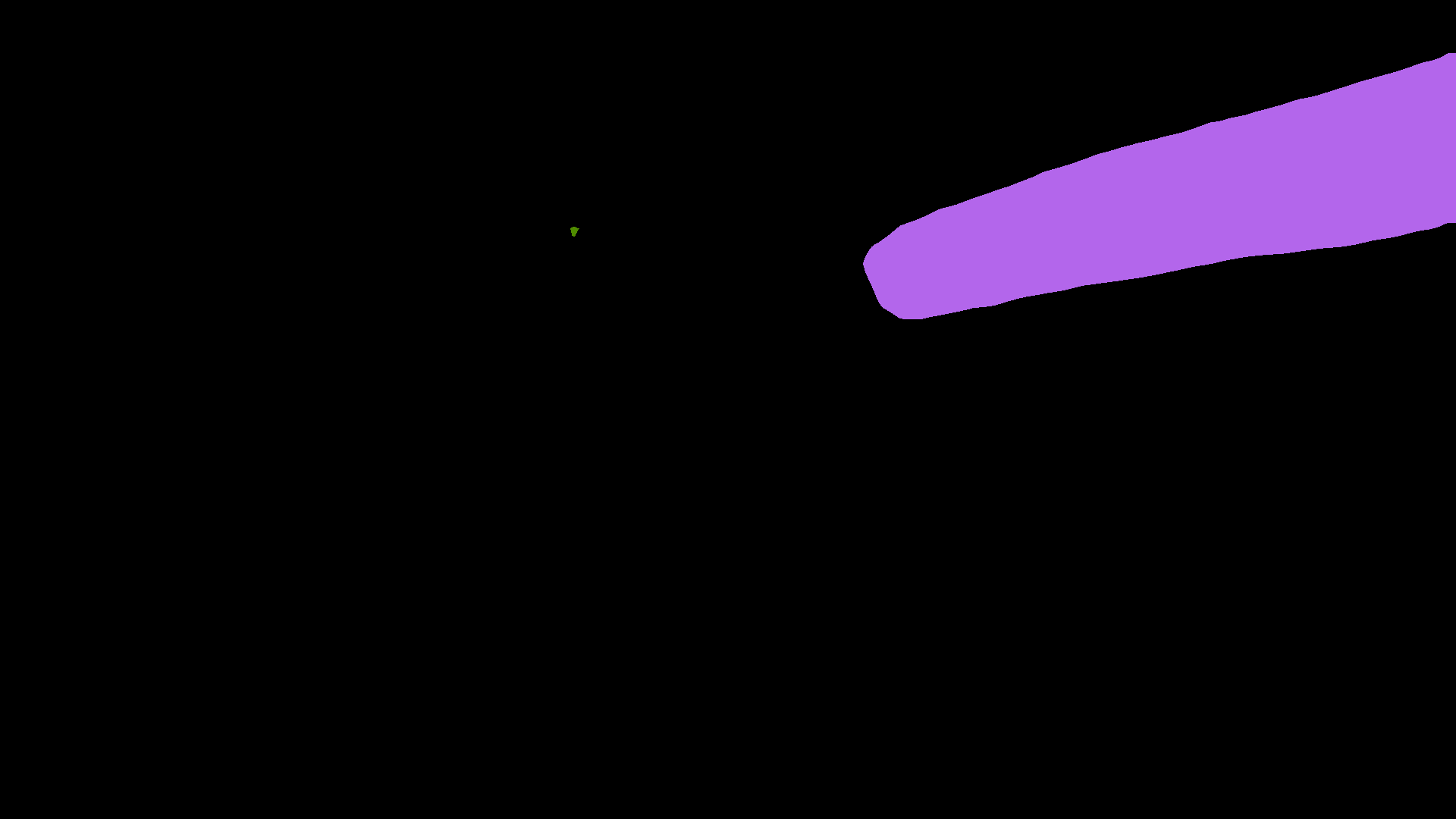}\\[-2pt]
    {\footnotesize 32.47 \(|\) 33.33 \(|\) 611.79}
\end{minipage}
&
\begin{minipage}{0.165\textwidth}
    \centering
    \includegraphics[width=\linewidth]{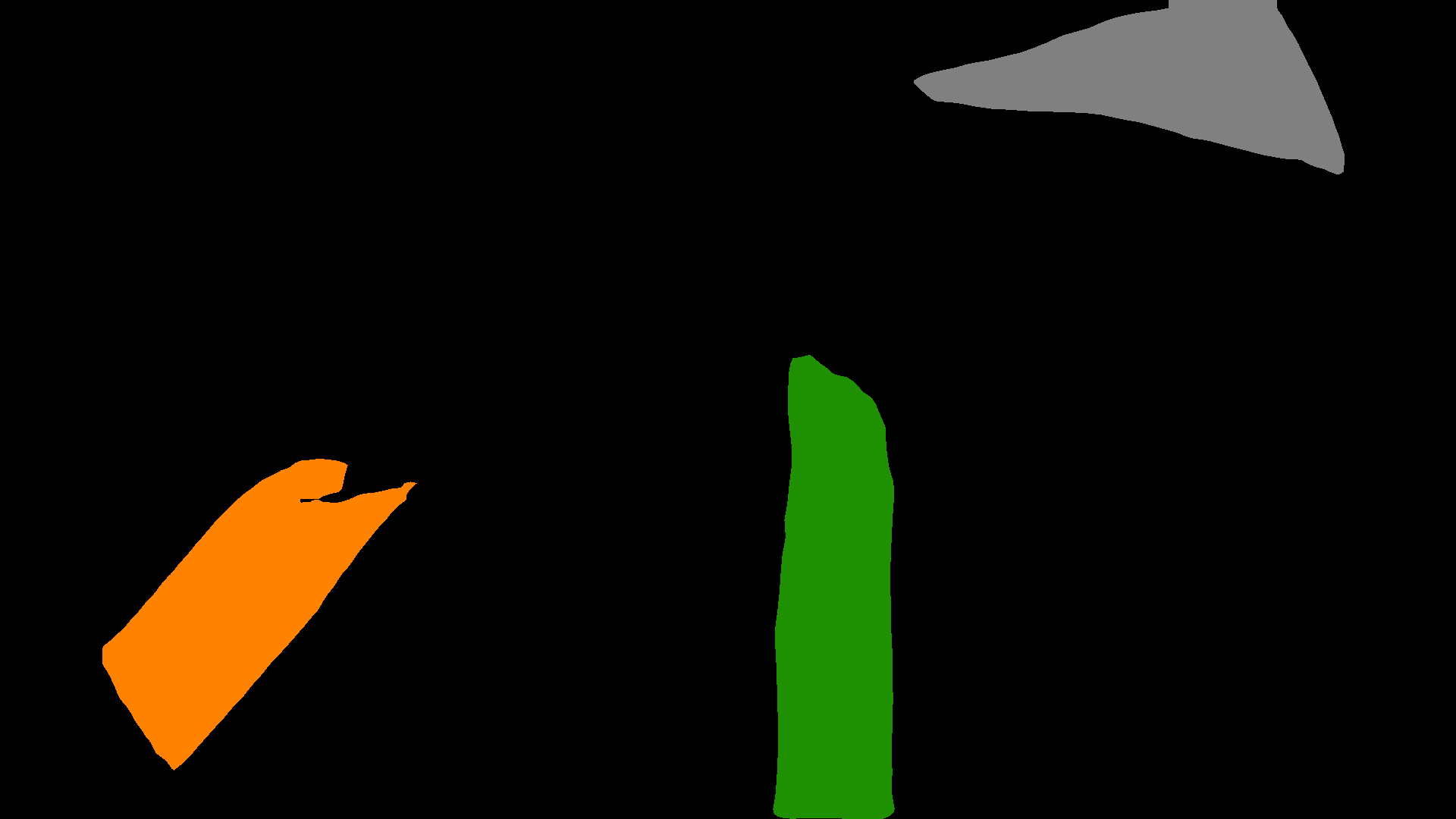}\\[-2pt]
    {\footnotesize 89.11 \(|\) 100.00 \(|\) 60.73}
\end{minipage}
\\
\multirow{2}{*}{sk} &
\begin{minipage}{0.165\textwidth}
    \centering
    \includegraphics[width=\linewidth]{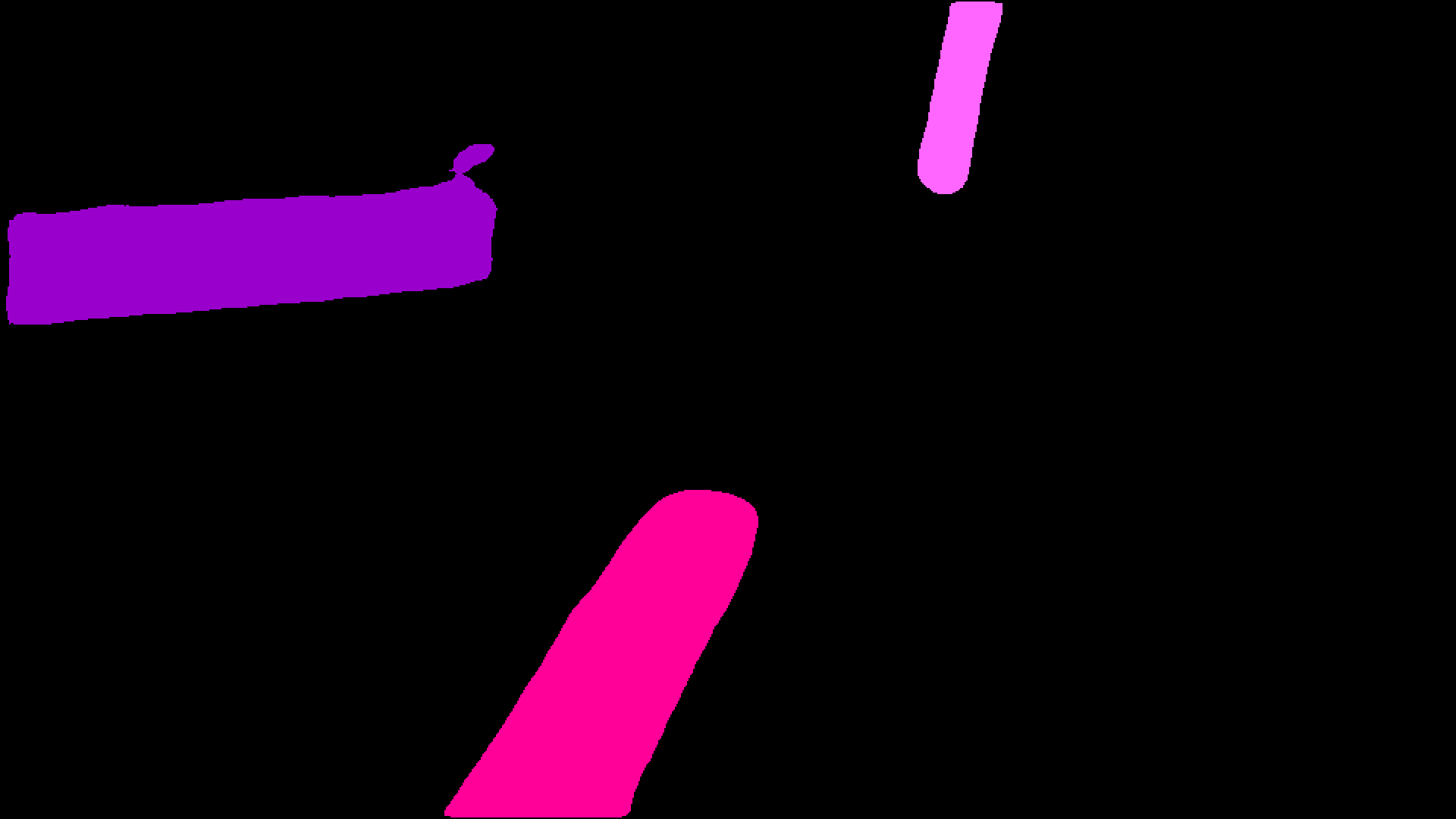}\\[-2pt]
    {\footnotesize 69.32 \(|\) 75.00 \(|\) 28.73}
\end{minipage}
&
\begin{minipage}{0.165\textwidth}
    \centering
    \includegraphics[width=\linewidth]{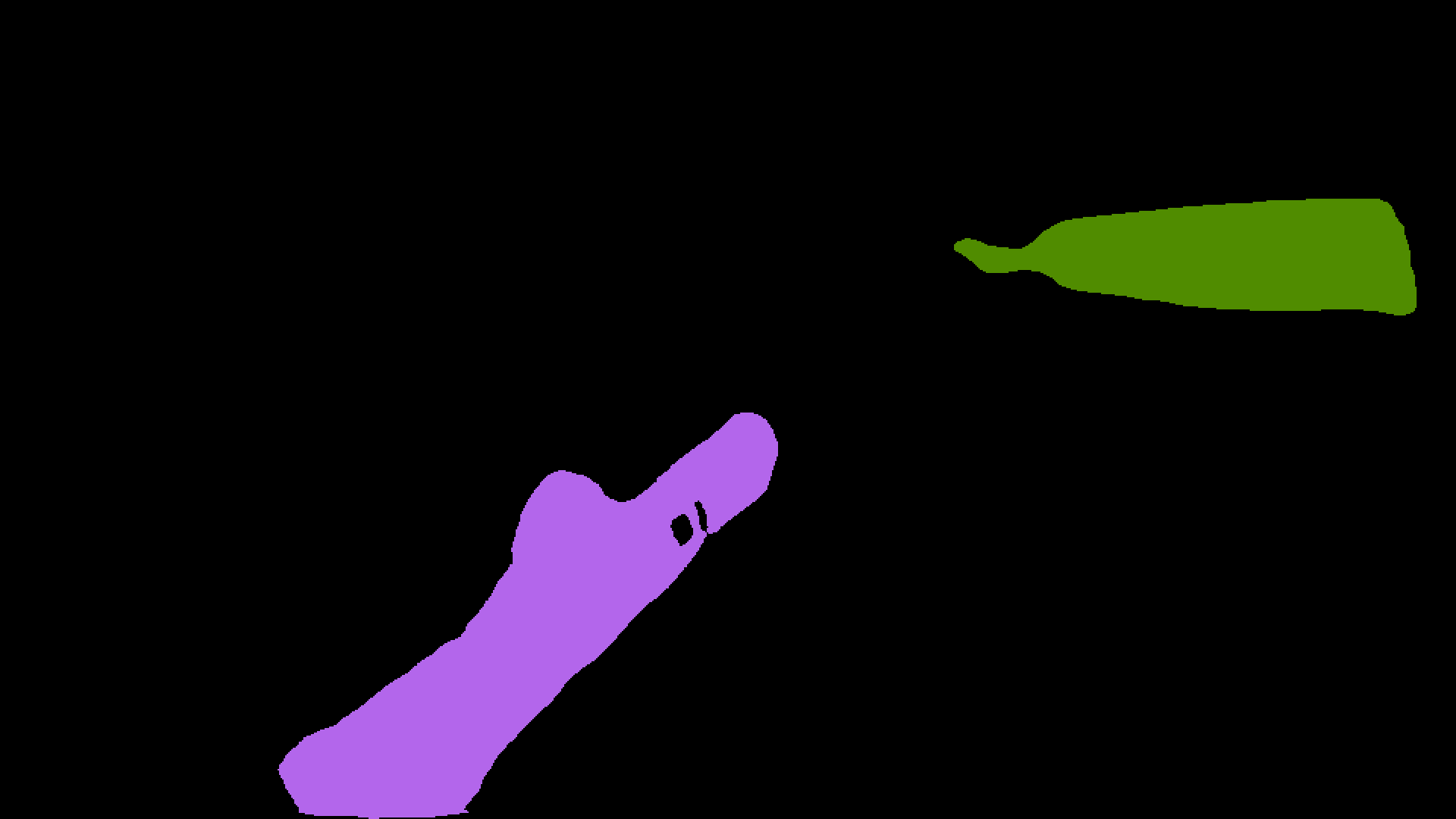}\\[-2pt]
    {\footnotesize 32.49 \(|\) 33.33 \(|\) 25.06}
\end{minipage}
&
\begin{minipage}{0.165\textwidth}
    \centering
    \includegraphics[width=\linewidth]{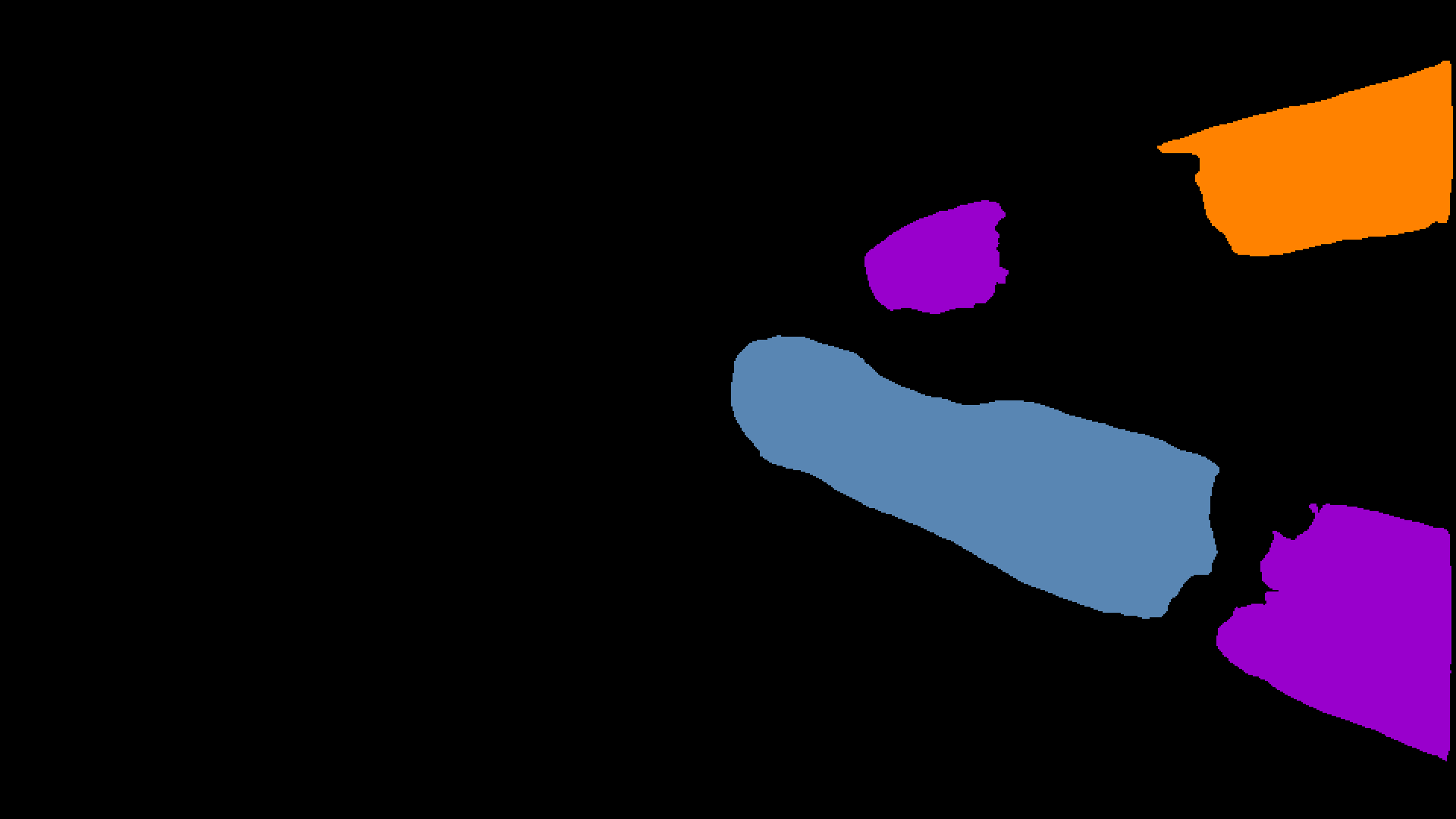}\\[-2pt]
    {\footnotesize 15.14 \(|\) 0.00 \(|\) 335.29}
\end{minipage}
&
\begin{minipage}{0.165\textwidth}
    \centering
    \includegraphics[width=\linewidth]{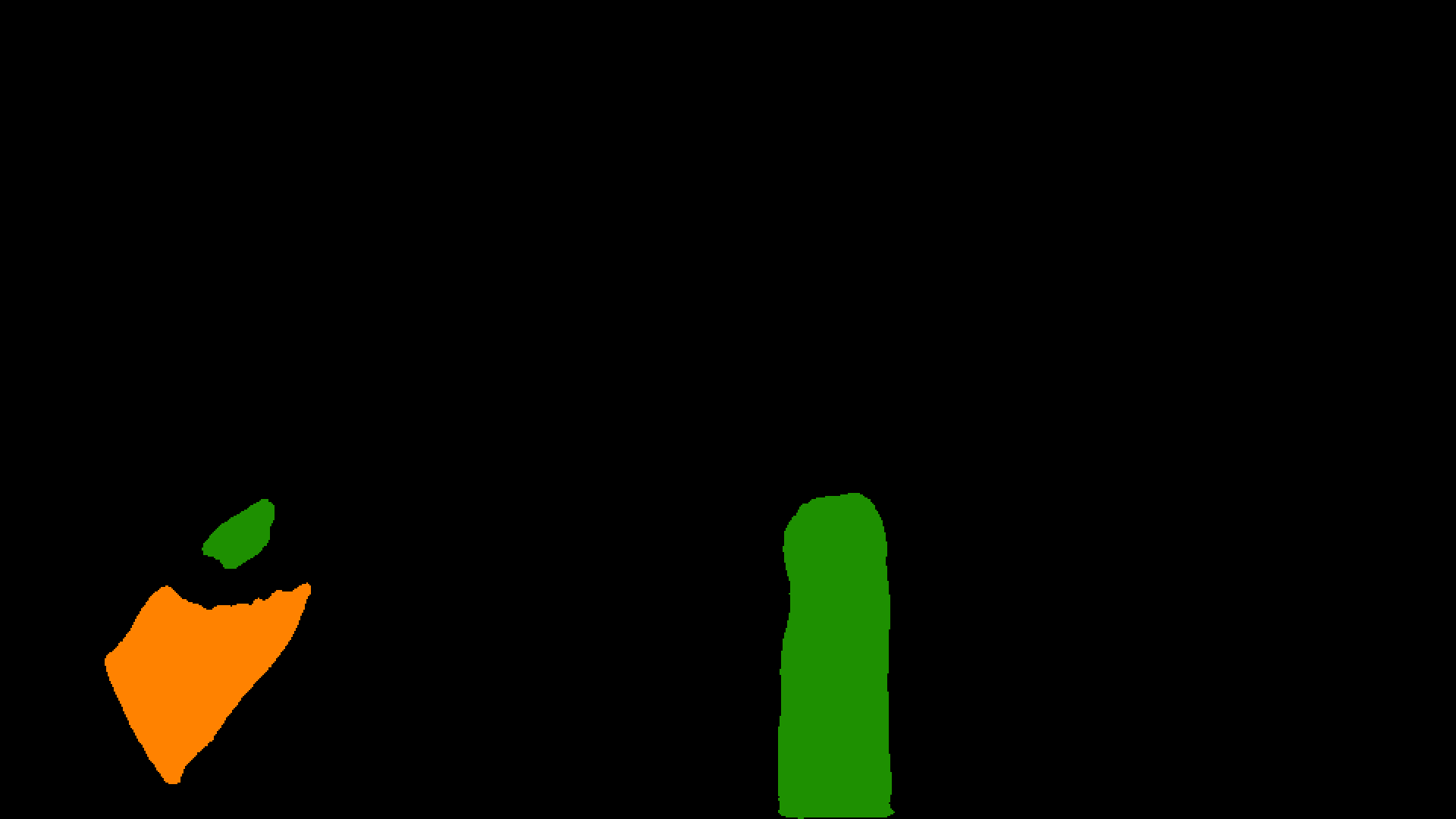}\\[-2pt]
    {\footnotesize 46.39 \(|\) 66.67 \(|\) 430.51}
\end{minipage}
\\
\multirow{2}{*}{goncalo} &
\begin{minipage}{0.165\textwidth}
    \centering
    \includegraphics[width=\linewidth]{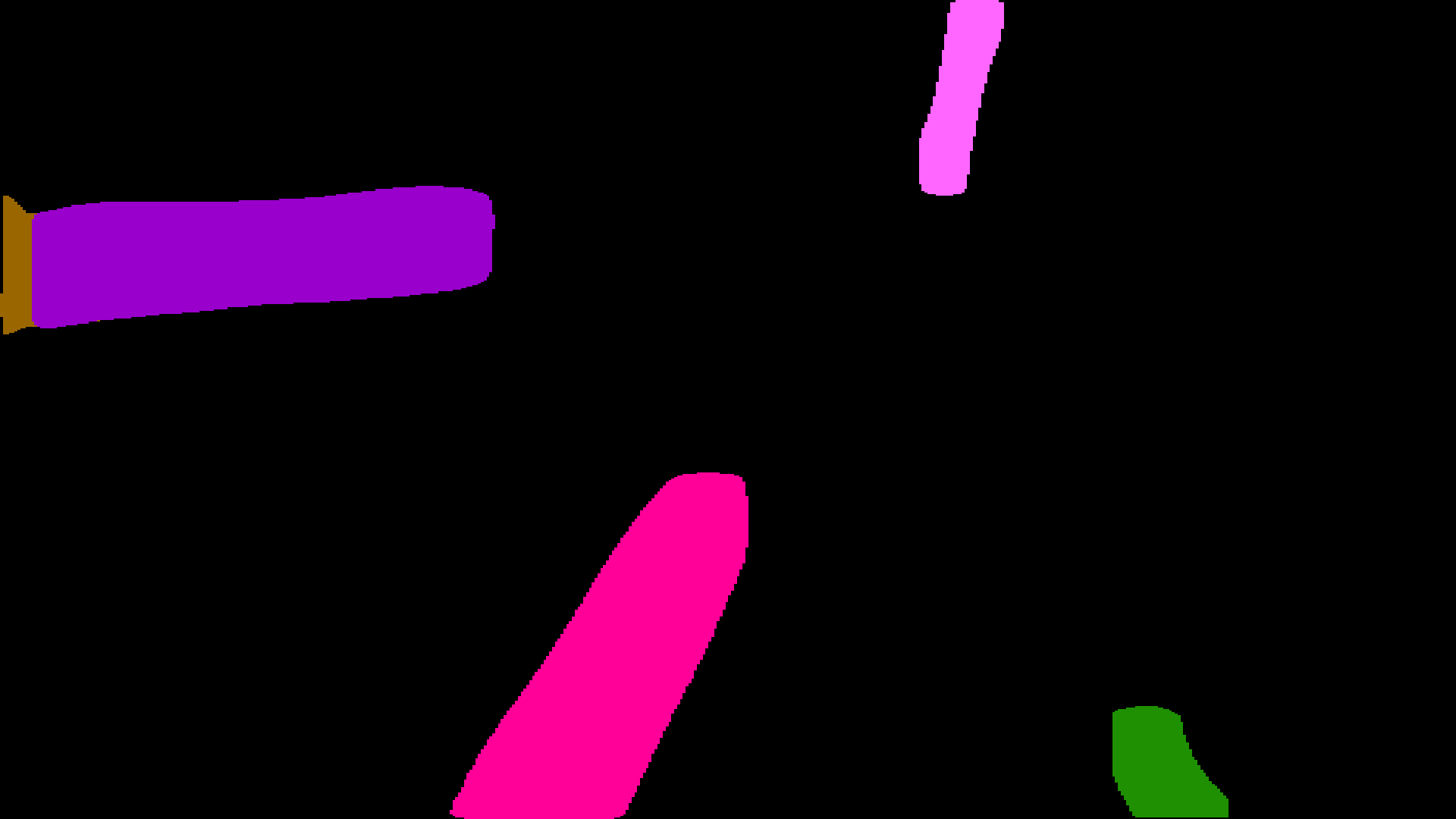}\\[-2pt]
    {\footnotesize 70.45 \(|\) 80.00 \(|\) 28.82}
\end{minipage}
&
\begin{minipage}{0.165\textwidth}
    \centering
    \includegraphics[width=\linewidth]{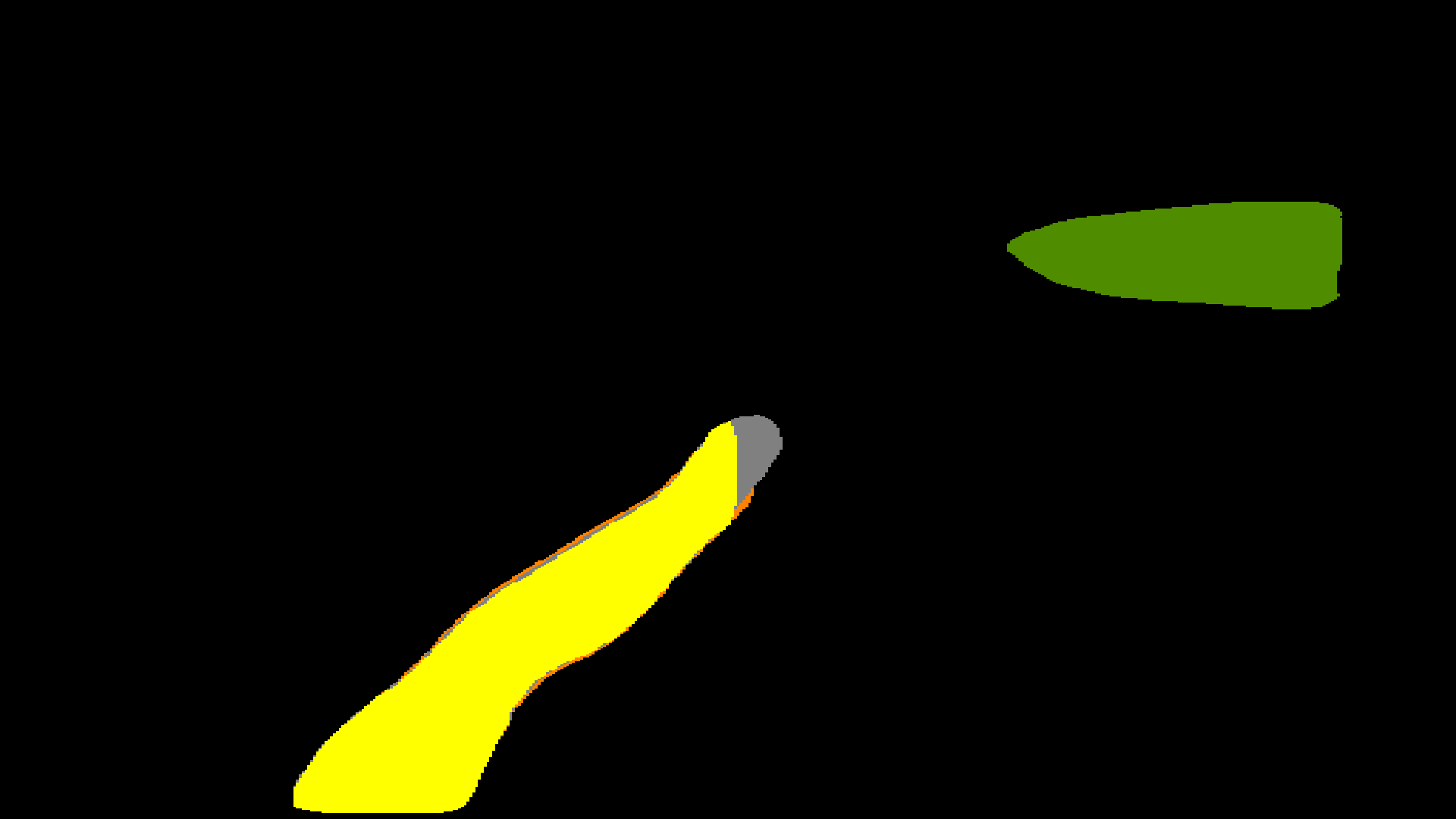}\\[-2pt]
    {\footnotesize 23.45 \(|\) 25.00 \(|\) 70.35}
\end{minipage}
&
\begin{minipage}{0.165\textwidth}
    \centering
    \includegraphics[width=\linewidth]{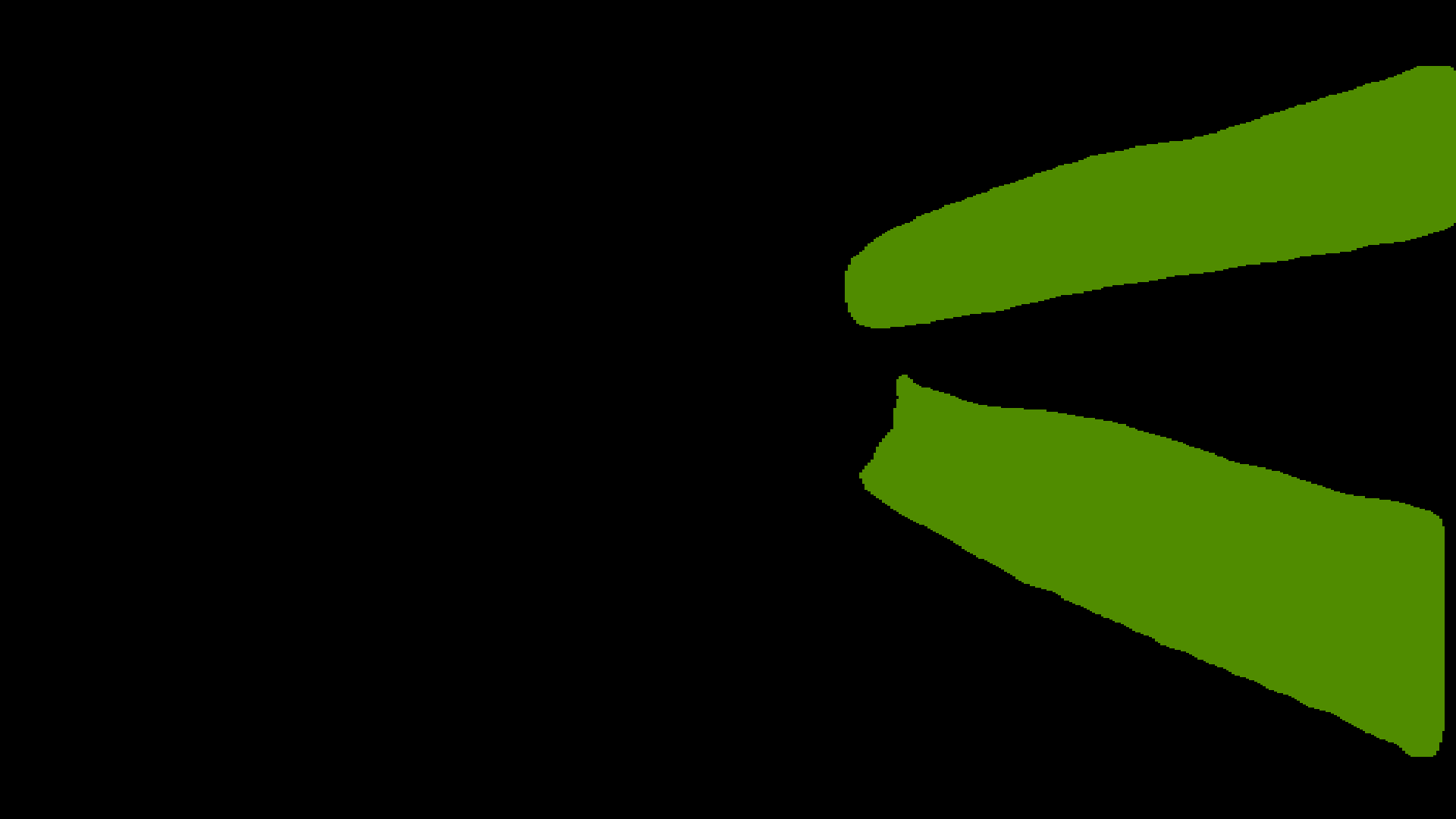}\\[-2pt]
    {\footnotesize 0.00 \(|\) 0.00 \(|\) 407.28}
\end{minipage}
&
\begin{minipage}{0.165\textwidth}
    \centering
    \includegraphics[width=\linewidth]{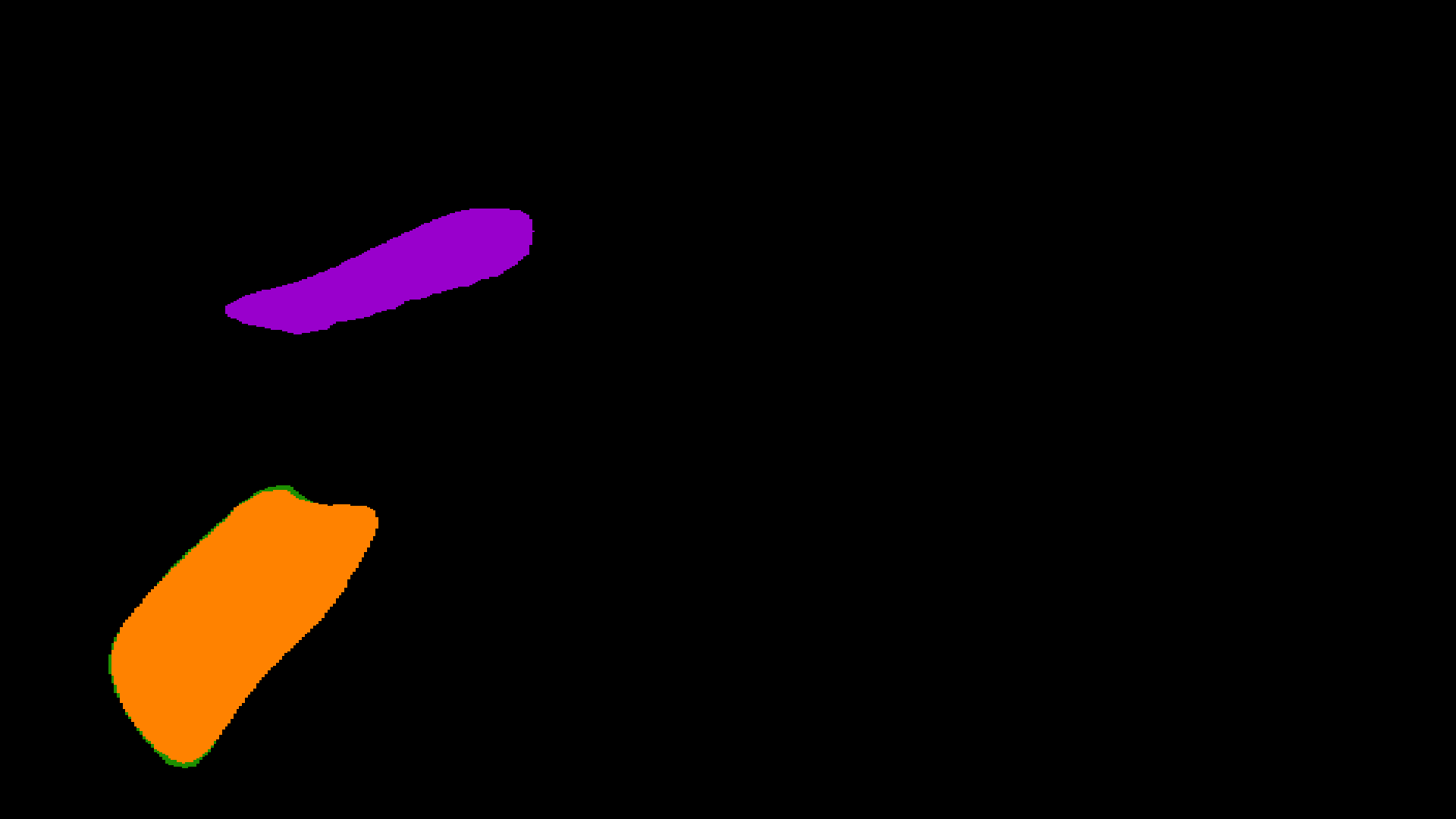}\\[-2pt]
    {\footnotesize 22.53 \(|\) 25.00 \(|\) 650.25}
\end{minipage}
\\
\multirow{2}{*}{hanglok} &
\begin{minipage}{0.165\textwidth}
    \centering
    \includegraphics[width=\linewidth]{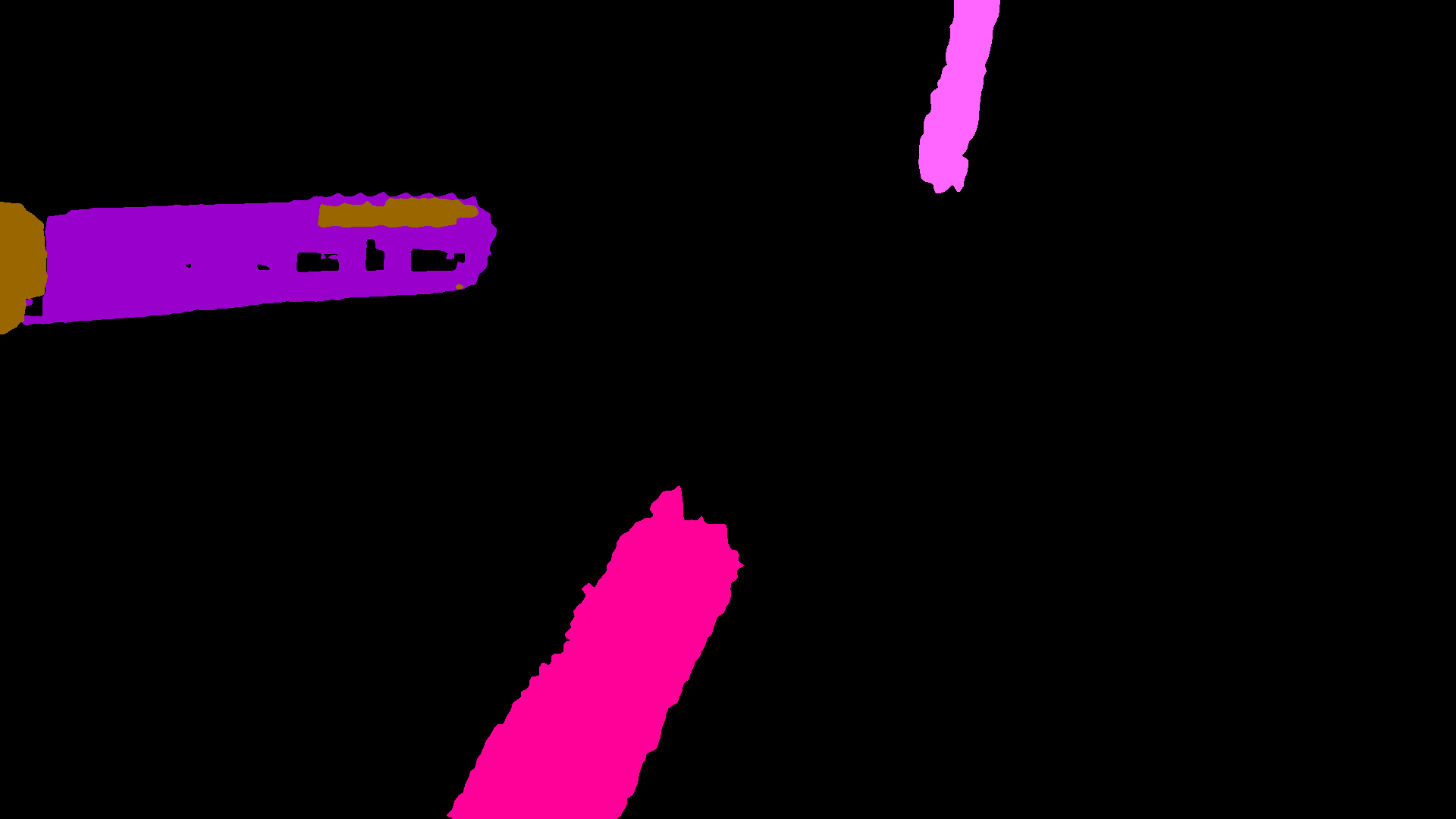}\\[-2pt]
    {\footnotesize 83.37 \(|\) 100.00 \(|\) 165.37}
\end{minipage}
&
\begin{minipage}{0.165\textwidth}
    \centering
    \includegraphics[width=\linewidth]{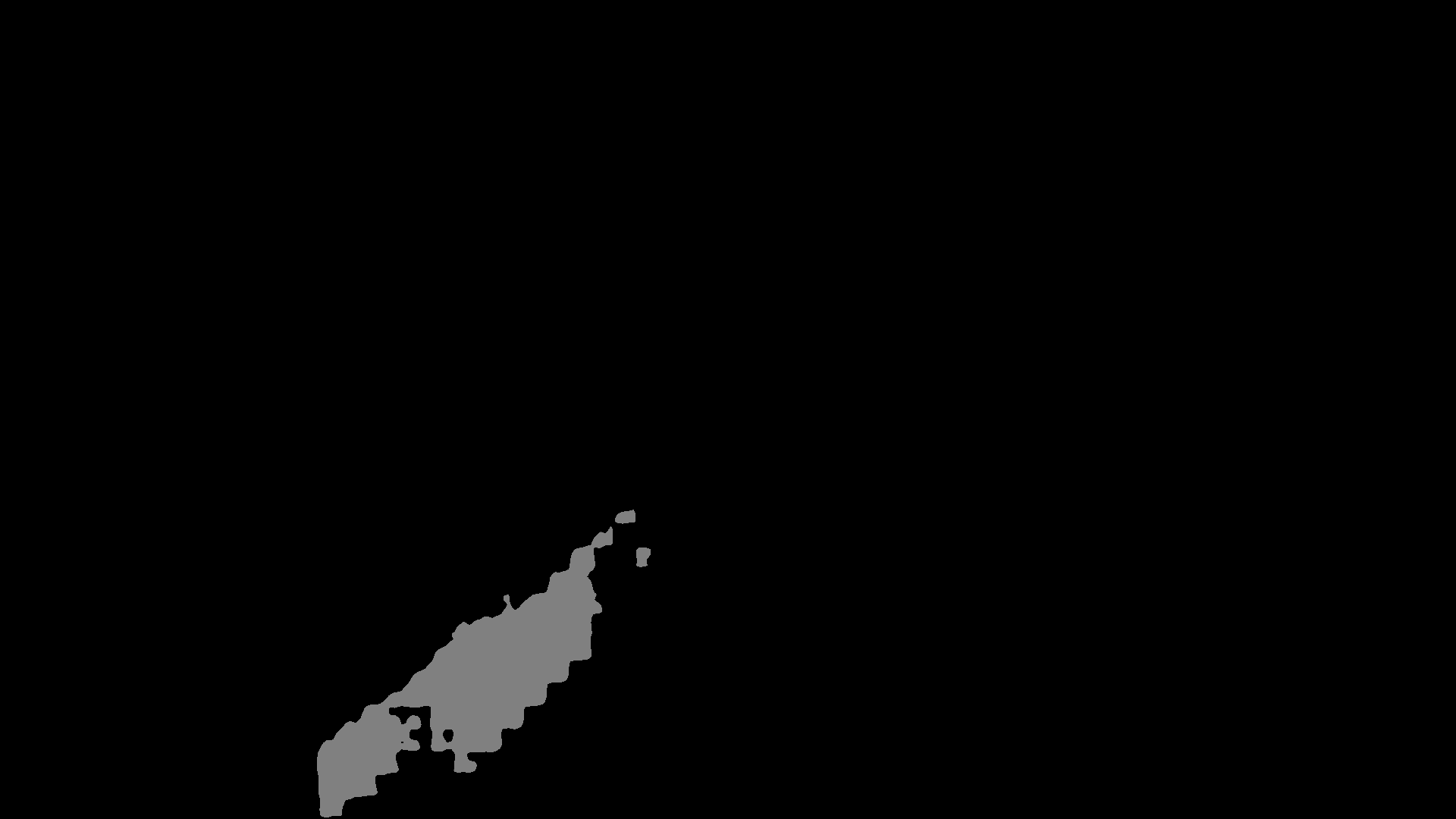}\\[-2pt]
    {\footnotesize 29.28 \(|\) 0.00 \(|\) 153.79}
\end{minipage}
&
\begin{minipage}{0.165\textwidth}
    \centering
    \includegraphics[width=\linewidth]{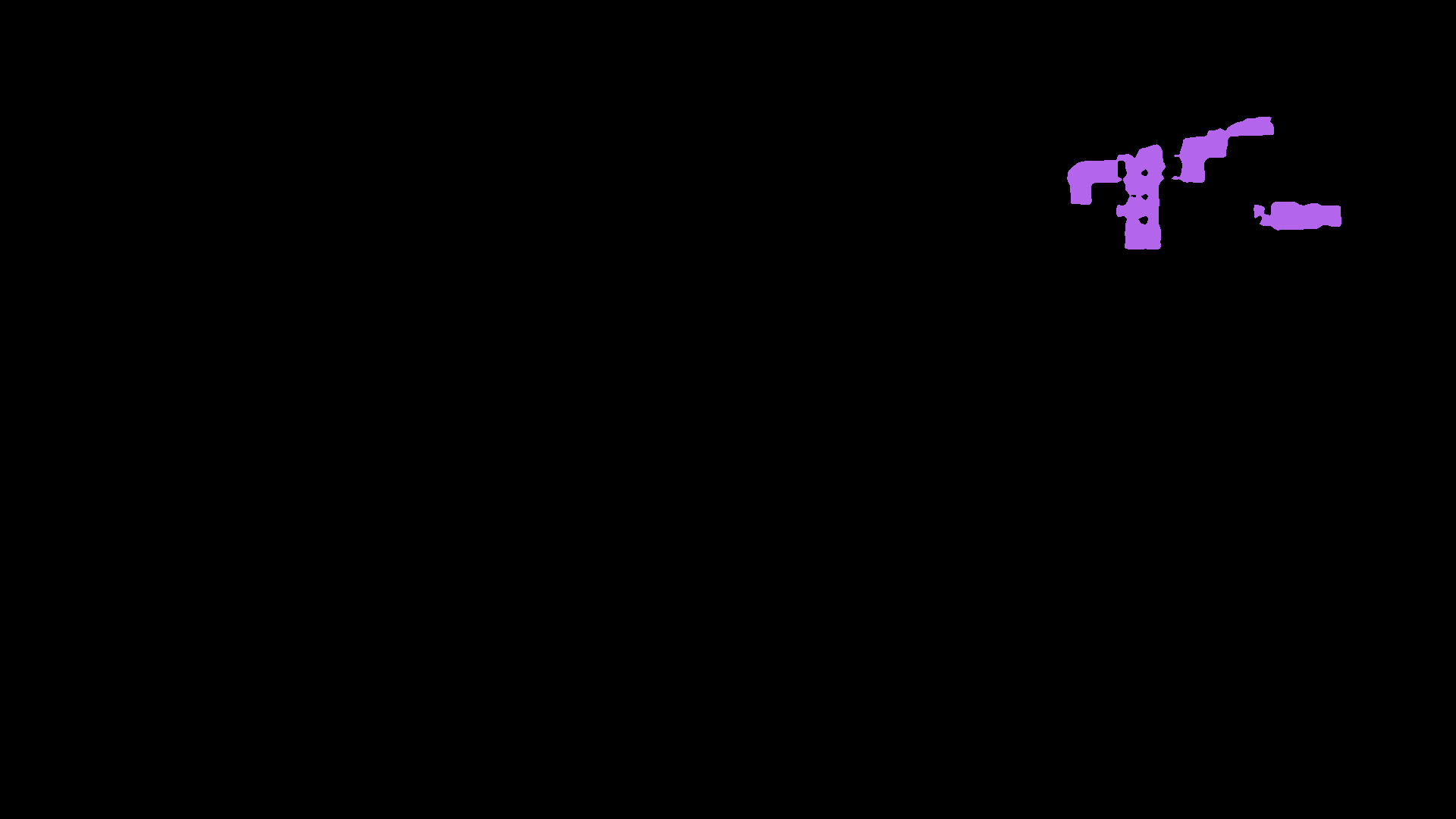}\\[-2pt]
    {\footnotesize 10.67 \(|\) 0.00 \(|\) 255.55}
\end{minipage}
&
\begin{minipage}{0.165\textwidth}
    \centering
    \includegraphics[width=\linewidth]{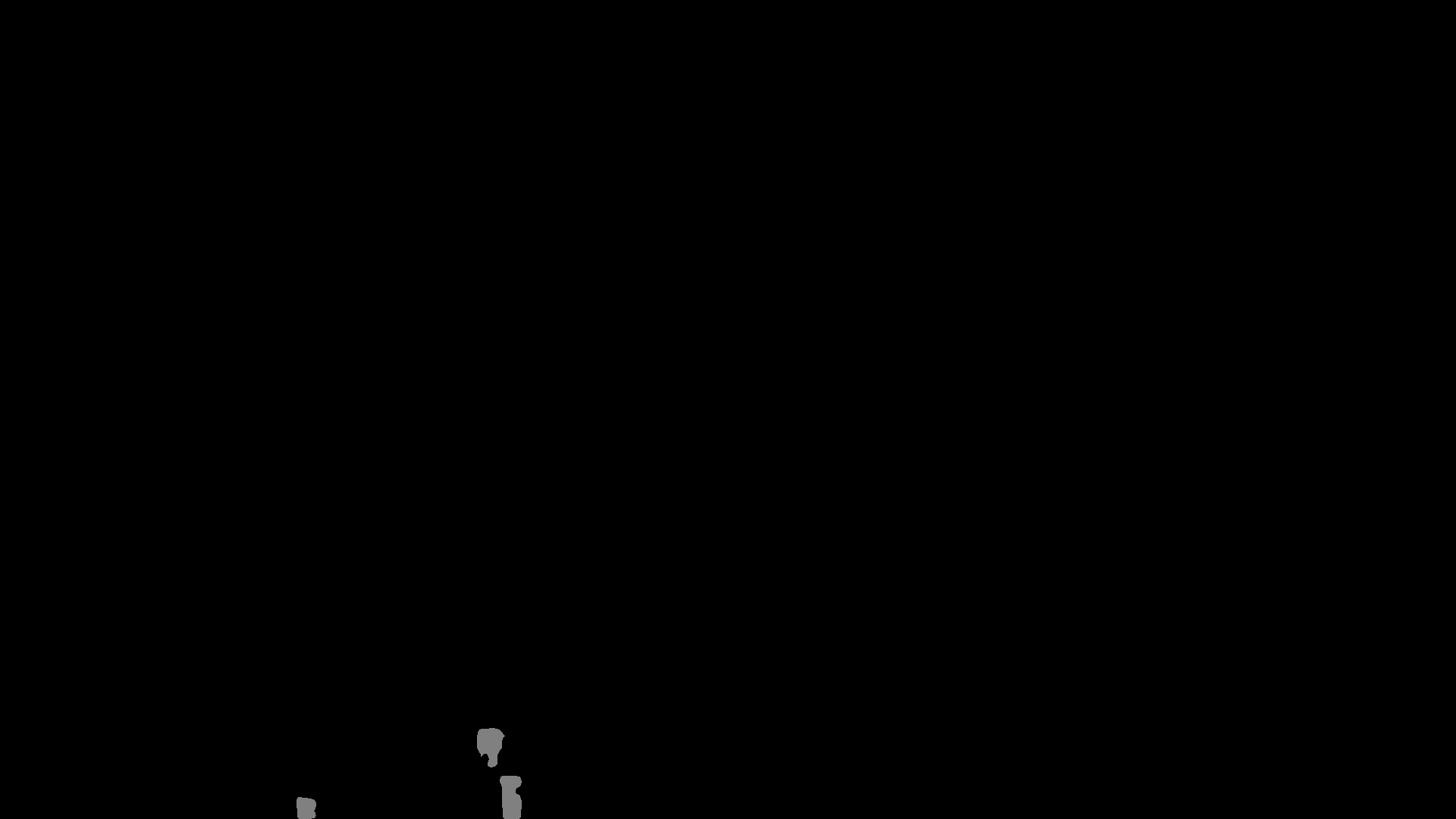}\\[-2pt]
    {\footnotesize 0.00 \(|\) 0.00 \(|\) 1386.63}
\end{minipage}
\\
\multirow{2}{*}{recogna} &
\begin{minipage}{0.165\textwidth}
    \centering
    \includegraphics[width=\linewidth]{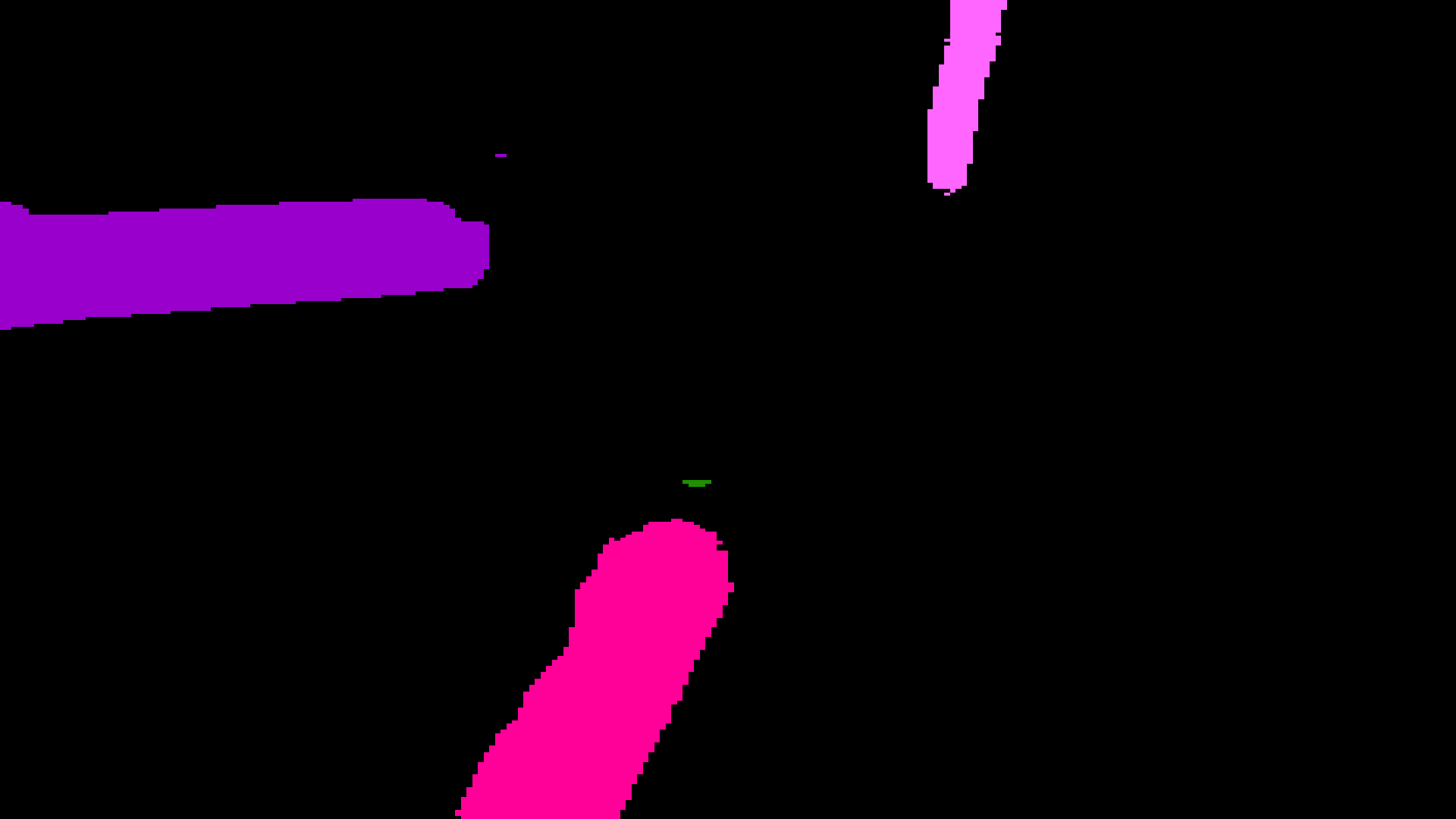}\\[-2pt]
    {\footnotesize 52.65 \(|\) 60.00 \(|\) 268.49}
\end{minipage}
&
\begin{minipage}{0.165\textwidth}
    \centering
    \includegraphics[width=\linewidth]{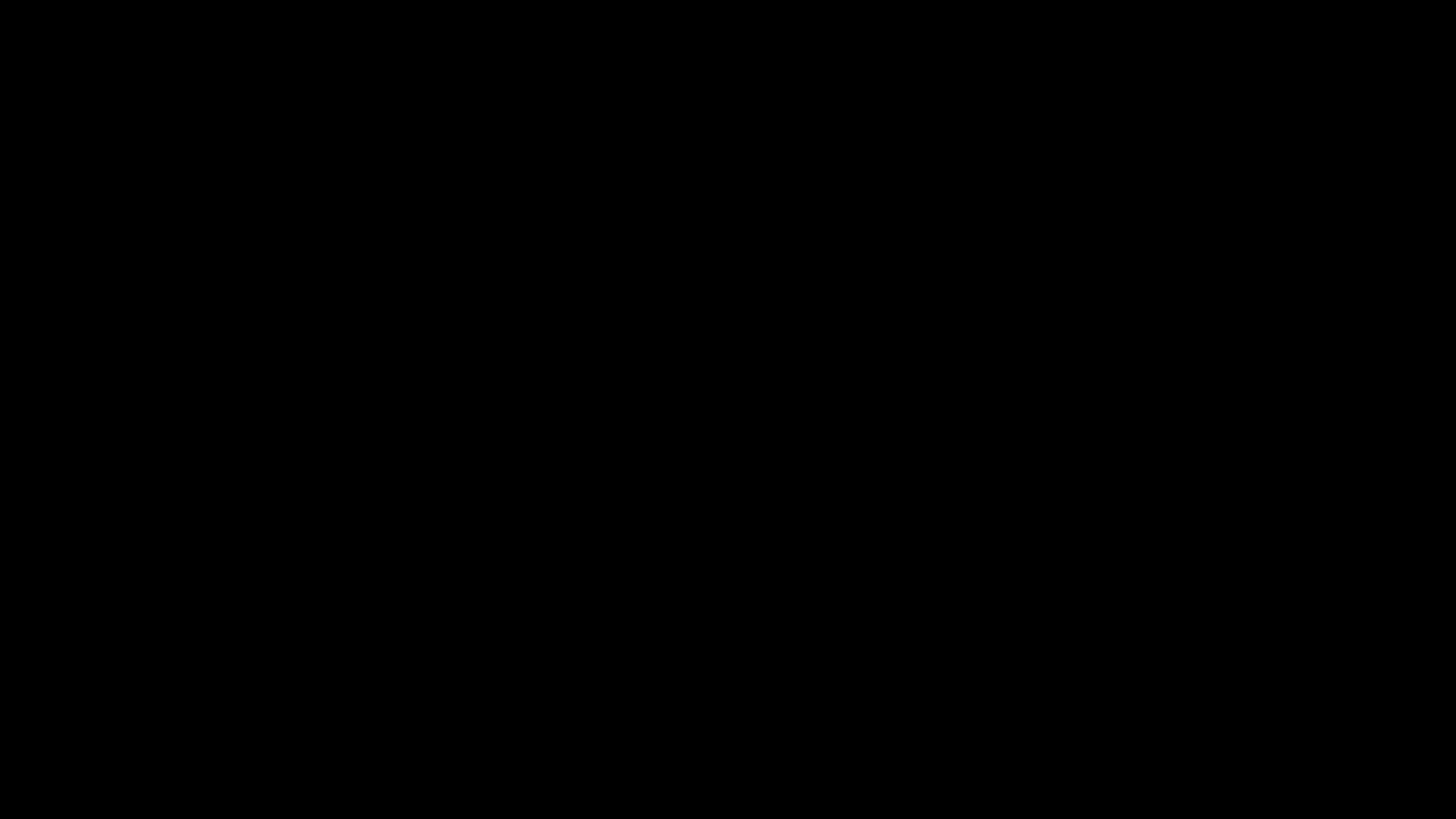}\\[-2pt]
    {\footnotesize 0.00 \(|\) 0.00 \(|\) nan}
\end{minipage}
&
\begin{minipage}{0.165\textwidth}
    \centering
    \includegraphics[width=\linewidth]{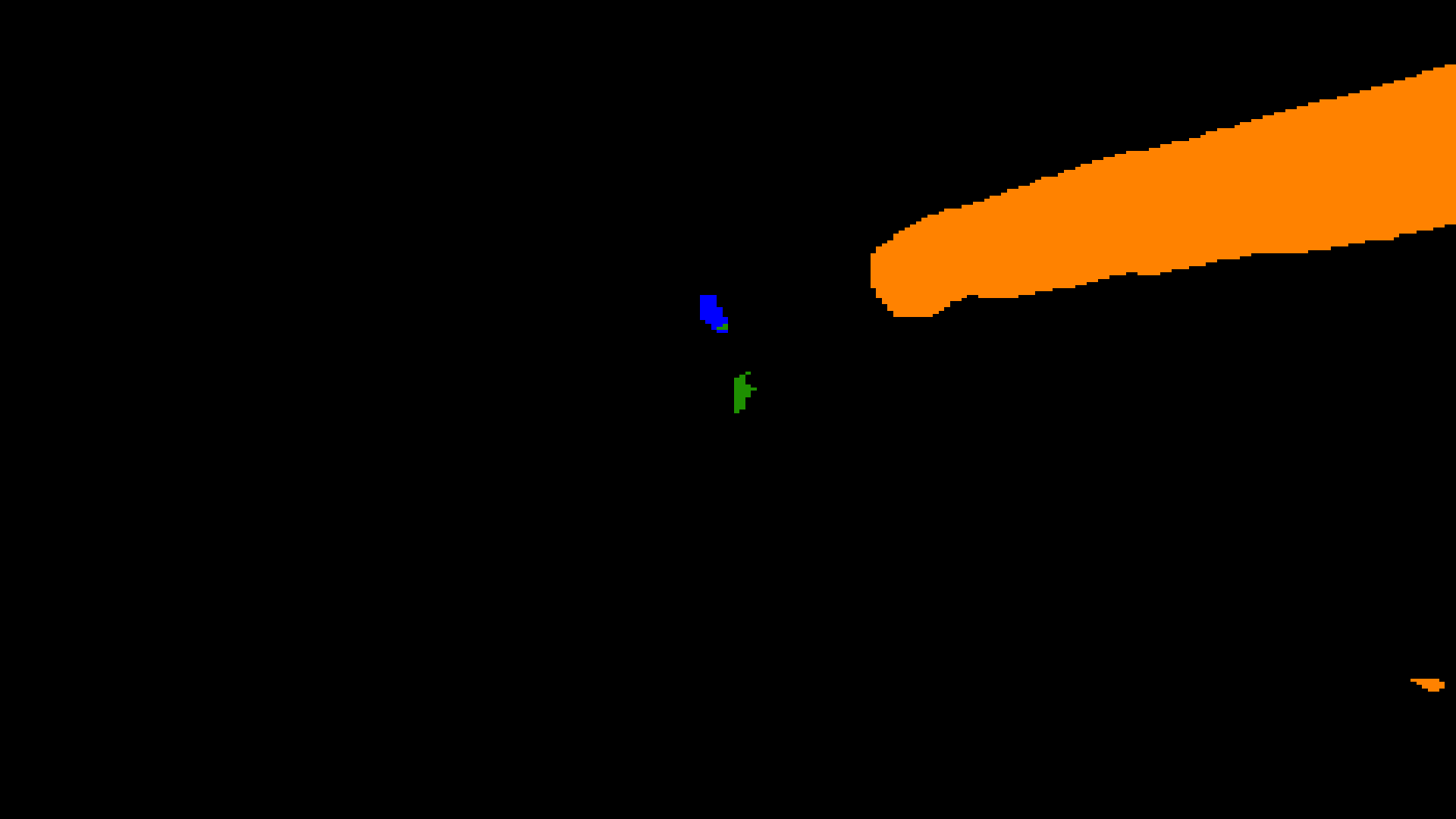}\\[-2pt]
    {\footnotesize 0.00 \(|\) 0.00 \(|\) 496.23}
\end{minipage}
&
\begin{minipage}{0.165\textwidth}
    \centering
    \includegraphics[width=\linewidth]{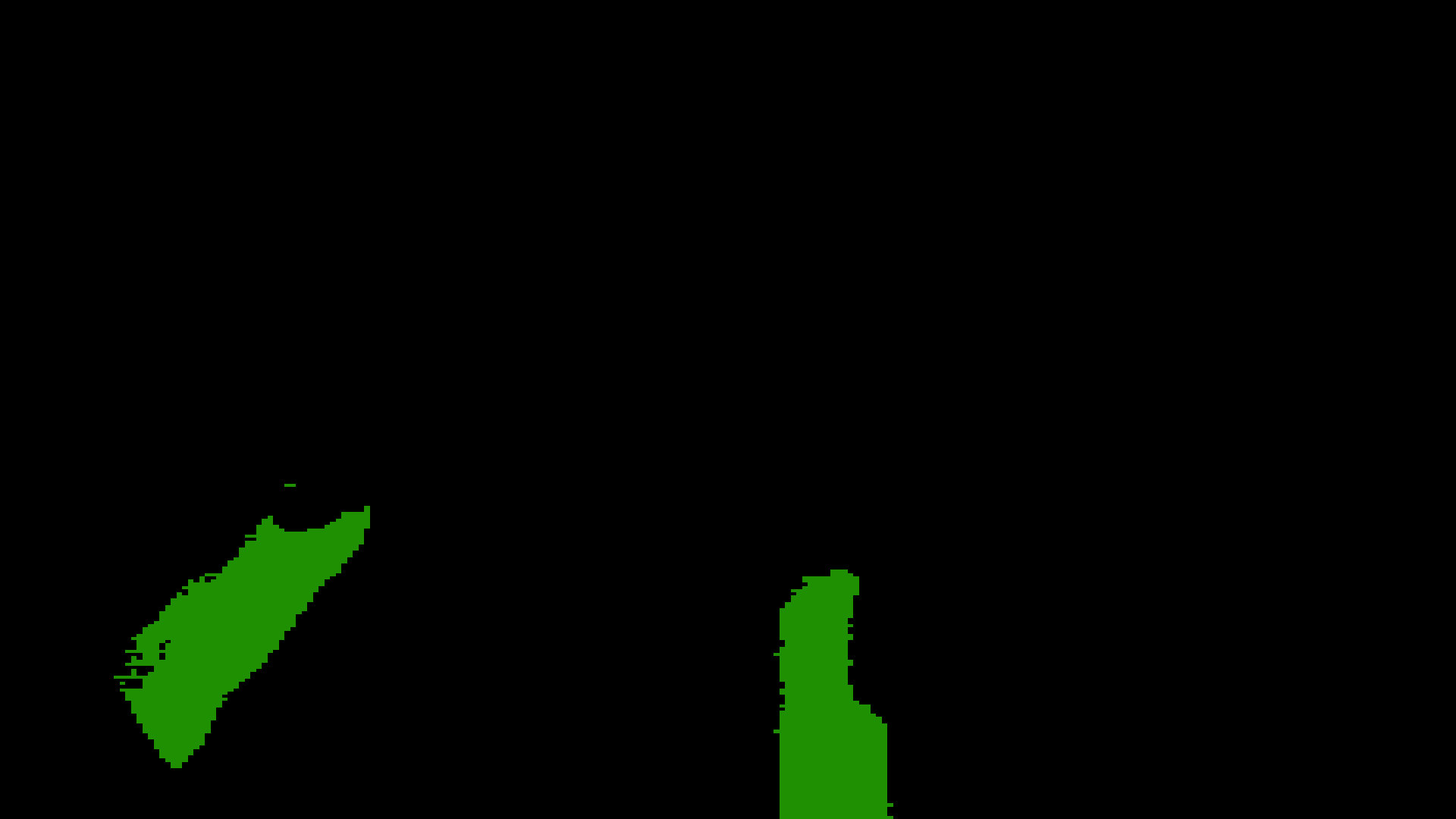}\\[-2pt]
    {\footnotesize 13.22 \(|\) 0.00 \(|\) 649.44}
\end{minipage}
\\

\bottomrule
\end{tabularx}
\label{results:instrument_seg:tab:qualitative_results_page_02}
\end{figure*}

\begin{figure*}[t]
    \centering
    \includegraphics[width=1.0\linewidth]{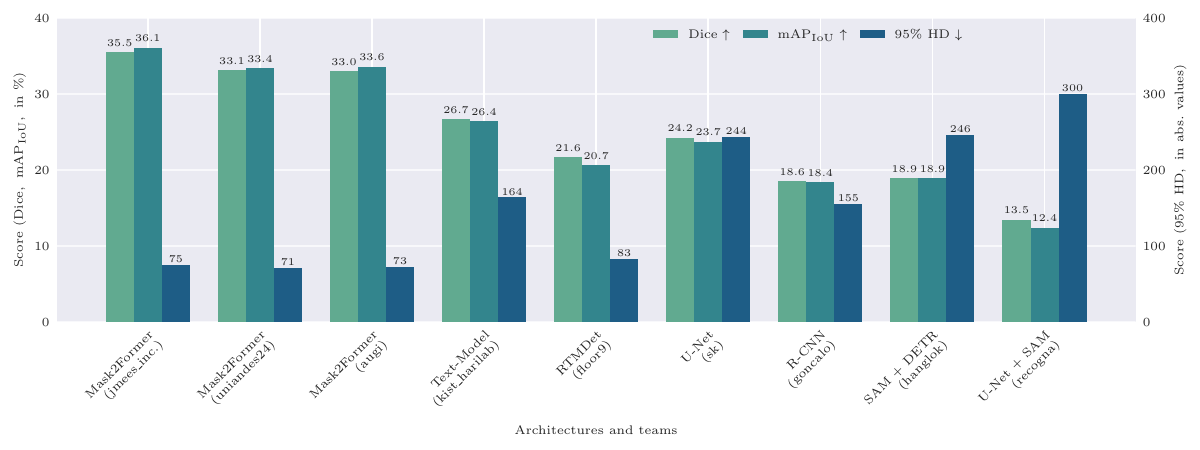}
    \caption{Submitted methods together with the corresponding participants and results with regard to the instrument instance segmentation task. If a methodology consists of several parts, the primary approach is listed first.}
    \label{results:instance_seg:method_comparison}
\end{figure*}

\begin{figure*}[b!]
    \small
    \hfill
    \begin{overpic}[width=0.9\textwidth]{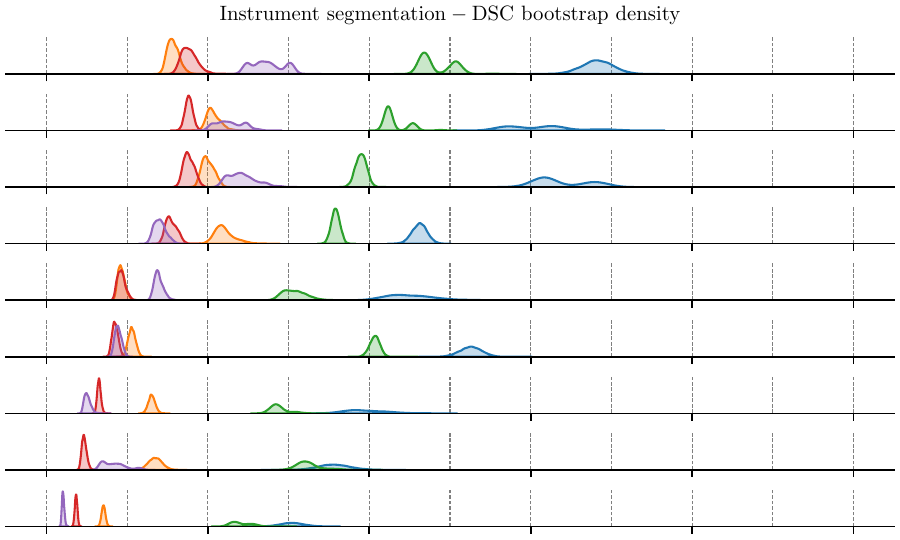}
    \put(-9, 54.0){\color{black} jmees\_inc.}
    \put(-9, 47.75){\color{black} uniandes24}
    \put(-9, 41.50){\color{black} augi}
    \put(-9, 35.25){\color{black} kist\_harilab}
    \put(-9, 29.0){\color{black} floor9}
    \put(-9, 22.75){\color{black} sk}
    \put(-9, 16.5){\color{black} goncalo}
    \put(-9, 10.25){\color{black} hanglok}
    \put(-9, 4){\color{black} recogna}
    \put(3.8, -1.5){\color{black} 0.0}
    \put(21.5, -1.5){\color{black} 0.2}
    \put(39.6, -1.5){\color{black} 0.4}
    \put(57.5, -1.5){\color{black} 0.6}
    \put(75.5, -1.5){\color{black} 0.8}
    \put(93.5, -1.5){\color{black} 1.0}
    
\end{overpic}
    \caption{KDE plot based on the bootstrap results for the participating teams of the instrument instance segmentation task with regard to the DSC values for the test set videos 6 \legendsquare{darkblue}, 8 \legendsquare{darkorange}, 9 \legendsquare{darkgreen}, 12 \legendsquare{darkred}, and 14 \legendsquare{darkpurple}.}
    \label{results:instance_seg:density_plot:dsc}
\end{figure*}

\begin{figure*}
    \small
    \hfill
    \begin{overpic}[width=0.9\textwidth]{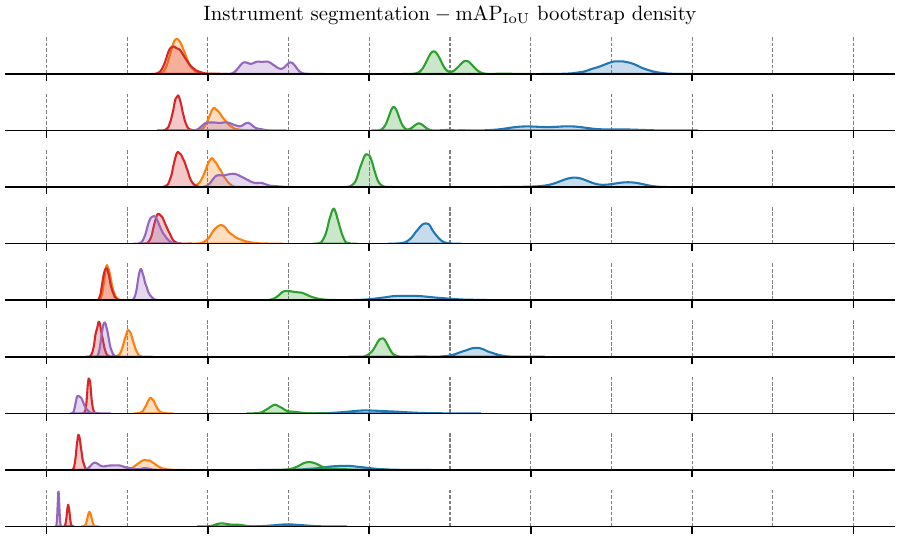}
    \put(-9, 54.0){\color{black} jmees\_inc.}
    \put(-9, 47.75){\color{black} uniandes24}
    \put(-9, 41.50){\color{black} augi}
    \put(-9, 35.25){\color{black} kist\_harilab}
    \put(-9, 29.0){\color{black} floor9}
    \put(-9, 22.75){\color{black} sk}
    \put(-9, 16.5){\color{black} goncalo}
    \put(-9, 10.25){\color{black} hanglok}
    \put(-9, 4){\color{black} recogna}
    \put(3.8, -1.5){\color{black} 0.0}
    \put(21.5, -1.5){\color{black} 0.2}
    \put(39.6, -1.5){\color{black} 0.4}
    \put(57.5, -1.5){\color{black} 0.6}
    \put(75.5, -1.5){\color{black} 0.8}
    \put(93.5, -1.5){\color{black} 1.0}
    
\end{overpic}
    \caption{KDE plot based on the bootstrap results for the participating teams of the instrument instance segmentation task with regard to the $\text{mAP}_{\text{IoU}}$ values for the test set videos 6 \legendsquare{darkblue}, 8 \legendsquare{darkorange}, 9 \legendsquare{darkgreen}, 12 \legendsquare{darkred}, and 14 \legendsquare{darkpurple}.}
    \label{results:instance_seg:density_plot:map}
\end{figure*}

\begin{figure*}
    \small
    \hfill
    \begin{overpic}[width=0.9\textwidth]{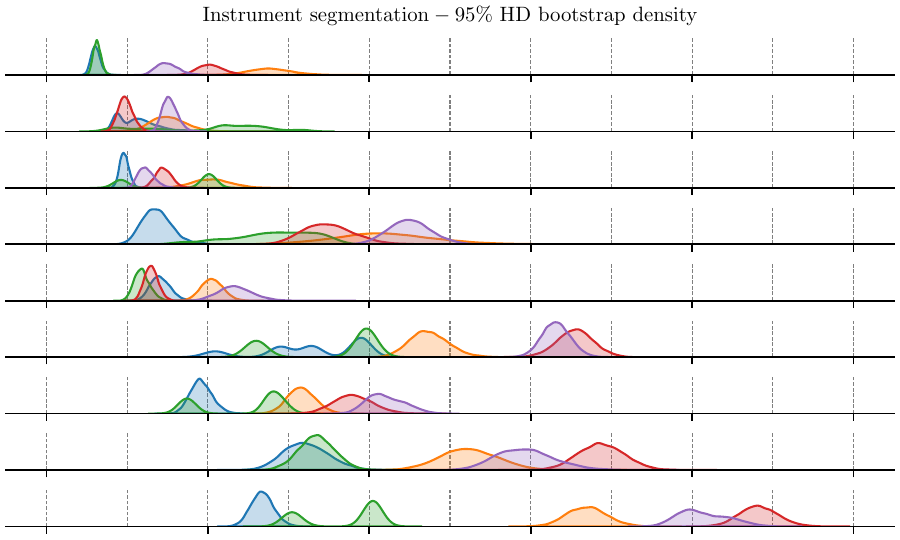}
    \put(-9, 54.0){\color{black} jmees\_inc.}
    \put(-9, 47.75){\color{black} uniandes24}
    \put(-9, 41.50){\color{black} augi}
    \put(-9, 35.25){\color{black} kist\_harilab}
    \put(-9, 29.0){\color{black} floor9}
    \put(-9, 22.75){\color{black} sk}
    \put(-9, 16.5){\color{black} goncalo}
    \put(-9, 10.25){\color{black} hanglok}
    \put(-9, 4){\color{black} recogna} 
    \put(4.7, -1.5){\color{black} 0}
    \put(20.9, -1.5){\color{black} 100}
    \put(38, -1.5){\color{black} 200}
    \put(55.2, -1.5){\color{black} 300}
    \put(72.4, -1.5){\color{black} 400}
    \put(89.6, -1.5){\color{black} 500}
    
\end{overpic}
    \caption{KDE plot based on the bootstrap results for the participating teams of the instrument instance segmentation task with regard to the 95\% HD values for the test set videos 6 \legendsquare{darkblue}, 8 \legendsquare{darkorange}, 9 \legendsquare{darkgreen}, 12 \legendsquare{darkred}, and 14 \legendsquare{darkpurple}.}
    \label{results:instance_seg:density_plot:hd}
\end{figure*}

\begin{table*}[t]
    \footnotesize
    \centering
    \caption{Results of the instrument instance segmentation task for each team and each individual video in the test dataset, together with the mean metric values for each video.}
    \begin{tabularx}{\textwidth}{p{1.25cm}Z{0.65cm}Z{0.65cm}Z{0.65cm}Z{0.65cm}Z{0.65cm}
    Z{0.65cm}Z{0.65cm}Z{0.65cm}Z{0.65cm}Z{0.65cm}
    Z{0.77cm}Z{0.77cm}Z{0.77cm}Z{0.77cm}Z{0.77cm}}
        \toprule
        \multirow{2}{*}{Team} 
        & \multicolumn{5}{c}{DSC (in \%) $\uparrow$} 
        & \multicolumn{5}{c}{$\text{mAP}_{\text{IoU}}$ (in \%) $\uparrow$} 
        & \multicolumn{5}{c}{95\% HD $\downarrow$} \\
        \cmidrule(lr){2-6} \cmidrule(lr){7-11} \cmidrule(l){12-16}
        & V\_06 & V\_08 & V\_09 & V\_12 & V\_14
        & V\_06 & V\_08 & V\_09 & V\_12 & V\_14
        & V\_06 & V\_08 & V\_09 & V\_12 & V\_14 \\
        \midrule
        \rowcolor{gray!10}
        jmees\_inc. & 68.26 & 15.70 & 48.35 & 17.70 & 27.40 & 70.90 & 16.34 & 49.57 & 16.23 & 27.21 & 30.50 & 138.61 & 31.38 & 101.83 & 75.09 \\
        uniandes24 & 61.24 & 20.68 & 43.17 & 17.65 & 22.74 & 63.26 & 21.17 & 43.88 & 16.27 & 22.38 & 55.79 & 75.71 & 100.37 & 49.02 & 76.49 \\
        \rowcolor{gray!10}
        augi & 63.85 & 20.02 & 38.98 & 17.64 & 24.24 & 67.42 & 20.68 & 39.70 & 16.55 & 23.30 & 48.19 & 102.31 & 80.46 & 72.57 & 61.97 \\
        kist\_harilab & 46.23 & 22.03 & 35.82 & 15.47 & 14.04 & 46.88 & 21.97 & 35.59 & 14.16 & 13.41 & 68.48 & 209.79 & 141.25 & 174.87 & 225.17 \\
        \rowcolor{gray!10}
        floor9 & 45.01 & 9.28 & 30.73 & 9.26 & 13.90 & 45.43 & 7.63 & 30.77 & 7.47 & 11.87 & 70.59 & 102.64 & 59.18 & 64.93 & 117.20 \\
        sk & 52.57 & 10.57 & 40.70 & 8.51 & 8.88 & 53.15 & 10.18 & 41.54 & 6.49 & 7.27 & 164.34 & 237.17 & 173.00 & 327.48 & 315.82 \\
        \rowcolor{gray!10}
        goncalo & 39.48 & 13.04 & 28.76 & 6.51 & 5.06 & 40.61 & 12.94 & 28.65 & 5.29 & 4.20 & 95.91 & 158.02 & 121.35 & 190.05 & 210.28 \\
        hanglok & 35.64 & 13.48 & 32.45 & 4.67 & 8.36 & 36.81 & 12.40 & 32.99 & 4.03 & 7.99 & 160.83 & 262.00 & 167.43 & 343.76 & 296.62 \\
        \rowcolor{gray!10}
        recogna & 30.43 & 7.05 & 24.09 & 3.66 & 2.05 & 30.03 & 5.32 & 22.45 & 2.69 & 1.48 & 133.71 & 334.87 & 183.70 & 441.37 & 406.95 \\
        \midrule
        \textbf{mean} & \textbf{49.19} & \textbf{14.65} & \textbf{35.89} & \textbf{11.23} & \textbf{14.07} & \textbf{50.50} & \textbf{14.29} & \textbf{36.13} & \textbf{9.91} & \textbf{13.23} & \textbf{92.04} & \textbf{180.12} & \textbf{117.57} & \textbf{196.21} & \textbf{198.40} \\
        \bottomrule
    \end{tabularx}
    \label{results:instance_seg:results_tab_individual_videos}
\end{table*}

\begin{table*}[b!]
    \small
    \centering
    \caption{Results and rankings for each individual team and metric regarding the surgical instrument keypoint estimation task, together with the averaged and overall rank for each team. For the $\text{mAP}_{\text{OKS}}$ metric, the mean value over the 10,000 bootstrap runs ($\text{Mean}_\text{BT}$), the 95\% confidence interval (95\% CI), and the metric-specific rank are shown, and it is indicated whether a higher or lower value is desired.}
    \begin{tabularx}{\textwidth}{XYYYYY}
        \toprule
        \multirow{2}{*}{Team} 
        & \multicolumn{3}{c}{$\text{mAP}_{\text{OKS}}$ (in \%) $\uparrow$} 
        & \multirow{2}{*}{Averaged}
        & \multirow{2}{*}{Global} \\
        \cmidrule(r){2-4}
        & $\text{Mean}_{\text{BT}}$ & 95\% CI & Rank & rank & rank \\
        \midrule
        \rowcolor{gray!10}
        sds-hd & 30.82 & $[29.17, 32.34]$ & 1 & 1.0 & \textbf{1st} \\
        alvaro & 15.29 & $[13.75, 16.75]$ & 2 & 2.0 & \textbf{2nd} \\
        \bottomrule
    \end{tabularx}
    \label{results:keypoint_estimation:results_tab}
\end{table*}

Qualitative results are presented in Tables~\ref{results:instrument_seg:tab:qualitative_results_page_01} and~\ref{results:instrument_seg:tab:qualitative_results_page_02}, showcasing eight representative test samples along with their respective ground truth segmentations.
For each sample, the predicted outputs of the participating teams are visualized, and the corresponding metric values are reported.
The dataset distribution is as follows: Samples one, four, five, and eight originate from hospital one, samples two and six from hospital two, and samples three and seven from hospital three, with hospital identifiers listed in Figure~\ref{fig:annotations}.
Instruments that appeared frequently in the training data (e.g., pe-forceps, grasper, palpation-probe, clip-applicator, HFcoag-probe) were generally segmented with higher accuracy.
In contrast, instruments with limited representation (e.g., scissor, blunt-grasper, blunt-grasper-spec., bipolar-clamp) were more prone to misclassification or missed detection.
Additionally, a notable limitation across many submissions was the lack of true instance segmentation.
Several teams instead applied semantic segmentation, failing to distinguish between multiple instances of the same instrument class.
This issue is particularly evident in sample three, where multiple algorithms generated reasonable segmentation masks for the blunt-grasper-spec.~class but did not differentiate between separate instruments positioned at the top of the frame.
This shortcoming negatively impacted performance metrics that rely on accurate instance separation.

Figure~\ref{results:instance_seg:method_comparison} provides a comparative visualization of team performances across the three evaluation metrics, accompanied by an overview of the primary method employed by each team.
In cases where a method comprised multiple components, only the principal approach is listed.
A clear trend emerges: the top three performing teams all implemented transformer-based architectures, specifically the Mask2Former model.
In contrast, teams that relied on convolutional neural networks, such as U-Net and R-CNN, demonstrated notably lower performance across all metrics.
Furthermore, teams utilizing Mask2Former achieved consistently strong results across all evaluation criteria, highlighting the architecture’s robustness and adaptability for the instrument instance segmentation task.
In comparison, alternative methods showed a discernible performance gap, suggesting the growing advantage of transformer-based approaches in medical image segmentation..
Figures~\ref{results:instance_seg:density_plot:dsc},~\ref{results:instance_seg:density_plot:map}, and~\ref{results:instance_seg:density_plot:hd}
display the KDE plots for the three evaluation metics: DSC, $\text{mAP}_{\text{IoU}}$, and 95\% HD.
Across all participating teams, video six consistently yielded the highest performance for both DSC and $\text{mAP}_{\text{IoU}}$ metrics, followed by video nine.
In contrast, the remaining three videos exhibit substantially lower scores, indicating greater segmentation difficulty.
While the KDE plot for the 95\% HD metric reveals a generally similar trend, it is characterized by larger fluctuations and a less clearly defined ranking pattern.
The broader Gaussian distributions observed for this metric -- across all videos and methods -- highlight its greater sensitivity to boundary discrepancies and higher variability compared to the DSC and $\text{mAP}_{\text{IoU}}$.

Table~\ref{results:instance_seg:results_tab_individual_videos} summarizes the quantitative performance per video.
\begin{itemize}
\item Video six achieved the best results overall (49.19\% DSC, 50.50\% $\text{mAP}_{\text{IoU}}$, 92.04\% 95\% HD),
\item followed by video nine (35.89\% DSC, 36.13\% $\text{mAP}_{\text{IoU}}$, 117.57\% 95\% HD),
\item and video eight (14.65\% DSC, 14.29\% $\text{mAP}_{\text{IoU}}$, 180.12\% 95\% HD).
\end{itemize}
Performance for videos 12 and 14 was more mixed:
\begin{itemize}
    \item video 14 yielded slightly better scores for DSC (14.07\%) and $\text{mAP}_{\text{IoU}}$ (13.23\%), 
    \item whereas video 12 performed marginally better with respect to the 95\% HD metric (196.21\%).
\end{itemize}
These findings underscore the video-specific variability in segmentation difficulty and highlight the importance of evaluating performance across diverse test scenarios.

\subsection{Task 3: Instrument keypoint estimation}
\label{results:keypoint_estimation}

Table~\ref{results:keypoint_estimation:results_tab} presents the results for the instrument keypoint estimation task.
Team sds-hd achieved the highest performance with a $\text{mAP}_{\text{OKS}}$ score of 30.82\%, clearly outperforming team alvaro, which reached 15.29\% $\text{mAP}_{\text{OKS}}$.
As only two teams participated, the average rank equaled the global rank, directly determining the final ranking outcome.

\begin{figure*}[t!]
\centering
    \includegraphics[width=0.4\textwidth]{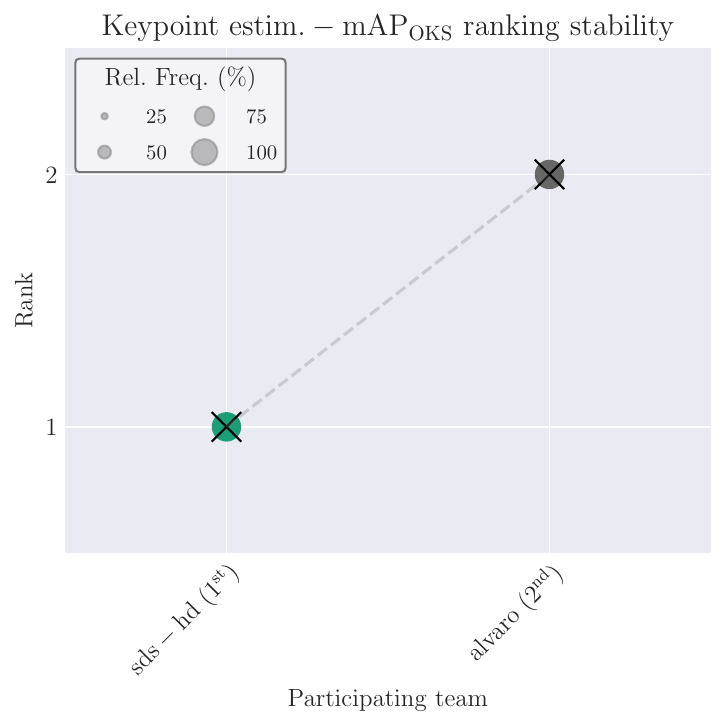}
\caption{
\label{fig:results:keypoint_estimation:ranking_robustness:blobs}%
Visualization of ranking stability based on bootstrapping for the instrument keypoint estimation task for the $\text{mAP}_{\text{OKS}}$ metric. The individual ranking stability for the $\text{mAP}_{\text{OKS}}$ metric is shown and the global rank is indicated after each team name.}
\end{figure*}

The ranking stability analysis shown in Figure~\ref{fig:results:keypoint_estimation:ranking_robustness:blobs} confirmed the robustness of these results.
Team sds-hd consistently ranked first in all 10{,}000 bootstrap iterations, demonstrating highly stable performance across the test dataset.
Since this task was evaluated using a single evaluation metric, the global ranking plot shown corresponds directly to this $\text{mAP}_{\text{OKS}}$ metric, and therefore only the ranking stability of $\text{mAP}_{\text{OKS}}$ is visualized.

\begin{figure*}[b!]
    \small
    \hfill
    \begin{overpic}[width=0.9\textwidth]{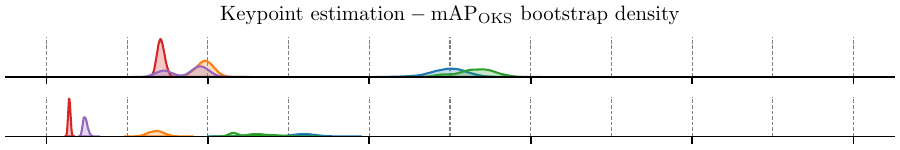}
    \put(-9, 10){\color{black} sds-hd}
    \put(-9, 3.25){\color{black} alvaro}
    \put(3.8, -1.5){\color{black} 0.0}
    \put(21.5, -1.5){\color{black} 0.2}
    \put(39.6, -1.5){\color{black} 0.4}
    \put(57.5, -1.5){\color{black} 0.6}
    \put(75.5, -1.5){\color{black} 0.8}
    \put(93.5, -1.5){\color{black} 1.0}
    
\end{overpic}
    \caption{KDE plot based on the bootstrap results for the participating teams of the instrument keypoint estimation task with regard to the $\text{mAP}_{\text{OKS}}$ values for the test set videos 6 \legendsquare{darkblue}, 8 \legendsquare{darkorange}, 9 \legendsquare{darkgreen}, 12 \legendsquare{darkred}, and 14 \legendsquare{darkpurple}.}
    \label{results:keypoint_estimation:density_plot:map}
\end{figure*}

Table~\ref{results:keypoint_estimation:tab:qualitative_results_page_01} presents a selection of qualitative results for the instrument keypoint estimation task, using the same image samples as in Tables~\ref{results:instrument_seg:tab:qualitative_results_page_01} and~\ref{results:instrument_seg:tab:qualitative_results_page_02}.
Given the limited number of participating teams, all results are consolidated into a single table.
The upper half displays the predicted keypoints and corresponding $\text{mAP}_{\text{OKS}}$ scores for the first four samples, while the lower half shows results for the remaining four.
In several cases -- specifically samples two, six, seven, and eight -- the algorithms clearly struggled with accurate instrument classification, highlighting the inherent difficulty of the task.
Consistent with the findings from the instance segmentation task, instruments with higher representation in the training set were generally estimated more accurately than those with fewer examples.

The KDE plot in Figure~\ref{results:keypoint_estimation:density_plot:map} provides further insight into performance variability. 
It shows that both submitted algorithms most accurately predicted videos six and nine, whereas the remaining three videos exhibited substantially lower performance.
The aggregated Gaussian distributions for team sds-hd appear slightly broader, indicating a greater spread of metric values and reflecting more variability in performance across different video samples compared to team alvaro.

Table~\ref{results:keypoint_estimation:results_tab_individual_videos} summarizes the average $\text{mAP}_{\text{OKS}}$ scores for each test video:
\begin{itemize}
    \item Video six achieved the highest average score (39.78\%),
    \item closely followed by video nine (39.28\%).
    \item Video eight saw a notable drop (16.58\%),
    \item while videos 14 (11.13\%) and 12 (8.49\%) yielded the lowest overall results.
\end{itemize}
These outcomes highlight the challenges of keypoint estimation, particularly for underrepresented instruments and more complex video sequences.

\section{Discussion}
\label{discussion}

In the following, we reflect on the key findings of our PhaKIR challenge and discuss potential limitations.  
We begin by discussing the organizational structure of the challenge~(Sec.~\ref{discussion:organization}), and examining the characteristics of the provided dataset~(Sec.~\ref{discussion:dataset}).  
We then outline the evaluation methodology used to assess the submitted approaches~(Sec.~\ref{discussion:assessment}), and analyze the performance of the submitted methods for each individual task~(Sec.~\ref{discussion:phase}, \ref{discussion:segmentation}, and \ref{discussion:keypoints}).  
We then derive overarching insights across all tasks~(Sec.~\ref{discussion:overall}), before concluding with perspectives on future research directions in this domain~(Sec.~\ref{discussion:future_research}).

\subsection{Challenge organization}
\label{discussion:organization}

\begin{figure*}[tbph!]
\footnotesize
\centering
\renewcommand{\arraystretch}{1.5}
\captionof{table}{Visualization of example participant predictions, together with quantitative results for $\text{mAP}_{\text{OKS}}$ (in \%) $\uparrow$ below each image. White-bordered dots indicate predicted hidden keypoints, red-bordered circles indicate keypoiknts for which coordinates were predicted but marked as not present. According to the metric description in Section~\ref{assessment:keypoint}, for metric calculation, only the visibility flag of the GT keypoints is taken into account. An assignment of the color codings to the respective instrument categories is provided in Section 4.1 in \ref{appendix:data_description}.}
\begin{tabularx}{\textwidth}{m{1.5cm} X X X X }
\toprule
\multicolumn{5}{c}{\textbf{Challenge data}} \\
\midrule
\textbf{Team} & \textbf{Sample 1} & \textbf{Sample 2} & \textbf{Sample 3} & \textbf{Sample 4} \\
\midrule
\multirow{2}{*}{Input} &
\begin{minipage}{0.165\textwidth}
    \centering
    \includegraphics[width=\linewidth]{03_results_instrument_seg_qualitative_01_frame.png}\\
\end{minipage}
&
\begin{minipage}{0.165\textwidth}
    \centering
    \includegraphics[width=\linewidth]{03_results_instrument_seg_qualitative_02_frame.png}\\
\end{minipage}
&
\begin{minipage}{0.165\textwidth}
    \centering
    \includegraphics[width=\linewidth]{03_results_instrument_seg_qualitative_03_frame.png}\\
\end{minipage}
&
\begin{minipage}{0.165\textwidth}
    \centering
    \includegraphics[width=\linewidth]{03_results_instrument_seg_qualitative_04_frame.png}\\
\end{minipage}
\\[1cm]
\multirow{2}{*}{\shortstack[l]{Ground \\ truth}} &
\begin{minipage}{0.165\textwidth}
    \centering
    \includegraphics[width=\linewidth]{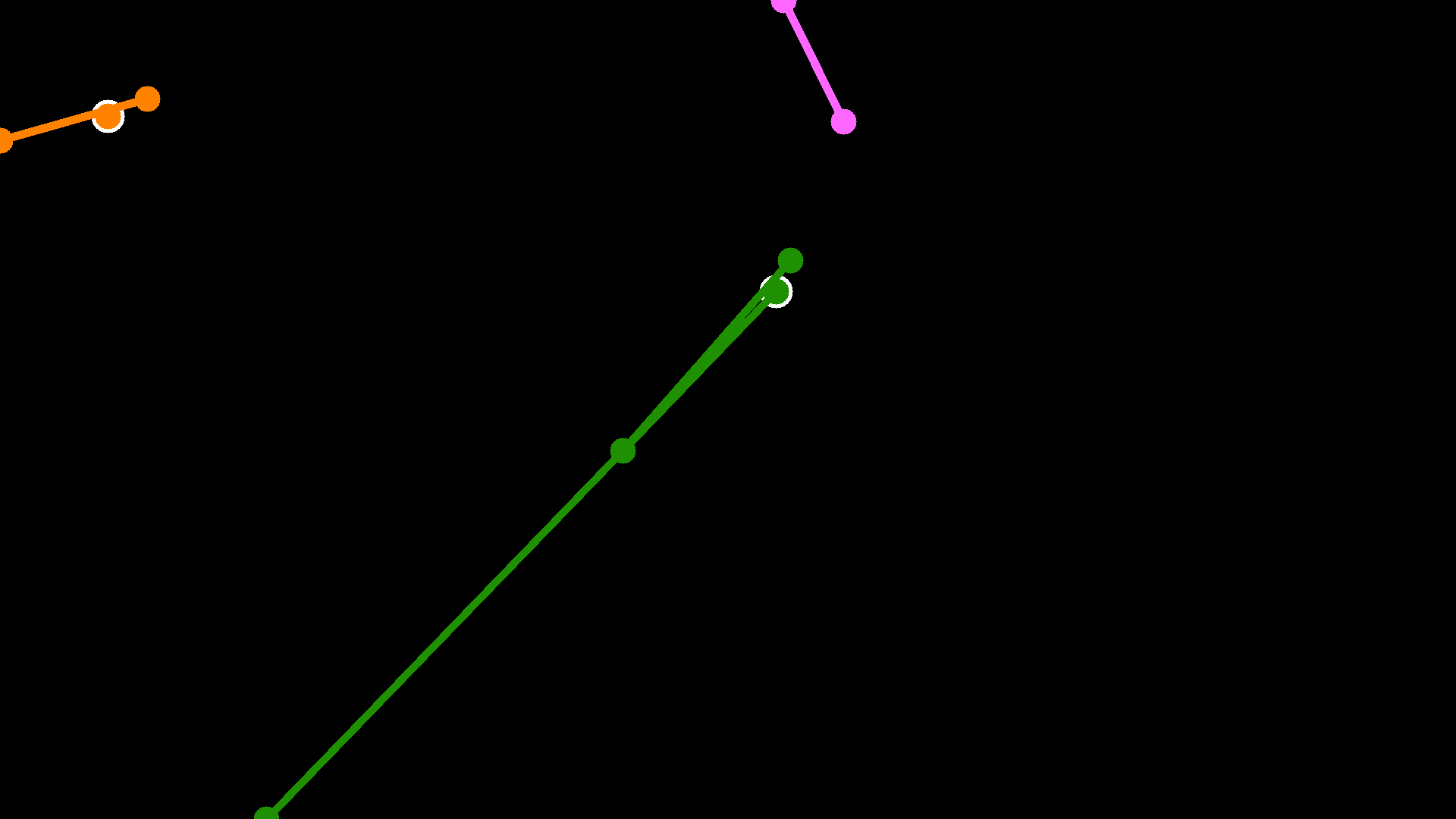}\\
\end{minipage}
&
\begin{minipage}{0.165\textwidth}
    \centering
    \includegraphics[width=\linewidth]{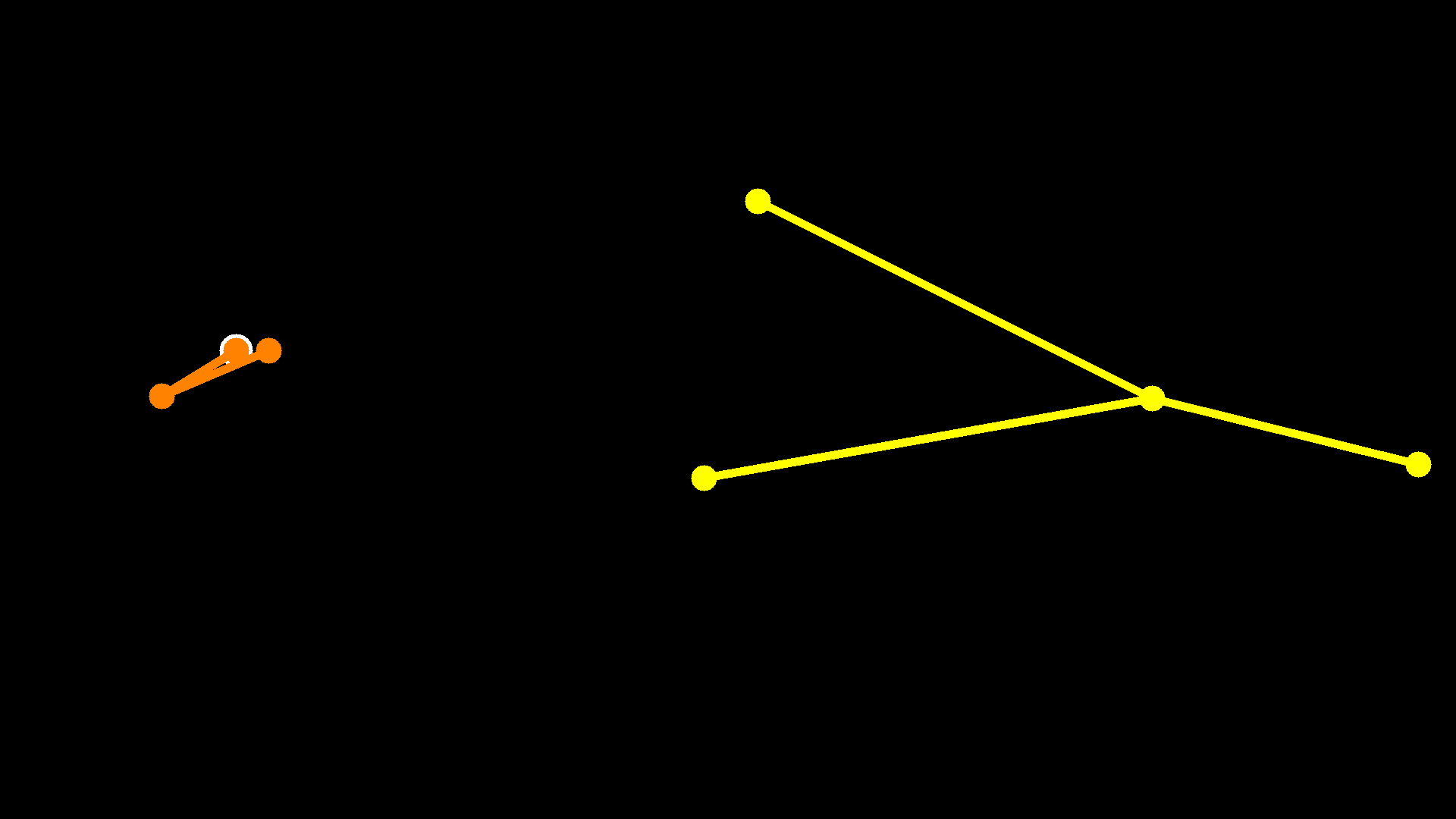}\\
\end{minipage}
&
\begin{minipage}{0.165\textwidth}
    \centering
    \includegraphics[width=\linewidth]{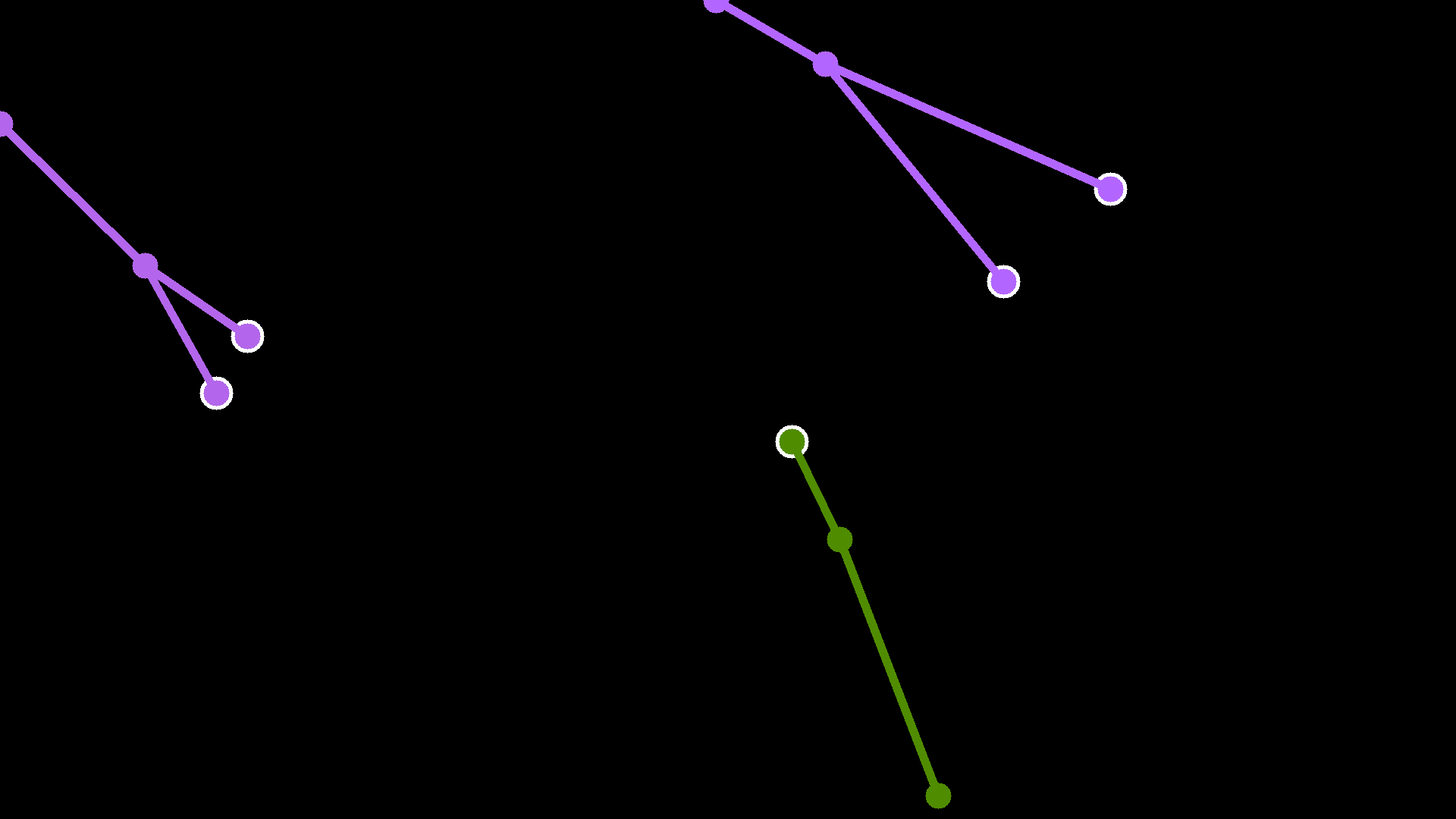}\\
\end{minipage}
&
\begin{minipage}{0.165\textwidth}
    \centering
    \includegraphics[width=\linewidth]{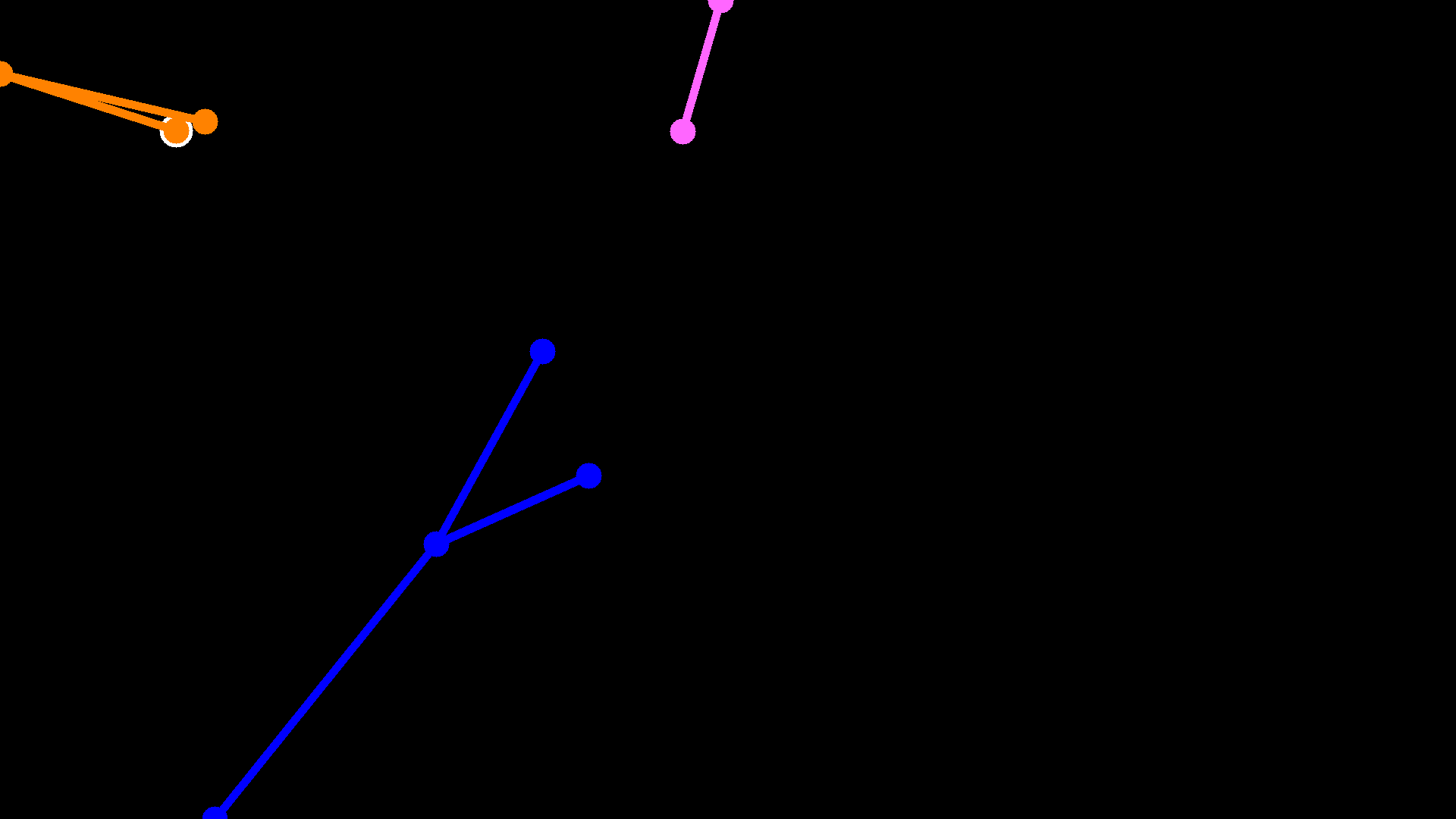}\\
\end{minipage}
\\
\midrule
\multicolumn{5}{c}{\textbf{Participant predictions}} \\
\midrule
\multirow{2}{*}{sds-hd} &
\begin{minipage}{0.165\textwidth}
    \centering
    \includegraphics[width=\linewidth]{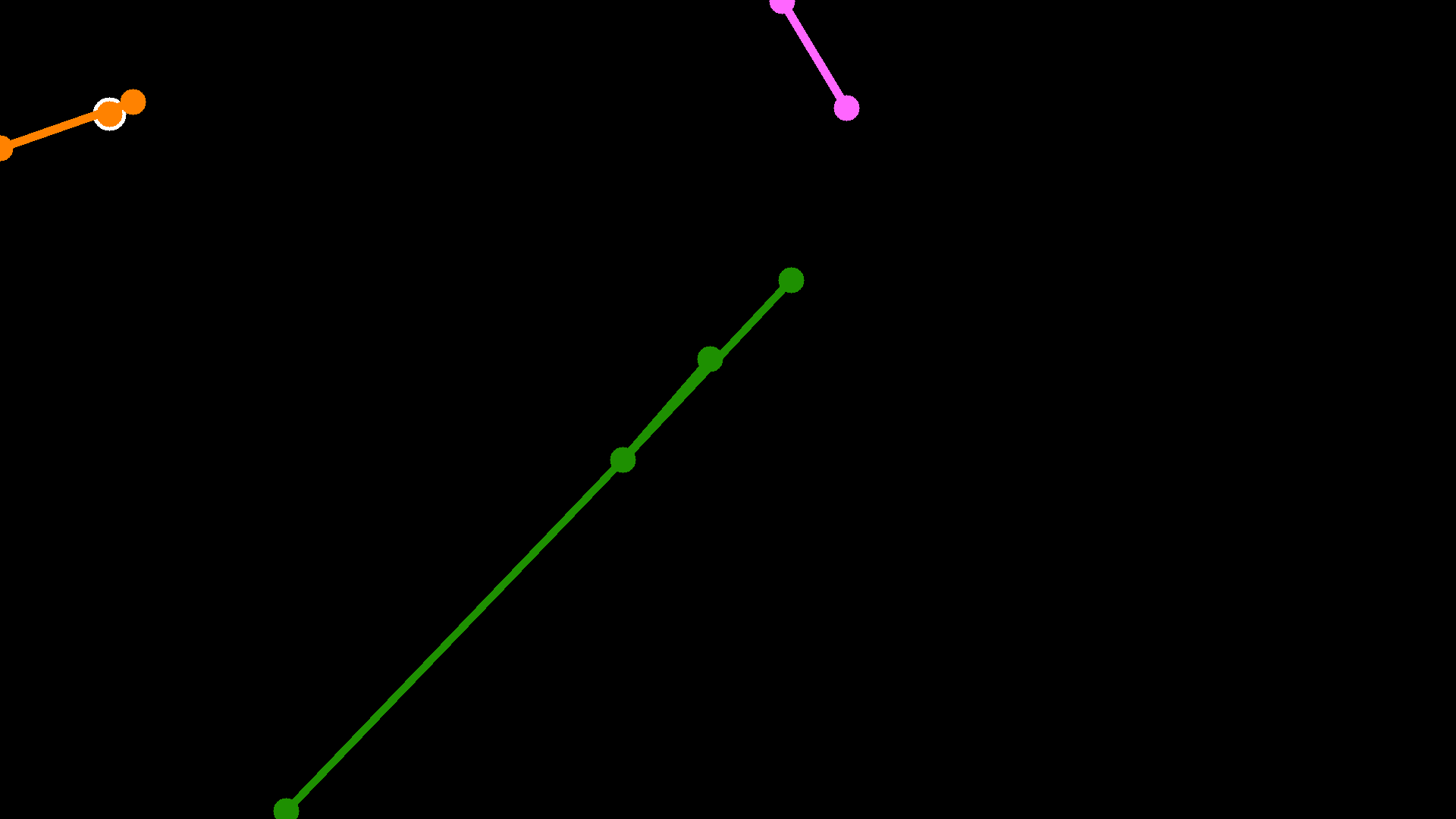}\\[-2pt]
    {\footnotesize 25.00}
\end{minipage}
&
\begin{minipage}{0.165\textwidth}
    \centering
    \includegraphics[width=\linewidth]{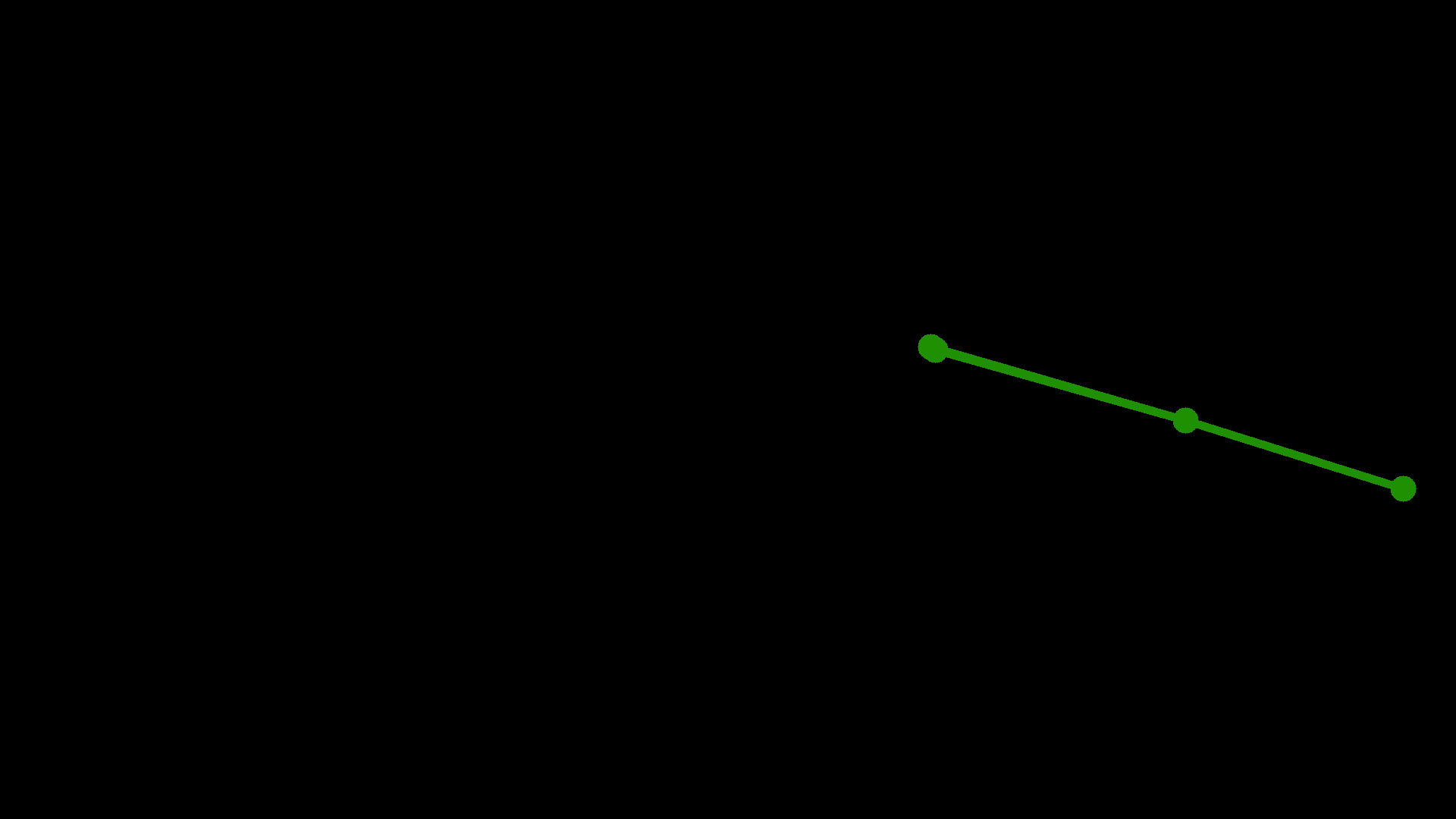}\\[-2pt]
    {\footnotesize 0.00}
\end{minipage}
&
\begin{minipage}{0.165\textwidth}
    \centering
    \includegraphics[width=\linewidth]{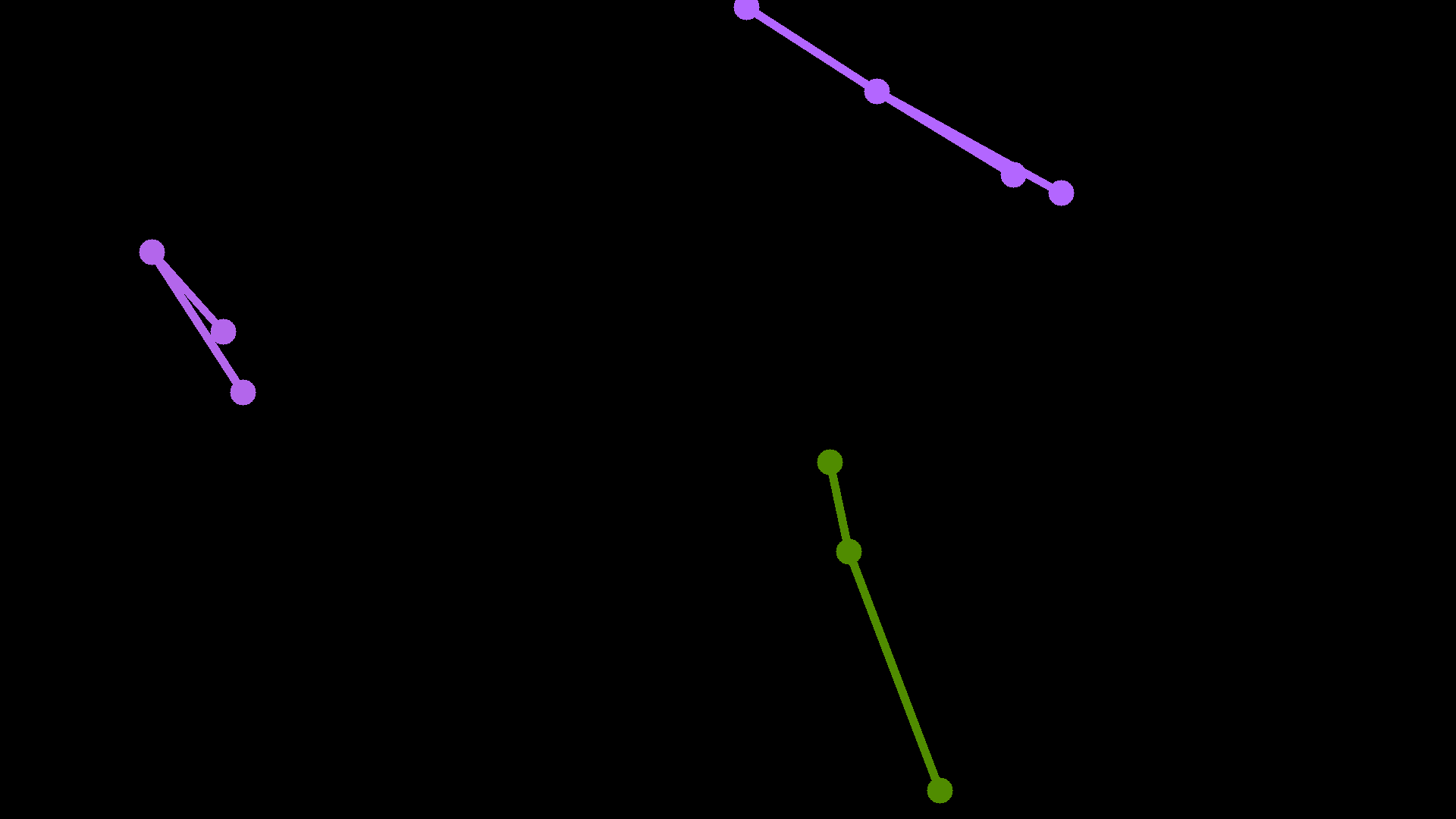}\\[-2pt]
    {\footnotesize 13.64}
\end{minipage}
&
\begin{minipage}{0.165\textwidth}
    \centering
    \includegraphics[width=\linewidth]{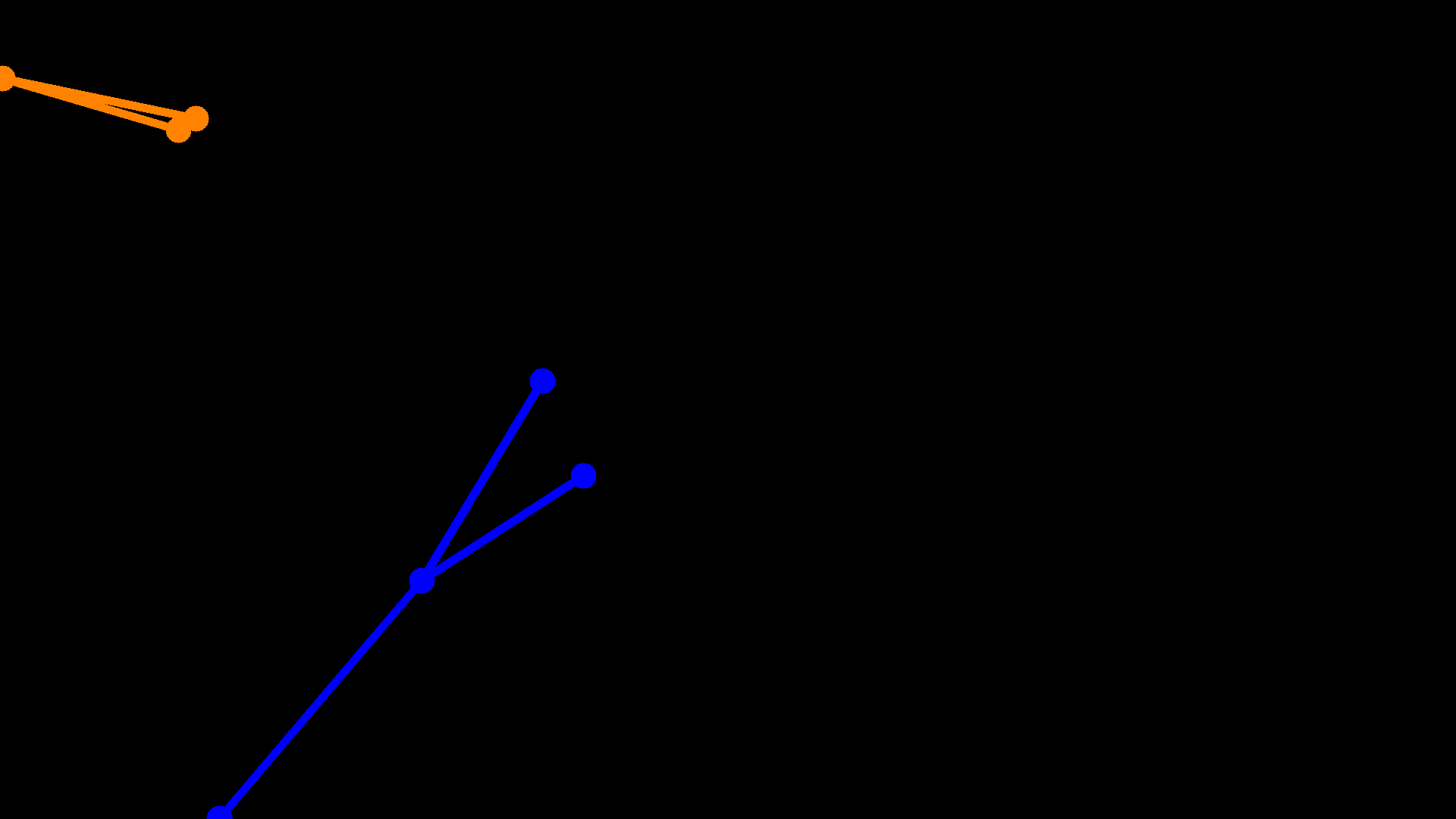}\\[-2pt]
    {\footnotesize 18.18}
\end{minipage}
\\
\\[-1.5em]
\multirow{2}{*}{alvaro} &
\begin{minipage}{0.165\textwidth}
    \centering
    \includegraphics[width=\linewidth]{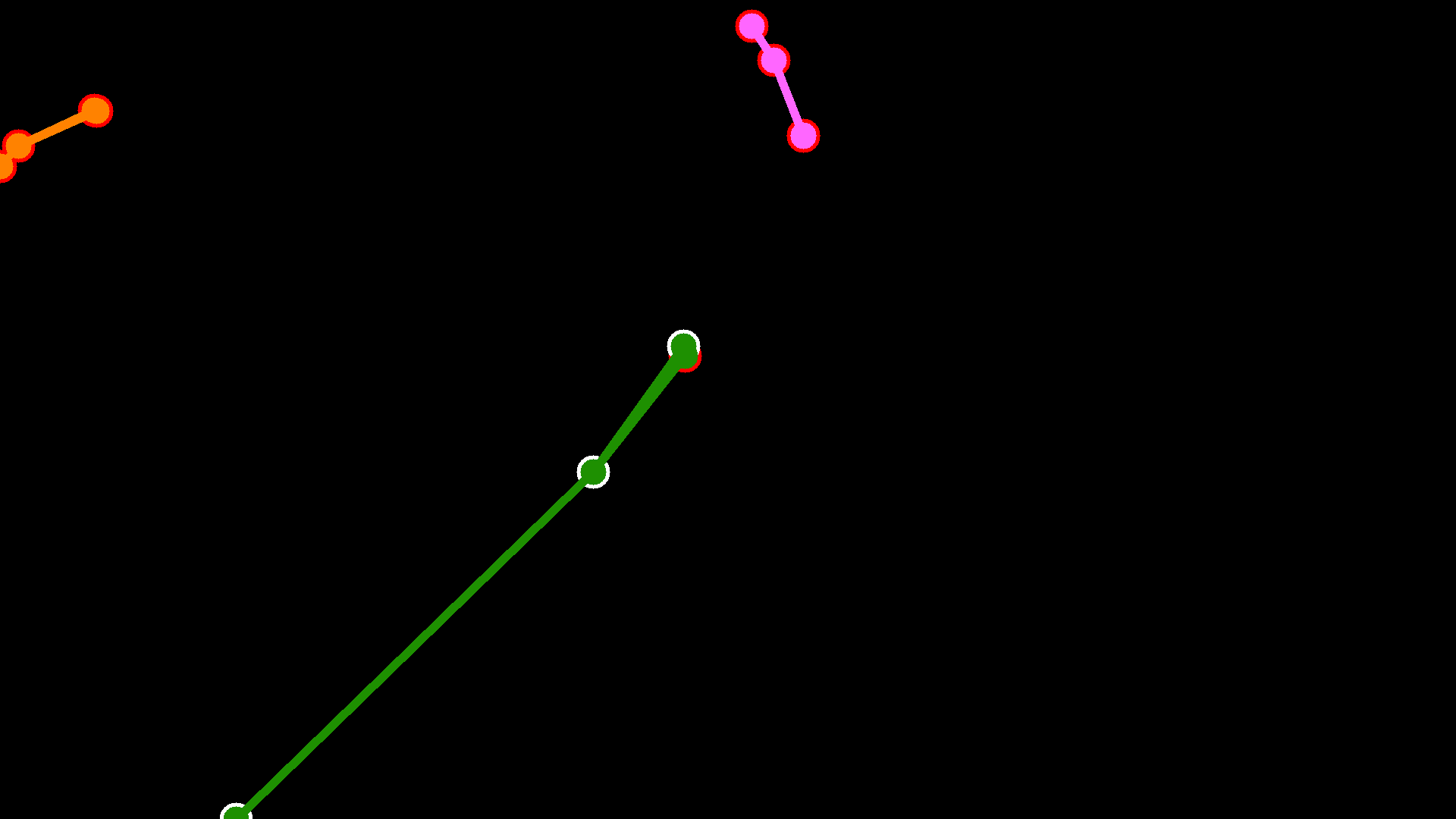}\\[-2pt]
    {\footnotesize 15.00}
\end{minipage}
&
\begin{minipage}{0.165\textwidth}
    \centering
    \includegraphics[width=\linewidth]{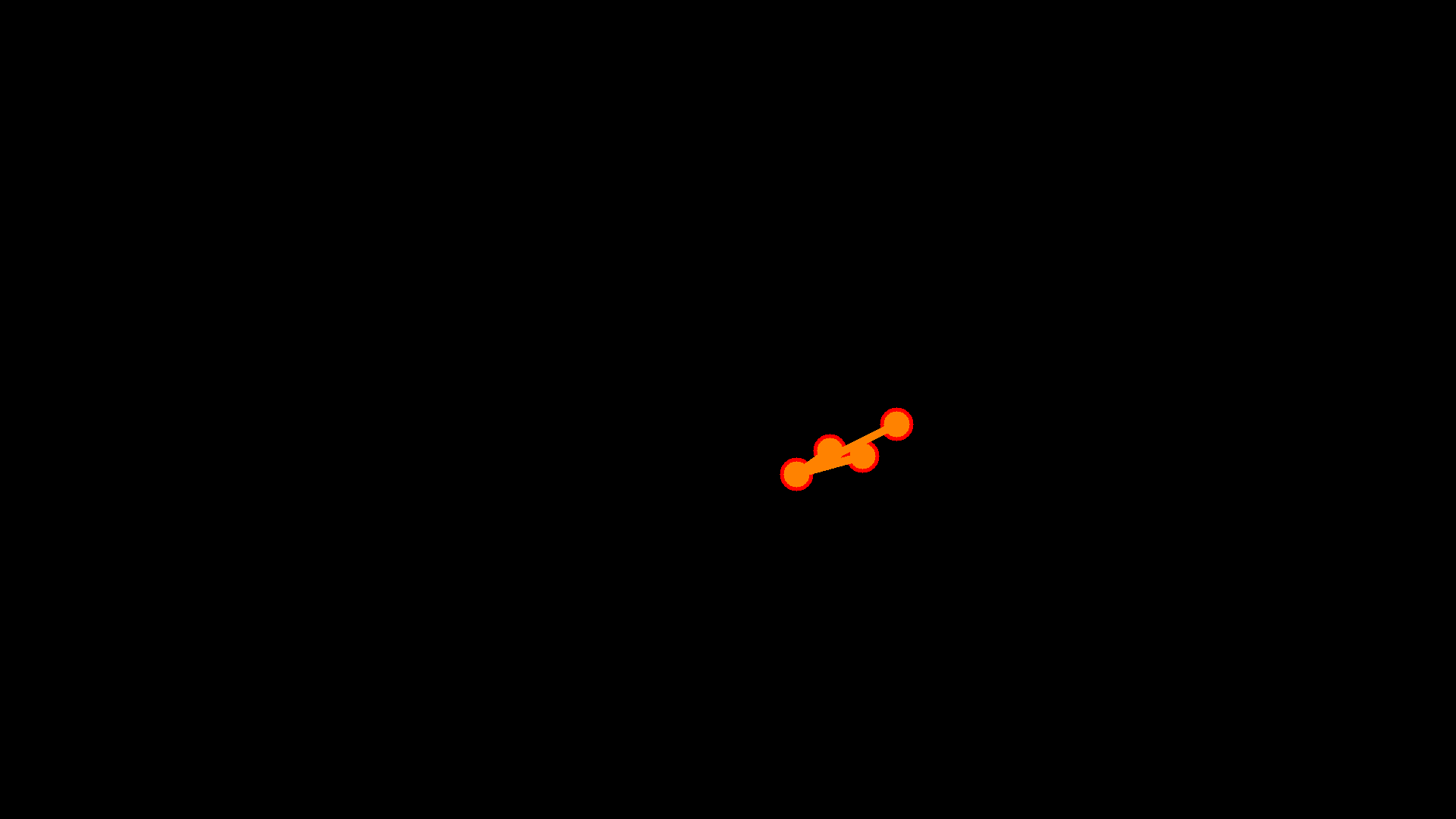}\\[-2pt]
    {\footnotesize 0.00}
\end{minipage}
&
\begin{minipage}{0.165\textwidth}
    \centering
    \includegraphics[width=\linewidth]{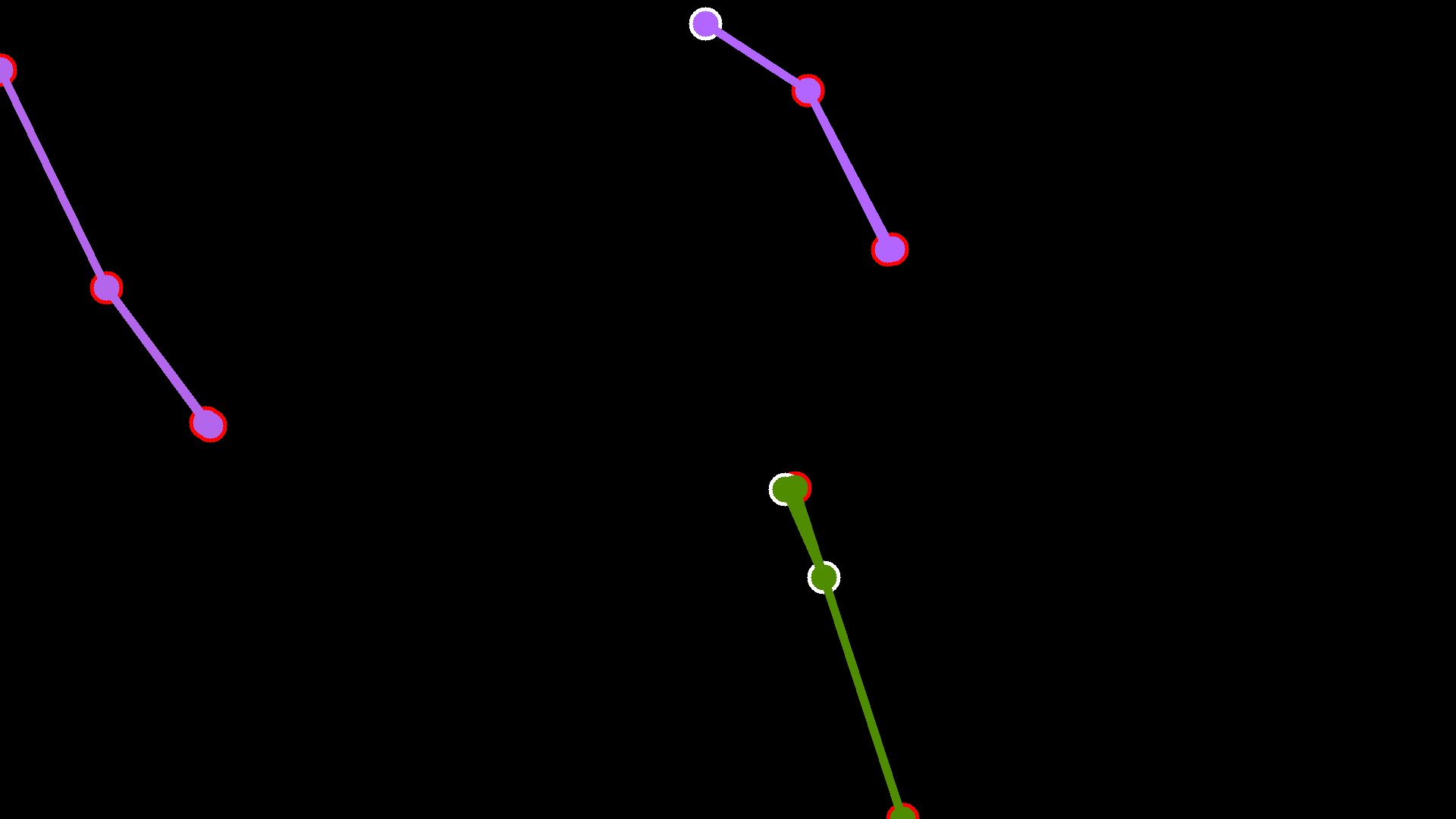}\\[-2pt]
    {\footnotesize 6.82}
\end{minipage}
&
\begin{minipage}{0.165\textwidth}
    \centering
    \includegraphics[width=\linewidth]{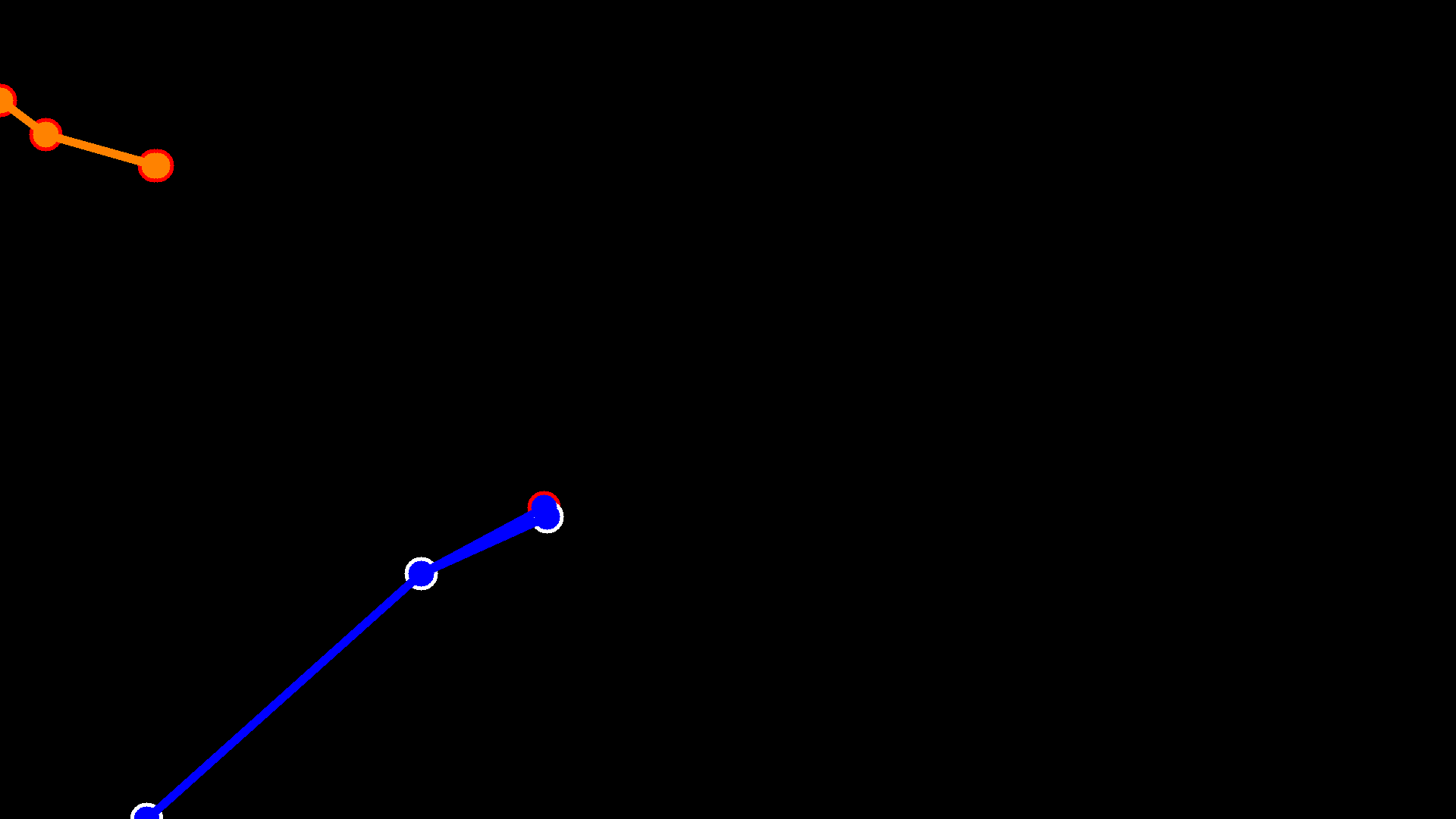}\\[-2pt]
    {\footnotesize 4.55}
\end{minipage}
\\
\midrule
\multicolumn{5}{c}{\textbf{Challenge data}} \\
\midrule
\textbf{Team} & \textbf{Sample 5} & \textbf{Sample 6} & \textbf{Sample 7} & \textbf{Sample 8} \\
\midrule
\multirow{2}{*}{Input} &
\begin{minipage}{0.165\textwidth}
    \centering
    \includegraphics[width=\linewidth]{03_results_instrument_seg_qualitative_05_frame.png}\\
\end{minipage}
&
\begin{minipage}{0.165\textwidth}
    \centering
    \includegraphics[width=\linewidth]{03_results_instrument_seg_qualitative_06_frame.png}\\
\end{minipage}
&
\begin{minipage}{0.165\textwidth}
    \centering
    \includegraphics[width=\linewidth]{03_results_instrument_seg_qualitative_07_frame.png}\\
\end{minipage}
&
\begin{minipage}{0.165\textwidth}
    \centering
    \includegraphics[width=\linewidth]{03_results_instrument_seg_qualitative_08_frame.png}\\
\end{minipage}
\\[1cm]
\multirow{2}{*}{\shortstack[l]{Ground \\ truth}} &
\begin{minipage}{0.165\textwidth}
    \centering
    \includegraphics[width=\linewidth]{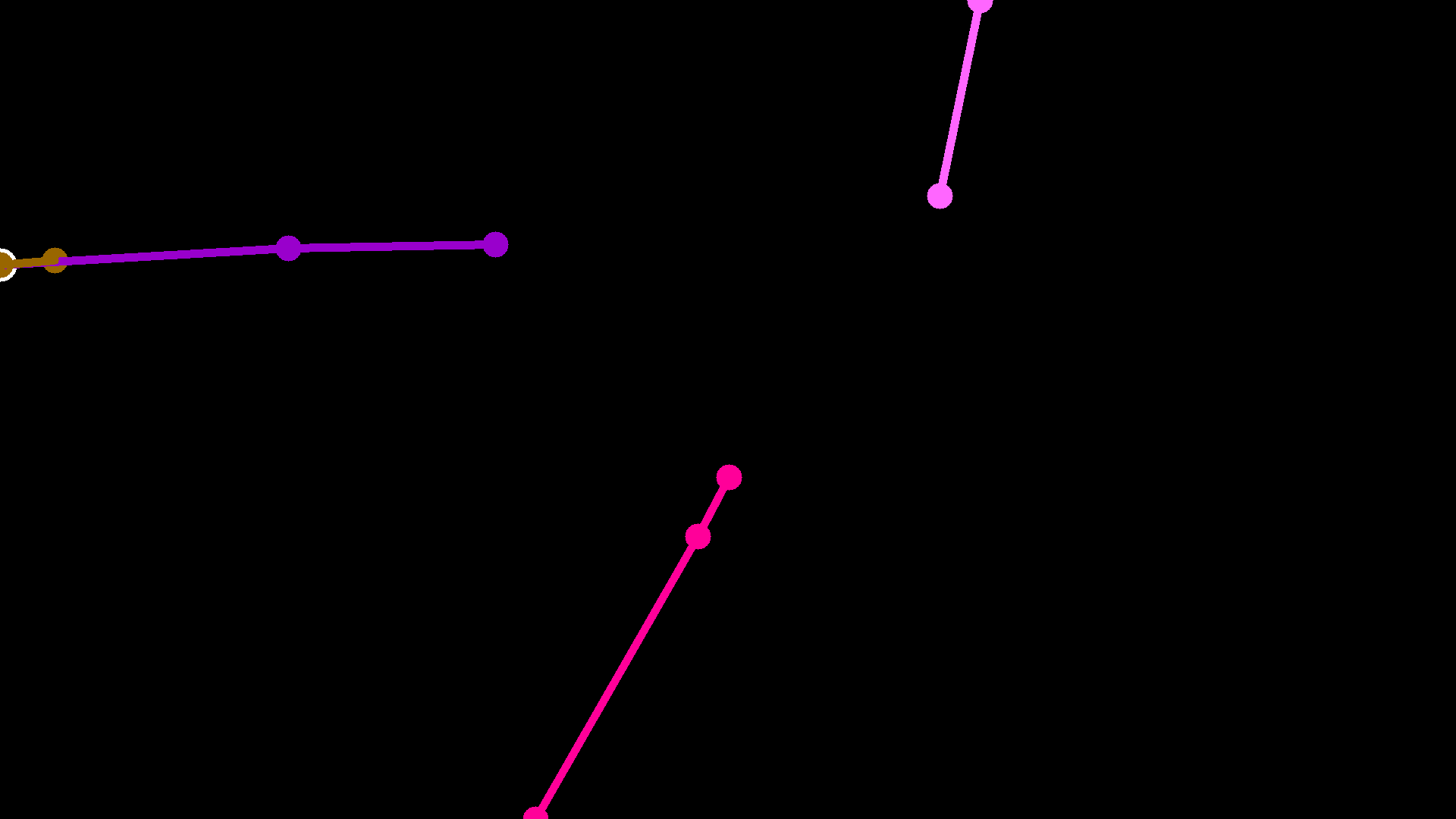}\\
\end{minipage}
&
\begin{minipage}{0.165\textwidth}
    \centering
    \includegraphics[width=\linewidth]{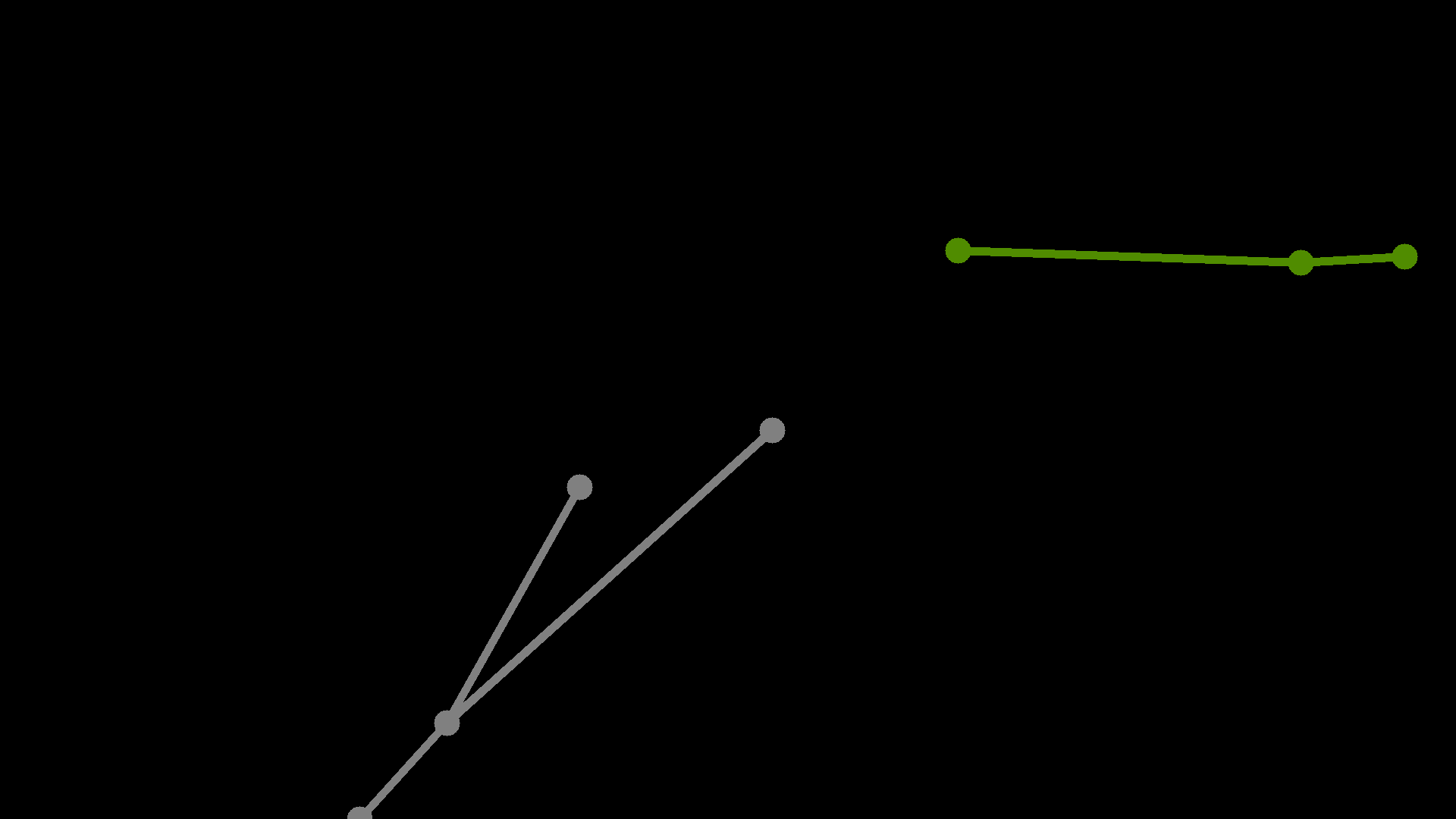}\\
\end{minipage}
&
\begin{minipage}{0.165\textwidth}
    \centering
    \includegraphics[width=\linewidth]{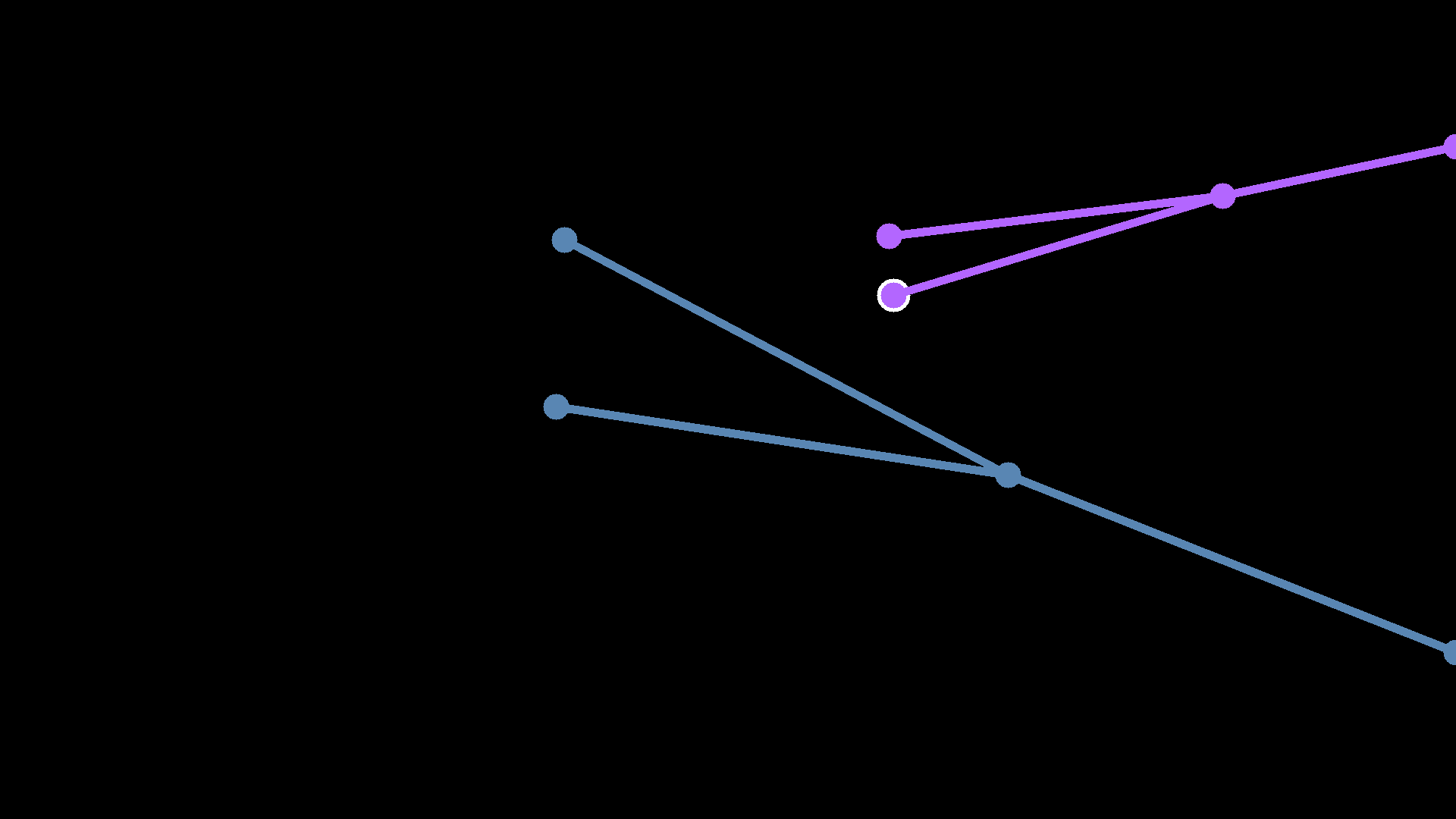}\\
\end{minipage}
&
\begin{minipage}{0.165\textwidth}
    \centering
    \includegraphics[width=\linewidth]{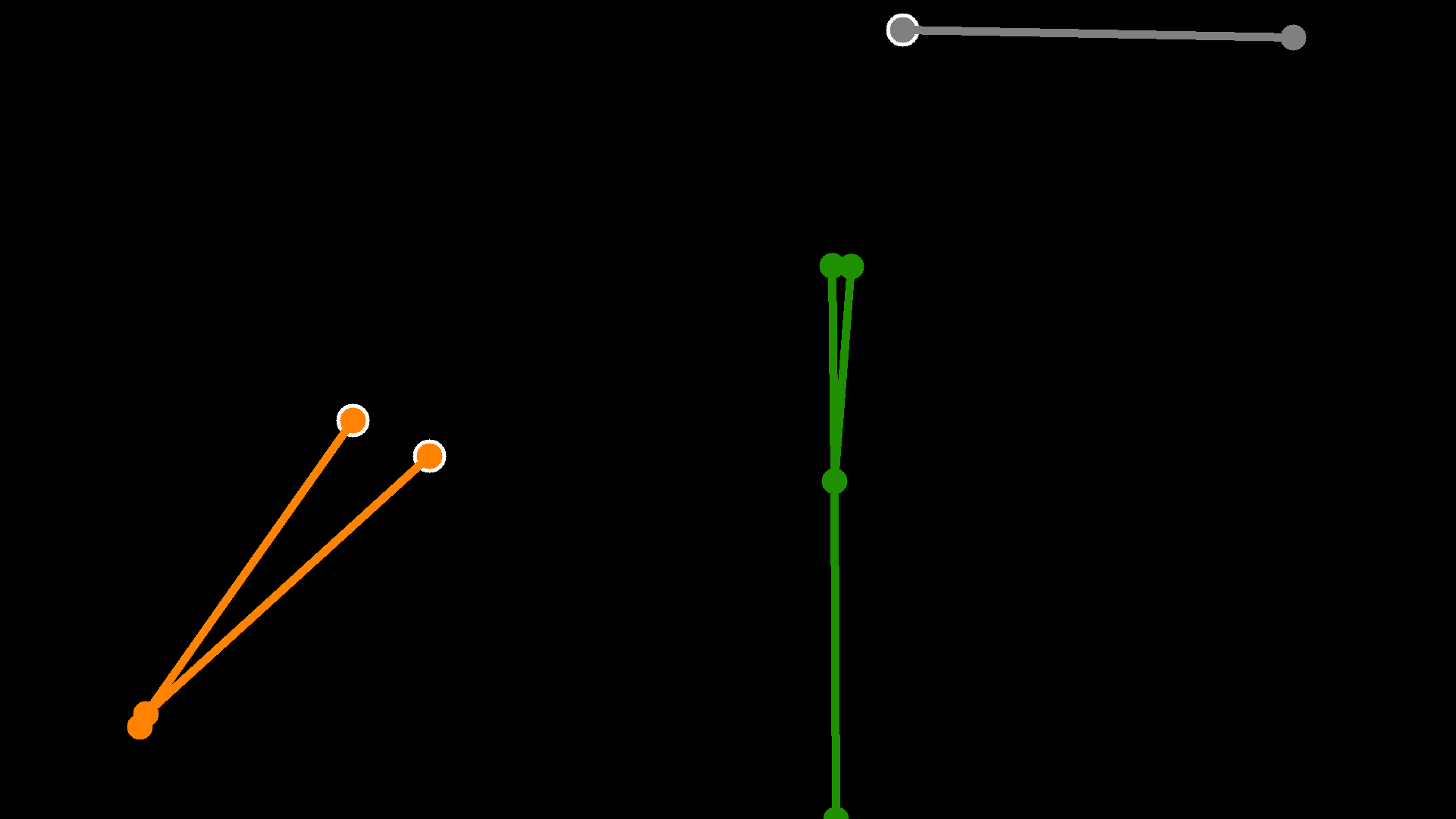}\\
\end{minipage}
\\
\midrule
\multicolumn{5}{c}{\textbf{Participant predictions}} \\
\midrule
\multirow{2}{*}{sds-hd} &
\begin{minipage}{0.165\textwidth}
    \centering
    \includegraphics[width=\linewidth]{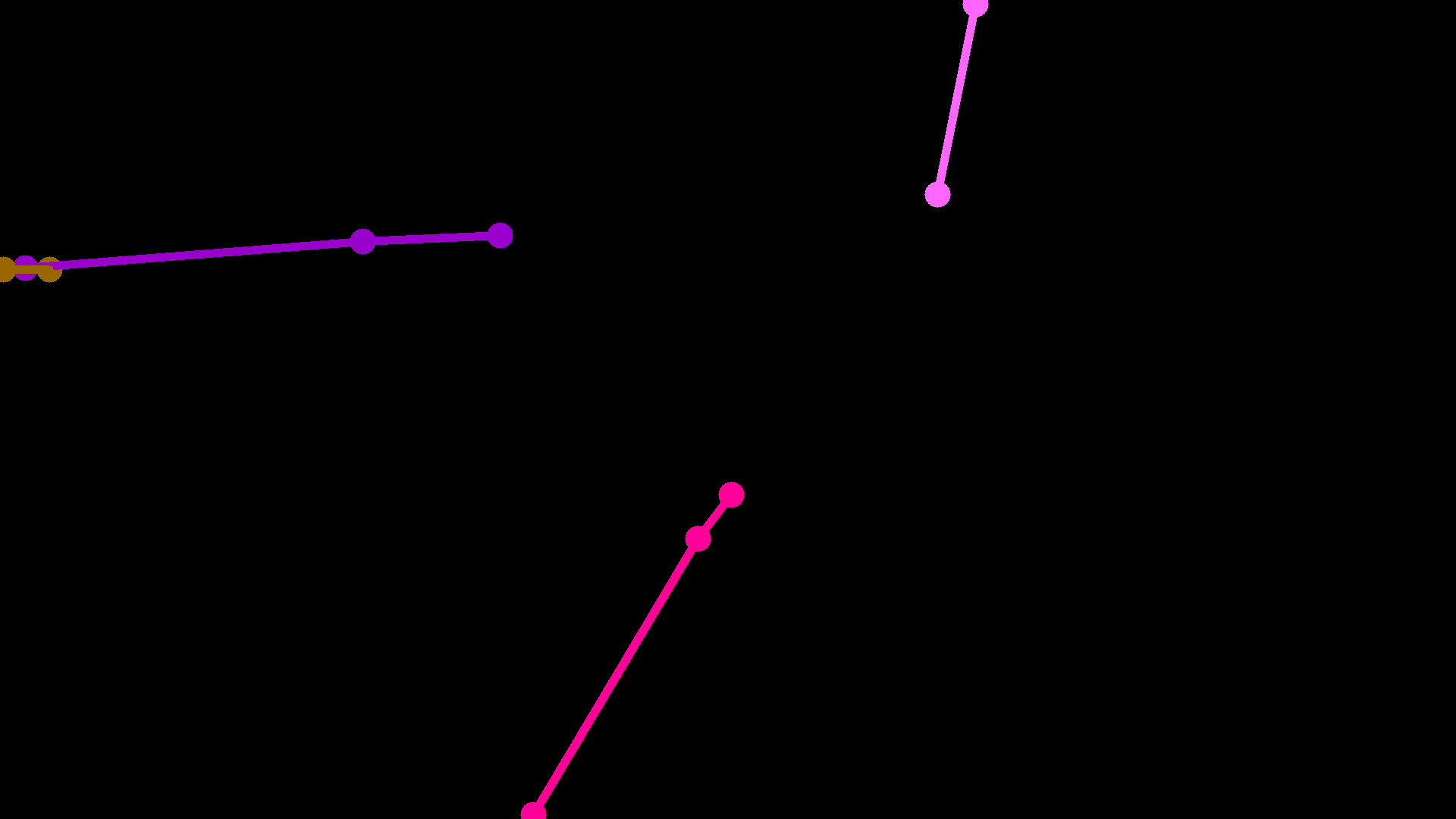}\\[-2pt]
    {\footnotesize 31.82}
\end{minipage}
&
\begin{minipage}{0.165\textwidth}
    \centering
    \includegraphics[width=\linewidth]{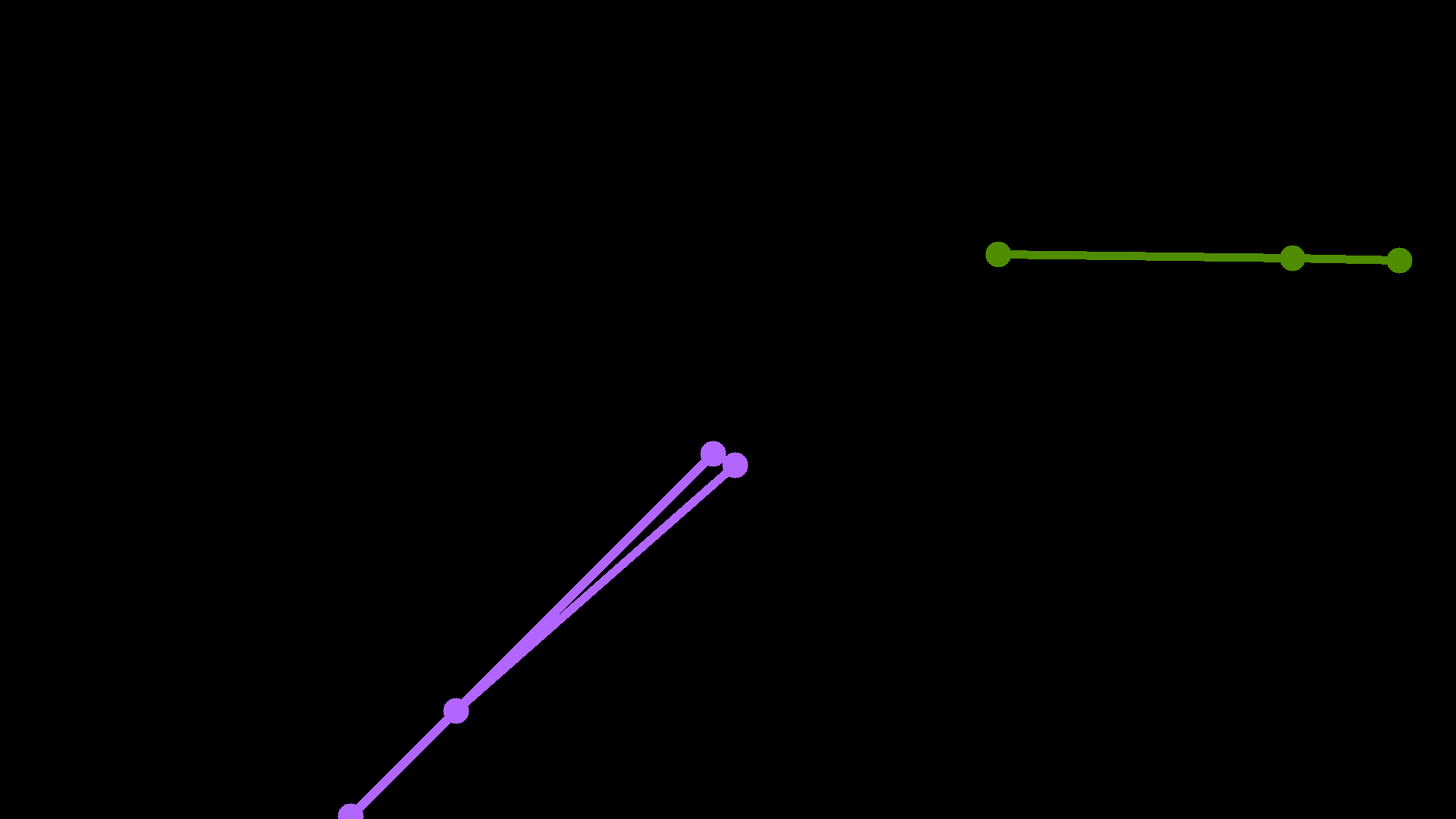}\\[-2pt]
    {\footnotesize 4.17}
\end{minipage}
&
\begin{minipage}{0.165\textwidth}
    \centering
    \includegraphics[width=\linewidth]{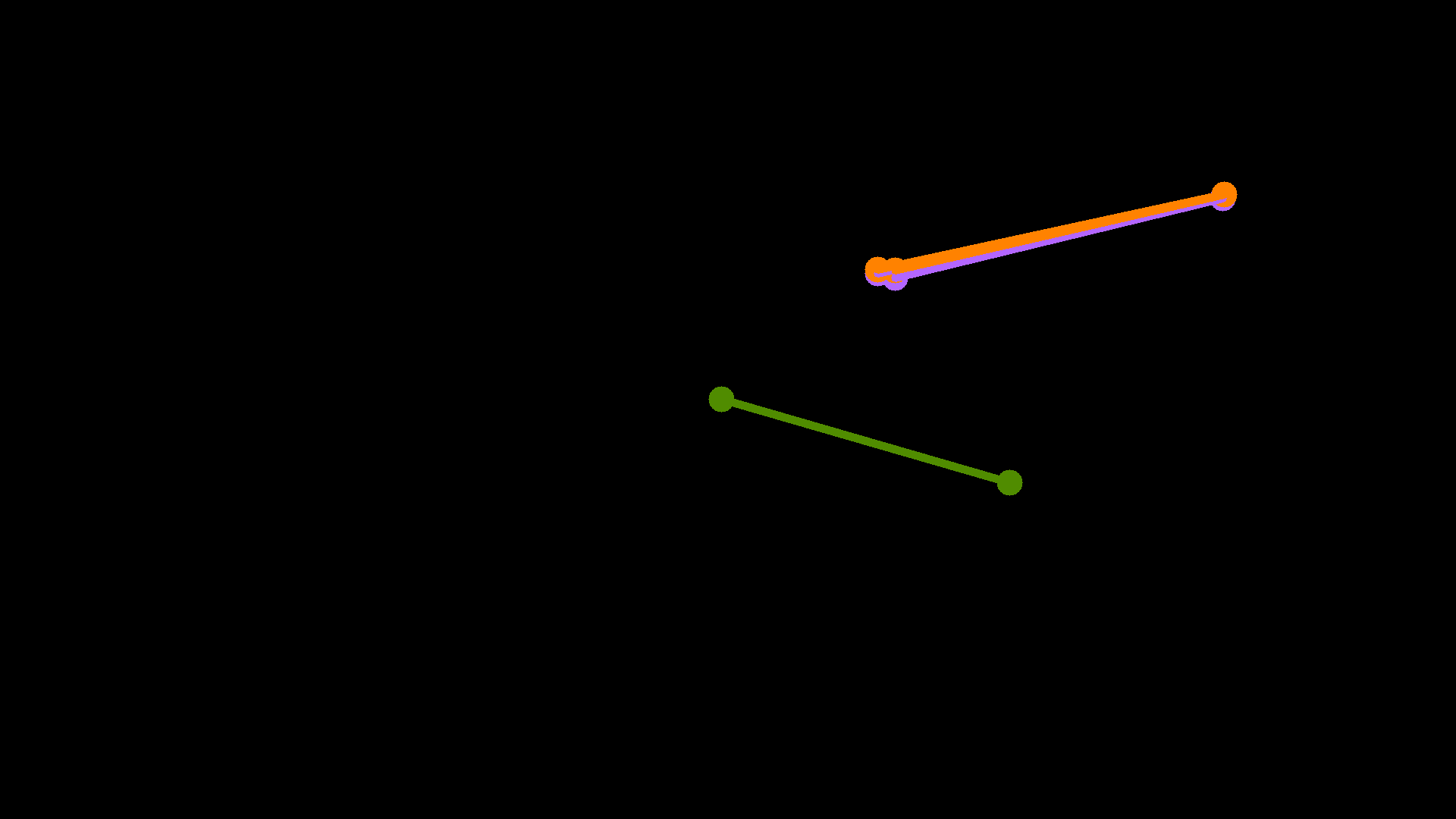}\\[-2pt]
    {\footnotesize 4.55}
\end{minipage}
&
\begin{minipage}{0.165\textwidth}
    \centering
    \includegraphics[width=\linewidth]{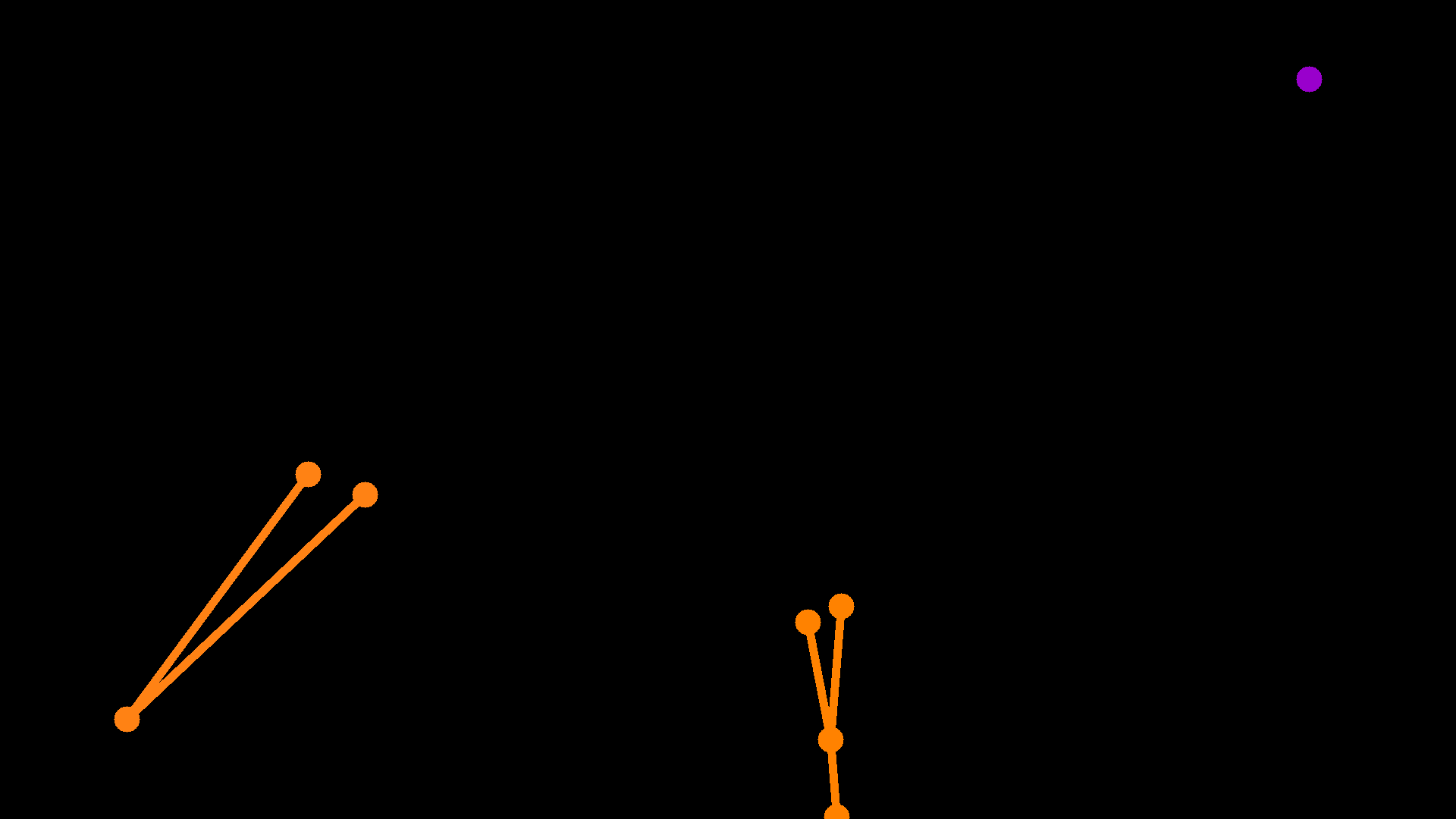}\\[-2pt]
    {\footnotesize 0.00}
\end{minipage}
\\
\\[-1.5em]
\multirow{2}{*}{alvaro} &
\begin{minipage}{0.165\textwidth}
    \centering
    \includegraphics[width=\linewidth]{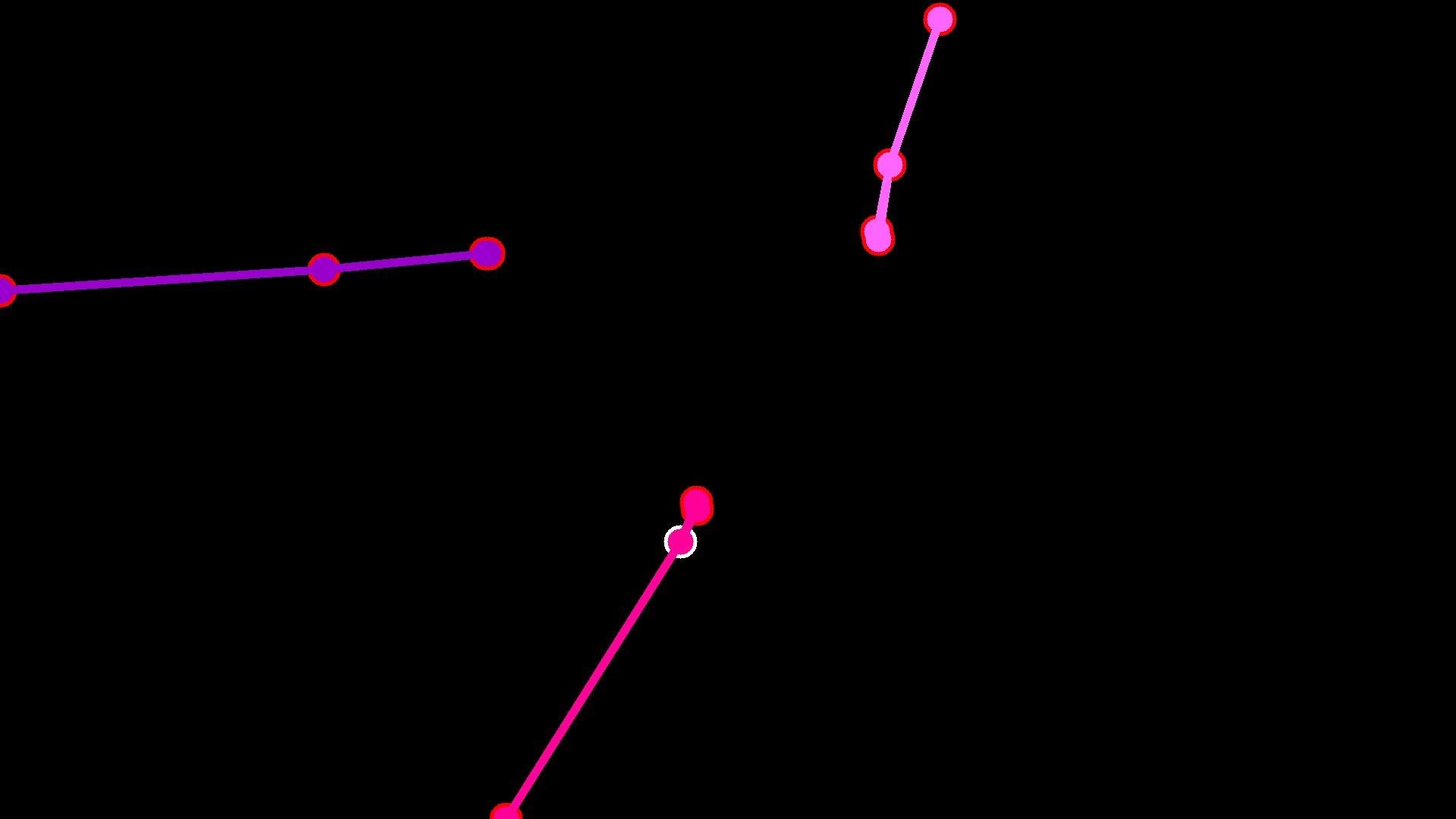}\\[-2pt]
    {\footnotesize 9.09}
\end{minipage}
&
\begin{minipage}{0.165\textwidth}
    \centering
    \includegraphics[width=\linewidth]{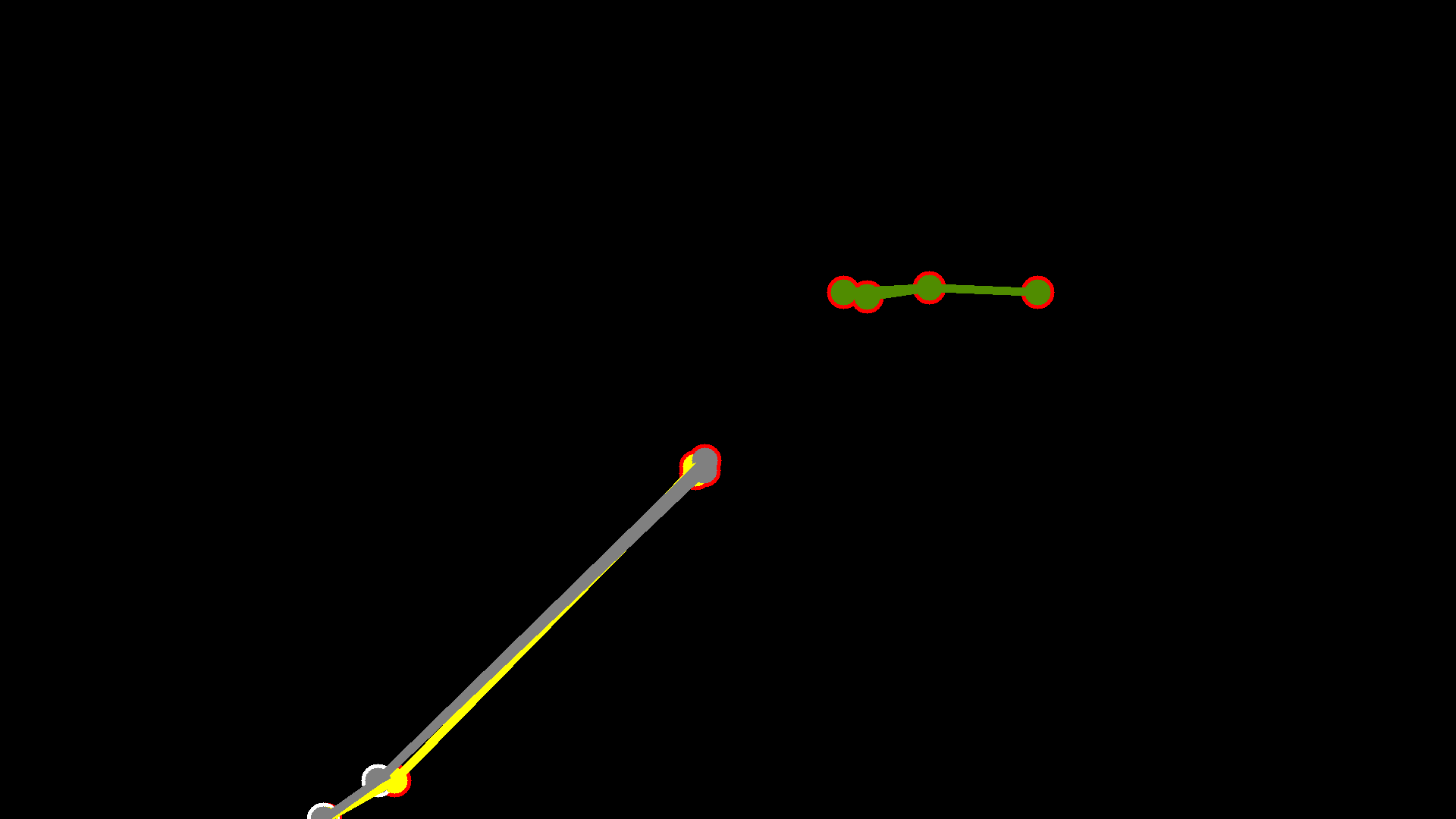}\\[-2pt]
    {\footnotesize 0.00}
\end{minipage}
&
\begin{minipage}{0.165\textwidth}
    \centering
    \includegraphics[width=\linewidth]{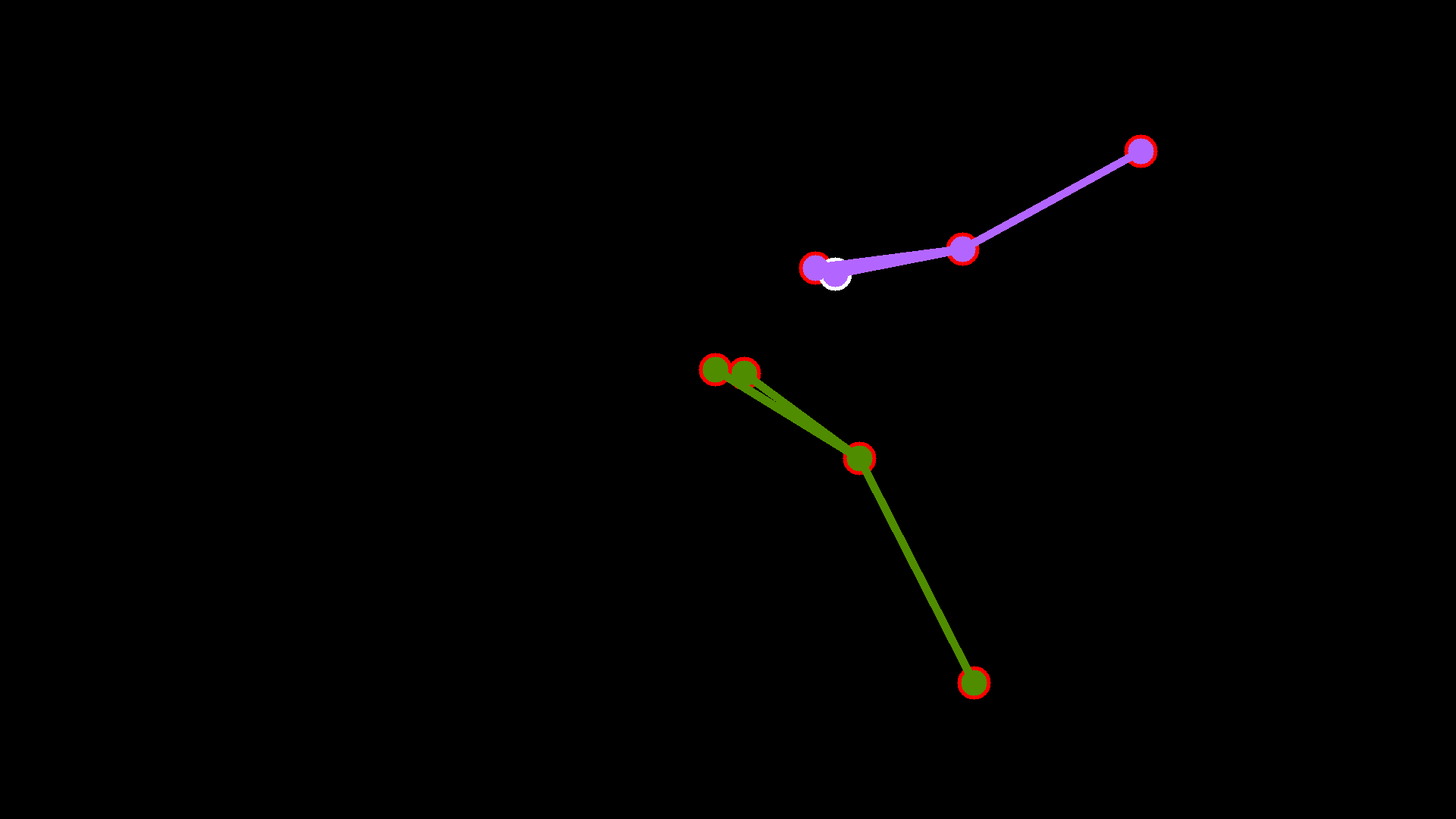}\\[-2pt]
    {\footnotesize 0.00}
\end{minipage}
&
\begin{minipage}{0.165\textwidth}
    \centering
    \includegraphics[width=\linewidth]{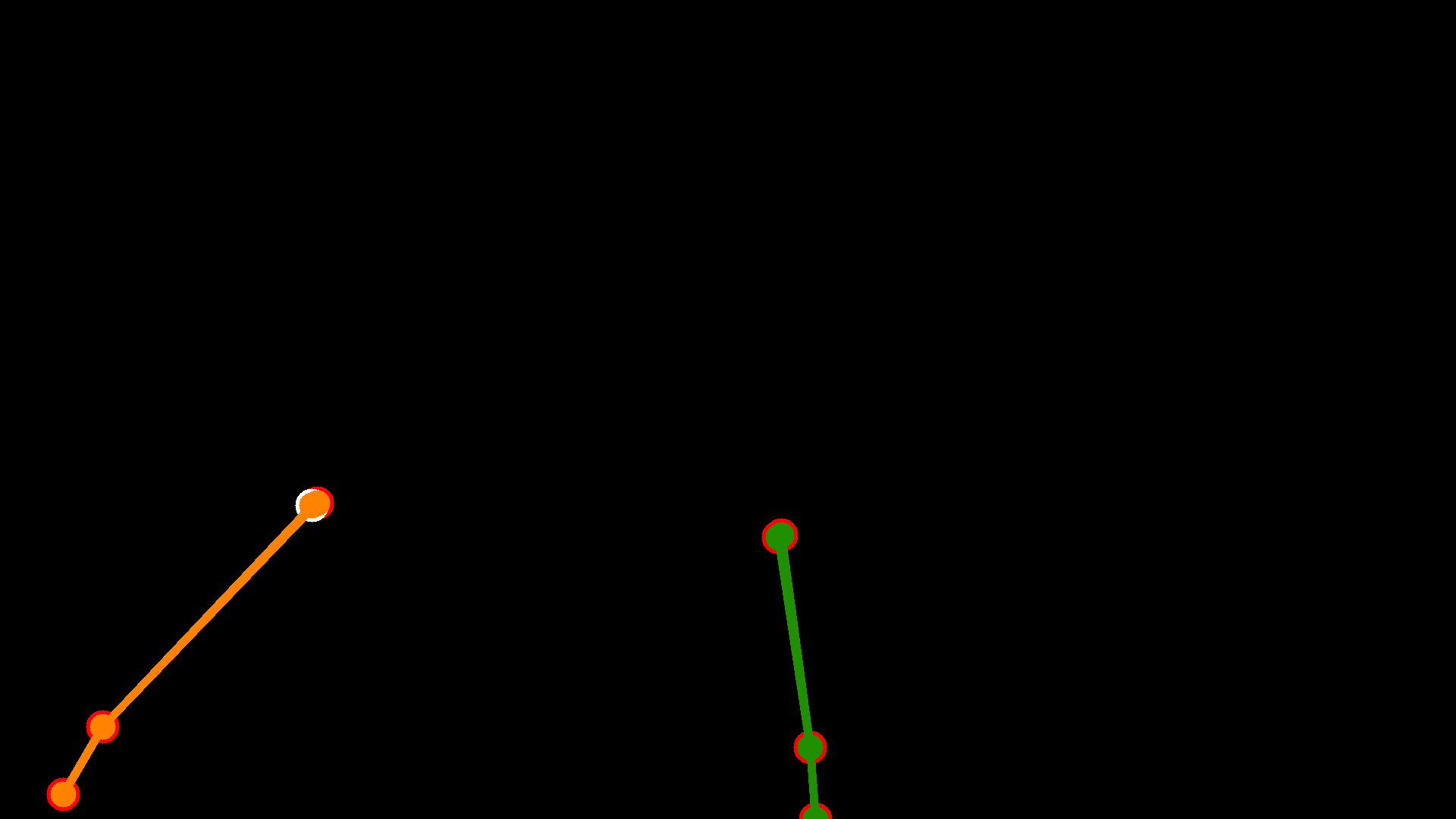}\\[-2pt]
    {\footnotesize 0.00}
\end{minipage}
\\

\bottomrule
\end{tabularx}
\label{results:keypoint_estimation:tab:qualitative_results_page_01}
\end{figure*}

\begin{table*}[t]
    \footnotesize
    \centering
    \caption{Results of the instrument keypoint estimation task for each team and each individual video in the test dataset, together with the mean metric values for each video}
    \begin{tabularx}{\textwidth}{XRRRRR}
        \toprule
        \multirow{2}{*}{Team} 
        & \multicolumn{5}{c}{$\text{mAP}_{\text{OKS}}$ (in \%) $\uparrow$} \\
        \cmidrule{2-6}
        & V\_06 & V\_08 & V\_09 & V\_12 & V\_14 \\
        \midrule
        \rowcolor{gray!10}
        sds-hd & 49.79 & 19.58 & 53.08 & 14.17 & 17.46 \\
        alvaro & 29.76 & 13.58 & 25.49 & 2.82 & 4.80 \\
        \midrule
        \textbf{mean} & \textbf{39.78} & \textbf{16.58} & \textbf{39.28} & \textbf{8.49} & \textbf{11.13} \\
        \bottomrule
    \end{tabularx}
    \label{results:keypoint_estimation:results_tab_individual_videos}
\end{table*}

As illustrated in Figure~\ref{organization:fig:endovis24}, the PhaKIR challenge attracted the highest number of registrations and submissions among all EndoVis-2024 sub-challenges, highlighting both the high relevance of the research questions and the strong interest from the scientific community.
Notably, PhaKIR was hosted on a bespoke platform developed according to the principles outlined in the ``DIY Challenge Blueprint'' publication~\cite{klausmann2025diy}.
Self-hosting provided several advantages that likely contributed to the high level of engagement.
The platform was fully customizable to the specific needs of the challenge, enabling streamlined submission workflows, tailored evaluation pipelines, and flexible task definitions.
Participants were continuously kept informed via email updates and news posts on the challenge website, reflecting an active communication strategy by the organizers.
This close interaction reduced participation barriers and further supported the recruitment of a large and diverse participant cohort.
In addition, the large dataset size and the high quality of annotations offered a uniquely rich benchmarking opportunity, enhancing the attractiveness of the challenge.
This combination of platform usability, dataset quality, task diversity, and proactive participant engagement distinguished PhaKIR from challenges hosted on standard portals such as Synapse or Grand Challenge.
It should be noted, however, that PhaKIR comprises three distinct tasks, whereas most other sub-challenges focused on a single task.
This difference should be considered when interpreting participation metrics.

\subsection{Challenge dataset}
\label{discussion:dataset}

Our PhaKIR challenge introduced a novel dataset specifically designed to overcome key limitations of existing datasets and previous challenges, as outlined in Chapter~\ref{introduction}.  
By collecting surgical videos from three different medical centers, we ensured a high degree of variability in both surgical phases and instrument usage -- variability that not only exists between institutions but also within a single hospital or procedure type.  
To maintain consistency with established benchmarks, we adopted the widely used surgical phase definitions from the Cholec80 dataset~\cite{twinanda2016endonet} for the surgical phase recognition task.
Only real clinical interventions were included, with no restrictions on instrument types or preselection of video material.  
Providing full-length video sequences of entire procedures allowed participants to leverage temporal context, a valuable yet underutilized aspect in surgical AI research~\cite{rueckert2024methods}.
To ensure high-quality annotations, we followed a rigorous three-step process involving two independent validation stages, all guided by a detailed annotation protocol.
This approach resulted in consistent and reliable labels across the dataset.

Nevertheless, several limitations remain.  
While the dataset contains a substantial number of annotated frames and complete procedural videos, it is based on only 13 patient cases, which may limit the variety of surgical scenarios.  
Furthermore, it focuses exclusively on minimally invasive cholecystectomies -- a specific procedure type that restricts broader applicability to other laparoscopic surgeries.  
All videos were recorded under uniform conditions (25 fps, $1920 \times 1080$ resolution), which may not reflect the diversity of imaging systems used in clinical practice and could impact the generalizability of results.  
To strengthen future evaluations, it will be important to address these aspects through more extensive and diverse data collection, as well as systematic robustness analyses~\cite{ross2021can}.

\subsection{Assessment method}
\label{discussion:assessment}

For the selection of evaluation metrics, we followed the recommendations of the Metrics Reloaded Framework by \cite{maier2024metrics}.
Accordingly, we employed the F1-score and BA for the surgical phase recognition task, DSC, $\text{mAP}_{\text{IoU}}$, and 95\% HD for the instrument instance segmentation task, and $\text{mAP}_{\text{OKS}}$ for the instrument keypoint estimation task.  
Metric values were first computed individually for each phase or instrument class within each frame.
These values were then aggregated across the sequence in two steps: 
first, we averaged across all phases or classes within a sequence using equal weights; second, we averaged across all sequences, again applying uniform weighting.  
This approach ensures that all categories contribute equally to the final metric score, preventing an overrepresentation of frequent phases or classes.  
An alternative approach -- averaging per frame before aggregating - was  
considered but ultimately not adopted, as it would have been more susceptible to distortions caused by class imbalances. 
As shown in Figure~\ref{fig:instrument_occurences}, the frequency of certain instruments varies widely.
Our aggregation strategy was deliberately chosen to minimize the impact of such skewed distributions and ensure fairer comparisons.

To further enhance the robustness of our results, we applied a bootstrapping approach with 10,000 iterations for all evaluations.  
By averaging the metric-specific ranks instead of the raw metric values, the global ranking reflects consistency across multiple performance dimensions.
This approach favors balanced methods and prevents single-metric dominance in the final ranking.

\subsection{Task 1: Surgical phase recognition}
\label{discussion:phase}

\begin{box_phase_recognition}
\section*{Task 1: Lessons learned}
    Key design choices:
\begin{itemize}
\itemsep0em
    \item Usage of various individual architectures
    \item Wide spread in performance across submissions
    \item Highly specialized methods outperform others
    \item Transformer-based models represent the majority
    \item Advanced strategies for temporal context processing are greatly beneficial
\end{itemize}
    Failure cases:
\begin{itemize}
\itemsep0em
    \item High number of misclassifications between phases 
    \item Poor generalization ability between hospitals due to varying quantities of training data
\end{itemize}
    Future research:
\begin{itemize}
\itemsep0em
    \item Exploration on reasons for superior performance of specialized models
    \item Investigation for improved temporal context usage
    \item Enhancing generalization capabilities between medical centers
\end{itemize}
\end{box_phase_recognition}

The results presented in Section~\ref{results:phase_recognition} reveal notable variability in the performance of the submitted methods, regardless of the evaluation metric used.  
This is evident from the quantitative differences between teams, with an average $\text{Mean}_\text{BT}$ difference of 18.89 percentage points for the F1-score, and 5.93 percentage points for the BA.
Ranking stability, as illustrated in Figure~\ref{fig:resuts:phase_recognition:ranking_robustness:blobs}, further underscores the spread in performance across submissions. 

A methodical comparison reveals that the architectures employed are highly individual, suggesting that there is currently no consensus within the community on a common model for this task. 
It can further be stated that the highly specialized phase recognition architecture of the winning team uniandes24 is superior to the general classification models used by the following ranked teams.
This result indicates that the development of task-specific methods has a significant impact on performance and is particularly influential in conjunction with the size and quality of the available dataset.
Furthermore, it can be assumed that the advanced use of temporal information in the task-specific architecture of the uniandes24 team also contributes to the better results.
Compared to the more straightforward approaches of the jmees\_inc. and yipingli teams to incorporate temporal context, it is evident that uniandes24's two-stage strategy for capturing temporal context across different time periods improves the results.

Our findings align with those of \cite{wagner2019comparative}, especially concerning the challenges of accurately identifying the ClCu phase, which once again proved to be the most difficult for participating teams.  
Moreover, consistent with \cite{wagner2019comparative}'s earlier challenge, all methods submitted to PhaKIR incorporated temporal information, lending further support to the hypothesis that temporal modeling offers clear advantages over frame-wise processing in the context of surgical phase recognition.  
A methodological comparison between both challenges reveals an important shift: whereas the earlier study exclusively featured convolutional neural network approaches, all but one team (ryze) in the PhaKIR challenge employed transformer-based architectures.
This highlights an emerging consensus in the field, reflecting the growing preference for transformer-based models in surgical workflow analysis.

\subsection{Task 2: Instrument instance segmentation}
\label{discussion:segmentation}

\begin{box_instance_segmentation}
\section*{Task 2: Lessons learned}
    Key design choices:
\begin{itemize}
\itemsep0em
    \item Top three methods based on Mask2Former
    \item Pseudo-label generation as key factor
    \item Superior results from transformer-based networks
    \item Usage of temporal information not decisive
    \item Individual pre-training strategies and datasets
\end{itemize}
    Failure cases:
\begin{itemize}
\itemsep0em
    \item Instrument instances rarely differentiated
    \item Underrepresented instruments often misclassified
    \item Poor generalization ability between hospitals due to varying quantities of training data
\end{itemize}
    Future research:
\begin{itemize}
\itemsep0em
    \item Investigation for improved temporal context usage
    \item Better identification of instrument instances
    \item Improved classification of underrepresented tools
    \item Enhancing generalization capabilities between medical centers
\end{itemize}
\end{box_instance_segmentation}

The analysis of results in Section~\ref{results:instrument_segmentation} highlights that the top three teams -- jmees\_inc., uniandes24, and augi -- achieved similarly strong performances, while a noticeable gap separates them from the remaining six teams.  
Rankings based on the DSC and $\text{mAP}_{\text{IoU}}$ metrics were substantially more stable than those derived from the 95\% HD, which showed greater variability.  

As detailed in Table~\ref{tab:methods:segmentation} and visualized in Figure~\ref{results:instance_seg:method_comparison}, the submitted methods demonstrated considerable architectural diversity.
Nevertheless, all three top-performing teams based their solutions on Mask2Former, underscoring its current dominance for instance segmentation in surgical scenes.
A comparison of the methodological approaches suggests that the semi-supervised learning strategy leveraging pseudo-label generation was likely the decisive factor behind the success of the jmees\_inc.~team. 
Furthermore, it can be observed that the advanced use of temporal context implemented by team uniandes24 did not confer a clear advantage over the two competitors who did not incorporate any temporal information. 

Despite this, the qualitative evaluation -- particularly sample 3 in Table~\ref{results:instrument_seg:tab:qualitative_results_page_01} -- revealed a key limitation: even the highest-ranking methods failed to distinguish between multiple instances of the same instrument class (of the blunt-grasper-spec.), underscoring the persistent challenge of true instance-level segmentation in cluttered surgical environments.  
Interestingly, only two of the nine teams incorporated temporal information into their approach, implying that most participants regarded temporal modeling as less beneficial for this task.  
One team even reported removing a temporal component after observing decreased performance during ablation testing.
This suggests that the advantages of temporal integration may be highly task-specific in the context of surgical instrument segmentation.

\subsection{Task 3: Instrument keypoint estimation}
\label{discussion:keypoints}

\begin{box_keypoint_estimation}
\section*{Task 3: Lessons learned}
    Key design choices:
\begin{itemize}
\itemsep0em
    \item YOLOv8 architecture represents current standard
    \item Human pose estimation networks may not be suitable for instrument recognition
    \item Uncertainty estimation techniques could not improve performance
    \item Substantial differences in results due to deviations in training procedure
\end{itemize}
    Failure cases:
\begin{itemize}
\itemsep0em
    \item Underrepresented instruments often misclassified
    \item Coordinate accuracy varies across tool types
    \item Poor generalization ability between hospitals due to varying quantities of training data
\end{itemize}
    Future research:
\begin{itemize}
\itemsep0em
    \item Investigation for improved temporal context usage
    \item Improved classification of underrepresented tools
    \item More accurate predictions of keypoint coordinates
    \item Enhancing generalization capabilities between medical centers
\end{itemize}
\end{box_keypoint_estimation}

The results of the instrument keypoint estimation task, as outlined in Section~\ref{results:keypoint_estimation}, reveal that only two teams participated.
Nonetheless, a substantial performance gap was observed, as reflected by the $\text{mAP}_{\text{OKS}}$ metric.
A methodological comparison of the two approaches indicates that the introduction of uncertainty estimation techniques by team alvaro was not a decisive factor in securing victory.
Although both teams built their approaches on the same underlying architecture, differences in training procedures contributed significantly to the observed performance discrepancy.
Despite these efforts, the overall $\text{mAP}_{\text{OKS}}$ scores remained low for both teams.
This can likely be attributed to the inherent limitations of pose estimation networks, which are typically optimized for human pose recognition involving a single class and a fixed number of keypoints.
In contrast, the PhaKIR dataset includes multiple instrument classes, each associated with a class-specific number of keypoints. 
This added considerable complexity, as accurate classification became a critical prerequisite for precise keypoint localization.

A closer look at Table~\ref{results:keypoint_estimation:tab:qualitative_results_page_01} reveals two central limitations of the submitted methods.
Firstly, analogous to the instrument instance segmentation task, the instrument classes are often confused, particularly in the case of infrequently occurring tools.
Secondly, the coordinates of the predicted keypoints are often rather inaccurate, which can also be seen especially in the case of rarer instruments.

\subsection{Overall findings}
\label{discussion:overall}

An analysis of team participation per task reveals a clear trend: significantly more teams submitted methods for the surgical phase recognition ($n = 7$) and instrument instance segmentation ($n = 9$) tasks compared to the instrument keypoint estimation task ($n = 2$).  
This imbalance suggests that the former two tasks currently attract greater attention within the research community.
However, it is important to note that the keypoint estimation task addresses clinically meaningful challenges, such as the localization of occluded instrument parts, not covered by the other tasks and highly relevant for computer-assisted surgery.  

In terms of temporal modeling, all teams in the surgical phase recognition task incorporated temporal context, whereas only two out of nine teams did so for the instrument instance segmentation task, and none for the instrument keypoint estimation task.  
These findings indicate that temporal context integration is currently more established and accepted in phase recognition, while its potential in the other areas remains underexplored.

A further trend emerges in the architecture choices: in the two tasks where transformer-based models were used, the top-performing teams consistently employed transformer-based approaches, suggesting a shift in preference towards such architectures for surgical video analysis.  

Finally, cross-hospital comparison of test video results reveals a performance bias: algorithms performed best on videos from hospital 1 ($n = 3$), which was also more prominently represented in the training data.
In contrast, performance on videos from hospital 2 ($n = 1$) and hospital 3 ($n = 1$) was consistently lower.  
These results indicate that the predominance of data from hospital 1 within the training set has a significant impact on test performance, enabling the models to achieve improved adaptability to this hospital’s data distribution.
This outcome also highlights a key limitation of all submissions -- the challenge of generalizability across unseen clinical environments remains unsolved.

\subsection{Future research directions}
\label{discussion:future_research}

Future research across the three task domains presented in this challenge should place a strong emphasis on improving the generalizability of algorithmic approaches.
None of the current methods demonstrated satisfactory performance when confronted with data from unfamiliar clinical environments -- a critical shortcoming that must be addressed to enable real-world applicability.
Tackling this challenge will require the development of more comprehensive and diverse datasets that include surgical procedures from a broad range of medical institutions and cover different types of surgeries.
Such datasets should be explicitly designed to support generalization across centers, equipment, and surgical styles.

Additionally, future work could investigate cross-hospital generalization by training models on one or two centers and testing on a completely unseen third.
Such experiments would help identify failure modes and provide insights into the robustness of methods under real-world distribution shifts.

Beyond dataset diversity, another key direction is the design of model architectures and algorithms that explicitly promote generalization. 
Architectural choices, regularization strategies, and training paradigms can all influence a model’s ability to transfer across institutions, equipment, and surgical styles. 
For instance, approaches that incorporate domain-invariant features, multi-scale representations, or advanced attention mechanisms may improve robustness to distribution shifts. 
Investigating such strategies in combination with diverse training data could therefore provide a powerful way to enhance real-world applicability of the developed models.

In particular, the accurate detection and localization of instrument keypoints holds significant promise for real-world clinical applications, such as robotic assistance or intraoperative guidance, but remains underexplored.
Advancing keypoint estimation methods tailored to the specific characteristics of surgical tools and contexts is therefore a key direction for future research. 

Moreover, temporal modeling has proven effective in the surgical phase recognition task, but is still rarely employed in the other two domains.
Exploring how temporal context can be harnessed in instance segmentation and keypoint estimation may lead to substantial performance gains and should be a focus of subsequent work.

Unfortunately, we did not have access to manually annotated metadata that would allow us to identify video sequences that are particularly challenging for state-of-the-art methods through an extended structural analysis described by~\cite{ross2023beyond}.
Conducting such a systematic analysis also represents a promising direction for future research, as it could provide valuable insights into the underlying conditions that lead to failure cases and inform the development of more robust methods.

\section{Conclusion}
\label{conclusion}

To overcome the limitations of existing datasets and earlier challenges, we organized the PhaKIR challenge -- covering surgical phase recognition, keypoint estimation, and instrument instance segmentation -- as part of the EndoVis event at MICCAI 2024.  
For this purpose, we introduced a novel multi-center dataset featuring complete real-world laparoscopic cholecystectomy sequences from 13 patients, annotated consistently for all three tasks.
Using this dataset, we systematically benchmarked the performance of submitted algorithms and conducted an in-depth evaluation across a range of metrics.  
Our findings reveal that surgical phase recognition and instrument instance segmentation currently receive greater attention in the research community than the more complex and underexplored task of instrument keypoint estimation.  

Temporal modeling was applied in all phase recognition approaches, confirming its value in this domain.
In contrast, temporal context was rarely exploited in keypoint estimation or instrument instance segmentation methods, highlighting untapped potential in these areas.  
A key challenge identified across all tasks is generalizability: none of the submitted approaches demonstrated robust performance across surgical videos from different medical centers.  
We believe that the dataset and challenge presented here make a valuable contribution to the field of computer- and robot-assisted minimally invasive surgery.
We hope the insights gained will stimulate further research, foster innovation, and support the development of more reliable and transferable AI methods in surgical contexts.

\section*{Declaration of competing interest}

The authors declare that they have no known competing financial interests or personal relationships that could have appeared to
influence the work reported in this paper.

\section*{Data availability statement}

The dataset was published under a Creative Commons  AttributionNonCommercial-ShareAlike (CC BY-NC-SA) license on Zenodo (\url{https://doi.org/10.5281/zenodo.15740620}), which means that it is publicly available for non-commercial usage. Should you wish to use or refer to this dataset, you must cite this paper and the associated dataset paper~\cite{rueckert2025video}. The licensing of new creations
must use the exact same terms as in the current version of the dataset. 

\section*{CRediT authorship contribution statement}

A draft was sent to all authors prior to submission, all comments were integrated and all authors agreed to publish the paper.
The CRediT authorship contributions are as follows:

\textbf{T. Rueckert:} Conceptualization, Data curation, Formal analysis, Methodology, Project administration, Software, Validation, Visualization, Writing – original draft, Writing - Review \& Editing.
\textbf{D. Rauber:} Conceptualization, Formal analysis, Methodology, Software, Validation, Visualization.
\textbf{R. Maerkl:} Conceptualization, Data Curation, Formal analysis, Methodology, Software, Validation.
\textbf{L. Klausmann:} Conceptualization, Methodology, Resources, Software, Validation.
\textbf{S. R. Yildiran:} Conceptualization, Methodology, Software.
\textbf{M. Gutbrod:} Conceptualization, Data curation, Formal analysis, Methodology, Software, Validation.
\textbf{D. W. Nunes:} Conceptualization, Methodology, Software.
\textbf{A. F. Moreno, I. Luengo, D. Stoyanov, N. Toussaint, E. Cho, H. B. Kim, O. S. Choo, K. Y. Kim, S. T. Kim, G. Arantes, K. Song, J. Zhu, J. Xiong, T. Lin, S. Kikuchi, H. Matsuzaki, A. Kouno, J. R. R. Manesco, J. P. Papa, T.-M. Choi, T. K. Jeong, J. Park, O. Alabi, M. Wei, T. Vercauteren, R. Wu, M. Xu, A. Wang, L. Bai, H. Ren, A. Yamlahi, J. Hennighausen, L. Maier-Hein, S. Kondo, S. Kasai, K. Hirasawa, S. Yang, Y. Wang, H. Chen, S. Rodríguez, N. Aparicio, L. Manrique, J. C. Lyons, O. Hosie, N. Ayobi, P. Arbeláez, Y. Li, Y. A. Khalil, S. Nasirihaghighi:} Data analysis, Software, Writing – review \& editing.
\textbf{S. Speidel:} Conceptualization, Methodology, Writing – review \& editing.
\textbf{D. Rueckert:} Conceptualization, Methodology, Supervision, Writing – review \& editing.
\textbf{H. Feussner, D. Wilhelm:} Conceptualization, Data Curation, Funding acquisition, Methodology, Supervision, Writing – review \& editing.
\textbf{C. Palm:} Conceptualization, Funding acquisition, Methodology, Project administration, Resources, Supervision, Writing – original draft, Writing - Review \& Editing.

\section*{Acknowledgements / Authors' notes}
\label{acknowledgements}

Funding: This work was supported by the bavarian research foundation [BFS, grant number AZ-1506-21], and the Bavarian Academic Forum [BayWISS].

During the preparation of this work the authors used \hbox{ChatGPT} (GPT-4o, OpenAI) for language refinement to improve the clarity and readability of the manuscript. After using this tool/service, the authors reviewed and edited the content as needed and take full responsibility for the content of the publication.


\appendix

\section{Challenge organization}
\label{appendix:challenge_orga}

Three parties were involved in the challenge organization.
The research group for Minimally Invasive Interdisciplinary Therapeutic Intervention (MITI) (\url{https://web.med.tum.de/en/miti/}) at the Technical University of Munich (TUM) obtained the source data from the three different medical centers and created the annotations.
The Regensburg Medical Image Computing (ReMIC) (\url{https://re-mic.de/}) laboratory at the Ost\-baye\-ri\-sche Technische Hochschule Regensburg (OTH Regensburg) carried out the technical implementation, which included the pre-processing and the provision of the data, the registration and administration of the participants, the submission process, and the evaluation and presentation of the results.
The company AKTORmed - Robotic Surgery (\url{https://aktormed.info/}) provided the prize money of €3,000 which was distributed equally across the three tasks.

\section{Data description and labeling instructions}
\label{appendix:data_description}

See Supplementary file S1.

\section{Submission instructions}
\label{appendix:submission_instructions}

Information on the submission process was provided to the participants both directly on the website (\url{https://phakir.re-mic.de/participation/}) and in a separate GitHub repository (\url{https://github.com/schnadoslin/PhaKIR_Submission_Template/}).

\section{Challenge design document}
\label{appendix:challenge_design}

See Supplementary file S2.

\section*{Supplementary material}
\label{appendix:supplementary_material}

Supplementary material associated with this article can be found at the arXiv page.

\bibliographystyle{unsrtnat}

\bibliography{PhaKIR_Challenge_Paper_arXiv}

\end{document}